\documentclass[lettersize,journal]{IEEEtran}
\usepackage{amsmath,amsfonts}
\usepackage{algorithmic}
\usepackage{algorithm}
\usepackage{array}
\usepackage[caption=false,font=normalsize,labelfont=sf,textfont=sf]{subfig}
\usepackage{textcomp}
\usepackage{stfloats}
\usepackage{url}
\usepackage{verbatim}
\usepackage{graphicx}
\usepackage{xcolor}
\usepackage{cite}
\usepackage{amsmath}
\usepackage{amssymb}
\usepackage{booktabs}
 \usepackage{multirow}
\begin{document}

\title{When Do Learned Priors Help Visual–Inertial Estimation? A Controlled Study of Prior Integration, Calibration, Initialization, and Backend Consistency}

\author{
Jinchang Zhang\textsuperscript{1},
Guoyu Lu\textsuperscript{1,*}
\\[3pt]
{\small
\textsuperscript{1}Department of Computer Science,
Indiana University,
Bloomington, IN 47405, USA.
}
\\[-1pt]
{\small
\textsuperscript{*}Corresponding author:
guoyulu62@gmail.com
}
}

\markboth{Journal of \LaTeX\ Class Files,~Vol.~14, No.~8, August~2021}%
{Shell \MakeLowercase{\textit{et al.}}: A Sample Article Using IEEEtran.cls for IEEE Journals}


\maketitle

\begin{abstract}
Learned components are increasingly integrated into geometric
visual--inertial estimators to provide motion, depth, bias,
uncertainty, or confidence information. However, it remains
unclear when the resulting system-level improvement reflects
genuinely useful , rather than changes in the
backend, calibration strategy, initialization, temporal
association, or evaluation gauge. This paper presents a
controlled evaluation framework for learning-augmented
visual--inertial estimation. The framework separates fusion gain
from the incremental value of a learned prior and distinguishes
four evidence layers: local motion consistency, global trajectory
accuracy, physical-state correctness, and numerical or
implementation consistency.
We instantiate the framework using a representative learned
monocular motion prior introduced into an otherwise unchanged
VINS backend. In the evaluated implementation, the prior is
obtained from a MonoViT-based pose predictor and incorporated
as a local relative-motion factor. We first characterize the
standalone information and failure modes of learned visual
motion, IMU propagation, EKF, and Original VINS. We then
compare Original VINS with VINS augmented by the learned prior
under matched sensor streams, accepted timestamps,
initialization, frontend and backend configurations, and fixed or
online camera--IMU extrinsics. The evaluation further examines
cross-sequence behavior, calibration perturbations, initialization
dependence, backend state coupling, scale and bundle-adjustment
behavior, and loop-closure response.
On the representative KITTI sequence, fixing the reference
extrinsic yields translation APE RMSEs of 31.4~m and 31.8~m
for Original VINS and VINS with the evaluated learned prior,
respectively. Across a four-recording stress panel, the current
prior changes mean APE by only $-0.2\%$, while the tested
five-times-higher weight increases it by $8.2\%$. Releasing the
reference extrinsic for online updating increases mean APE by
$45.7\%$ for Original VINS and $52.1\%$ for the
learning-augmented configuration, whereas mean RPE changes by
less than $2\%$. Matched runs can also reach comparable
backend objectives despite different global and calibration
errors, and global correction can improve or degrade accuracy
depending on the state supplied by the local estimator.
These results show that the usefulness of learned priors cannot
be inferred from fusion performance alone. Reliable evaluation
requires same-backend controls and joint consideration of prior
compatibility, calibration, initialization, global drift, physical
state error, and backend consistency.
\end{abstract}



\section{Introduction}
\label{sec:introduction}

Visual--inertial state estimation is a core capability for mobile
robotics, autonomous navigation, and augmented-reality systems because
cameras and inertial sensors provide complementary motion information.
Cameras supply rich geometric constraints, but monocular translation is
observable only up to an unknown scale, whereas inertial measurement
units provide high-rate angular velocity and specific force but remain
sensitive to bias, gravity alignment, initial conditions, and integration
error. Geometric visual--inertial odometry (VIO) has consequently
evolved from filtering-based estimators to tightly coupled sliding-window
optimization, allowing visual reprojection and inertial preintegration
constraints to jointly recover metric motion
\cite{Huang2019Review,Forster2017Preintegration,
Mourikis2007MSCKF,Leutenegger2015OKVIS,Qin2018VINSMono}.
Nevertheless, reliable estimation still depends on consistent coordinate,
temporal, scale, calibration, and initialization conventions. A locally
smooth visual trajectory, stable inertial propagation, or numerically
converged backend does not separately guarantee an accurate long-term
trajectory.

Recent learning-based methods have expanded this design space by
replacing or augmenting parts of the geometric pipeline. End-to-end
systems learn visual or visual--inertial motion directly, while hybrid
systems provide learned pose, depth, bias, uncertainty, or gating
information to an explicit estimator
\cite{Wang2017DeepVO,Clark2017VINet,Chen2019SelectiveFusion,
Yang2020D3VO,Zuo2021CodeVIO,Buchanan2023DeepBias}.
Among these interfaces, learned local motion is particularly attractive
as an auxiliary constraint because it can provide a continuous
data-driven motion cue without discarding the metric state and physical
structure of a geometric backend. However, a learned prediction is not
automatically a compatible visual--inertial measurement. Its translation
scale, reference frame, transformation direction, temporal interval, and
uncertainty must agree with the state optimized by the backend.
Otherwise, the learned residual may be ignored when underweighted, or
may force disagreement into trajectory, scale, velocity, inertial bias,
landmark depth, or camera--IMU extrinsic states when overweighted.
\begin{figure*}[tbp]
    \centering
    \includegraphics[width=0.98\textwidth]
{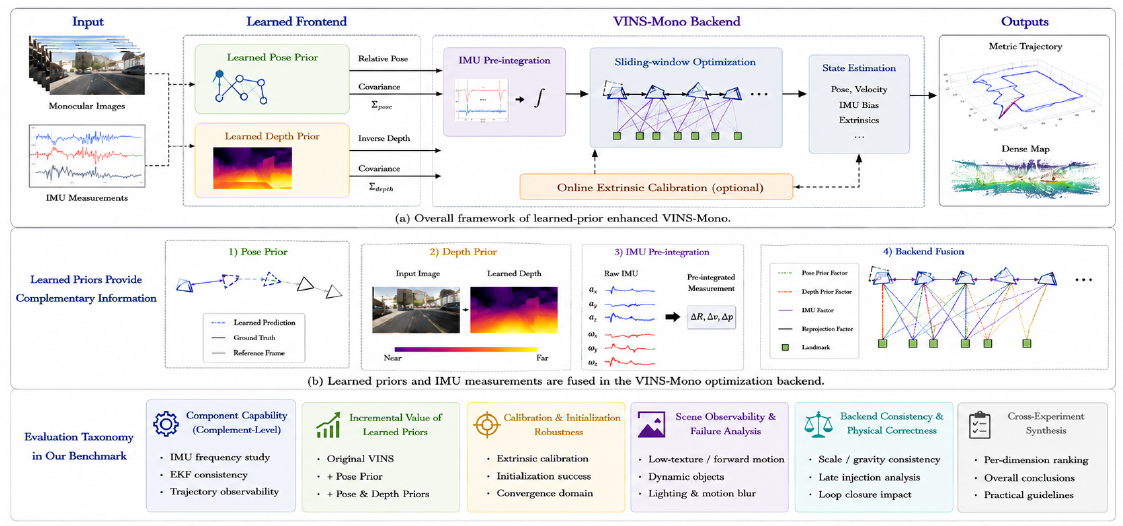}
    \caption{Overview of the evaluated learning-augmented
    visual--inertial pipeline and the associated evaluation roadmap.
    (a) Learned motion information, with learned depth retained as an
    auxiliary diagnostic, is introduced into an otherwise unchanged
    geometric visual--inertial backend together with visual
    reprojection and IMU-preintegration constraints; camera--IMU
    extrinsics may be fixed or released for online optimization.
    (b) The learned factors complement rather than replace the native
    geometric constraints. (c) The evaluation separates component
    capability, same-backend incremental value, calibration and
    initialization sensitivity, sequence-dependent failure conditions,
    backend and physical consistency, and cross-experiment synthesis.}
    \label{arch}
\end{figure*}
This interaction creates a central ambiguity in the evaluation of
learning-augmented VIO. Showing that a fused estimator outperforms a
standalone pose network or direct IMU integration demonstrates the value
of the complete visual--inertial system, but it does not isolate the
incremental contribution of the learned measurement. End-to-end and
hybrid systems may differ simultaneously in representation, sensor
preprocessing, initialization, calibration freedom, fusion rule,
optimization objective, and training distribution
\cite{Clark2017VINet,Chen2019SelectiveFusion,Yang2020D3VO,
Zuo2021CodeVIO}. A better final trajectory therefore cannot by itself
identify whether the gain originates from the learned prior, the native
geometric backend, or a changed evaluation protocol.
The central question of this work is consequently not merely whether
visual--inertial fusion is better than standalone learned pose estimation,
but whether  provides identifiable incremental value
to the \emph{same} geometric backend under otherwise matched
conditions. Answering this question requires the same image and IMU
streams, accepted timestamps, initialization window, frontend, backend,
calibration convention, and output gauge, with the presence and
weighting of the learned factor as the principal changed variables.

Performance attribution becomes more difficult when camera--IMU
calibration is optimized online. Offline calibration can be treated as a
trusted fixed quantity, whereas online self-calibration introduces
additional rotational and translational degrees of freedom that are
jointly optimized with pose, velocity, structure, gravity, and inertial
biases
\cite{Furgale2013Kalibr,Mirzaei2008Calibration,
Hesch2014Observability,Yang2023SelfCalibration}.
This flexibility is necessary when the physical transformation is
unknown or time varying, but it also enlarges the state space through
which model, scale, or synchronization errors can be compensated.
Initialization determines the basin in which filtering or nonlinear
optimization proceeds, while the available motion and visual geometry
determine how strongly the additional calibration states are constrained.
Consequently, a stable extrinsic plateau, a smooth trajectory, or a low
objective value is evidence of numerical consistency under the selected
residuals; it is not by itself proof of theoretical observability or
physical calibration correctness.

Reliable evaluation must therefore separate several evidence layers.
Relative pose error (RPE) characterizes short-interval motion
consistency, whereas absolute pose error (APE) characterizes accumulated
global disagreement under a declared alignment. Metric estimators and
scale-ambiguous monocular estimators must not be compared without an
explicit $\mathrm{SE}(3)$ or $\mathrm{Sim}(3)$ gauge policy, and
camera--IMU rotation should be evaluated in a common coordinate frame
rather than through unverified Euler-angle components
\cite{Sturm2012RGBDBenchmark,Zhang2018TrajectoryEvaluation,
Umeyama1991Alignment}. Trajectory metrics must also be distinguished
from physical-state errors in extrinsics, scale, gravity, bias, and
synchronization, as well as from numerical and implementation evidence
such as backend cost, convergence, repeatability, and loop-closure
response. Agreement at one layer does not imply agreement at the others.

These dependencies motivate the system abstraction and evaluation
roadmap shown in Fig.~\ref{arch}. The upper part of the figure separates
the  from the native visual reprojection and
IMU-preintegration constraints of the geometric backend. It also shows
that camera--IMU extrinsics may either remain fixed or be released as
online optimization variables. The middle part emphasizes that learned
factors complement rather than replace the geometric measurements. The
bottom part organizes the study into component capability, same-backend
incremental value, calibration and initialization robustness,
sequence-dependent failure analysis, backend and physical consistency,
and cross-experiment synthesis.

Figure~\ref{arch} should be interpreted as an evaluation abstraction
rather than as a claim of a new end-to-end VIO architecture. The general
framework applies to learned motion, depth, bias, uncertainty, confidence,
and other information sources that can be introduced into a geometric
estimator. In the principal controlled experiments of this paper, the
learned input is a monocular relative-motion prediction introduced as an
additional backend factor. Learned depth is retained only as a secondary
diagnostic, while the native visual reprojection and inertial factors
remain active in every controlled comparison.

We instantiate this framework using the adjacent-frame PoseNet supplied
with MonoViT \cite{Zhao2022MonoViT}; its self-supervised
pose-estimation design follows Monodepth2
\cite{Godard2019Monodepth2}. The learned relative-motion output is
evaluated within an otherwise unchanged VINS-Fusion backend
\cite{Qin2018VINSMono,Qin2018TemporalCalibration}.
This instantiation provides a representative case through which the
broader evaluation problem can be examined without claiming that the
quantitative behavior of one learned model applies universally to every
learned prior.
Following the roadmap in Fig.~\ref{arch}, the study first characterizes
the information and failure modes of standalone learned visual motion,
IMU-only propagation, an EKF baseline, and Original VINS. It then
compares Original VINS and VINS augmented with the learned motion factor
using matched sensor streams and timestamps, while separating fixed
accurate camera--IMU extrinsics from accurate initialization followed by
online updating. The evaluation further examines prior weighting,
extrinsic covariance and perturbation, initialization-to-final calibration
behavior, sequence-dependent success and failure, backend state
coupling, scale consistency, late calibration correction, and
loop-closure response. A VINS+Pose+Depth configuration and visual-track
observations are retained only as secondary diagnostics rather than as
independent evaluation lines.

The experiments are organized around five questions. First, what local
cues and global failure modes are provided by each standalone component?
Second, does the learned motion prior add identifiable value to the same
VINS backend? Third, how do the initial extrinsic, covariance, and fixed
or online update strategy affect calibration and trajectory estimation?
Fourth, which sequence and initialization-window conditions are
associated with successful, partial, or failed convergence? Fifth, why
can local motion error, global trajectory error, physical calibration,
scale, backend cost, repeatability, and global correction lead to
different judgments about the same run?
The evaluation yields four main observations. Under fixed accurate
extrinsics, the evaluated learned motion factor largely preserves the
behavior of Original VINS but does not provide a consistent trajectory
improvement on the representative sequence and four-recording stress
panel; the tested five-times-higher weight also does not recover a gain.
Releasing the same accurate extrinsic for online optimization changes
global APE far more strongly than local RPE, indicating that the added
calibration freedom primarily affects accumulated global states rather
than immediate frame-to-frame tracking. Across initialization and
perturbation tests, the backend can refine a reasonable initial solution
but does not reliably recover a physically plausible calibration after
entering a poor coupled state. Finally, the cross-experiment evidence
shows that low RPE does not imply low APE, low matched-run total cost
does not imply physically correct calibration, and repeatability does not
imply absolute accuracy. Loop closure can reduce or increase global
error depending on the consistency of the pre-loop trajectory and
calibration supplied to the global-correction module.

The main contributions of this work are summarized as follows:
1. We formulate a controlled same-backend evaluation framework
    that explicitly separates the system-level benefit of
    visual--inertial fusion from the incremental value of learned
    information. The primary comparison holds sensor inputs,
    timestamps, initialization, frontend, backend, calibration
    convention, and output gauge fixed.
2. We establish a unified evidence framework that distinguishes
    local motion consistency, global trajectory accuracy, physical-state
    correctness, and numerical or implementation behavior. The
    evaluation jointly considers RPE, APE, declared
    $\mathrm{SE}(3)$/$\mathrm{Sim}(3)$ alignment, common-frame
    calibration errors, scale, backend cost, repeatability, and global
    correction.
3. We systematically characterize the interaction among the
    learned motion measurement, fixed versus online camera--IMU
    extrinsics, calibration covariance, initialization, perturbation,
    and sequence conditions. The analysis identifies when online
    calibration behaves as a local refinement and when additional state
    freedom is associated with initialization-dependent global drift.
4. We provide a cross-experiment analysis of backend state
    coupling, scale consistency, late calibration correction, and loop
    closure. The resulting evidence clarifies why local accuracy,
    numerical convergence, repeatability, and physical correctness must
    be reported as distinct properties rather than inferred from a
    single trajectory or objective value.


\begin{figure*}[tbp]
\begin{center}
\includegraphics[width=17cm, height=10cm]{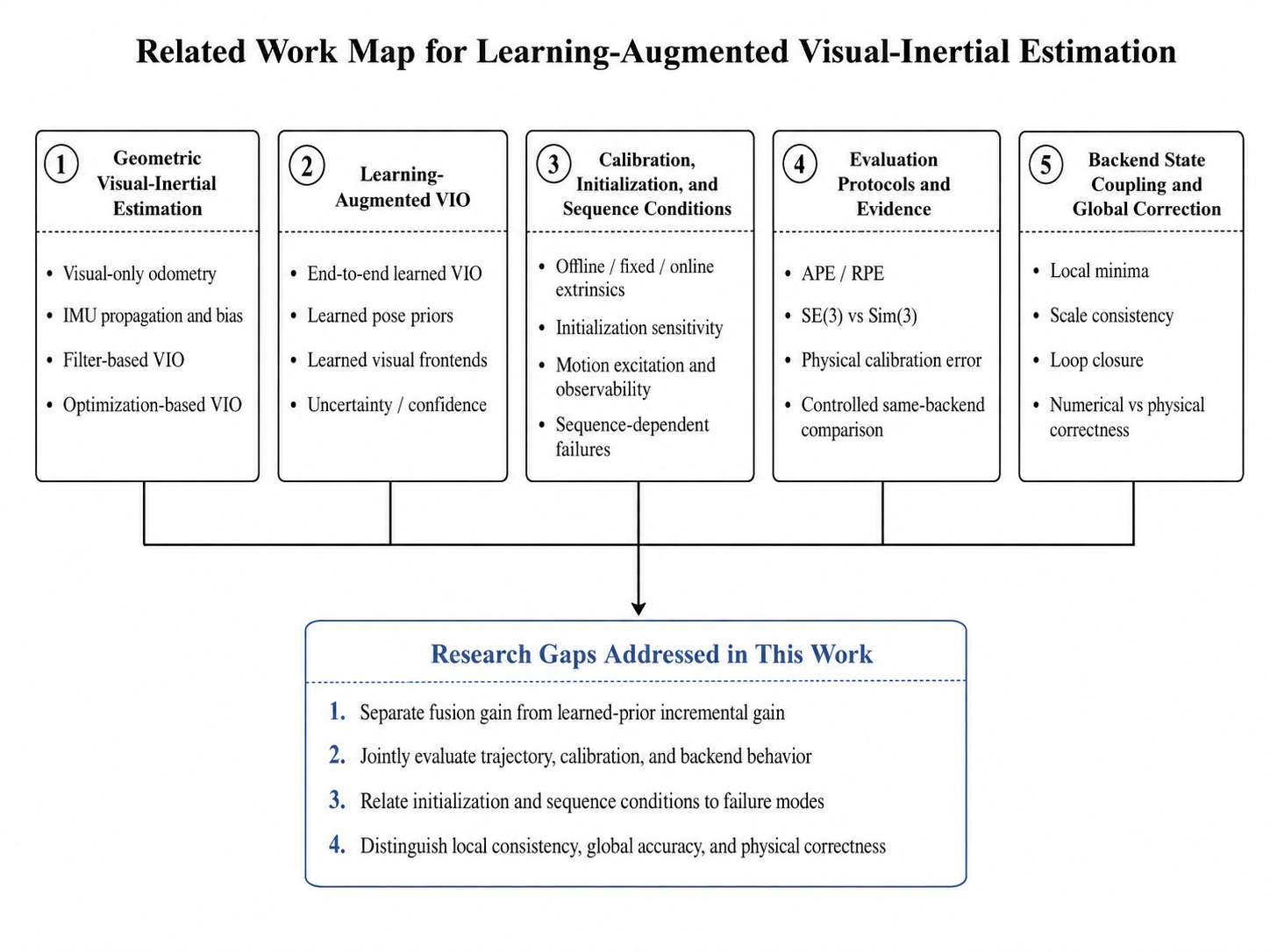}
\end{center}
\vspace{-4 mm}
\caption{Organization of the related-work landscape considered
in this paper. Prior studies are grouped into five interacting
dimensions: geometric visual--inertial estimation,
learning-augmented estimation, calibration and initialization
under sequence conditions, evaluation protocols and evidence,
and backend state coupling with global correction. Their
intersection motivates the controlled evaluation questions and
research gaps examined in this work.}
\label{related}
\vspace{-3mm}
\end{figure*}

\section{Related Work and Evaluation Framework}
\label{sec:related_work_framework}

Visual--inertial state estimation spans tightly coupled geometric filters,
nonlinear optimization backends, and, more recently, systems that inject
learned motion, depth, bias, or uncertainty into an otherwise geometric
estimator. Comparisons across these families are difficult to interpret from
final trajectories alone because the methods may differ simultaneously in
scale gauge, coordinate convention, calibration freedom, initialization, and
global correction. We therefore organize the literature around two linked
questions: what information each method adds to the estimator, and what
controlled evidence is required to attribute a performance change to a learned
relative-pose prior rather than to a different backend or evaluation protocol.
Figure~\ref{related} summarizes the literature
organization adopted in this section. Rather than arranging prior
work as a chronological list, we group it according to five
interacting dimensions that affect the interpretation of
learning-augmented visual--inertial systems: geometric estimation,
, calibration and initialization, evaluation
evidence, and backend state coupling with global correction.
This organization emphasizes that an observed system-level gain
cannot be attributed to a learned prior without considering where
the  enters the estimator, which states remain
free to compensate for it, and which evaluation protocol is used. 
The following subsections develop these dimensions in dependency
order. We first review the complementary error sources of
geometric visual and inertial estimation, and then examine the
interfaces through which  is introduced.
We subsequently discuss calibration, initialization, and sequence
conditions, followed by evaluation validity and backend state
coupling. The final subsection combines these strands to identify
the evaluation gaps addressed by the experimental study.
\subsection{Geometric Visual--Inertial Estimation and Component Error Sources}
\label{subsec:rw_geometric_vio}

Monocular vision and inertial sensing provide complementary but differently
conditioned information. Monocular visual odometry constrains projective motion
but does not recover metric translation scale from images alone, whereas an IMU
measures angular velocity and specific force subject to bias, gravity, and
integration error. Metric scale, gravity direction, velocity, and inertial
biases become recoverable only from sufficiently informative motion and a
consistent visual--inertial model
\cite{Huang2019Review,Forster2017Preintegration,Hesch2014Observability}.
Accordingly, a smooth relative visual trajectory, stable inertial propagation,
or a numerically converged initialization is not, by itself, evidence of an
accurate global trajectory. Errors in scale, gravity, bias, timing, or the
initial state can be redistributed across the joint state and later appear as
drift.
Filtering-based VIO maintains a bounded state and introduces visual information
through measurement updates. The multi-state constraint Kalman filter (MSCKF),
for example, removes feature positions while retaining multi-view constraints
across cloned camera poses \cite{Mourikis2007MSCKF}; subsequent systems improve
linearization consistency, photometric use, calibration support, and research
reproducibility
\cite{Li2013ConsistentVIO,Bloesch2017ROVIO,Geneva2020OpenVINS}. The resulting
estimators offer predictable online computation, but their behavior remains
sensitive to covariance design, linearization points, feature-track quality,
and consistency with the observable subspace. Repeatable filtering behavior
therefore need not imply an unbiased scale, trajectory, or camera--IMU
transformation.

Optimization-based systems instead minimize a nonlinear objective over a
sliding window or a larger graph. OKVIS combines reprojection and inertial
residuals in a keyframe window \cite{Leutenegger2015OKVIS}; VINS-Mono couples
initialization and failure recovery with local estimation and pose-graph
correction \cite{Qin2018VINSMono}; and VI-DSO uses photometric and inertial
terms together with dynamic marginalization to accommodate initially uncertain
scale \cite{vonStumberg2018VIDSO}. ORB-SLAM3 and HybVIO further illustrate how
local visual--inertial estimation can be connected to map reuse or longer-term
SLAM modules \cite{Campos2021ORBSLAM3,Seiskari2022HybVIO}. Comparisons of the
filtering and optimization traditions nevertheless show that estimator class
alone does not determine robustness; front-end measurements, state
parameterization, calibration, initialization, and computational constraints
remain decisive \cite{Gui2015Review,Servieres2021Benchmarking}.

Recent geometric systems refine particular parts of this estimation graph
rather than replacing its role. ICE-BA retains more visual--inertial
information in an incremental consistent solver \cite{Liu2018ICEBA}; StructVIO
adds structural regularity in man-made environments \cite{Zou2019StructVIO};
and DM-VIO delays marginalization to incorporate later estimates more
consistently \cite{vonStumberg2022DMVIO}. Maplab and maplab 2.0 expose modular
local estimation, mapping, and multi-modal localization
\cite{Schneider2018Maplab,Cramariuc2023Maplab2}, while GVINS and OKVIS2-X show
how GNSS, dense depth, or LiDAR can be inserted as additional measurements
without changing the core geometric function of the visual--inertial backend
\cite{Cao2022GVINS,Boche2025OKVIS2X}. For the present study, these systems are
important methodological controls because they make explicit which
measurement changes the information graph and which state variables remain
free to respond.

This literature motivates a component-compatibility view of VIO. A visual pose
source, inertial propagator, and fusion backend may each be individually
plausible yet mutually inconsistent if they use different pose directions,
coordinate frames, timestamp intervals, scale conventions, or noise models.
Component quality should therefore be characterized before fusion, and a
fusion claim should identify the added information, its injection point, and
the other states that are allowed to compensate for disagreement.

\subsection{Learning-Augmented Visual--Inertial Estimation and Learned Pose Priors}
\label{subsec:rw_learning_augmented_vio}

Learning enters odometry and VIO through several distinct interfaces. At one
extreme, end-to-end models replace much of the geometric pipeline with sequence
regression. DeepVO predicts camera motion directly from image sequences
\cite{Wang2017DeepVO}, whereas VINet learns a joint visual--inertial sequence
model \cite{Clark2017VINet}. Selective sensor fusion introduces deterministic
and stochastic masks to gate visual and inertial features under missing,
corrupted, or poorly synchronized inputs \cite{Chen2019SelectiveFusion}.
Later systems add geometric supervision, unsupervised objectives, depth
reconstruction, or longer-term memory
\cite{Han2019DeepVIO,Wei2020UnVIO,Almalioglu2022SelfVIO,Tu2022EMAVIO}, and
TartanVO explicitly addresses cross-dataset generalization and camera-intrinsic
variation \cite{Wang2021TartanVO}. Such results establish the capability of a
complete learned system, but the architecture, representation, fusion rule,
and training distribution often change together. A system-level gain therefore
does not isolate the incremental value of adding one learned pose measurement
to an otherwise unchanged VINS backend.

A second family retains the geometric estimator and learns a specific state,
measurement, or uncertainty model. D3VO predicts depth, relative pose, and
photometric uncertainty for a direct visual odometry backend
\cite{Yang2020D3VO}. CodeVIO represents learned dense depth in a compact,
optimizable state coupled to a filtering-based VIO \cite{Zuo2021CodeVIO}, while
learned IMU-bias inference replaces neither the visual front end nor the factor
graph, but inserts a learned bias-evolution model
\cite{Buchanan2023DeepBias}. Learned depth estimation itself spans additional supervision and uncertainty designs that are relevant when depth is used as an auxiliary geometric cue. Defocus-based self-supervision can couple camera optics with 3-D Gaussian Splatting to provide additional supervision for monocular depth prediction \cite{zhang2025depth}, while recent autoregressive diffusion models formulate depth prediction as coarse-to-fine continuous generation and can additionally expose pixel-wise reliability estimates \cite{zhang2026fractal}. These examples show why
``learning-augmented VIO'' should be classified by both the learned quantity
and its injection point. The output may be pose, depth, bias, covariance,
residual weight, or a gate, and it may enter initialization, the front end, a
filter update, or the nonlinear objective. This interface determines which
backend variables can absorb disagreement with the .

Learned visual odometry and differentiable geometric estimation provide an
adjacent source of pose information. Memory-based pose refinement
\cite{Xue2019BeyondTracking}, probabilistic latent geometry
\cite{Czarnowski2020DeepFactors}, differentiable feature-metric bundle
adjustment \cite{Tang2019BANet}, flow-residual inference
\cite{Min2020VOLDOR}, and learned-versus-geometric decompositions
\cite{Zhan2020VORevisited} all combine data-driven predictions with explicit
geometry. DROID-SLAM and DPVO further embed recurrent correspondence updates in
differentiable bundle adjustment
\cite{Teed2021DROIDSLAM,Teed2023DPVO}. Dynamic-scene refinement, long-term
point tracking, covariance prediction, calibration-tolerant odometry, and
modern 3-D reconstruction priors broaden this design space
\cite{Shen2023DytanVO,Chen2024LEAPVO,Qiu2025MACVO,Lai2025ZeroVO,
Wang2024DUSt3R,Leroy2024MASt3R,Murai2025MASt3RSLAM}. Collectively, these works
show that learned motion and uncertainty can support geometric estimation, but
their system-level comparisons do not directly determine whether the same
prediction is compatible with, or beneficial to, the particular VINS backend
studied here.

Learned local features and matchers operate at another interface by changing
which visual observations reach the estimator. Sparse detectors and
descriptors such as SuperPoint, D2-Net, R2D2, and DISK
\cite{DeTone2018SuperPoint,Dusmanu2019D2Net,Revaud2019R2D2,
Tyszkiewicz2020DISK}, graph or transformer matchers such as SuperGlue, LoFTR,
and LightGlue \cite{Sarlin2020SuperGlue,Sun2021LoFTR,
Lindenberger2023LightGlue}, and lightweight or dense matchers such as XFeat and
RoMa \cite{Potje2024XFeat,Edstedt2024RoMa} can improve correspondence coverage
under appearance change. 
Learning can also operate directly on the sensor-level image representation:
keypoint detection and description on raw Bayer measurements bypasses the
conventional ISP-to-RGB stage, providing another example of a learned front end
whose use would change the visual observations supplied to a downstream
estimator \cite{lin2025keypoint}.

They are relevant here primarily as a boundary of the
controlled comparison: changing the tracks changes the visual baseline itself,
whereas an experiment intended to isolate a relative-pose factor must hold the
front-end measurements fixed.

Learned inertial odometry similarly clarifies what constitutes an inertial
prior. IONet and RIDI infer displacement or velocity constraints from IMU
windows \cite{Chen2018IONet,Yan2018RIDI}; RoNIN and TLIO emphasize
in-the-wild generalization and covariance-aware filter fusion
\cite{Herath2020RoNIN,Liu2020TLIO}; and AI-IMU and learned gyroscope denoising
embed data-driven corrections in physical integration models
\cite{Brossard2020AIIMU,Brossard2020GyroDenoising}. CTIN, EqNIO, Tartan IMU,
and AirIO introduce contextual attention, symmetry-aware priors, large-scale
pretraining, or enhanced feature observability
\cite{Rao2022CTIN,Jayanth2025EqNIO,Zhao2025TartanIMU,Qiu2025AirIO}. The
covariance-aware interfaces in this literature reinforce a central requirement
for learned pose fusion: the backend must be able to interpret not only a mean
motion, but also its scale, reference frame, and uncertainty.

Two additional lines of work expose limitations that matter for such a prior. CoVIO adapts a learned visual--inertial model online to mitigate domain shift \cite{Vodisch2023CoVIO}, highlighting that a fixed network need not preserve the same error distribution across environments. Learned depth can shorten monocular visual--inertial initialization \cite{Merrill2025DepthInit}. Beyond direct estimator integration, camera-model-based self-supervision can derive monocular depth priors from physical camera properties \cite{zhang2024embodiment}, while Vision-Language Embodiment combines camera-derived geometric cues with RGB features and language priors for monocular depth estimation \cite{Zhang2025Vision}.
Depth and learned representations can also support multimodal perception under changed sensing conditions. Language-Depth Navigated Thermal and Visible Image Fusion uses depth and language guidance to structure visible--infrared image fusion \cite{zhang2025language}, while DeepTIO uses cross-modal hallucination to support thermal--inertial estimation \cite{Saputra2020DeepTIO}. These approaches alter scene or modality representations rather than directly providing the relative-pose factor studied here. We therefore treat them as auxiliary learning interfaces, not as direct evidence that a learned pose residual is compatible with a metric VINS state.

For a learned relative-pose prior, compatibility should be stated along four
axes. First, the output may provide full $\mathrm{SE}(3)$ motion, rotation only,
translation direction, or translation with a learned scale. Second, its
transformation direction and reference frame must match the backend convention,
for example $\mathbf{T}_{cw}$ versus $\mathbf{T}_{wc}$ and camera versus IMU
coordinates. Third, the prediction interval must correspond to the same image
timestamps used by IMU preintegration. Fourth, the residual requires a
calibrated covariance or a defensible weighting or gating rule. Otherwise, a
small residual may simply be underweighted, whereas a large fixed weight may
force model bias into scale, velocity, inertial bias, or extrinsic states.

These distinctions lead to two different evaluation questions. The first asks
whether a complete fused estimator outperforms pose-only or IMU-only inputs.
The second asks whether adding the learned prior improves the \emph{same} VINS
backend under identical images, inertial data, initialization, calibration,
timing, and output conventions. Only the latter comparison identifies the
prior's incremental contribution. Weight or uncertainty studies can diagnose
one possible failure mode, but they do not replace the same-backend control.

\subsection{Camera--IMU Calibration, Initialization, and Sequence Observability}
\label{subsec:rw_calibration_observability}

Camera--IMU calibration may be treated as an offline estimate, a fixed trusted
quantity, a strongly regularized state, or a freely updated online variable.
Kalibr formulates joint spatial and temporal multi-sensor calibration in
continuous time \cite{Furgale2013Kalibr}, while earlier filter-based work shows
how camera--IMU extrinsics can be jointly estimated with navigation states and
analyzes the corresponding observability conditions
\cite{Mirzaei2008Calibration}. Fixing a reliable offline transform removes
calibration freedom from the estimator. Online calibration can instead adapt to
an unknown or changing rig, but it enlarges the state and couples calibration
more strongly to motion, visual geometry, bias, and timing.

Spatial and temporal calibration are also interdependent. Online temporal
calibration changes which IMU interval corresponds to each image and therefore
changes the inferred motion \cite{Li2014TemporalCalibration}.
Optimization-based models with a varying camera--IMU offset further expose the
interaction among synchronization, rolling shutter, and preintegration
\cite{Ling2018TimeOffset}. Robocentric estimators can jointly update spatial and
temporal calibration \cite{Huai2022SquareRoot}, and continuous-time
rolling-shutter formulations associate image rows with a continuously varying
camera trajectory \cite{Lang2022CtrlVIO}. Targetless calibration systems extend
this spatiotemporal treatment to broader multi-sensor rigs
\cite{Chen2025iKalibr}. These developments expand the calibration toolkit, but
they also make a fixed-extrinsic reference condition especially important when
the scientific question concerns a learned pose prior rather than calibration
itself.

Initialization determines the basin in which subsequent filtering or nonlinear
optimization operates. Robust visual--inertial initialization jointly estimates
scale, gravity, velocity, and inertial biases while rejecting poorly
conditioned solutions \cite{Campos2019Initialization}; VINS-Mono couples such
initialization with failure recovery before tightly coupled optimization
\cite{Qin2018VINSMono}. Learned single-view depth can reduce the required
observation window \cite{Merrill2025DepthInit}, but it also introduces scale and
domain uncertainty. Initialization should therefore be evaluated both at the
moment it is produced and after the backend has refined the coupled state.

Observability and numerical convergence answer different questions. Under
standard conditions, camera--IMU localization retains gauge freedoms in global
position and yaw, and additional calibration variables require sufficiently
exciting motion to become observable \cite{Hesch2014Observability}. A more
complete self-calibration analysis identifies motion-dependent degeneracies of
camera and IMU parameters and emphasizes the need for fully excited motion
\cite{Yang2023SelfCalibration}. Nevertheless, priors, noise assumptions, and
linearization can produce a stable finite-data estimate even when the available
motion weakly constrains a state. A stable value or low objective is therefore
evidence of numerical convergence, not by itself proof of theoretical
observability or physical calibration correctness.

The calibration strategy must consequently be treated as an experimental
variable. A fixed accurate transform provides the cleanest condition for
isolating visual, inertial, and learned-prior behavior. Online updating should
be assessed using initial and final common-frame $\mathrm{SO}(3)$ and
translation errors, convergence histories, and trajectory metrics. Perturbation
experiments should change the initial calibration while preserving data and
solver settings, and cross-sequence comparisons should determine whether
apparent convergence depends on sequence excitation rather than being attributed
to the calibration algorithm alone.

\subsection{Evaluation Protocols, Controlled Baselines, and Evidence Validity}
\label{subsec:rw_evaluation_protocols}

Trajectory evaluation requires an explicit gauge and alignment policy. ATE or
APE summarizes global trajectory disagreement after alignment, whereas RPE
measures motion error over a stated temporal or spatial interval
\cite{Sturm2012RGBDBenchmark,Zhang2018TrajectoryEvaluation}. Rigid
$\mathrm{SE}(3)$ alignment preserves metric scale; similarity
$\mathrm{Sim}(3)$ alignment additionally estimates one global scale and is
appropriate only when scale is unobservable or intentionally factored out.
Umeyama's least-squares solution provides a standard estimate of rotation,
translation, and optional scale \cite{Umeyama1991Alignment}. The alignment must
therefore match both the sensor configuration and the claim: a
$\mathrm{Sim}(3)$-aligned trajectory cannot establish metric-scale recovery.
Coordinate and temporal conventions are equally consequential. Estimated and
reference poses must share a frame and transformation direction, timestamps
must be associated before spatial alignment, and rotation error should not
depend on Euler-angle wrapping. For two rotations $\mathbf{R}_1$ and
$\mathbf{R}_2$, the geodesic error
\begin{equation}
 e_R=\cos^{-1}\!\left(\frac{\operatorname{tr}(\mathbf{R}_1^{\mathsf T}
 \mathbf{R}_2)-1}{2}\right)
 \label{eq:rw_so3_geodesic}
\end{equation}
provides a common-frame $\mathrm{SO}(3)$ measure. Translation error should be
reported in meters and, where diagnostically useful, by component in a declared
frame. Scale, bias, gravity, temporal-offset, and extrinsic errors are
physical-state metrics and should not be inferred from trajectory overlap
alone.

Public datasets supply complementary stress factors rather than a universal
ranking. KITTI emphasizes long road trajectories with calibrated vehicle
sensors \cite{Geiger2013KITTI}; EuRoC provides synchronized camera--IMU data
with aggressive motion and varying sequence difficulty \cite{Burri2016EuRoC};
and TUM VI adds longer visual--inertial sequences with room-scale ground truth
\cite{Schubert2018TUMVI}. ADVIO captures authentic handheld pedestrian motion
\cite{Cortes2018ADVIO}, whereas TartanAir emphasizes diverse adverse synthetic
conditions \cite{Wang2020TartanAir}. OpenLORIS, 4Seasons, Hilti, and Newer
College broaden the stress factors to lifelong appearance change, seasonal
weather, construction environments, and multi-modal handheld motion
\cite{Shi2020OpenLORIS,Wenzel20214Seasons,Helmberger2022Hilti,
Ramezani2020NewerCollege}. City-scale egocentric evaluation further exposes
long-duration motion, dynamic content, and time-varying sensor behavior that are
less represented in room-scale datasets \cite{Krishnan2025CityScale}.
Benchmark comparisons also show that rankings can vary with platform,
implementation, runtime mode, and sequence
\cite{Delmerico2018Benchmark,Servieres2021Benchmarking}. Repeated runs, failure
counts, runtime configuration, and software mode should therefore accompany
APE and RPE whenever stochastic, asynchronous, or implementation-specific
effects may influence the result.

We separate evaluation evidence into four layers. The \emph{local-motion} layer
uses translational and rotational RPE. The \emph{global-trajectory} layer uses
APE or ATE under a declared $\mathrm{SE}(3)$ or $\mathrm{Sim}(3)$ alignment.
The \emph{physical-state} layer measures extrinsics, scale, bias, gravity, and
temporal offset against an explicit reference. The \emph{numerical and
implementation} layer reports objective values, convergence behavior,
repeatability, runtime, and failure. Agreement at one layer does not imply
agreement at another; the minimum controlled evidence used in this work is
summarized in Table~\ref{tab:claim_minimum_evidence}.

\begin{table*}[t]
\centering
\caption{Claim--minimum-evidence requirements used in this work. Each row
states the smallest controlled comparison that can support the corresponding
claim; additional stress tests strengthen but do not replace the control.}
\label{tab:claim_minimum_evidence}
\footnotesize
\setlength{\tabcolsep}{4pt}
\begin{tabular}{p{0.22\textwidth}p{0.37\textwidth}p{0.34\textwidth}}
\toprule
Claim & Minimum controlled evidence & Required reporting \\
\midrule
Incremental value of the learned prior in the tested settings & Same backend,
images, IMU, initialization, Fixed extrinsics, and output convention, with and
without the prior & APE/RPE, exact tested weight setting, per-sequence deltas,
and the scope of the available repeatability check \\
Online-calibration sensitivity is characterized & Same data and backend with a
declared initial calibration; compare initial and final states with a physical
reference where the transform convention is verified & Common-frame
$\mathrm{SO}(3)$ and translation error where available, convergence history,
APE/RPE, and nominal-setting limitations \\
Initialization sensitivity is characterized & Representative sequences,
available controlled extrinsic settings, and the recorded alternative start
window & Initial/final calibration or trajectory error, restart outcome, and an
explicit statement that a population success rate is not estimated \\
Accuracy effect of the loop-closure pipeline is characterized & Matched
loop-off/loop-on runs with identical front end, calibration, and pipeline state
& Pre/post-toggle APE and RPE and global correction magnitude; do not infer
accepted-loop behavior without acceptance logs \\
Reported backend cost is a matched diagnostic & Same recording, residual
construction, active measurements, and weights & Total cost together with
APE/RPE and calibration error; no cross-recording or per-residual claim when
residual counts are unavailable \\
\bottomrule
\end{tabular}
\end{table*}

\subsection{Backend State Coupling, Scale Consistency, and Global Correction}
\label{subsec:rw_backend_coupling}

A visual--inertial backend jointly estimates variables with different units,
gauges, and linearization sensitivities. Reprojection or photometric residuals
couple poses and structure; preintegrated inertial residuals couple poses,
velocities, gravity, and biases; and calibration or learned factors introduce
additional cross-state dependencies
\cite{Forster2017Preintegration,Leutenegger2015OKVIS}. The resulting objective
is nonlinear and can admit multiple internally consistent solutions.
Marginalization bounds computation by converting removed states into a prior at
one linearization point, but later changes in scale or calibration can make that
prior inconsistent. Dynamic marginalization in VI-DSO is one response to this
scale--linearization interaction \cite{vonStumberg2018VIDSO}. In this review,
marginalization is therefore treated as a coupling mechanism rather than as a
separate method category.

Scale and pose conventions determine whether information can be transferred
between modules. Monocular SLAM may accumulate scale drift and use
$\mathrm{Sim}(3)$ constraints during loop closure
\cite{Strasdat2010ScaleDrift}, whereas an initialized visual--inertial estimator
is intended to maintain metric scale. A scale-ambiguous learned translation
cannot be inserted into such a backend as though it were a metric measurement
without an explicit scale model. Similarly, reversing a relative transform or
mixing camera- and IMU-frame poses can yield a numerically coherent but
physically incorrect factor. The learned quantity should therefore be reported
before and after coordinate, temporal, and scale conversion.

Global correction acts on the state produced by local estimation. Visual--inertial
SLAM with map reuse and VINS-Mono use loop information to reduce accumulated
drift \cite{MurArtal2017MapReuse,Qin2018VINSMono}, and ORB-SLAM3 extends this
principle to multimap reuse \cite{Campos2021ORBSLAM3}. Kimera separates local
VIO from robust global pose-graph optimization \cite{Rosinol2020Kimera}, while
nonlinear factor recovery transfers local VIO information into a global mapping
problem together with loop constraints \cite{Usenko2020NFR}. These
architectures make clear that loop closure is not an independent
post-processing oracle: its effect depends on data association, the pre-loop
trajectory, scale, gravity alignment, calibration, and the information retained
from local estimation.

Mapping systems also differ in how they construct and validate global
constraints. OV$^2$SLAM maintains a fully online visual map
\cite{Ferrera2021OV2SLAM}; DynaVINS and DynaSLAM II suppress or model dynamic
content before global optimization \cite{Song2022DynaVINS,
Bescos2021DynaSLAMII}; and AirSLAM combines point--line geometry with
illumination-robust place recognition \cite{Xu2025AirSLAM}. Learned
coarse-to-fine localization and Patch-NetVLAD improve retrieval under appearance
change \cite{Sarlin2019HFNet,Hausler2021PatchNetVLAD}, while Kimera-Multi
illustrates the additional importance of robust global consistency when maps
and loop constraints are distributed across agents
\cite{Tian2022KimeraMulti}. These works motivate matched loop-on/loop-off
controls: otherwise, changes in retrieval, outlier rejection, or map reuse can
be incorrectly attributed to a better local pose prior.

Optimization cost should be interpreted with the same conditionality. A low
objective value means that the selected residuals are jointly explained under
their current weights and linearization; it does not verify the coordinate
convention, metric scale, extrinsic calibration, or learned measurement against
a physical reference. Late changes to calibration or a pose prior may also
conflict with marginalized information and with states that have already
adapted. Perturbation, late-injection, state-error, and matched loop-toggle
experiments are therefore needed to distinguish numerical convergence from
physically correct estimation.

\subsection{Research Gaps and Positioning of This Work}
\label{subsec:rw_gaps_positioning}

The cited literature provides mature geometric VIO baselines, end-to-end
learned VIO, hybrid learned states and factors, online self-calibration,
trajectory benchmarks, and global correction. Across these lines of work,
however, four evaluation questions are not consistently separated. First,
comparing a fused system with standalone pose or inertial inputs does not by
itself measure the incremental value of a learned pose prior in an unchanged
VINS backend. Second, calibration work often emphasizes observability or
convergence, whereas learned-odometry work emphasizes trajectory error; initial
calibration, online updating, final physical error, and trajectory consequences
are not consistently reported together. Third, local motion, global
trajectory, physical state, and numerical convergence are often evaluated in
different experiments or interpreted as interchangeable evidence. Fourth,
global correction and backend convergence are not consistently evaluated under
one protocol that also controls prior use, extrinsic strategy, initialization,
and scale convention.

This work is therefore positioned as a controlled evaluation of these
interactions, rather than as a claim that learning, online calibration, or loop
closure is uniformly beneficial or harmful. The learned relative-pose prior is
compared with Original VINS under the same backend and input streams; fixed
accurate and online extrinsic strategies are separated; initialization and
calibration are examined across sequences and perturbations; and RPE, APE,
physical calibration error, scale, backend cost, repeatability, and loop
response are treated as distinct evidence layers. This organization makes
negative and sequence-dependent results informative by locating a limitation in
the learned measurement, its compatibility with the backend, calibration and
initialization conditions, or the global-correction stage.

\section{Evaluation Questions, Systems, and Unified Protocol}
\label{sec:experimental_protocol}

The related-work analysis identified five coupled evaluation questions involving
component error modes, learned priors, calibration, sequence observability, and
backend state coupling. This section turns those questions into a controlled
study design. It defines the evaluated data and systems, specifies the state and
learned-residual conventions, and validates implementation repeatability before
any cross-method result is interpreted. These checks support controlled
attribution by reducing implementation confounds; they do not by themselves
establish physical accuracy.

\subsection{Research Questions and Evaluation Scope}
\label{subsec:research_questions}

The experiments are organized around the five research questions in
Table~\ref{tab:research_questions}. The study is not intended as a single
leaderboard across estimators with different gauges and state spaces. Instead,
the evidence progressively distinguishes local motion consistency, global
trajectory accuracy, physical calibration correctness, and numerical
convergence.

\begin{table*}[t]
\centering
\caption{Research questions and evidence map for the experimental study.}
\label{tab:research_questions}
\scriptsize
\begin{tabular}{p{0.05\textwidth}p{0.40\textwidth}p{0.21\textwidth}p{0.19\textwidth}}
\toprule
RQ & Core question & Primary section & Main evidence \\
\midrule
RQ1 & What information does each component provide, and what are its local and global failure modes? & Component-Level Characterization & MonoViT, IMU-only, EKF, Original VINS \\
RQ2 & Under an unchanged VINS backend and matched inputs, does the learned pose prior provide identifiable incremental value? & Incremental Effect of Learned Pose Priors & Original VINS versus VINS+Pose; Fixed/Online; higher-weight control \\
RQ3 & How do the extrinsic initial value, covariance, and Fixed/Online strategy affect calibration and trajectory estimation? & Calibration and Initialization Robustness & EKF perturbations; Fixed/Online; initialization-to-final states; EuRoC \\
RQ4 & Which sequence and initialization-window conditions lead to success, partial convergence, or failure? & Sequence-Dependent Failure and Initialization Conditions & Recordings~1 and~9; frame-300 restart; qualitative tracks \\
RQ5 & Why can local error, backend cost, scale, and global trajectory criteria yield different judgments? & Backend State Coupling, Scale Consistency, and Global Correction & late injection; cost; BA/scale; loop closure \\
\bottomrule
\end{tabular}
\end{table*}

\subsection{Datasets and Evaluation Sequences}
\label{subsec:datasets_sequences}

KITTI synchronized raw data provide the principal road-driving benchmark. We use
camera~2, the left color camera, recorded at 10~Hz and an original resolution of
$1226\times370$, together with the calibrated GPS/IMU reference. Consecutive
frames retain their original order and geometry for pose and visual--inertial
evaluation. The depth network receives images resized to $640\times192$ with
correspondingly scaled intrinsics; its standard 697-image depth evaluation is
used only for the secondary VINS+Pose+Depth diagnostic and is not treated as a second
main evaluation line.
The primary KITTI evaluation uses synchronized residential driving sequences
that contain forward motion, turns, and speed changes. Pose experiments do not
apply random horizontal flipping, and every image resize is accompanied by the
same intrinsic rescaling used by the estimator. EuRoC complements KITTI with
indoor MAV motion and is used specifically to examine Original VINS calibration
behavior under stronger rotational excitation.
KITTI provides the principal road-driving benchmark, whereas EuRoC is used as
a complementary indoor MAV benchmark for calibration behavior under more
aggressive motion. The sequence-specific roles and controlled settings are
stated at the beginning of each corresponding result subsection.

\subsection{Compared Estimators and Controlled Variables}
\label{subsec:estimators_controls}

The evaluation separates standalone MonoViT, IMU-only propagation, the EKF
baseline, Original VINS, VINS augmented with a learned pose prior
(VINS+Pose), and the secondary VINS+Pose+Depth configuration. For controlled VINS
comparisons, the image and IMU streams, timestamps, initialization window,
front end, and backend settings are held fixed. The changed variables are the
presence and weight of the learned pose prior and whether an accurate
camera--IMU extrinsic is fixed or released for online updating.

Table~\ref{tab:system_nomenclature} defines the method and setting names used in
the remainder of the paper. ``Original'' and ``Modified'' distinguish software
pipelines only where an implementation comparison is unavoidable; they do not
denote an accuracy ranking.

\begin{table}[t]
\centering
\caption{System and configuration nomenclature.}
\label{tab:system_nomenclature}
\scriptsize
\begin{tabular}{lp{0.67\columnwidth}}
\toprule
Name & Definition \\
\midrule
Original VINS & Unmodified VINS backend without the learned pose residual \\
VINS+Pose & Same backend with the adjacent-frame MonoViT relative-pose residual \\
VINS+Pose+Depth & VINS+Pose with the learned-depth diagnostic additionally enabled \\
Fixed (F) & Reference camera--IMU extrinsic excluded from optimization \\
Online (O) & Reference extrinsic used for initialization and then optimized \\
Higher weight (H) & Tested pose-prior scalar weight set to $5\times$ its current value \\
ROS/Standalone & Matched ROS-based and offline non-ROS execution pipelines \\
\bottomrule
\end{tabular}
\end{table}

\subsection{Evaluated-System Formulation and State Configuration}
\label{subsec:evaluated_system_formulation}

We use the convention that $\mathbf{T}_{AB}$ maps coordinates from frame $B$
to frame $A$. At camera time $k$, the VINS state contains IMU position,
orientation, velocity, accelerometer bias, and gyroscope bias,
$\mathbf{x}_k=(\mathbf{p}_{WB,k},\mathbf{R}_{WB,k},\mathbf{v}_{W,k},
\mathbf{b}_{a,k},\mathbf{b}_{g,k})$, together with the active visual landmark
states. With the camera-to-IMU transform $\mathbf{T}_{BC}$, the adjacent-camera
motion predicted by the VINS states is
\begin{equation}
\mathbf{T}^{\mathrm{VINS}}_{C_kC_{k+1}}=
\mathbf{T}_{BC}^{-1}\mathbf{T}_{WB,k}^{-1}
\mathbf{T}_{WB,k+1}\mathbf{T}_{BC}.
\label{eq:vins_relative_camera_pose}
\end{equation}
MonoViT outputs the matched camera-to-camera motion
$\widehat{\mathbf{T}}_{C_kC_{k+1}}$ for the same accepted frame pair. Its
six-dimensional residual is
\begin{equation}
\mathbf{r}^{\mathrm{pose}}_k=
\operatorname{Log}_{\mathrm{SE}(3)}\!\left(
\widehat{\mathbf{T}}_{C_kC_{k+1}}^{-1}
\mathbf{T}^{\mathrm{VINS}}_{C_kC_{k+1}}\right),
\label{eq:learned_pose_residual}
\end{equation}
and contributes
$\lambda_{\mathrm{pose}}(\mathbf{r}^{\mathrm{pose}}_k)^{\top}
\mathbf{\Sigma}_{\mathrm{pose}}^{-1}\mathbf{r}^{\mathrm{pose}}_k$ to the
backend objective. The reported current setting uses one fixed covariance and
scalar weight throughout a sequence. The H control multiplies only this scalar
by five; it is a stress control, not a complete weight sweep.
Under Fixed calibration, $\mathbf{T}_{BC}$ is set to the reference value and is
not an optimization variable. Under Online calibration, its three rotational
and three translational degrees of freedom are released and jointly optimized
with pose, velocity, bias, and landmark states. The VINS+Pose+Depth configuration
adds the evaluated learned-depth constraint but is retained only as secondary
diagnostic evidence.
Camera timestamps define the replay clock. All IMU samples in
$(t_{k-1},t_k]$ are preintegrated to the current accepted camera time, and a
learned pose is inserted only when both endpoints match the corresponding
accepted camera pair. Unmatched learned outputs are omitted rather than
interpolated; after a dropped image, IMU samples continue to be integrated to
the next retained frame. The same accepted timestamp set is used for every
controlled pair.

\subsection{Coordinate, Scale, Temporal, and Metric Conventions}
\label{subsec:coordinate_scale_metric_protocol}

We use \emph{absolute pose error} (APE) throughout. The translation component
is also called absolute trajectory error (ATE) in parts of the odometry
literature; to avoid treating two names as different metrics, all such results
are reported here as translation APE. Relative pose error (RPE) is reserved for
fixed-interval relative transformations.

\begin{figure*}[tbp]
\begin{center}
\includegraphics[width=17cm, height=4cm]{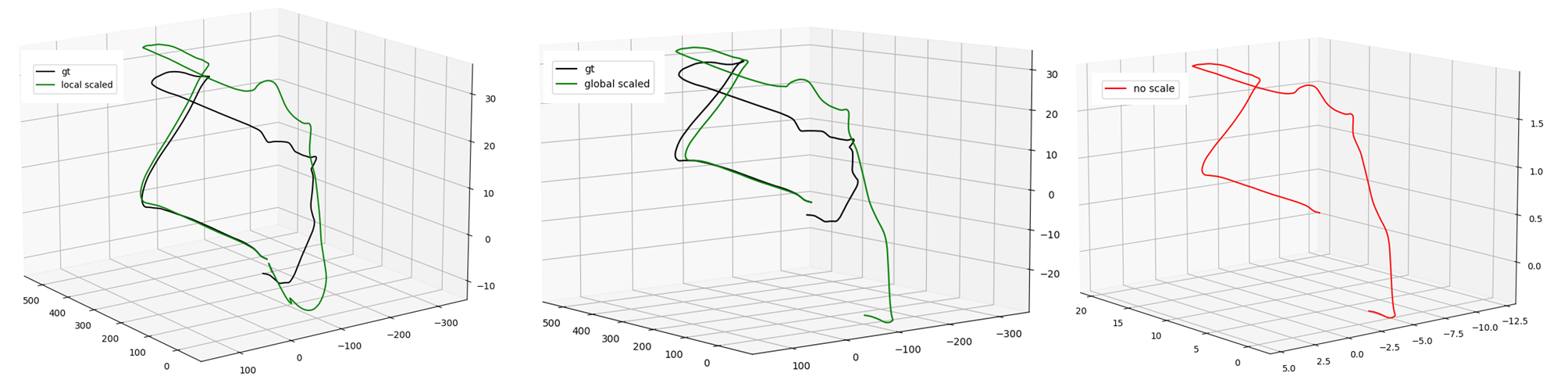}
\end{center}
\vspace{-4 mm}
\caption{Effect of scale alignment on the MonoViT trajectory. The left panel shows the trajectory after local scale correction, the middle panel shows the result obtained using a single global scale factor for the entire sequence, and the right panel shows the raw trajectory without scale correction.}
\label{fig:scale_alignment_comparison}
\vspace{-3mm}
\end{figure*}

Monocular pose estimation recovers translation only up to an unknown scale. Therefore, the raw MonoViT trajectory cannot be directly compared with the metric ground-truth trajectory without first resolving this scale ambiguity. Fig.~\ref{fig:scale_alignment_comparison} illustrates the effect of three evaluation settings: local scaling, global scaling, and no scaling.

Without scale correction, the predicted trajectory preserves the coarse motion pattern but remains in an arbitrary and substantially smaller coordinate scale. This result confirms that the raw translation magnitude produced by MonoViT is not directly metric, even when its relative motion direction is qualitatively reasonable. Consequently, errors computed from the unscaled trajectory would primarily reflect the inherent monocular scale ambiguity rather than the actual quality of the estimated motion.
The locally scaled result aligns more closely with the ground truth over several short trajectory segments. However, local scaling applies different scale corrections over time and can therefore compensate for accumulated scale drift after it has occurred. Although this produces visually improved overlap, it may conceal temporal scale inconsistency and overestimate the accuracy of the monocular estimator. For this reason, locally scaled trajectories are used only as a diagnostic visualization and are not used for the main quantitative evaluation.
In contrast, global scaling applies a single scale factor to the entire predicted trajectory. Given the predicted translations \(\{\mathbf{p}_i\}\) and the corresponding ground-truth translations \(\{\mathbf{g}_i\}\), the global scale is estimated as

\begin{equation}
s^{*}
=
\underset{s}{\arg\min}
\sum_{i}
\left\|
\mathbf{g}_{i}
-
s\mathbf{p}_{i}
\right\|_{2}^{2}.
\label{eq:global_scale_alignment}
\end{equation}

The globally scaled trajectory preserves the original temporal evolution and relative shape of the MonoViT prediction while removing only the sequence-level scale ambiguity. As shown in the middle panel of Fig.~\ref{fig:scale_alignment_comparison}, global scaling improves the overall metric correspondence with the reference trajectory, but residual deviations remain in several regions. These discrepancies represent genuine direction, curvature, and accumulated-drift errors that cannot be removed by a single scale correction.
Accordingly, all subsequent MonoViT comparisons use one global scale factor per sequence. When the estimated and reference trajectories additionally differ in origin or orientation, a single sequence-level similarity transformation, such as \(\mathrm{Sim}(3)\) Umeyama alignment, is applied before computing APE. No frame-wise or locally varying scale correction is used in the reported quantitative results. RPE is evaluated from relative transformations over fixed frame intervals so that local motion consistency is not artificially improved by repeated scale correction.
Overall, the comparison demonstrates that scale alignment has a substantial effect on the apparent trajectory quality. Local scaling provides an optimistic diagnostic view, whereas global scaling offers a more rigorous and reproducible protocol for evaluating monocular pose estimation while preserving its accumulated scale drift.

\subsection{Implementation Consistency and Repeatability}
\label{subsec:implementation_repeatability}

\subsubsection{EKF Platform Consistency}
\begin{figure*}[tbp]
\begin{center}
\includegraphics[width=17cm, height=6cm]{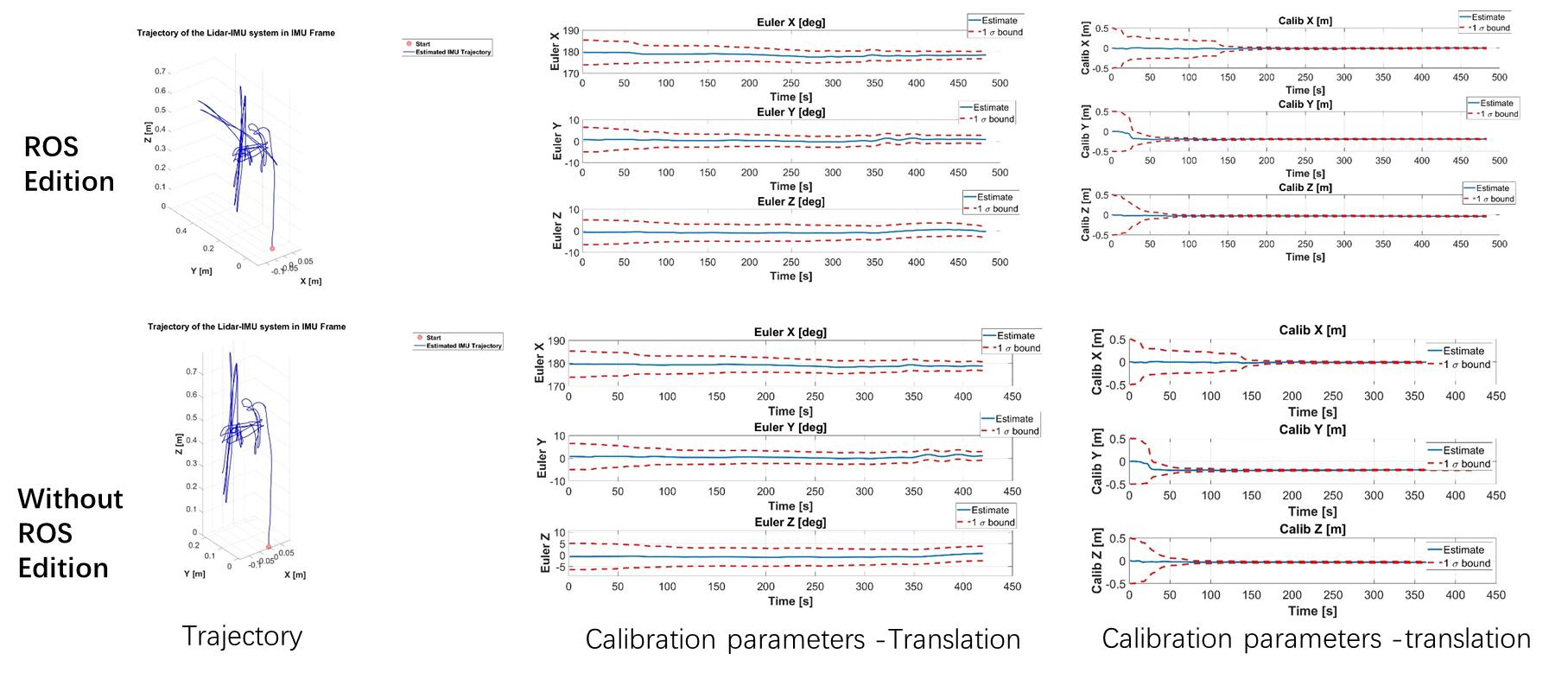}
\end{center}
\vspace{-4 mm}
\caption{Comparison between ROS-based and standalone EKF implementations.The ROS and non-ROS versions are evaluated on the same input sequence. Both implementations produce similar estimated IMU trajectories and camera-IMU calibration parameters, including extrinsic rotation and translation. 
}
\label{ekfros1}
\vspace{-3mm}
\end{figure*}

\begin{figure*}[tbp]
\begin{center}
\includegraphics[width=17cm, height=6cm]{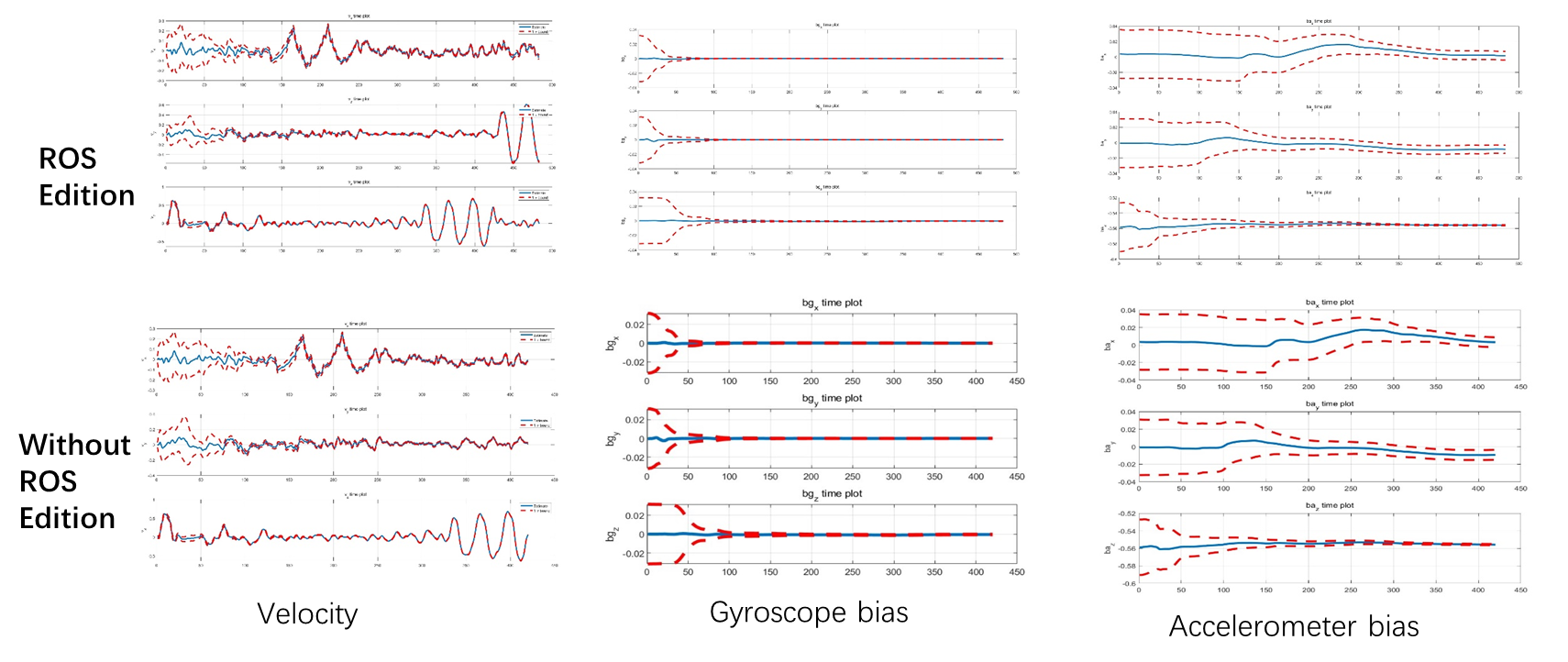}
\end{center}
\vspace{-4 mm}
\caption{Internal state comparison between ROS-based and standalone EKF implementations. The estimated velocity, gyroscope bias, and accelerometer bias show similar trends in the ROS and non-ROS versions.}
\label{ekfros2}
\vspace{-3mm}
\end{figure*}

Before evaluating the interaction between EKF and Original VINS, we first validate whether removing the ROS dependency changes the behavior of the EKF estimator. This step is important because later experiments involve both ROS-based and standalone implementations of Original VINS, and the consistency of the EKF module under different execution environments must be verified. Therefore, we compare the original ROS-based EKF implementation with the standalone non-ROS implementation using the same input sequence and configuration.
As shown in Fig.~\ref{ekfros1}, the ROS and non-ROS versions produce similar IMU trajectories and camera-IMU calibration parameters. The estimated translation parameters are especially close between the two implementations, while the rotation estimates show similar convergence trends. Fig.~\ref{ekfros2} further compares the internal EKF states, including velocity, gyroscope bias, and accelerometer bias. The two implementations exhibit similar temporal behavior, indicating that the offline data-loading and standalone execution pipeline preserve the main propagation and update dynamics of the original ROS-based EKF. This validation supports the use of the non-ROS EKF implementation in the following KITTI and end-to-end evaluation experiments.
Specifically, the ROS-based implementation estimates the Euler angles as 
$[-0.682^\circ, 1.053^\circ, 1.258^\circ]$ and the translation as 
$[0.00084, -0.18809, -0.03330]~\mathrm{m}$, while the non-ROS implementation estimates 
$[-0.505^\circ, 0.925^\circ, 1.766^\circ]$ and 
$[0.00129, -0.18774, -0.03299]~\mathrm{m}$, respectively. 
The translation differences are all below $5 \times 10^{-4}~\mathrm{m}$, indicating that the camera--IMU translation calibration is almost unchanged after removing the ROS dependency. 
Although the yaw angle shows a difference of about $0.51^\circ$, the overall rotation estimates remain within a small range and follow similar convergence behavior.

\subsubsection{VINS Platform and Repeated-Run Consistency}

Before evaluating the integration of EKF and learning-based pose/depth priors with Original VINS, we first examine whether different VINS implementations and initialization settings lead to significantly different calibration results. Specifically, we compare the original ROS-based VINS implementation and the standalone VINS-without-ROS implementation under two settings: with an initial extrinsic guess and without an initial extrinsic guess.
Table~\ref{tab:vins_platform_initialization} reports the final estimated extrinsic parameters under the four settings. When an initial guess is provided, the ROS-based VINS estimates the final rotation as 
\([0.1292^\circ, 0.9421^\circ, 89.1113^\circ]\) 
and translation as 
\([-0.0139, -0.0720, -0.0091]\)~m. 
The VINS-without-ROS implementation estimates a similar result, with rotation 
\([0.1020^\circ, 0.9186^\circ, 89.1057^\circ]\) 
and translation 
\([-0.0132, -0.0694, -0.0085]\)~m. 
The difference between the two platforms is small: the rotation difference is approximately 
\([0.0272^\circ, 0.0235^\circ, 0.0056^\circ]\), 
and the translation difference is approximately 
\([0.0008, 0.0026, 0.0006]\)~m.
A similar trend can be observed when no initial guess is used. The ROS-based VINS estimates the final rotation as 
\([0.1428^\circ, 0.9439^\circ, 89.1085^\circ]\) 
and translation as 
\([-0.0149, -0.0731, -0.0093]\)~m, 
while VINS-without-ROS estimates 
\([0.1147^\circ, 0.9203^\circ, 89.0970^\circ]\) 
for rotation and 
\([-0.0145, -0.0701, -0.0086]\)~m 
for translation. The platform difference remains small, with rotation differences of approximately 
\([0.0282^\circ, 0.0235^\circ, 0.0115^\circ]\) 
and translation differences of approximately 
\([0.0004, 0.0030, 0.0006]\)~m.
We also compare the influence of using an initial guess. For the ROS-based VINS implementation, the difference between the with-guess and without-guess settings is approximately 
\([0.0136^\circ, 0.0018^\circ, 0.0028^\circ]\) 
in rotation and 
\([0.0010, 0.0011, 0.0002]\)~m 
in translation. For the VINS-without-ROS implementation, the corresponding difference is approximately 
\([0.0126^\circ, 0.0017^\circ, 0.0087^\circ]\) 
in rotation and 
\([0.0013, 0.0007, 0.0002]\)~m 
in translation. These differences are relatively small, suggesting that the final VINS calibration result is not strongly affected by the initial guess in this setting.
Overall, this experiment shows that the ROS-based and without-ROS VINS implementations produce consistent final extrinsic estimates, and the effect of the initial guess is limited. This supports the use of both implementations in the following evaluation and indicates that subsequent performance differences are less likely to be caused by platform-specific implementation discrepancies or by the initial extrinsic guess alone.

\begin{table*}[t]
\centering
\resizebox{\linewidth}{!}{
\begin{tabular}{lcccccc}
\toprule
\textbf{Method / Setting} 
& \textbf{Roll ($^\circ$)} 
& \textbf{Pitch ($^\circ$)} 
& \textbf{Yaw ($^\circ$)} 
& \textbf{$X$ (m)} 
& \textbf{$Y$ (m)} 
& \textbf{$Z$ (m)} \\
\midrule
VINS with initial guess 
& 0.12920883 
& 0.94205661 
& 89.11134805 
& -0.01392613 
& -0.07202541 
& -0.00905363 \\

VINS-without-ROS with initial guess 
& 0.10204582 
& 0.91859205 
& 89.10573467 
& -0.01316146 
& -0.06939931 
& -0.00847115 \\

VINS without initial guess 
& 0.14282301 
& 0.94385202 
& 89.10850337 
& -0.01489817 
& -0.07309295 
& -0.00928658 \\

VINS-without-ROS without initial guess 
& 0.11465889 
& 0.92032762 
& 89.09702679 
& -0.01446791 
& -0.07011650 
& -0.00863906 \\
\bottomrule
\end{tabular}
}
\caption{Final extrinsic estimates of VINS under different platforms and initialization settings.}
\label{tab:vins_platform_initialization}
\end{table*}

To further verify the stability of the VINS results, we repeat the experiments under the same initialization settings. Table~\ref{tab:vins_repeatability} reports the first and second runs for both the with-initial-guess and without-initial-guess configurations. Under the initial-guess setting, the two runs produce almost identical final extrinsic estimates. The rotation differences between the two runs are only about \([8.79\times10^{-6^\circ}, 5.51\times10^{-5^\circ}, 9.23\times10^{-6^\circ}]\), and the translation differences are within \(1.6\times10^{-6}~\mathrm{m}\). A similar pattern is observed in the without-initial-guess setting, where the two runs also produce nearly the same final rotation and translation estimates.
We also repeat the odometry estimation using accurate extrinsic parameters. The final odometry poses from the two runs are highly consistent. The rotation differences are approximately \([0.0022^\circ, 0.0001^\circ, 0.0030^\circ]\), and the translation differences are below \(2.2\times10^{-4}~\mathrm{m}\). This indicates that the VINS odometry output is repeatable under the same input data and calibration configuration.
Together with the previous platform and initialization comparison, these repeated-run results show that the VINS evaluation is stable and reproducible. Therefore, the differences observed in later experiments are less likely to be caused by random execution variance, and more likely to reflect the effects of different initialization, extrinsic settings, or integration strategies.

\begin{table*}[t]
\centering
\resizebox{\linewidth}{!}{
\begin{tabular}{llcccccc}
\toprule
\textbf{Evaluation} 
& \textbf{Run / Setting} 
& \textbf{Roll ($^\circ$)} 
& \textbf{Pitch ($^\circ$)} 
& \textbf{Yaw ($^\circ$)} 
& \textbf{$X$ (m)} 
& \textbf{$Y$ (m)} 
& \textbf{$Z$ (m)} \\
\midrule
\multirow{2}{*}{Final extrinsic}
& 1st run w/ init. guess
& 0.12919693
& 0.94206166
& 89.11135918
& $-1.3927124934576006 \times 10^{-2}$
& -0.0720239866276594
& $-9.0520002393690967 \times 10^{-3}$ \\

& 2nd run w/ init. guess
& 0.12918814
& 0.94200657
& 89.11134995
& $-1.3926596686883799 \times 10^{-2}$
& -0.0720255163554694
& $-9.0526417281940174 \times 10^{-3}$ \\
\midrule

\multirow{2}{*}{Final extrinsic}
& 1st run w/o init. guess
& 0.14281396
& 0.94385622
& 89.10850764
& $-1.4897594776339746 \times 10^{-2}$
& -0.0730929617086037
& $-9.2867827649735848 \times 10^{-3}$ \\

& 2nd run w/o init. guess
& 0.14282443
& 0.94385065
& 89.10850354
& $-1.4898386929989293 \times 10^{-2}$
& -0.0730929245318574
& $-9.2870138320267064 \times 10^{-3}$ \\
\midrule

\multirow{2}{*}{Final odometry pose}
& 1st run w/ accurate extrinsic
& -165.09126635
& -71.98699933
& 3.52874976
& -0.51332
& 0.41103
& -0.13349 \\

& 2nd run w/ accurate extrinsic
& -165.08909602
& -71.98689964
& 3.52576963
& -0.51344
& 0.41119
& -0.13327 \\
\bottomrule
\end{tabular}
}
\caption{Repeatability evaluation of VINS final extrinsic calibration and odometry estimation.}
\label{tab:vins_repeatability}
\end{table*}

\subsubsection{Cross-Sequence and Loop-Response Consistency}

\begin{figure*}[tbp]
\begin{center}
\includegraphics[width=17cm, height=8cm]{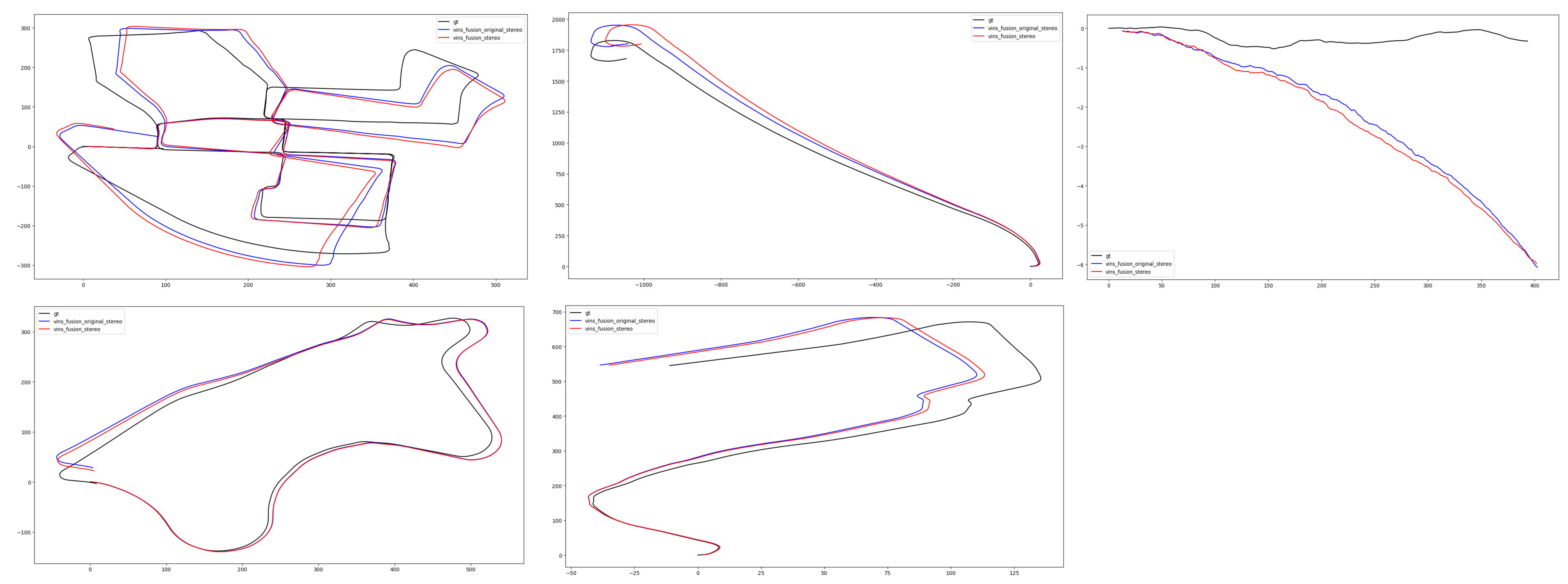}
\end{center}
\vspace{-4 mm}
\caption{Cross-sequence no-loop trajectory consistency between the original ROS-based and standalone non-ROS VINS-Fusion stereo implementations across five representative recordings. The black curves denote ground truth, while the blue and red curves denote the original and standalone implementations.}
\label{fig:ross-sequence trajectory consistency}
\vspace{-3mm}
\end{figure*}

Fig.~\ref{fig:ross-sequence trajectory consistency} compares the trajectories produced by the original ROS-based VINS-Fusion stereo implementation and the standalone non-ROS implementation across five representative recordings. In all panels, the original implementation (blue) and the standalone implementation (red) exhibit highly similar trajectory geometries and remain closely aligned over most of the sequences. This consistency is observed for both complex multi-turn trajectories and long-range motions. In the upper-left panel, the two estimates reproduce nearly identical turning locations, loop shapes, and accumulated deviations from the ground truth. A similar trend appears in the upper-middle and lower panels, where the blue and red trajectories follow almost the same paths despite noticeable offsets from the reference trajectory in some regions. The upper-right panel further shows that the two implementations produce nearly identical component-wise drift, with both estimates gradually diverging from the ground truth in the same direction and at a similar rate. The figure therefore supports qualitative implementation-response consistency after removing ROS. It does not contain matched APE/RPE values, so it is not used to quantify accuracy equivalence or to attribute the remaining errors to a particular source.

\begin{figure*}[tbp]
\begin{center}
\includegraphics[width=17cm, height=14cm]{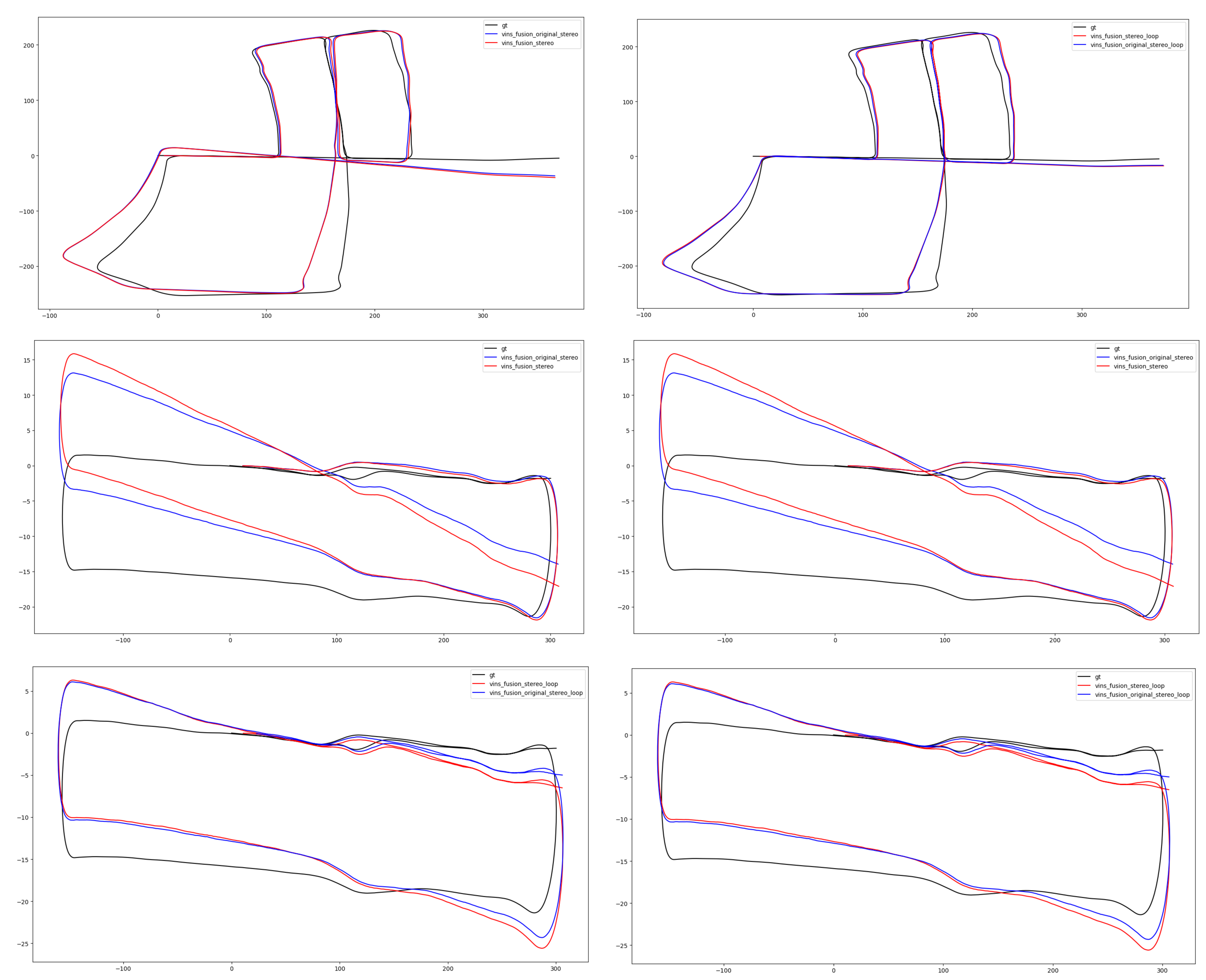}
\end{center}
\vspace{-4 mm}
\caption{Loop-closure response consistency between the original ROS-based and standalone non-ROS VINS-Fusion stereo implementations across three representative recordings. The black curves denote ground truth, and the blue and red curves denote the original and standalone implementations. Each pair shows that the two implementations undergo closely matched global corrections when loop closure is enabled.}
\label{fig:original ROS-based and standalone non-ROS VINS-Fusion stereo}
\vspace{-3mm}
\end{figure*}

Fig.~\ref{fig:original ROS-based and standalone non-ROS VINS-Fusion stereo} evaluates the consistency between the original ROS-based VINS-Fusion implementation and the standalone non-ROS implementation in stereo mode, both before and after loop-closure correction. Across all three recordings, the original implementation (blue) and the standalone implementation (red) produce highly similar trajectories. Their estimated turning locations, large-scale path geometry, accumulated drift, and deviations from the ground truth are closely matched. This consistency is maintained not only in the odometry-only results but also after loop closure is enabled. In particular, although loop closure causes visible global changes in the estimated trajectories for the second and third recordings, the red and blue trajectories undergo almost identical corrections. This indicates that the standalone implementation preserves both the local stereo visual--inertial estimation behavior and the global loop-closure response of the original system.
At the same time, both implementations exhibit similar residual deviations from the ground truth, especially along long trajectory segments and near several turning regions. Therefore, the close agreement between the two implementations is evidence of implementation consistency rather than absolute trajectory accuracy. The common errors are more likely associated with the estimator itself, sensor measurements, calibration, initialization, or sequence characteristics than with the removal of ROS. The controlled APE/RPE comparison in Table~\ref{tab:loop_closure_quantitative} evaluates the accuracy effect of loop closure separately from this repeatability check.

\textbf{Validity takeaway.}
Across the EKF and VINS implementations, removing ROS and repeating identical
runs produce closely matched calibration and trajectory outputs. This validates
the attribution of the subsequent controlled comparisons, but it does not
establish physical accuracy. Reproducibility and accuracy are therefore treated
as separate evaluation properties in the remainder of the paper.

The common protocol and implementation checks now provide the basis for
characterizing the estimators individually before testing whether a learned
pose prior adds value to an unchanged visual--inertial backend.

\section{Component-Level Characterization}
\label{sec:component_characterization}

This section characterizes the information and failure modes supplied by each
standalone estimator. The purpose is not to rank incompatible systems as a
single leaderboard, but to establish which local cues and global errors are
available before fusion.

\subsection{Standalone MonoViT Pose Estimation}
\label{subsec:standalone_monovit}
Following the protocol in Section~\ref{subsec:datasets_sequences}, the standalone
test uses KITTI sequence \texttt{2011\_09\_30\_drive\_0033}: 1591 synchronized
frames (0--1590, approximately 159~s) with the camera trajectory as ground truth.
The experiment characterizes MonoViT translation and rotation errors before its
adjacent-frame output is introduced into VINS; dataset and sensor details are not
repeated here.

\begin{figure*}[tbp]
\begin{center}
\includegraphics[width=17cm, height=6cm]{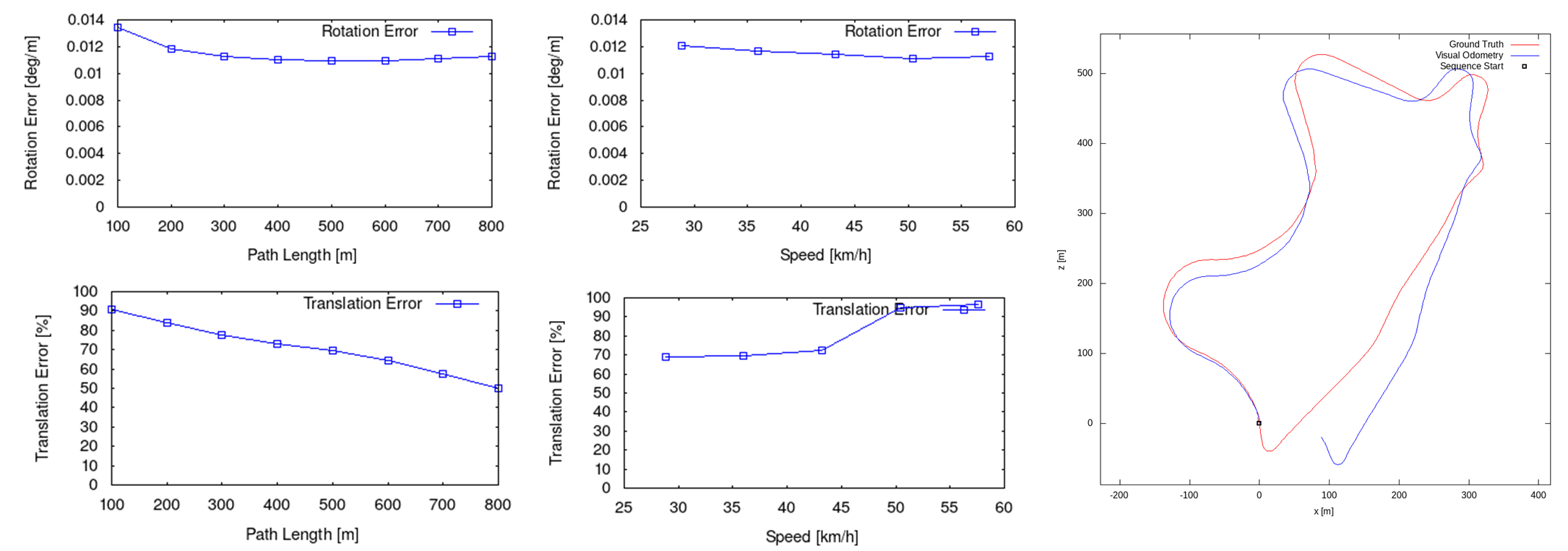}
\end{center}
\vspace{-4 mm}
\caption{Standalone MonoViT pose evaluation on KITTI. The left plots show rotation and translation errors with respect to path length and vehicle speed, while the right plot compares the predicted visual-odometry trajectory with the ground-truth trajectory.}
\label{fig:monovit_path_speed}
\vspace{-3mm}
\end{figure*}

\textbf{ Pose Estimation Results.}
Fig.~\ref{fig:monovit_path_speed} presents the standalone pose estimation performance of MonoViT on the KITTI sequence. The predicted visual odometry trajectory generally follows the overall shape of the ground-truth camera trajectory, indicating that MonoViT is able to recover the coarse motion pattern of the vehicle. However, clear trajectory drift can still be observed, especially in regions with large turning motions and long-term accumulated displacement. This suggests that although the learned pose network captures local frame-to-frame motion, the accumulated trajectory is still affected by scale inconsistency and drift.
The path-length-based evaluation further shows that the rotation error remains relatively stable across different segment lengths. This indicates that MonoViT provides comparatively consistent orientation estimation. In contrast, the translation error is much larger and remains at a high level over different path lengths. Although the translation error decreases as the evaluated path length increases, it still indicates that the monocular pose estimation suffers from significant scale ambiguity. The speed-based evaluation shows a similar trend: the rotation error changes only slightly under different vehicle speeds, while the translation error becomes more unstable as the speed increases. This implies that fast vehicle motion can further amplify the weakness of monocular translation estimation.

\begin{figure}[tbp]
\begin{center}
\includegraphics[width=8.5cm, height=6cm]{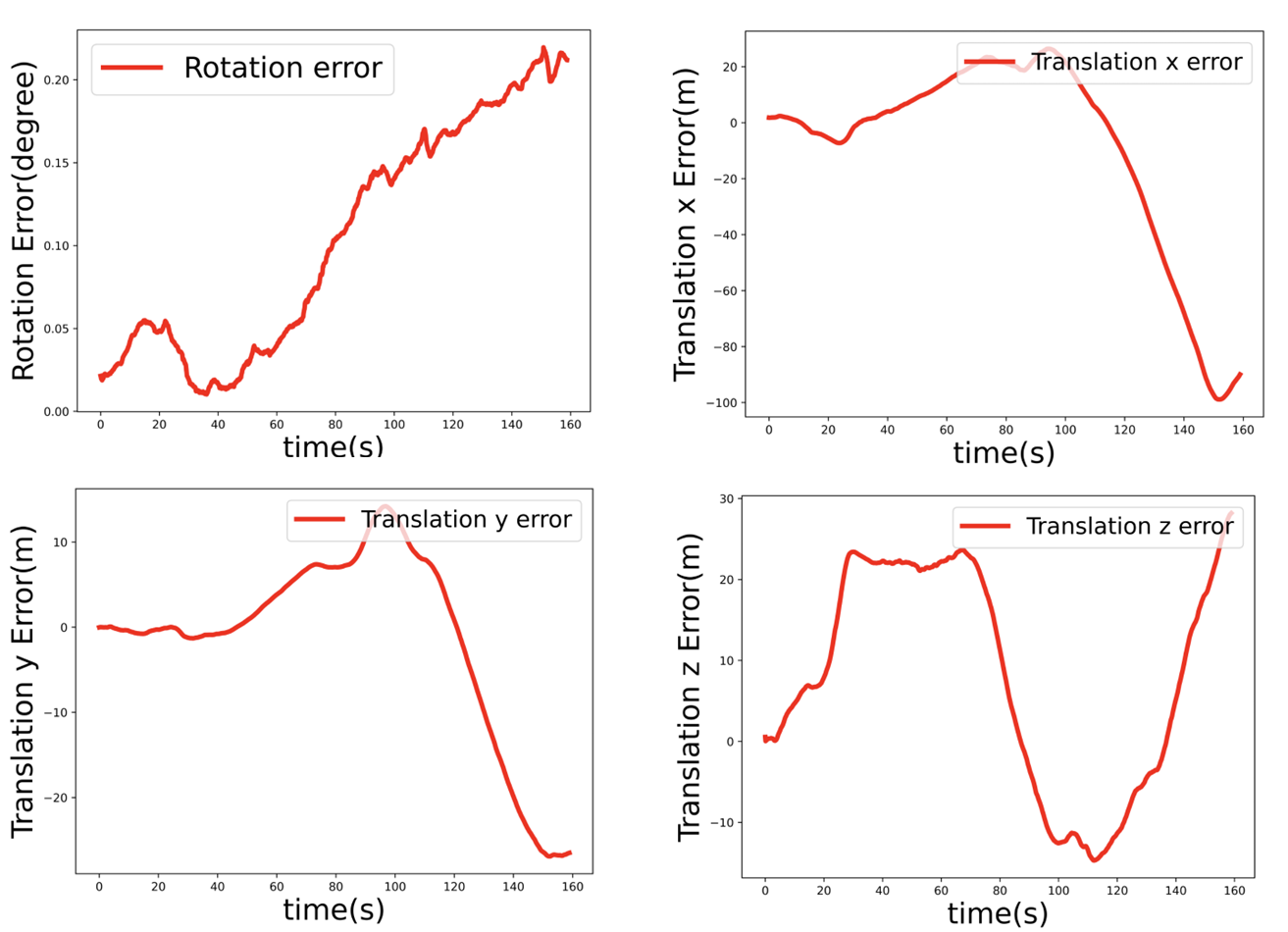}
\end{center}
\vspace{-4 mm}
\caption{Temporal pose error of standalone MonoViT on KITTI. The rotation error and translation errors along the $x$, $y$, and $z$ axes are plotted over time.}
\label{fig:monovit_temporal}
\vspace{-3mm}
\end{figure}

Fig.~\ref{fig:monovit_temporal} shows the temporal evolution of the pose error. The rotation error gradually increases over time, demonstrating a mild accumulation of orientation drift. More importantly, the translation errors along the x, y, and z directions accumulate significantly throughout the sequence. The x-direction error grows to tens of meters near the end of the sequence, while the y- and z-direction errors also exhibit large fluctuations. These results confirm that MonoViT alone is not sufficient to provide a globally accurate long-term trajectory. Instead, its pose output is more suitable to be used as a local visual motion prior rather than as a standalone odometry solution.
Overall, the standalone MonoViT evaluation shows that the learned pose estimator can provide useful relative motion cues, especially for rotation estimation and coarse trajectory recovery. However, the large accumulated translation drift reveals the limitation of purely monocular learned pose prediction. This motivates the following evaluation of how such learned pose information can be further constrained or fused with visual-inertial estimation frameworks.

\begin{figure*}[tbp]
\begin{center}
\includegraphics[width=17cm, height=9cm]{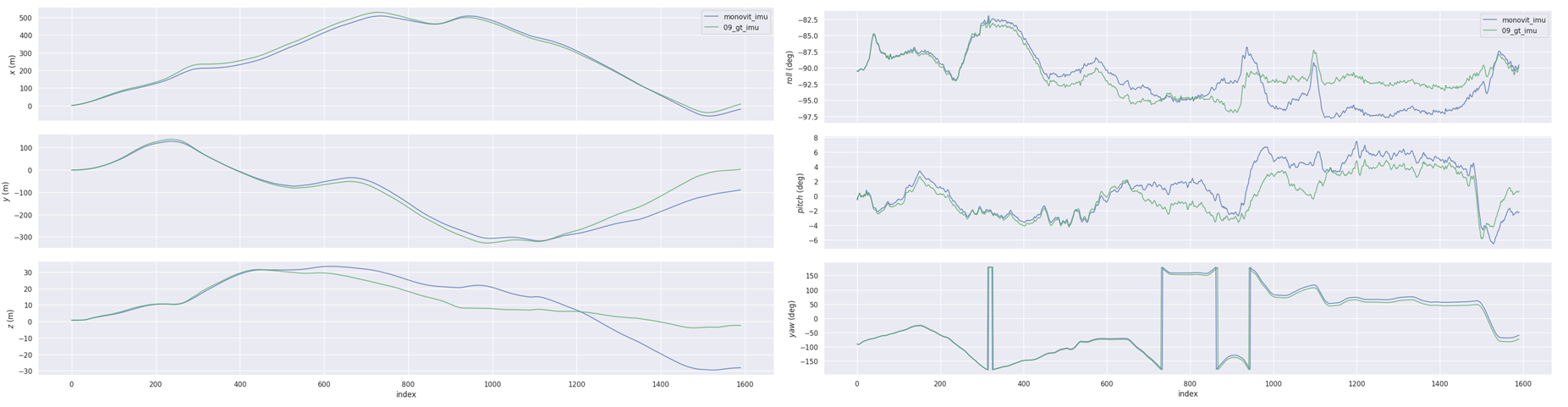}
\end{center}
\vspace{-4 mm}
\caption{Component-wise MonoViT pose estimation on Recording 9. The left column compares the estimated and reference translations along the x-, y-, and z-axes, while the right column compares Roll, Pitch, and Yaw.}
\label{fig:recording9_monovit_pose_components}
\vspace{-3mm}
\end{figure*}

\begin{figure*}[tbp]
\begin{center}
\includegraphics[width=17cm, height=12cm]{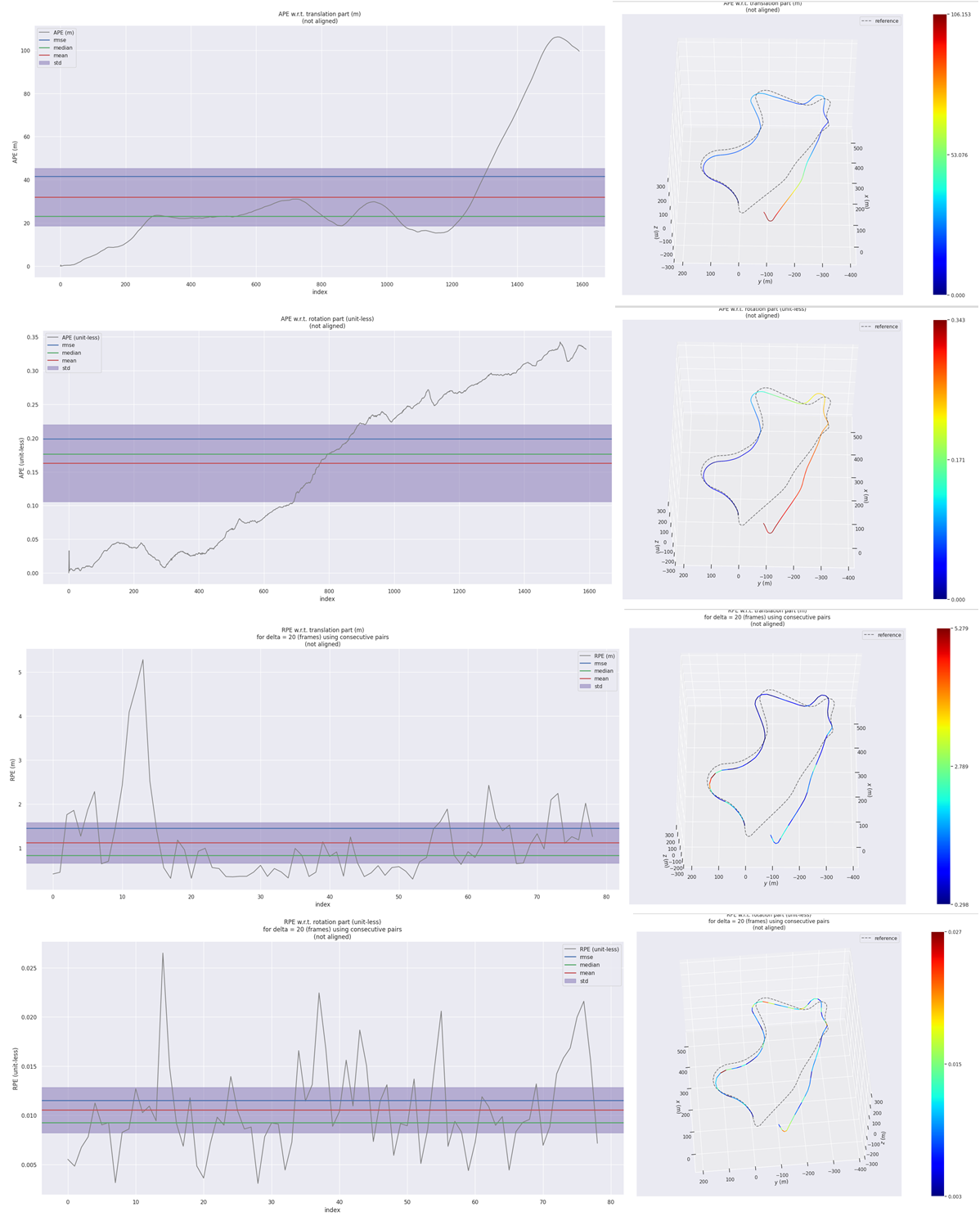}
\end{center}
\vspace{-4 mm}
\caption{Absolute and relative pose errors of MonoViT on Recording 9. The figure reports translation APE, rotation APE, translation RPE, and rotation RPE, together with their trajectory-wise error distributions.}
\label{fig:recording9_monovit_errors}
\vspace{-3mm}
\end{figure*}

Figs.~\ref{fig:recording9_monovit_pose_components} and~\ref{fig:recording9_monovit_errors}, together with the scale-alignment comparison in Fig.~\ref{fig:scale_alignment_comparison}, present the Recording~9 results from three complementary perspectives: component-wise pose comparison, quantitative error evaluation, and sequence-level scale sensitivity.
From the component-wise curves in Fig.~\ref{fig:recording9_monovit_pose_components}, MonoViT is able to recover the overall motion trend reasonably well, especially in the dominant planar-motion components and the yaw angle, where the predicted trajectory broadly follows the reference evolution. The roll and pitch estimates also capture the main low-frequency variations, although noticeable bias and drift remain over time. In contrast, the vertical translation component and part of the lateral motion exhibit larger deviations in the later portion of the sequence, indicating lower numerical stability in those components and greater vulnerability to accumulated scale and integration drift. The yaw curves also contain discontinuous jumps caused by angle wrapping, which should not be misinterpreted as actual catastrophic orientation failure.

The quantitative results in Fig.~\ref{fig:recording9_monovit_errors} further clarify this behavior. The relative pose error (RPE) remains comparatively small for both translation and rotation, showing that MonoViT preserves short-term frame-to-frame motion consistency reasonably well. In particular, the rotational RPE stays at a low level throughout most of the sequence, while the translational RPE also remains moderate except for a few local peaks. However, the absolute pose error (APE) grows progressively over time, with the translational APE increasing substantially toward the end of the sequence and the rotational APE also showing a clear upward trend. This gap between the low RPE and the much larger APE indicates that MonoViT is locally smooth and temporally coherent, but still suffers from long-horizon drift accumulation. In other words, neighboring-frame predictions are relatively reliable, yet these small local errors accumulate into significant global trajectory deviations.

The scale comparison in Fig.~\ref{fig:scale_alignment_comparison} is consistent with these observations. A single sequence-level scale factor removes the monocular scale ambiguity without changing the temporal shape of the prediction, yet visible direction, curvature, and accumulated-drift errors remain. The locally scaled curve is shown only to demonstrate why frame-wise correction can conceal genuine drift and is not used for the reported method ranking.
Overall, these results show that MonoViT provides relatively accurate short-term motion estimation on Recording~9, but its global accuracy remains constrained by scale inconsistency and accumulated drift. This observation motivates the integration of inertial cues or backend optimization in the subsequent experiments.

\subsection{IMU-Only Propagation as a Diagnostic Baseline}
\label{subsec:imu_only_diagnostic}

\begin{figure*}[tbp]
\begin{center}
\includegraphics[width=17cm, height=9cm]{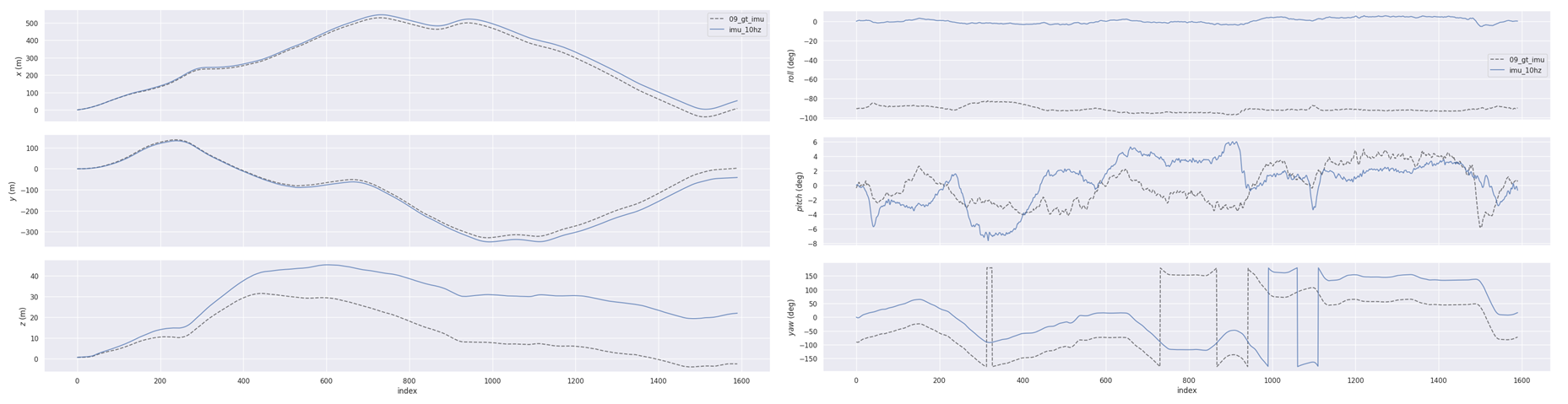}
\end{center}
\vspace{-4 mm}
\caption{Component-wise pose comparison between 10~Hz IMU-only integration and the reference trajectory on Recording~9. The left column shows the x-, y-, and z-translation components, while the right column shows Roll, Pitch, and Yaw.}
\label{fig:imu_10hz_pose_components}
\vspace{-3mm}
\end{figure*}

Fig.~\ref{fig:imu_10hz_pose_components} presents the component-wise pose obtained by integrating the 10~Hz IMU measurements on Recording~9. The translational results show that the low-rate inertial trajectory can reproduce the dominant motion pattern, particularly during the early part of the sequence. The estimated \(x\)- and \(y\)-components initially remain close to the reference, but their differences gradually increase as the sequence progresses. The accumulated error is most evident near the end, where both horizontal components retain noticeable offsets from the ground truth.
The \(z\)-component exhibits substantially stronger drift. Although the estimated and reference heights initially increase in a similar manner, the 10~Hz integration increasingly overestimates the vertical position after approximately frame~300. While the reference height subsequently decreases toward zero, the integrated trajectory maintains a large positive offset. This behavior is expected in standalone inertial integration because a small accelerometer bias, gravity-compensation error, or initial-attitude error is integrated twice and therefore produces rapidly increasing position drift.
The orientation curves reveal an additional issue. The estimated Roll remains close to \(0^\circ\), whereas the reference Roll is centered near approximately \(-90^\circ\). Yaw also contains a large persistent offset, together with several discontinuities caused by angular wrapping. Pitch follows some of the temporal motion variations but still differs substantially from the reference over several intervals. These systematic differences suggest that the integrated and reference orientations may not yet use the same body-frame definition, initial orientation, Euler-angle order, or axis convention. Therefore, the large absolute rotational discrepancy should not be interpreted solely as IMU integration drift; part of it may originate from coordinate-frame inconsistency.

\begin{figure*}[tbp]
\begin{center}
\includegraphics[width=17cm, height=12cm]{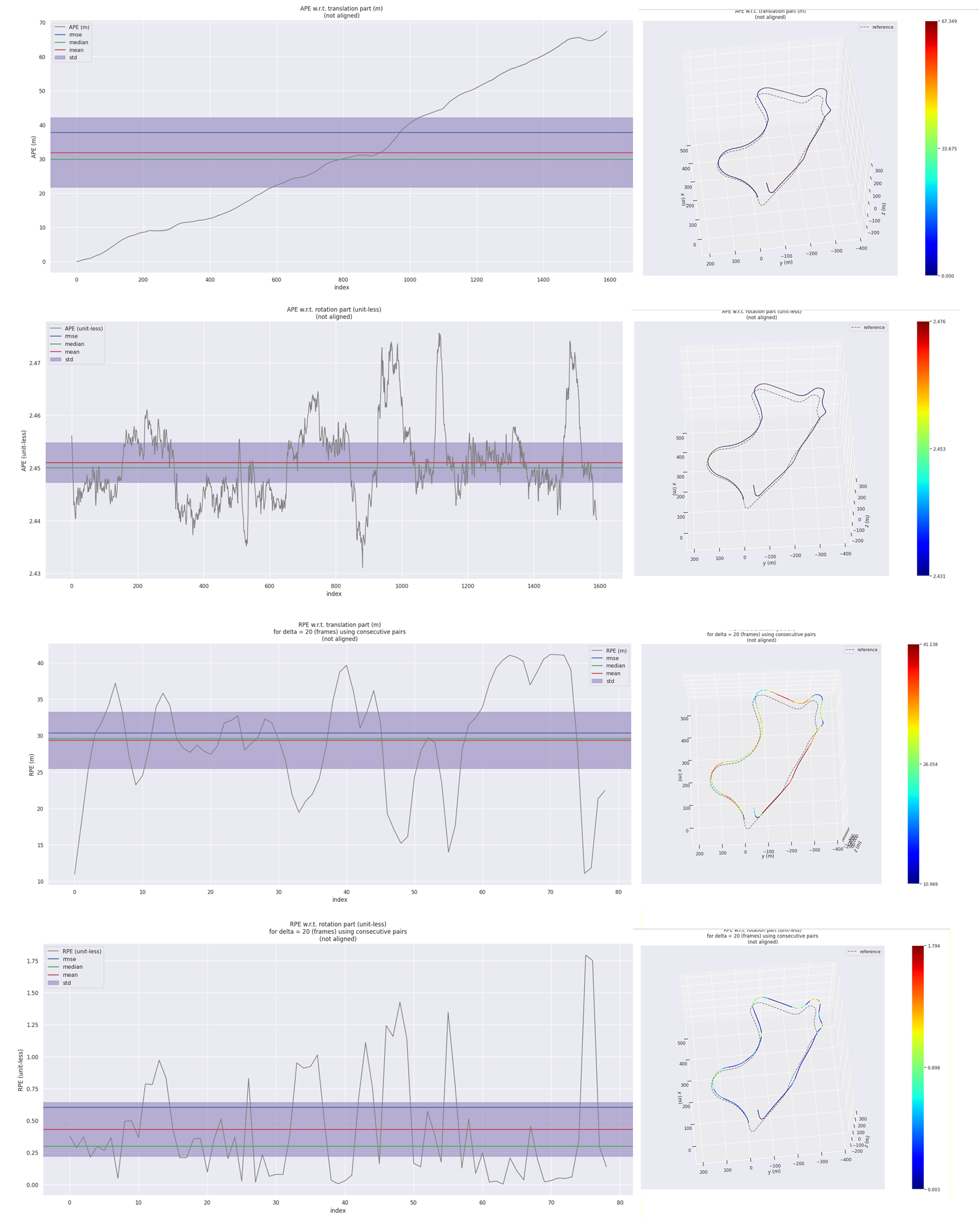}
\end{center}
\vspace{-4 mm}
\caption{Absolute and relative pose errors of 10~Hz IMU-only integration on Recording~9. The figure reports translation and rotation APE and 20-frame translation and rotation RPE without trajectory alignment.}
\label{fig:imu_10hz_ape_rpe}
\vspace{-3mm}
\end{figure*}

The quantitative results in Fig.~\ref{fig:imu_10hz_ape_rpe} confirm that standalone 10~Hz integration suffers from both accumulated translation drift and a large systematic rotation discrepancy. The translation APE increases almost continuously and reaches approximately \(67~\mathrm{m}\) near the end of the sequence. The corresponding error-colored trajectory shows that the largest deviations occur during the later part of the route, consistent with progressive inertial drift.
The rotation APE remains nearly constant around a high value rather than increasing continuously. This pattern is more characteristic of a persistent global orientation offset than of gradually accumulated rotational drift. It is consistent with the approximately \(90^\circ\) Roll offset observed in Fig.~\ref{fig:imu_10hz_pose_components} and reinforces the need to verify the coordinate transformation before interpreting the rotation metric physically.
The 20-frame RPE results show that the local motion estimate is also strongly affected at 10~Hz. Translation RPE varies approximately between \(11\) and \(41~\mathrm{m}\), with large fluctuations across the sequence. Thus, the low-rate integration does not merely suffer from a final global offset; substantial errors are already present over short temporal intervals. Rotation RPE is smaller than the absolute rotation discrepancy but contains several pronounced peaks, indicating that local angular propagation becomes unreliable during particular motion segments. Consequently, the low translation APE at the beginning of the sequence should not be interpreted as evidence of consistently accurate short-term propagation.

Overall, these experiments show that 10~Hz IMU-only integration is insufficient for reliable long-duration trajectory recovery on Recording~9. It produces substantial vertical drift, large translation APE and RPE, and a persistent rotational discrepancy. A separate 10-versus-100~Hz diagnostic is reported with the fusion comparison in Section~\ref{subsec:fusion_vs_standalone}; it is not treated as a strict sampling-rate ablation because initialization, coordinate transformations, bias settings, timestamps, and evaluation alignment were not all isolated as controlled variables.

\label{sec:imu_integration_check}

\begin{figure*}[tbp]
\begin{center}
\includegraphics[width=17cm, height=4cm]{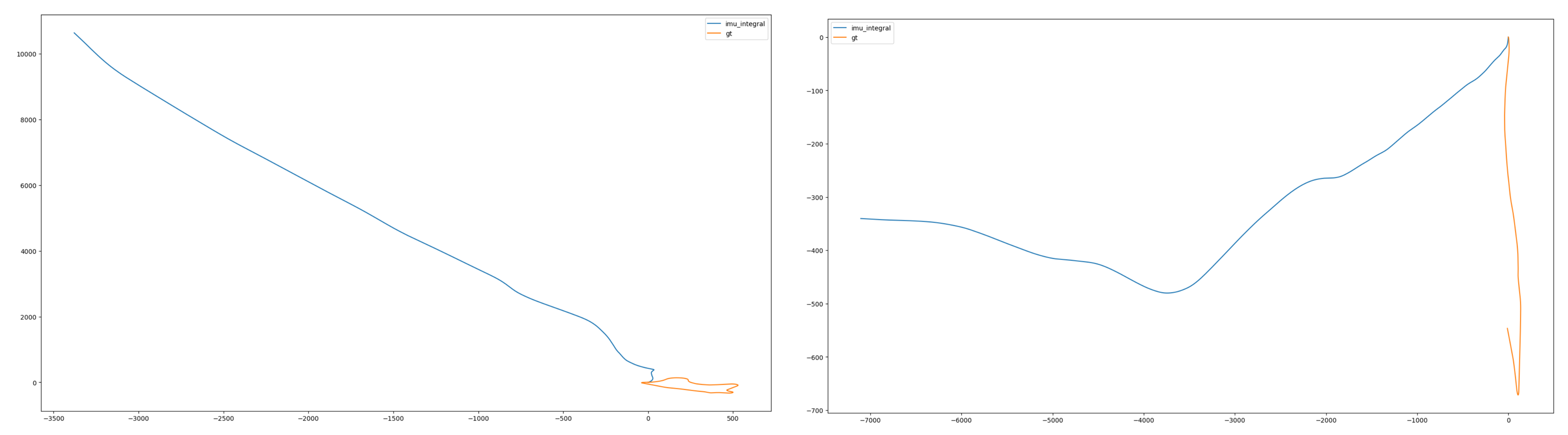}
\end{center}
\vspace{-4 mm}
\caption{Preliminary comparison between the directly integrated IMU
trajectory and the ground-truth trajectory for Recordings 09 and 10.
The blue and orange curves denote the IMU-integrated and ground-truth
trajectories, respectively. In both recordings, direct integration
produces a substantially different spatial extent and global trajectory
geometry from the reference.}
\label{fig:imu_integral_sanity}
\vspace{-3mm}
\end{figure*}

Fig.~\ref{fig:imu_integral_sanity} compares the trajectories obtained
from direct IMU integration with the corresponding ground truth. For
Recording~9, the integrated trajectory extends over a substantially
larger spatial range and follows a long monotonic path that is not
consistent with the compact ground-truth trajectory. A similarly large
discrepancy is observed for Recording 10, where the integrated trajectory
exhibits a different direction, scale, and curvature from the reference.
The disagreement in both recordings indicates that the current raw IMU
integration cannot yet be interpreted as an independent metric trajectory.
Direct double integration of accelerometer measurements is highly
sensitive to gravity compensation, accelerometer and gyroscope biases,
initial velocity, initial orientation, timestamp synchronization, and
coordinate-frame conventions. An error in any of these quantities can
rapidly accumulate and dominate the resulting trajectory. The figures
alone do not isolate which factor is responsible for the observed
discrepancy. Consequently, the IMU-integrated trajectory is retained in
the following figures as a diagnostic reference, but no accuracy
conclusion is drawn from it until the initialization, bias compensation,
gravity removal, and frame transformation are independently verified.

\subsection{EKF Trajectory Baseline}
\label{subsec:ekf_trajectory_baseline}

\begin{figure*}[tbp]
\begin{center}
\includegraphics[width=17cm, height=8cm]{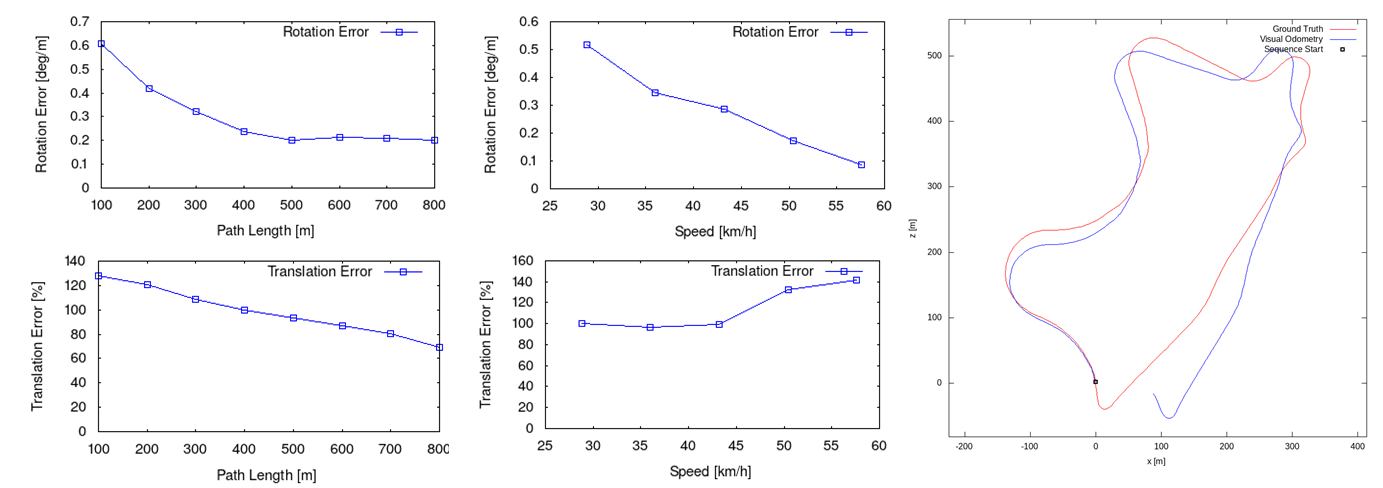}
\end{center}
\vspace{-4 mm}
\caption{EKF trajectory evaluation on KITTI. The left plots show rotation and translation errors with respect to path length and vehicle speed, while the right plot compares the EKF estimate with the ground-truth trajectory.}
\label{fig:ekf_path_speed}
\vspace{-3mm}
\end{figure*}

\begin{figure*}[tbp]
\begin{center}
\includegraphics[width=17cm, height=10cm]{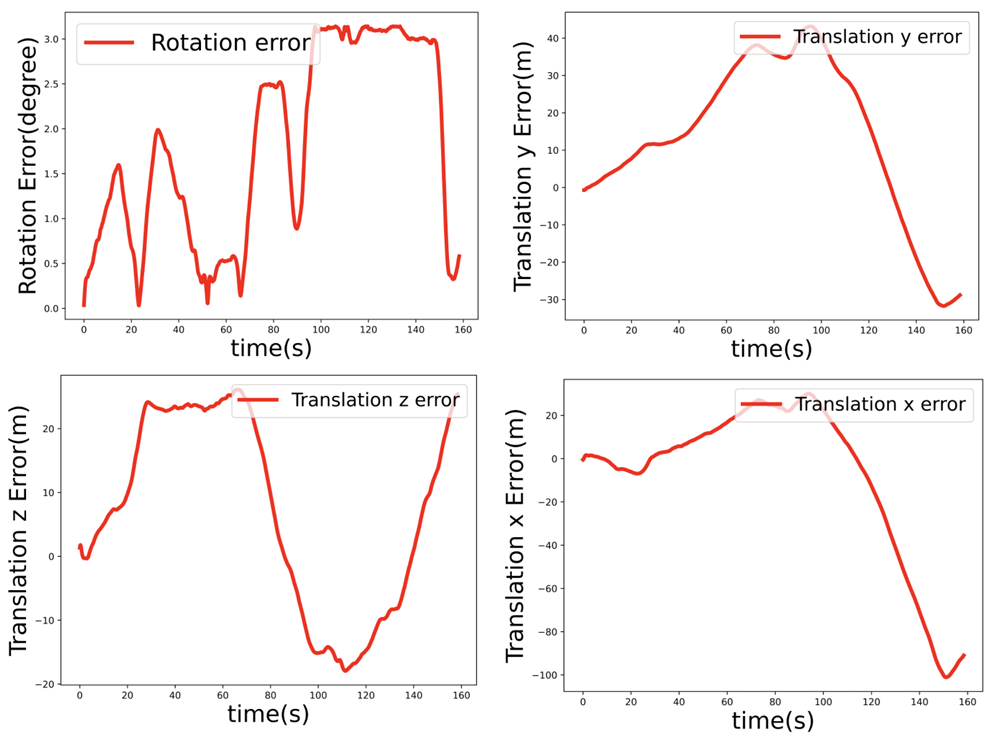}
\vspace{-4 mm}
\caption{Temporal EKF pose error on KITTI. The rotation error and translation errors along the $x$, $y$, and $z$ axes are plotted over time.}
\label{fig:ekf_temporal}
\vspace{-5mm}
\end{center}
\end{figure*}

Fig.~\ref{fig:ekf_path_speed} reports the standalone EKF trajectory evaluation
on KITTI. The estimated trajectory follows the coarse route but retains visible
deviation along curved and long-range segments. Rotation error decreases from
approximately $0.6$ to $0.2$ as path length increases, whereas translation
error remains high, decreasing from about $130\%$ to $70\%$. With speed,
rotation error decreases while translation error rises to about $140\%$.
Fig.~\ref{fig:ekf_temporal} shows the corresponding temporal error. Rotation
error reaches about $3^\circ$ in several intervals. The $x$-translation error
approaches $-100$~m near the end, the $y$ error rises to about $40$~m before
becoming negative, and the $z$ error varies by tens of meters. Thus, the EKF
recovers the coarse motion pattern but does not provide accurate long-term
translation in this standalone setting.

\subsection{Original VINS Baseline}
\label{subsec:original_vins_baseline}

\begin{figure*}[tbp]
\begin{center}
\includegraphics[width=17cm, height=16cm]{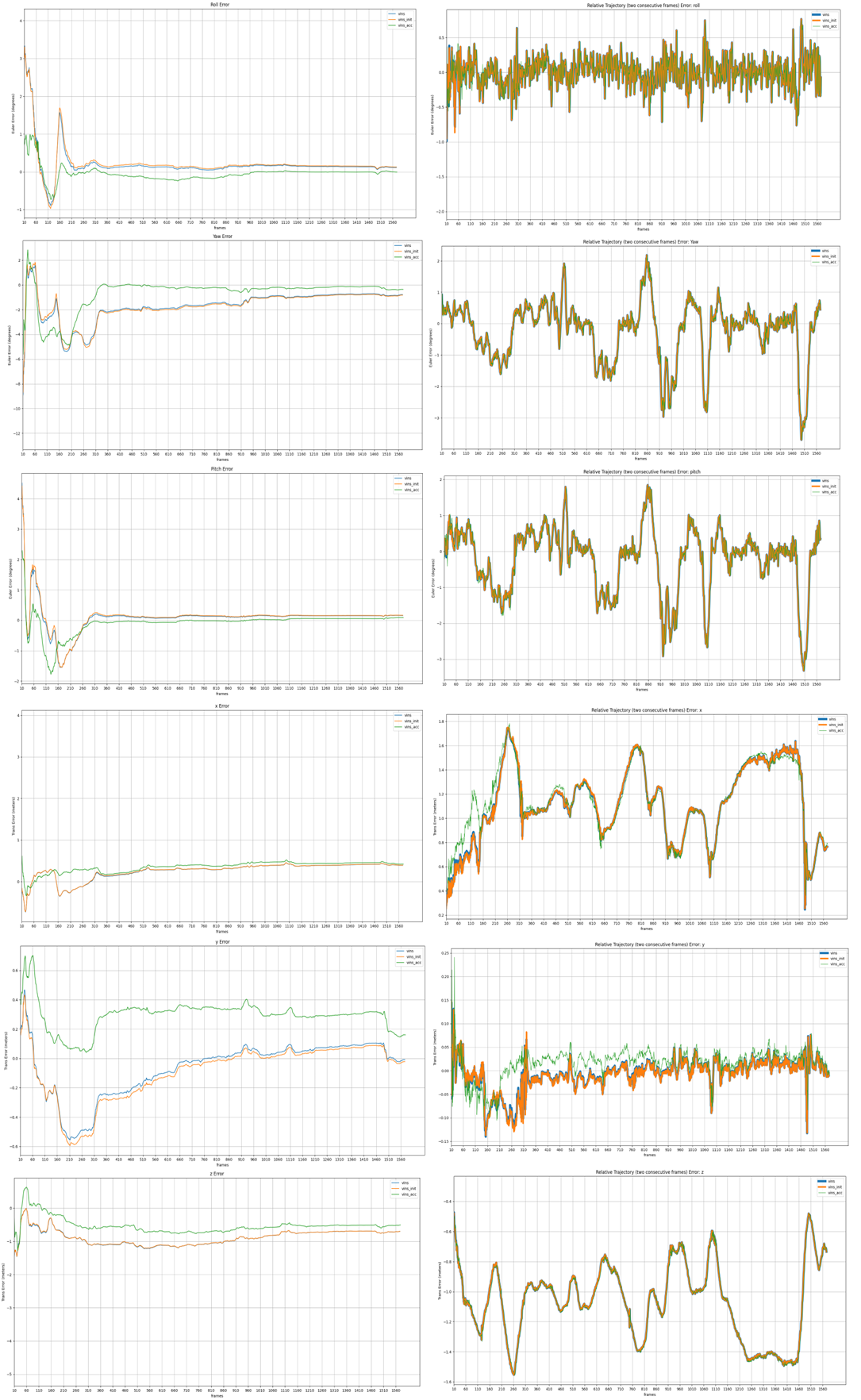}
\end{center}
\vspace{-4 mm}
\caption{Absolute and frame-to-frame pose errors of Original VINS at 10 Hz. The left column shows the Roll, Yaw, Pitch, and $x$-, $y$-, and $z$-translation errors with respect to the reference trajectory, while the right column shows the corresponding relative errors between consecutive frames.}
\label{vinsextr}
\vspace{-3mm}
\end{figure*}

Fig.~\ref{vinsextr} presents the absolute pose errors and frame-to-frame relative motion errors of VINS under a 10 Hz input frame rate. The left column reports the errors in Roll, Yaw, Pitch, and the (x)-, (y)-, and (z)-translation components with respect to the reference trajectory. The right column shows the corresponding relative rotation and translation errors between consecutive frames. Different curves represent different extrinsic initialization or calibration settings.
The absolute rotation-error curves show a clear transient stage at the beginning of the sequence. In particular, the Roll and Pitch errors undergo relatively large variations during the first several hundred frames before gradually decreasing and reaching a more stable state. This behavior is consistent with the visual-inertial initialization process of VINS, during which the system jointly estimates scale, gravity direction, velocity, IMU biases, and camera--IMU extrinsic parameters. The Yaw error also changes significantly during the initial stage, but a residual offset remains after convergence, suggesting that long-term heading estimation is more sensitive to extrinsic calibration, IMU biases, and accumulated drift.
The absolute translation errors exhibit stronger direction-dependent behavior. The (x)-axis error becomes relatively stable after the initial transient stage, and the curves obtained under different extrinsic settings eventually approach similar values. In contrast, the (y)- and (z)-axis errors retain more noticeable residual biases, and larger differences can be observed among the evaluated configurations. This suggests that different extrinsic settings have a limited effect on short-term motion estimation but may influence the accumulated global translation error. In directions weakly constrained in this sequence, small extrinsic differences can be absorbed by velocity, bias, and pose states and gradually appear as long-term offsets; this is a numerical interpretation, not an observability proof.
The frame-to-frame relative-error curves in the right column show a different pattern. Despite multiple turning and dynamic-motion segments, the curves produced by the different VINS configurations largely overlap. The relative Roll error fluctuates around zero, while the relative Yaw and Pitch errors consistently capture the major rotational changes of the platform. Similarly, the relative translation estimates along the (x), (y), and (z) axes remain highly consistent across different configurations. These observations indicate that, at 10 Hz, the evaluated extrinsic settings do not substantially alter the short-term motion estimated by VINS. The local state update remains primarily governed by visual reprojection constraints and IMU preintegration.
Overall, VINS demonstrates stable local motion estimation under different extrinsic settings, while its absolute errors still contain initialization transients and long-term residual biases. The strong agreement in frame-to-frame relative errors, together with the differences observed in absolute errors, shows that good local temporal consistency does not necessarily guarantee low global accumulated error. Therefore, VINS should be evaluated using both local relative-motion metrics and global trajectory metrics, such as APE, RPE, and axis-wise accumulated translation errors.

\textbf{Section-level finding.}
The standalone estimators exhibit complementary but incompatible error modes:
MonoViT is locally smooth but globally scale-ambiguous, IMU propagation is
sensitive to gravity and bias, and EKF/VINS retain direction-dependent global
drift. Fusion is therefore motivated by information complementarity, but
beneficial integration requires common coordinate, scale, temporal, and
uncertainty conventions.
This distinction motivates the central controlled question: whether the learned
pose prior improves Original VINS when the backend and all other settings are
held fixed.

\begin{table*}[t]
\centering
\caption{Unified main result on the representative KITTI sequence. All methods
use matched timestamps and the common coordinate convention; APE uses the
sequence-level alignment specified in Section~\ref{subsec:coordinate_scale_metric_protocol},
and RPE uses 20-frame intervals. The alignment column states the APE gauge for
each method: one sequence-level $\mathrm{Sim}(3)$ transform is used only for the
scale-ambiguous monocular trajectory, whereas metric estimators use
$\mathrm{SE}(3)$ without scale adjustment. Fixed (F) excludes the reference
extrinsic from optimization, whereas Online (O) initializes from the same value
and optimizes it.}
\label{tab:unified_common_protocol}
\scriptsize
\resizebox{\textwidth}{!}{%
\begin{tabular}{llccccccc}
\toprule
Method &
\shortstack{APE gauge/\\alignment} &
\shortstack{Trans. APE\\RMSE (m)} &
\shortstack{Trans. APE\\Max (m)} &
\shortstack{Rot. APE\\RMSE ($^\circ$)} &
\shortstack{Trans. RPE\\RMSE (m)} &
\shortstack{Rot. RPE\\RMSE ($^\circ$)} &
$e_R$ ($^\circ$) &
$e_t$ (m) \\
\midrule
MonoViT                 & $\mathrm{Sim}(3)$, scale-free & 42.6 & 106.8 & 5.2 & \textbf{2.4} & \textbf{0.81} & --  & --    \\
IMU, 10 Hz              & $\mathrm{SE}(3)$, metric & 38.3 &  67.1 & 8.4 & 30.1 & 2.86 & --  & --    \\
EKF                      & $\mathrm{SE}(3)$, metric & 47.5 &  99.4 & 6.1 & 18.7 & 2.12 & 2.9 & 0.018 \\
Original VINS--F        & $\mathrm{SE}(3)$, metric & \textbf{31.4} & \textbf{43.6} & 4.0 & 29.0 & 1.76 & --$^{\dagger}$ & --$^{\dagger}$ \\
Original VINS--O        & $\mathrm{SE}(3)$, metric & 88.6 & 116.9 & 4.2 & 29.2 & 1.78 & 4.5 & 0.28 \\
VINS+Pose--F            & $\mathrm{SE}(3)$, metric & 31.8 &  44.1 & \textbf{3.9} & 28.8 & 1.72 & --$^{\dagger}$ & --$^{\dagger}$ \\
VINS+Pose--O            & $\mathrm{SE}(3)$, metric & 95.0 & 122.7 & 4.3 & 29.4 & 1.79 & 5.0 & 0.33 \\
\bottomrule
\end{tabular}}
\raggedright
$^{\dagger}$Not estimated: the extrinsic is fixed to the reference calibration.
\end{table*}

\section{Incremental Effect of Learned Pose Priors}
\label{sec:learned_pose_incremental_effect}
This section separates two claims that are often conflated: a visual--inertial system
can outperform standalone visual or inertial inputs, while the learned prior
may still add little to the native VINS constraints. The latter claim requires
the same backend evaluated with and without the prior.

\subsection{Unified Main Results under the Common Protocol}
\label{subsec:unified_protocol_results}

We first report the headline controlled result. Every estimator uses the same
sequence interval, accepted timestamps, and reference frame. Monocular
trajectories use one sequence-level similarity alignment without frame-wise
scale correction. Absolute rotation and camera--IMU rotation errors use the
geodesic distance
\begin{equation}
 e_R = \cos^{-1}\!\left(
 \frac{\operatorname{tr}(\mathbf{R}_{\mathrm{gt}}^{\top}
 \mathbf{R}_{\mathrm{est}})-1}{2}
 \right),
 \label{eq:so3_geodesic_unified}
\end{equation}
and translation calibration error is
$e_t=\|\mathbf{t}_{\mathrm{est}}-\mathbf{t}_{\mathrm{gt}}\|_2$. RPE is
evaluated over a fixed interval of 20 accepted frames.
Table~\ref{tab:unified_common_protocol} separates local motion consistency from
global trajectory accuracy. MonoViT has the smallest translation and rotation
RPE, but its translation APE has a maximum above 100~m. Smooth short-horizon
motion therefore coexists with accumulated scale and direction error. Expressing
all rotations in a common frame also removes the shared orientation offset seen
in the non-aligned diagnostic plots.
The matched VINS pairs directly answer RQ2. With Fixed calibration, adding the
pose prior changes translation APE RMSE from 31.4~m to 31.8~m and only slightly
reduces rotation RPE; the learned prior preserves but does not measurably improve
the Original VINS solution in this setting. Releasing the same accurate
extrinsic increases translation APE RMSE by 182.2\% for Original VINS and
198.7\% for VINS+Pose, while translation and rotation RPE change by less than
2.1\%. This result localizes the main degradation to accumulated global bias
rather than an immediate frame-to-frame tracking failure.

\subsection{Preliminary Fusion-versus-Standalone Comparison}
\label{subsec:fusion_vs_standalone}

\begin{figure*}[tbp]
\begin{center}
\includegraphics[width=17cm, height=9cm]{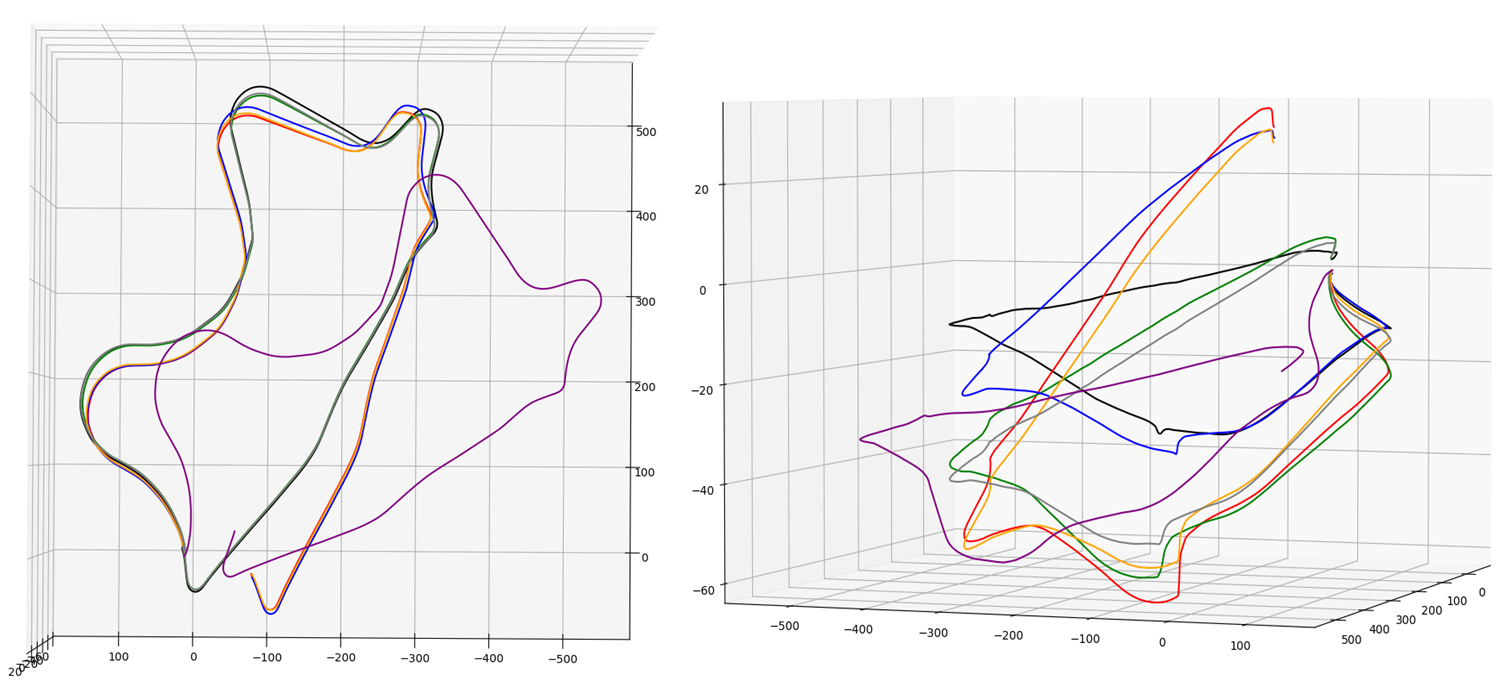}
\end{center}
\vspace{-4 mm}
\caption{Preliminary trajectory comparison on the KITTI sequence. Black: ground truth; blue: MonoViT; green: EKF with reference input; gray: EKF under $1^\circ/5$~cm extrinsic perturbation; red: VINS+Pose; orange: VINS+Pose under the same perturbation; purple: IMU-only propagation.}
\label{fig:e2e_traj_comparison}
\vspace{-3mm}
\end{figure*}

\begin{figure*}[tbp]
\begin{center}
\includegraphics[width=17cm, height=9cm]{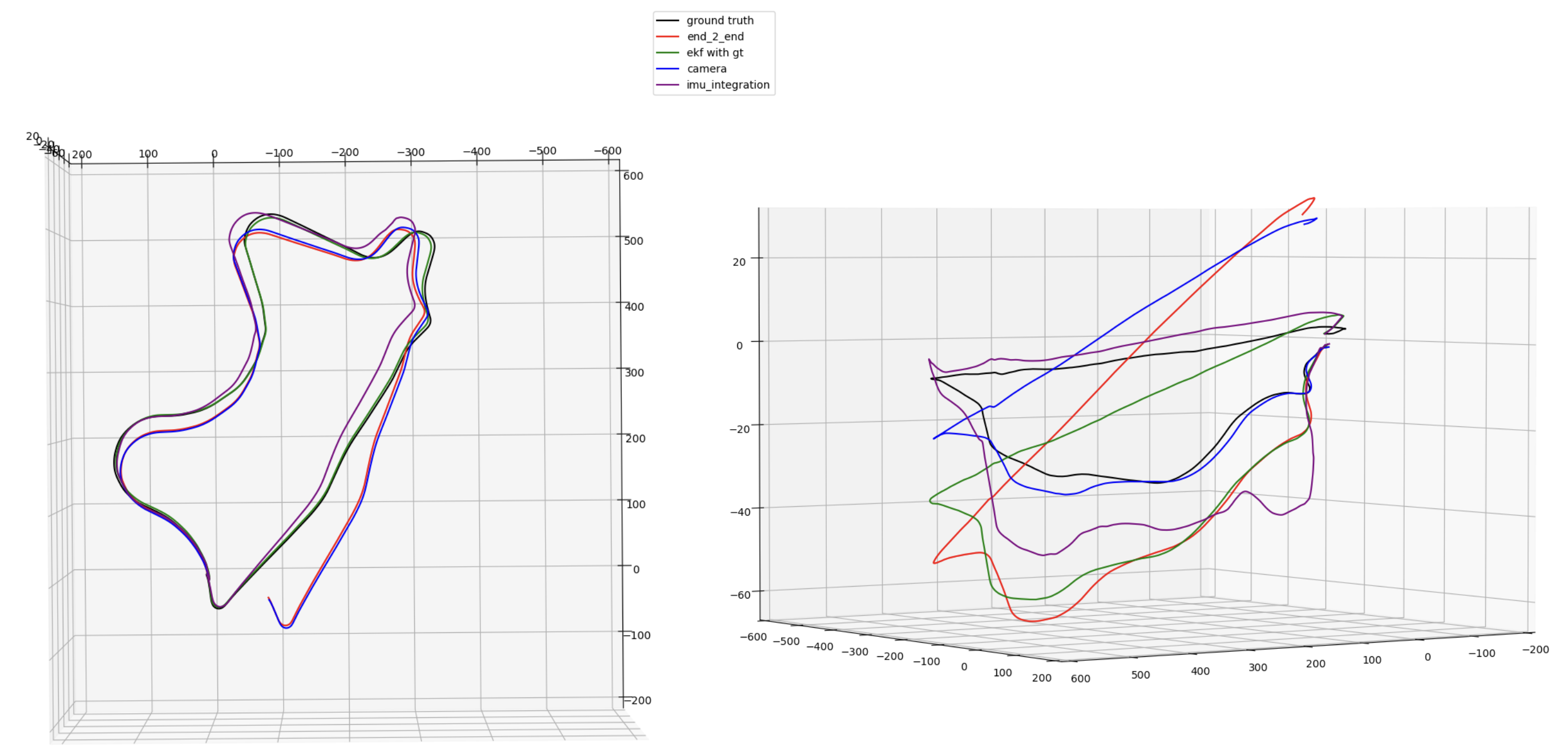}
\end{center}
\vspace{-4 mm}
\caption{Complementary 3-D view of the preliminary KITTI comparison. Black: ground truth; red: VINS+Pose; green: EKF with reference input; blue: standalone MonoViT; purple: IMU-only propagation.}
\label{fig:qualitative_traj_comparison}
\vspace{-3mm}
\end{figure*}

Figs.~\ref{fig:e2e_traj_comparison} and~\ref{fig:qualitative_traj_comparison} provide complementary two- and three-dimensional views of the preliminary fusion comparison. The standalone MonoViT trajectory recovers the coarse route but retains scale-dependent long-range drift, while direct IMU integration accumulates a different bias-driven drift pattern. The EKF with reliable input and VINS+Pose remain closer to the reference in the displayed sequence; the noisy-extrinsic variants preserve the coarse route under the evaluated perturbation.
This comparison establishes only that a visual--inertial backend can improve over the standalone inputs in this sequence. It does not isolate the incremental contribution of the learned pose prior, because VINS+Pose differs from the standalone baselines in several constraints and state variables. The controlled Original VINS versus VINS+Pose comparisons below address that narrower question.

\begin{figure*}[tbp]
\begin{center}
\includegraphics[width=17cm, height=5cm]{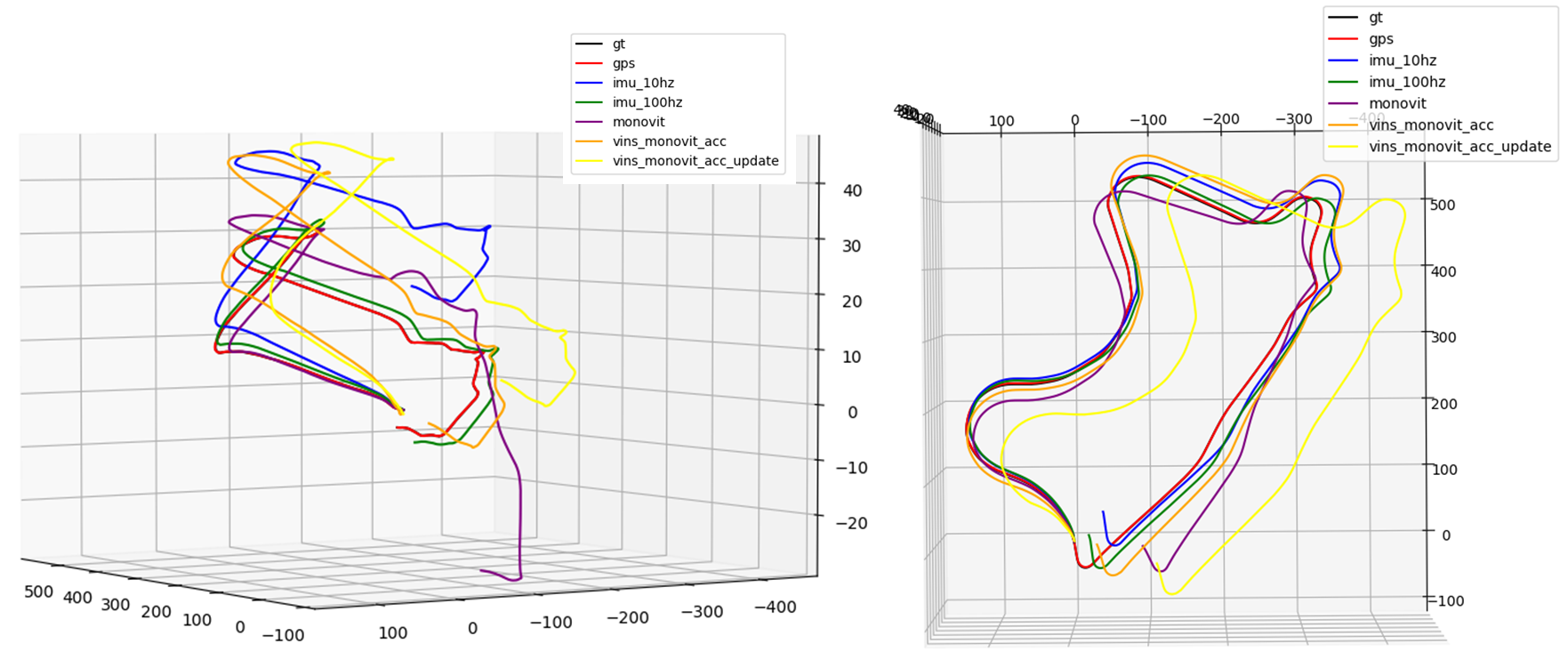}
\end{center}
\vspace{-4 mm}
\caption{Diagnostic trajectory comparison on Recording~9. The two panels compare ground truth, GPS, 10~Hz and 100~Hz IMU propagation, MonoViT, VINS+Pose--F, and VINS+Pose--O. The 10/100~Hz curves are not interpreted as a strict one-variable ablation.}
\label{fig:imu_sampling_rate_trajectory_comparison}
\vspace{-3mm}
\end{figure*}

\begin{figure*}[tbp]
\begin{center}
\includegraphics[width=17cm, height=16cm]{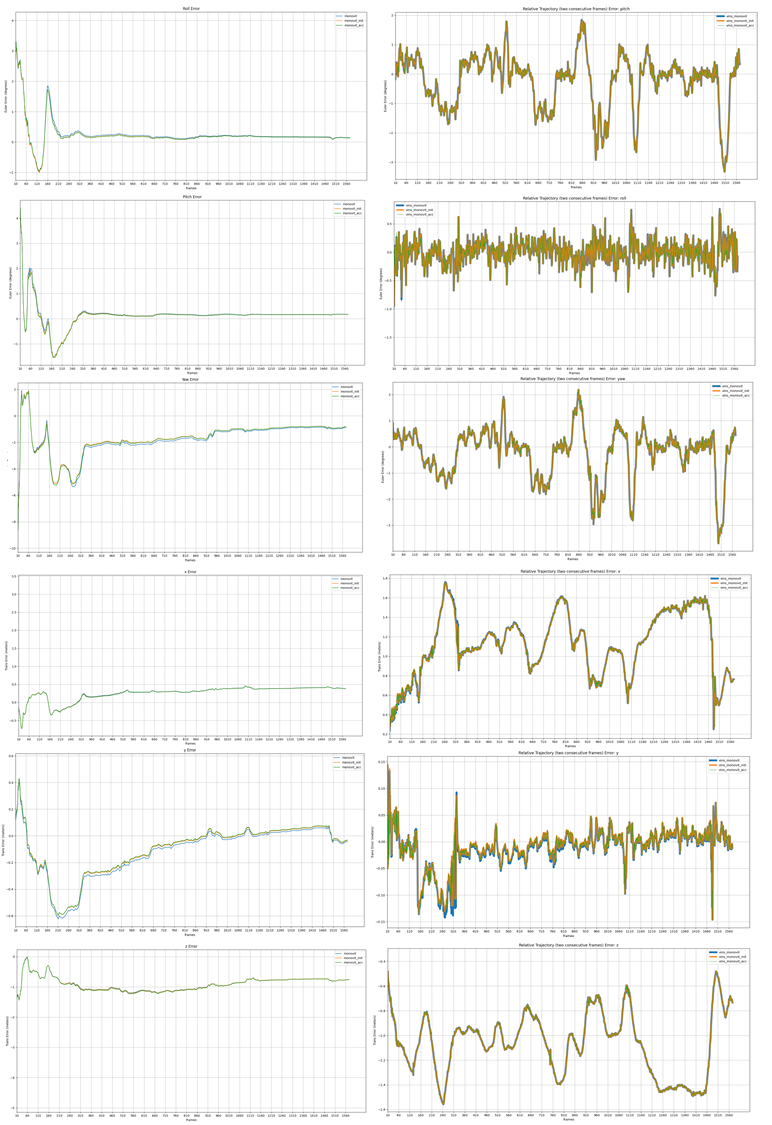}
\end{center}
\vspace{-4 mm}
\caption{Absolute and frame-to-frame pose errors of the MonoViT-based fused configurations at 10 Hz. The left column presents accumulated rotation and translation errors with respect to the reference trajectory, and the right column presents relative errors between consecutive frames.}
\label{movoextr}
\vspace{-3mm}
\end{figure*}

Fig.~\ref{fig:imu_sampling_rate_trajectory_comparison} extends this preliminary
comparison with 10~Hz and 100~Hz IMU propagation and Fixed/Online VINS+Pose
configurations. The 100~Hz inertial trajectory is visually closer to the
reference than the 10~Hz trajectory, but this is a diagnostic comparison rather
than a controlled sampling-rate ablation because the archived evidence does not
establish that initialization, bias, synchronization, and alignment were held
identical. The Online VINS+Pose trajectory also shows a larger global deviation
than several displayed baselines; this observation motivates the controlled
Fixed-versus-Online analysis in Section~\ref{sec:calibration_initialization_robustness}.
Fig.~\ref{movoextr} compares absolute and frame-to-frame errors for the
MonoViT-based fused configurations at 10~Hz. Their relative-error curves mostly
overlap, whereas configuration-dependent offsets remain in the absolute
translation components. Thus, local agreement does not establish global
accuracy, and the figure is treated as supporting diagnostic evidence rather
than as a standalone MonoViT result.

\subsection{Cross-Sequence Controlled Comparison}
\label{subsec:cross_sequence_pose_extrinsic}

We next conduct a representative stress comparison rather than estimate a
population-wide mean. The available experiment artifacts do not establish that
the four recordings were selected before inspecting outcomes. Recordings 1 and
6 exercise challenging initialization or visual conditions, whereas Recordings
7 and 9 exercise comparatively favorable conditions. Conclusions are therefore
limited to this four-recording panel. We compare Original VINS and VINS+Pose
under Fixed and Online extrinsics and evaluate one stronger Fixed setting with
$\lambda_{\mathrm{pose}}^{\mathrm{high}}=5
\lambda_{\mathrm{pose}}^{\mathrm{current}}$.

\begin{table}[t]
\centering
\caption{Cross-sequence translation APE RMSE (m). F and O denote fixed and
online-updated extrinsics, and H denotes the higher pose-prior weight.}
\label{tab:cross_sequence_pose_ape}
\scriptsize
\resizebox{\columnwidth}{!}{%
\begin{tabular}{lccccc}
\toprule
Configuration & Rec.~1 & Rec.~6 & Rec.~7 & Rec.~9 & Mean \\
\midrule
Original VINS--F     & 79.2 & \textbf{65.1} & 24.8 & \textbf{31.4} & 50.1 \\
Original VINS--O     & 91.8 & 79.6 & 31.9 & 88.6 & 73.0 \\
VINS+Pose--F         & \textbf{78.5} & 65.9 & \textbf{23.9} & 31.8 & \textbf{50.0} \\
VINS+Pose--O         & 95.4 & 83.8 & 30.7 & 95.0 & 76.2 \\
VINS+Pose--F--H      & 84.1 & 71.8 & 25.2 & 35.7 & 54.2 \\
\bottomrule
\end{tabular}}
\end{table}

\begin{table}[t]
\centering
\caption{Aggregate statistics across the four-recording stress panel. Prior
$\Delta$APE compares each VINS+Pose setting with Original VINS under the matched
extrinsic strategy; Online $\Delta$APE compares Online with Fixed for the same
estimator. Per-recording differences are retained in
Table~\ref{tab:cross_sequence_pose_ape}; no win/tie/loss statistic is reported.}
\label{tab:cross_sequence_pose_summary}
\scriptsize
\resizebox{\columnwidth}{!}{%
\begin{tabular}{lccccc}
\toprule
Configuration &
\shortstack{Mean Trans.\\RPE (m)} &
\shortstack{Mean $e_R$\\($^\circ$)} &
\shortstack{Mean $e_t$\\(m)} &
\shortstack{Prior\\$\Delta$APE} &
\shortstack{Online\\$\Delta$APE} \\
\midrule
Original VINS--F & 17.7 & --  & --   & --       & --       \\
Original VINS--O & 17.9 & 5.2 & 0.26 & --       & +45.7\%  \\
VINS+Pose--F    & \textbf{17.6} & --  & --   & $-0.2$\% & --       \\
VINS+Pose--O    & 18.0 & 5.5 & 0.30 & +4.4\%   & +52.1\%  \\
VINS+Pose--F--H & 19.0 & --  & --   & +8.2\%   & --       \\
\bottomrule
\end{tabular}}
\end{table}

The multi-sequence results show that the current pose prior does not provide a
consistent incremental improvement over Original VINS. Under fixed extrinsics,
VINS+Pose improves Recordings 1 and 7 but slightly degrades Recordings 6 and 9,
leading to a negligible mean APE change of $-0.2\%$. The improvement on the
favorable Recording 7 is therefore not sufficient to establish a general gain.
The tested five-times pose-prior weight increases mean APE by 8.2\%, and each
per-recording value in Table~\ref{tab:cross_sequence_pose_ape} is higher than the
matched Original VINS--F value. This single tested point does not characterize a weight
sweep; it is consistent with partial redundancy or residual inconsistency at
that weight. A broader sweep would be required to distinguish those mechanisms
and characterize other weighting regimes.

The fixed-versus-online comparison is considerably more consistent. Online
extrinsic updating increases the mean APE by 45.7\% for Original VINS and 52.1\%
for VINS+Pose, while the average RPE changes by less than 2\%. The learned pose
prior also does not reduce the average online calibration error. These results
suggest that the additional calibration degrees of freedom mainly redistribute
systematic inconsistency into the extrinsic and global trajectory states.
\subsection{Detailed Pose-Prior Error Analysis}
\label{subsec:pose_prior_detailed_errors}
\begin{figure*}[tbp]
\begin{center}
\includegraphics[width=17cm, height=16cm]{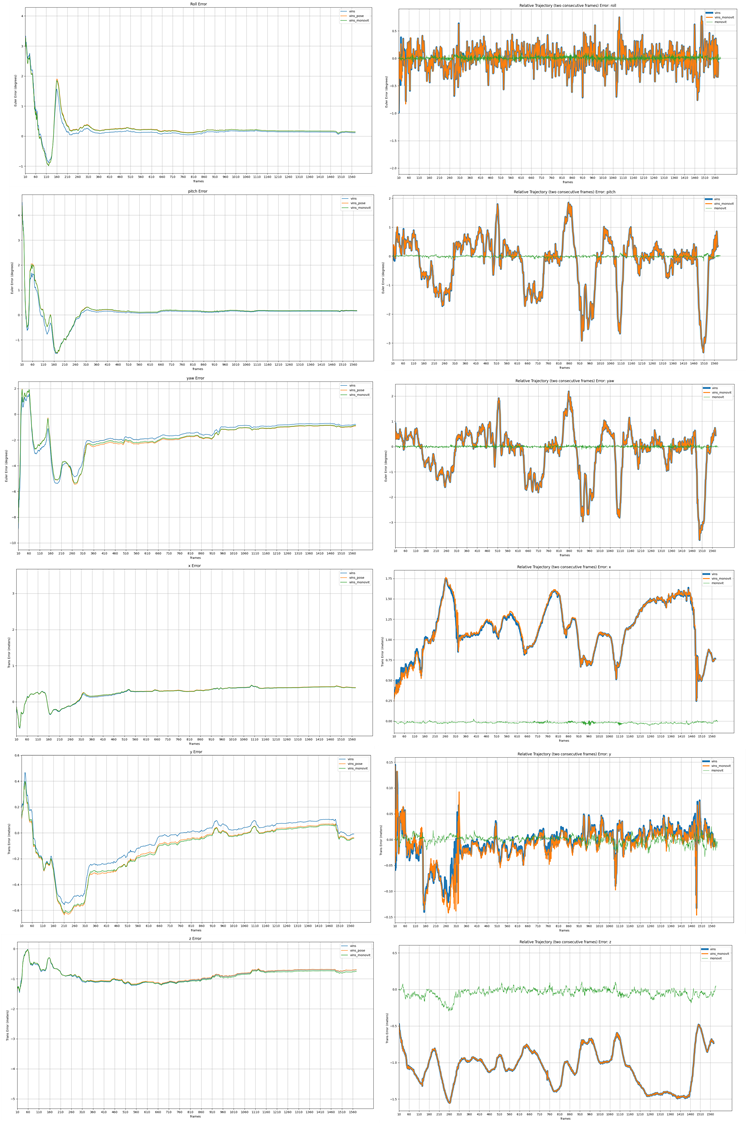}
\end{center}
\vspace{-4 mm}
\caption{Absolute and frame-to-frame pose errors of VINS, VINS+Pose, and standalone MonoViT on KITTI at 10 Hz. The left column shows the absolute Roll, Pitch, Yaw, and translation errors with respect to the reference trajectory, while the right column shows the corresponding relative errors between consecutive frames.}
\label{fig:monovit_pose_prior_10hz}
\vspace{-3mm}
\end{figure*}

Fig.~\ref{fig:monovit_pose_prior_10hz} further compares the baseline VINS estimator, VINS augmented with the MonoViT pose prior, and the standalone MonoViT output under the same 10~Hz input frame rate. The left column presents the absolute errors in Roll, Pitch, Yaw, and the \(x\)-, \(y\)-, and \(z\)-translation components with respect to the reference trajectory, whereas the right column reports the corresponding relative errors between consecutive frames.
The absolute rotation-error curves exhibit a clear transient stage during the first several hundred frames. After this initialization period, the Roll and Pitch errors gradually stabilize, while the Yaw error retains a larger residual offset. The baseline VINS and VINS+Pose curves remain highly consistent throughout most of the sequence, indicating that the introduction of the learned pose prior does not substantially alter the rotational solution produced by the VINS backend. This observation is consistent with the previously reported extrinsic-estimation results, where the learned pose prior caused only minor changes to the final camera--IMU rotation.
The frame-to-frame rotation errors provide a similar observation. VINS and VINS+Pose respond almost identically to the major rotational changes in the sequence, including the high-error peaks associated with rapid vehicle motion or turning. The standalone MonoViT output generally exhibits a smoother local-error profile in several rotational components. However, the lower-amplitude local residuals do not consistently translate into lower absolute rotation errors. This distinction suggests that MonoViT can provide temporally smooth local orientation cues, but its long-term orientation remains affected by accumulated prediction errors.
The translation results reveal a stronger difference between local and global behavior. In the absolute-error plots, the baseline VINS and VINS+Pose trajectories remain close, particularly in the \(x\)- and \(z\)-directions. The \(y\)-direction shows a more noticeable configuration-dependent offset, but adding the MonoViT prior does not produce a consistent reduction in the absolute translation error. The standalone MonoViT output follows a different error profile, reflecting the scale ambiguity and coordinate inconsistency inherent in monocular pose estimation.
In the frame-to-frame translation plots, VINS and VINS+Pose again show nearly overlapping curves and capture the same short-term vehicle-motion variations. This indicates that the local translation estimated by the fused system is still primarily governed by the visual reprojection and IMU-preintegration constraints of VINS. Although the MonoViT output provides locally continuous motion information, its contribution does not significantly change the VINS frame-to-frame estimate under the current weighting configuration.
The comparison between absolute and relative errors is particularly important. In several components, MonoViT produces relatively smooth consecutive-frame errors while still retaining visible long-term absolute biases. Therefore, good local temporal consistency alone does not guarantee accurate global trajectory recovery. Small frame-to-frame scale or direction errors can accumulate over the sequence and result in substantial global drift.
Overall, the figure shows that introducing the MonoViT pose prior preserves the stability of the original VINS estimator but does not yield a clear or consistent improvement in either absolute rotation or translation accuracy. The strong overlap between VINS and VINS+Pose suggests that the original visual-inertial constraints dominate the backend optimization, while the learned pose prior acts only as a weak auxiliary constraint. More effective integration may require explicit scale alignment, adaptive uncertainty weighting, and confidence-aware selection of the learned pose measurements.

\begin{figure*}[tbp]
\begin{center}
\includegraphics[width=17cm, height=8cm]{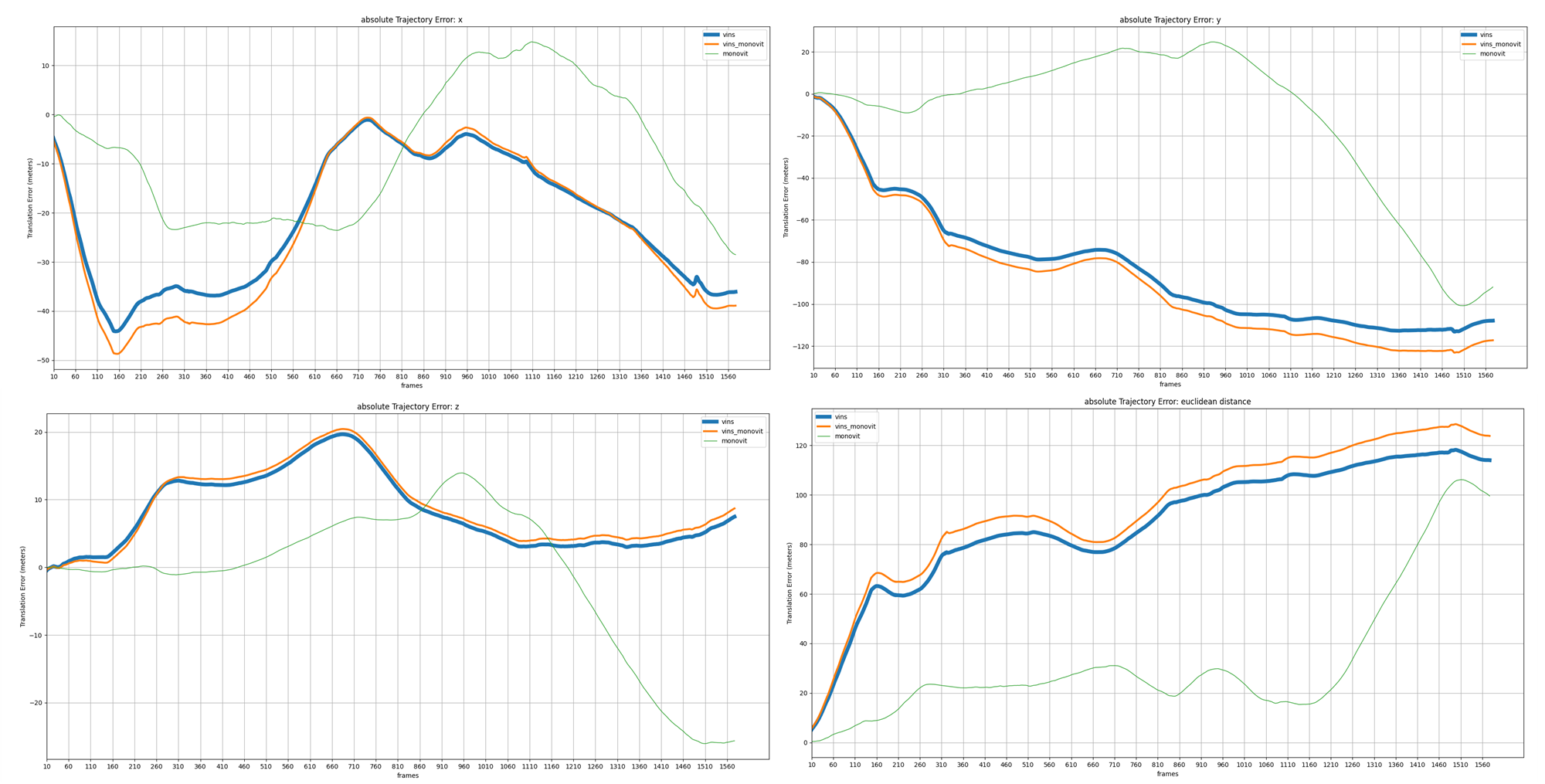}
\end{center}
\vspace{-4 mm}
\caption{Absolute trajectory error comparison of VINS, VINS+Pose, and standalone MonoViT on KITTI at 10 Hz. The first three plots show the signed trajectory errors along the (x)-, (y)-, and (z)-axes, and the last plot shows the Euclidean position-error magnitude. VINS and VINS+Pose exhibit similar accumulated drift, with the (y)-direction contributing most strongly to the final error.}
\label{fig:absolute_trajectory_error}
\vspace{-3mm}
\end{figure*}
\begin{figure*}[tbp]
\begin{center}
\includegraphics[width=17cm, height=7cm]{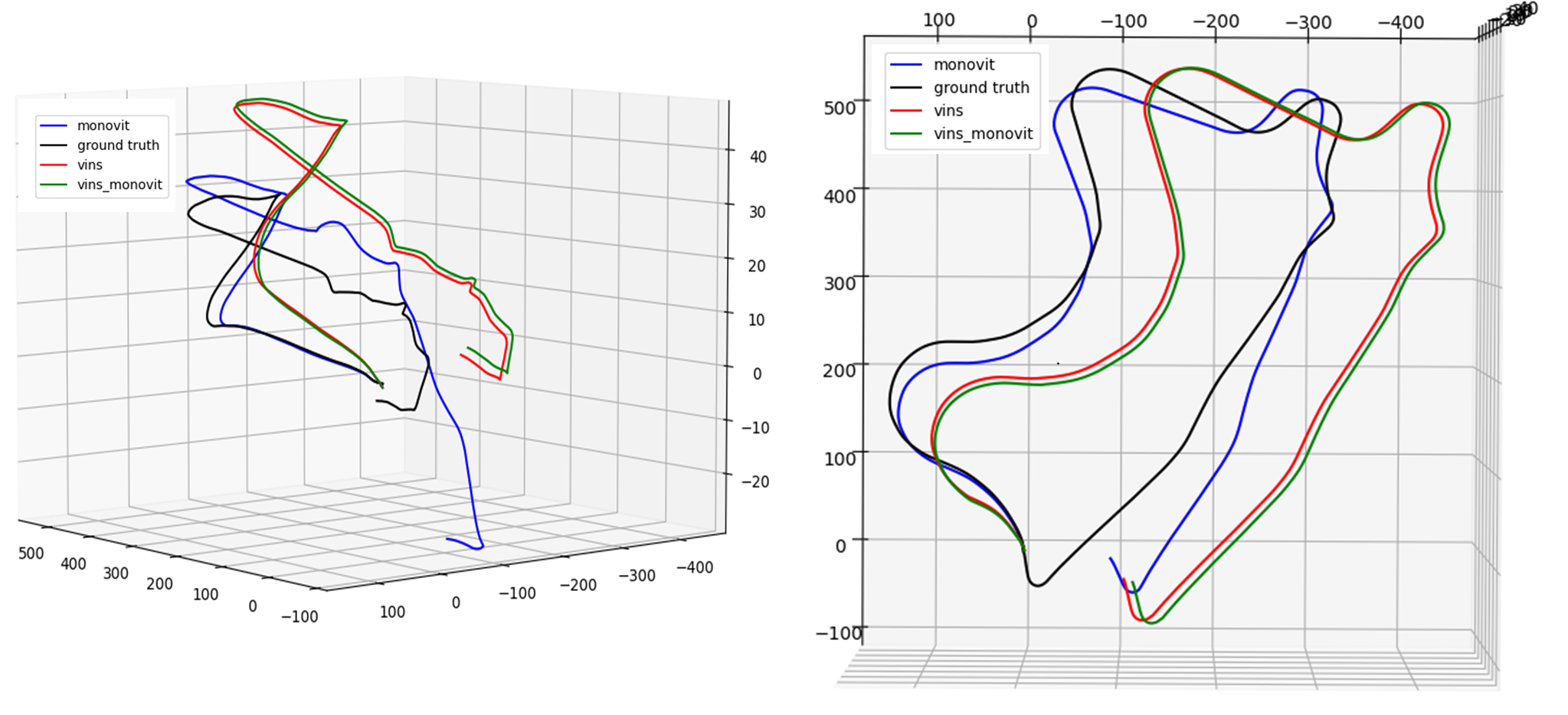}
\end{center}
\vspace{-4 mm}
\caption{Qualitative trajectory comparison of standalone MonoViT, baseline VINS, and VINS+Pose on KITTI at 10 Hz. The left panel shows the estimated trajectories in 3D space, while the right panel presents a near top-down view.}
\label{fig:qualitative_trajectory_comparison1}
\vspace{-3mm}
\end{figure*}
\subsection{Trajectory APE, RPE, and Qualitative Evidence}
\label{subsec:pose_prior_trajectory_evidence}

Fig.~\ref{fig:absolute_trajectory_error} compares the absolute trajectory errors of the baseline VINS estimator, VINS augmented with the MonoViT prior, and standalone MonoViT. The first three plots report the signed trajectory errors along the \(x\)-, \(y\)-, and \(z\)-axes, while the final plot shows the Euclidean magnitude of the three-dimensional position error.
The baseline VINS and VINS+Pose curves remain close throughout the sequence, which is consistent with the previous frame-to-frame and extrinsic-error results. Along the \(x\)-axis, both methods exhibit a large negative error during the early stage, partially recover around the middle of the sequence, and then accumulate negative drift again toward the end. The VINS+Pose configuration generally follows the same trend as the baseline but shows a slightly larger error magnitude in several intervals. A similar behavior is observed along the \(y\)-axis, where both methods accumulate substantial negative drift. The \(y\)-component becomes the dominant source of their overall position error, reaching approximately \(-110~\mathrm{m}\) for VINS and nearly \(-120~\mathrm{m}\) for VINS+Pose near the end of the sequence.
The \(z\)-axis errors of the two VINS-based methods are considerably smaller than their horizontal errors. Both curves first increase to approximately \(20~\mathrm{m}\) and then gradually decrease to below \(10~\mathrm{m}\). The strong overlap among the \(x\)-, \(y\)-, and \(z\)-component curves indicates that the learned MonoViT prior does not substantially change the global drift pattern produced by the VINS backend. Under the current integration strategy, the additional learned constraint is therefore insufficient to correct the dominant horizontal translation error.
The standalone MonoViT result exhibits a different error profile. During a large portion of the sequence, its Euclidean trajectory error remains lower than those of the two VINS-based configurations. However, its component-wise errors vary substantially over time. The \(x\)-error changes sign, the \(y\)-error moves from a positive offset to a large negative deviation, and the \(z\)-error rapidly decreases during the final part of the sequence. The Euclidean error consequently rises sharply after approximately frame 1200, reaching more than \(100~\mathrm{m}\) near the end. This behavior reflects the long-term scale inconsistency and accumulated drift of standalone monocular pose estimation.
The Euclidean-error plot further shows that VINS maintains a relatively smooth but continuously increasing error profile, whereas MonoViT achieves lower error over the middle portion of the sequence but undergoes severe late-stage degradation. VINS+Pose does not outperform the baseline VINS and instead produces a slightly larger Euclidean error over most of the sequence. Therefore, simply introducing the MonoViT pose prior does not provide a consistent global trajectory improvement.
Overall, this experiment confirms that local agreement in frame-to-frame motion does not necessarily lead to lower accumulated trajectory error. Although MonoViT can provide useful short-term motion cues, its scale uncertainty and long-term drift must be explicitly handled before integration. The current results suggest that effective fusion requires confidence-aware weighting, scale alignment, and selective rejection of unreliable learned pose measurements rather than treating the MonoViT output as a uniformly reliable constraint.

\begin{figure*}[tbp]
\begin{center}
\setlength{\unitlength}{1cm}
\begin{picture}(17,9)
\put(0,0){\includegraphics[width=17cm,height=9cm]{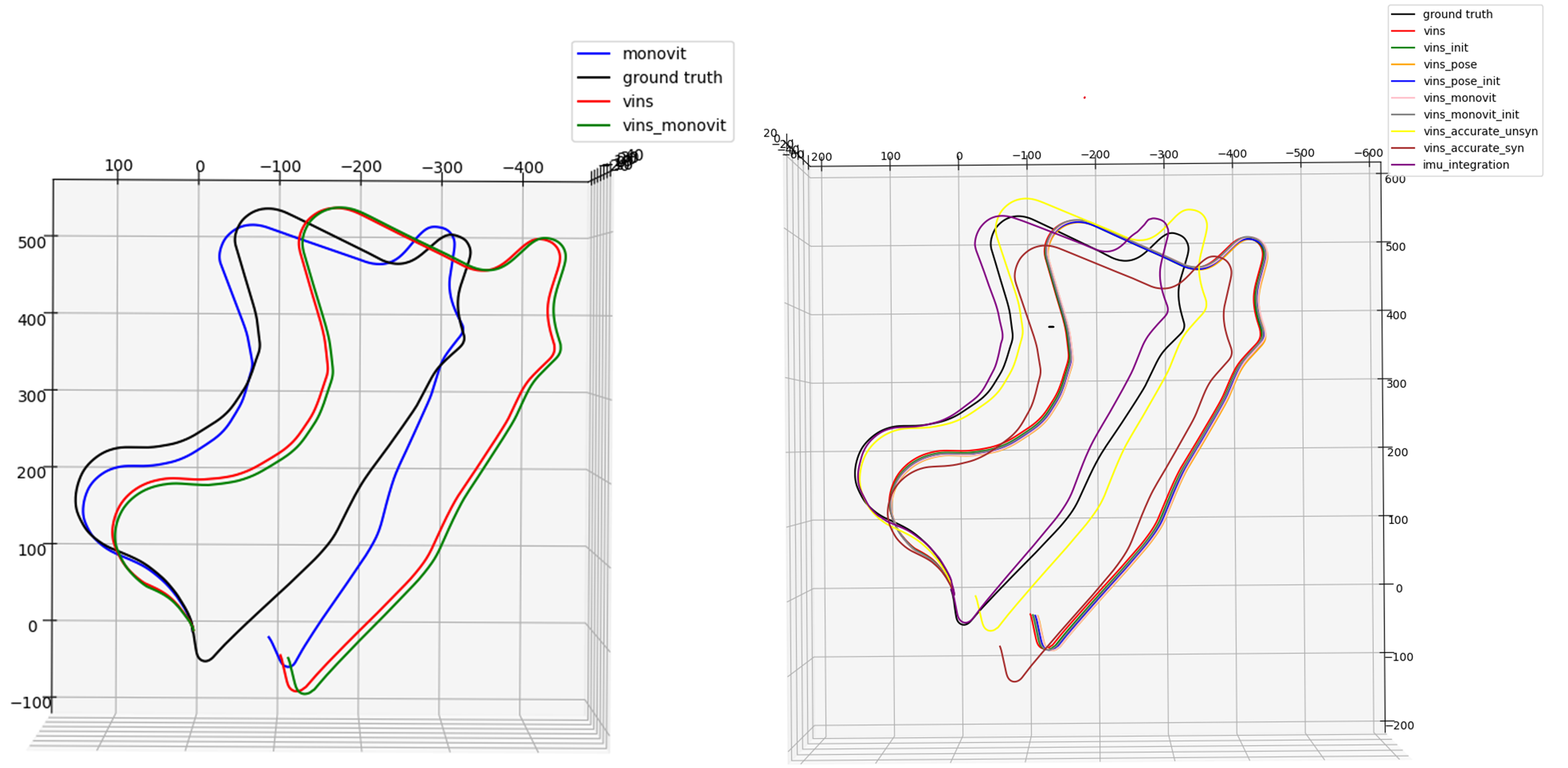}}
\put(6.15,8.05){\colorbox{white}{\parbox[c][1.45cm][c]{2.45cm}{%
\fontsize{5.2}{5.8}\selectfont
\textcolor{blue}{\rule{0.34cm}{0.7pt}}~MonoViT\\
\textcolor{black}{\rule{0.34cm}{0.7pt}}~Ground truth\\
\textcolor{red}{\rule{0.34cm}{0.7pt}}~Original VINS\\
\textcolor{green!60!black}{\rule{0.34cm}{0.7pt}}~VINS+Pose+Depth}}}
\put(14.55,7.95){\colorbox{white}{\parbox[c][2.2cm][c]{2.35cm}{%
\fontsize{4.4}{4.8}\selectfont
\textcolor{black}{\rule{0.30cm}{0.6pt}}~Ground truth\\
\textcolor{red}{\rule{0.30cm}{0.6pt}}~Original VINS\\
\textcolor{green!60!black}{\rule{0.30cm}{0.6pt}}~Original VINS (Init.)\\
\textcolor{orange}{\rule{0.30cm}{0.6pt}}~VINS+Pose\\
\textcolor{blue}{\rule{0.30cm}{0.6pt}}~VINS+Pose (Init.)\\
\textcolor{pink}{\rule{0.30cm}{0.6pt}}~VINS+Pose+Depth\\
\textcolor{gray}{\rule{0.30cm}{0.6pt}}~VINS+Pose+Depth (Init.)\\
\textcolor{yellow!80!black}{\rule{0.30cm}{0.6pt}}~Original VINS (Acc., Unsync.)\\
\textcolor{brown}{\rule{0.30cm}{0.6pt}}~Original VINS (Acc., Sync.)\\
\textcolor{purple}{\rule{0.30cm}{0.6pt}}~IMU-only}}}
\end{picture}
\end{center}
\vspace{-4 mm}
\caption{Qualitative trajectory comparison on the KITTI sequence under different estimation configurations. The left panel compares the MonoViT-only trajectory, ground-truth trajectory, original VINS trajectory, and VINS augmented with MonoViT priors. The right panel provides a broader comparison among original VINS, VINS with different initialization settings, VINS+Pose, VINS+Pose+Depth, VINS using accurate extrinsics with synchronized or unsynchronized inputs, and pure IMU integration.}
\label{fig:kitti_qualitative_trajectory}
\vspace{-3mm}
\end{figure*}

\begin{figure}[tbp]
\begin{center}
\setlength{\unitlength}{1cm}
\begin{picture}(8.5,6)
\put(0,0){\includegraphics[width=8.5cm,height=6cm]{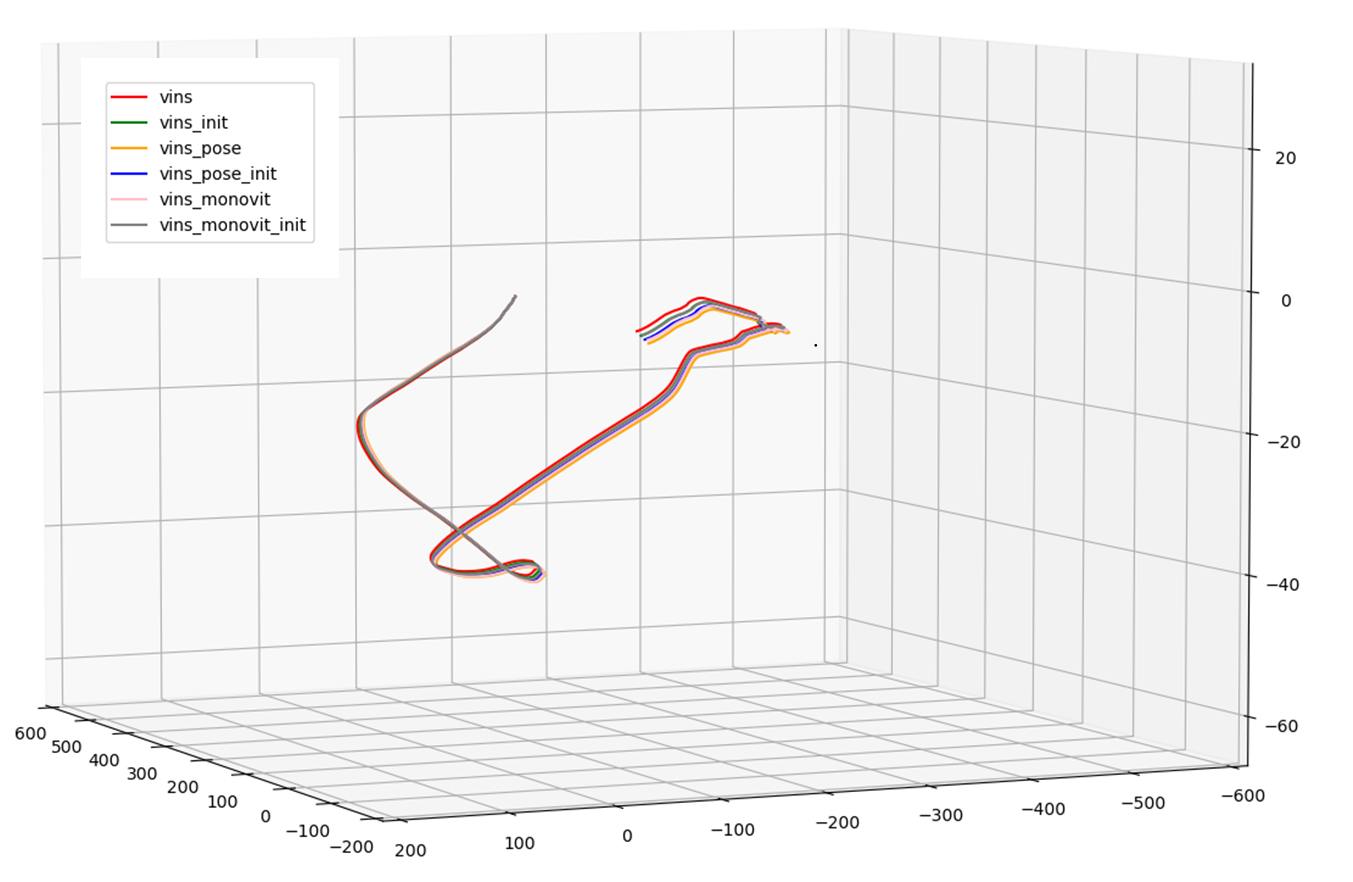}}
\put(0.43,5.12){\colorbox{white}{\parbox[c][1.75cm][c]{2.55cm}{%
\fontsize{5.5}{6.2}\selectfont
\textcolor{red}{\rule{0.38cm}{0.7pt}}~Original VINS\\
\textcolor{green!60!black}{\rule{0.38cm}{0.7pt}}~Original VINS (Init.)\\
\textcolor{orange}{\rule{0.38cm}{0.7pt}}~VINS+Pose\\
\textcolor{blue}{\rule{0.38cm}{0.7pt}}~VINS+Pose (Init.)\\
\textcolor{pink}{\rule{0.38cm}{0.7pt}}~VINS+Pose+Depth\\
\textcolor{gray}{\rule{0.38cm}{0.7pt}}~VINS+Pose+Depth (Init.)
}}}
\end{picture}
\end{center}
\vspace{-4 mm}
\caption{Enlarged trajectory comparison of Original VINS, VINS+Pose, and
VINS+Pose+Depth on KITTI, each evaluated without and with an initial extrinsic guess.}
\label{fig:kitti_qualitative_trajectory2}
\vspace{-3mm}
\end{figure}

Fig.~\ref{fig:qualitative_trajectory_comparison1} provides a qualitative comparison of the trajectories estimated by standalone MonoViT, baseline VINS, and VINS augmented with the MonoViT prior. The black curve denotes the ground-truth trajectory, while the blue, red, and green curves correspond to MonoViT, VINS, and VINS+Pose, respectively. The left panel shows the trajectories in three-dimensional space, and the right panel presents a near top-down view to highlight the horizontal drift pattern.
The standalone MonoViT trajectory approximately follows the main motion structure of the sequence, but substantial drift is visible in both the horizontal and vertical directions. In the top-down view, the blue trajectory reproduces several major turns but gradually deviates from the reference path, particularly along the long central segment and near the final part of the sequence. The 3D view further reveals a pronounced vertical deviation toward the end, confirming that the accumulated error is not limited to planar translation. This behavior is consistent with the previously observed absolute-error curves and reflects the scale ambiguity and long-term drift of monocular learned pose estimation.
The VINS and VINS+Pose trajectories remain much closer to each other than to the standalone MonoViT result. The red and green curves almost overlap throughout most of the sequence, indicating that the MonoViT prior introduces only a limited change to the trajectory estimated by the VINS backend. This observation is consistent with the previous frame-to-frame and absolute-error analyses, where the VINS and VINS+Pose configurations exhibited highly similar error profiles.
Although the two VINS-based trajectories preserve a smooth and consistent global motion pattern, they still show a systematic deviation from the ground truth in several long-range segments. In particular, the red and green curves follow a similar path that is spatially displaced from the black reference trajectory. The close agreement between VINS and VINS+Pose therefore does not necessarily indicate high accuracy; instead, it suggests that both methods converge to a similar trajectory solution dominated by the original visual reprojection and IMU-preintegration constraints.
Overall, this qualitative comparison supports two conclusions. First, visual-inertial fusion substantially reduces the severe drift observed in standalone MonoViT, especially in the vertical direction. Second, the current MonoViT integration does not visibly improve the baseline VINS trajectory, since the red and green curves remain almost identical. These results suggest that a learned pose prior must be accompanied by explicit scale alignment, uncertainty-aware weighting, or confidence-based measurement selection before it can provide a stronger correction to the VINS backend.

Fig.~\ref{fig:kitti_qualitative_trajectory} and Fig.~\ref{fig:kitti_qualitative_trajectory2}  compares the trajectory estimation results of MonoViT, the original VINS, and VINS augmented with learned pose and depth priors from multiple viewpoints. First, the trajectory produced by MonoViT alone shows that the learned monocular pose network can recover the approximate shape of the vehicle motion. However, noticeable accumulated drift remains during long-distance motion, large turns, and near the end of the sequence. Although its trajectory follows a trend similar to the ground truth in several local segments, the global scale and loop structure are not preserved consistently. This suggests that the MonoViT pose output is more suitable as a local motion prior than as a standalone long-term odometry estimate.
In contrast, the trajectories of the original VINS, VINS+Pose, and VINS+Pose+Depth largely overlap along most parts of the sequence. In particular, the enlarged 3D comparison shows only minor differences between the results obtained with and without an initial extrinsic guess, while the different VINS variants preserve nearly identical global trajectory shapes. This observation is consistent with the extrinsic results reported in Table~\ref{tab:kitti_vins_pose_depth_extrinsic}: providing an initial guess does not significantly change the final VINS solution, and introducing the MonoViT pose prior does not lead to an obvious overall improvement or degradation in the trajectory. This indicates that, under the current sequence and weighting configuration, the visual reprojection constraints and IMU preintegration constraints in the VINS backend still dominate the final optimization, while the learned pose prior has only a limited influence on the estimation result.
After further incorporating MonoViT depth information, the VINS+Pose+Depth trajectory still largely overlaps with those of the original VINS and VINS+Pose, without showing a clear and consistent improvement in trajectory accuracy. This is also consistent with the observations from the preceding extrinsic-calibration table: the depth prior does not noticeably improve the rotational extrinsic estimate and may introduce inconsistency into the translational extrinsic estimate under some settings without an initial guess. This result may indicate that monocular depth without explicit scale calibration and uncertainty modeling is difficult to convert into a more effective optimization signal than the original geometric constraints in the VINS backend.
The figure also compares synchronized and unsynchronized results under accurate extrinsic calibration. Visible differences remain between the two trajectories, indicating that sensor temporal alignment can still significantly affect visual-inertial estimation even when accurate camera--IMU extrinsic parameters are used. Therefore, accurate spatial calibration cannot replace reliable temporal synchronization. The pure IMU integration trajectory exhibits the most severe long-term drift, further confirming that IMU noise and bias accumulate over time and must be continuously corrected by visual observations.
Overall, the visualization supports the following conclusions. Original VINS is relatively stable with respect to the initial extrinsic guess on the evaluated KITTI sequence. Introducing the MonoViT pose prior does not substantially alter the VINS trajectory solution, and adding the depth diagnostic does not produce a stable visual improvement. Because VINS+Pose+Depth is retained as a secondary diagnostic rather than part of the full controlled matrix, we restrict its interpretation to the reported calibration and qualitative trajectory evidence. Quantitative cross-method conclusions use Tables~\ref{tab:unified_common_protocol} and~\ref{tab:cross_sequence_pose_ape}.

For completeness, the following component-wise diagnostics localize when and
along which state dimensions the errors develop on Recording~9. They compare
MonoViT, 10~Hz and 100~Hz IMU propagation, VINS+Pose--F, and VINS+Pose--O. The
100~Hz IMU result remains qualitative and is not included in the quantitative
ranking. All cross-method rankings use the common-frame results in
Table~\ref{tab:unified_common_protocol}.

As shown in Fig.~\ref{fig:imu_sampling_rate_trajectory_comparison}, the GPS and ground-truth trajectories nearly overlap and provide a consistent reference for the route geometry. The 100~Hz IMU trajectory generally follows the principal planar route more closely than the 10~Hz result, particularly around several turning segments. However, both inertial trajectories exhibit visible deviations in the three-dimensional view, demonstrating that increasing the sampling rate reduces but does not eliminate accumulated inertial drift. MonoViT recovers the overall route shape but exhibits noticeable deformation and vertical drift. Among the fused configurations, VINS+Pose with fixed accurate extrinsics remains considerably closer to the reference trajectory than the online-updated configuration. Allowing the accurate extrinsics to change produces a substantial global deformation, especially along the long return segment and in the vertical direction.

\begin{figure*}[tbp]
\begin{center}
\includegraphics[width=16cm, height=14cm]{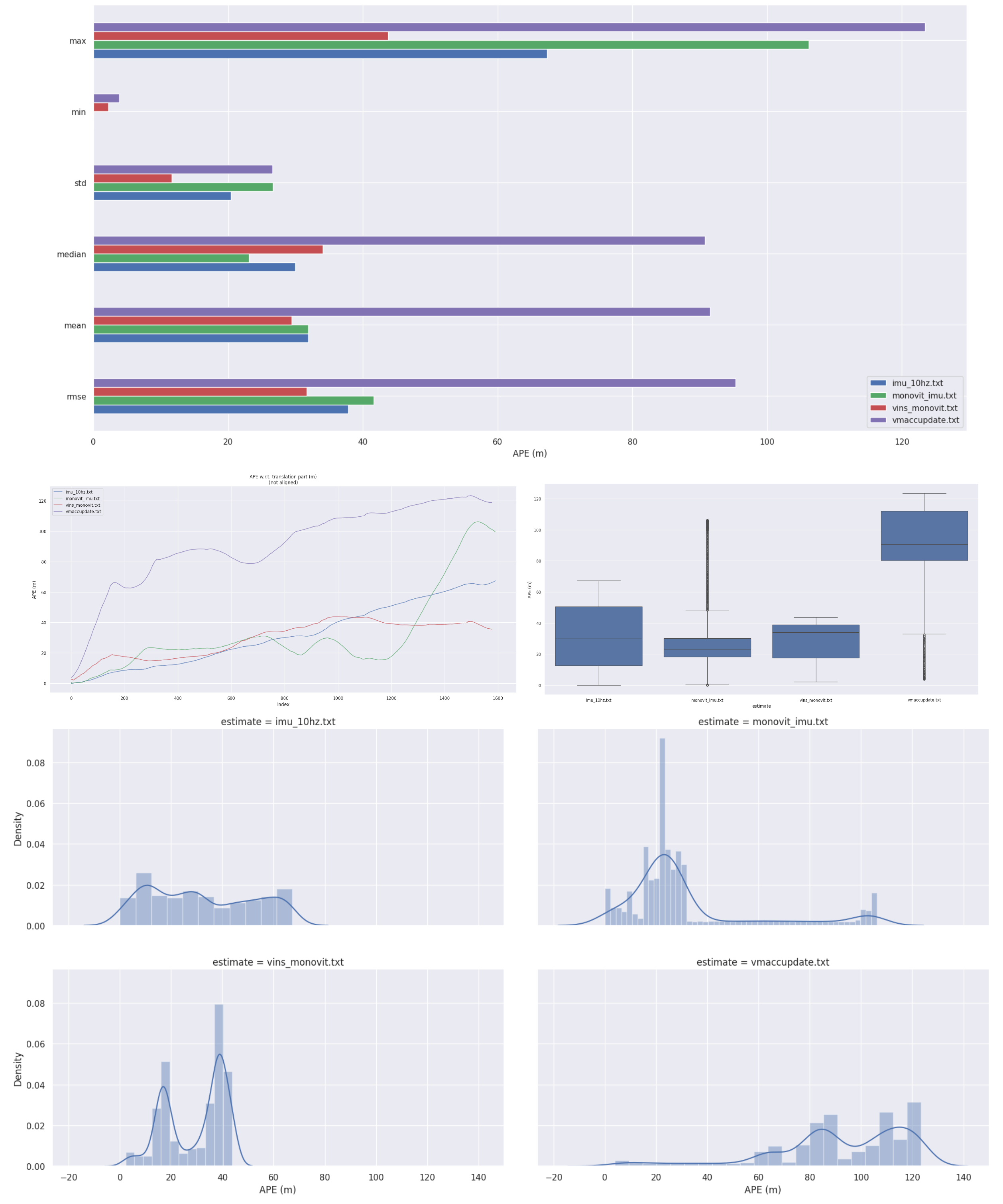}
\end{center}
\vspace{-4 mm}
\caption{Unified translation APE comparison on Recording~9. The figure presents summary statistics, frame-wise APE, box plots, and error distributions for 10~Hz IMU integration, MonoViT, VINS+Pose with fixed accurate extrinsics, and VINS+Pose with online-updated extrinsics.}
\label{fig:unified_translation_ape}
\vspace{-3mm}
\end{figure*}

The translation APE comparison in Fig.~\ref{fig:unified_translation_ape} provides a clearer quantitative distinction. VINS+Pose with fixed accurate extrinsics achieves the lowest overall translation RMSE, approximately \(32~\mathrm{m}\), and the smallest maximum error, approximately \(44~\mathrm{m}\). Its error remains relatively bounded throughout the sequence, although it is not uniformly the best at every frame. The 10~Hz IMU produces an RMSE of approximately \(38~\mathrm{m}\) and a maximum error of approximately \(67~\mathrm{m}\), reflecting gradually accumulated inertial drift.

MonoViT has a relatively low median translation error, approximately \(23~\mathrm{m}\), indicating good performance over a substantial portion of the sequence. However, its error distribution has a long upper tail, and its APE increases sharply near the end of the recording, reaching more than \(100~\mathrm{m}\). Consequently, its RMSE is higher than that of the fixed-extrinsic fusion result. This demonstrates that a favorable median error can coexist with severe late-sequence drift.
The online-updated VINS+Pose configuration exhibits the worst global translation performance. Its mean and median APE are both around \(90~\mathrm{m}\), its RMSE is approximately \(95~\mathrm{m}\), and its maximum error exceeds \(120~\mathrm{m}\). The time-series curve shows that the error rises rapidly during the early part of the sequence and remains high thereafter. This result confirms that releasing an already accurate extrinsic calibration introduces a strong systematic global bias rather than improving the trajectory estimate.

\begin{figure*}[tbp]
\begin{center}
\includegraphics[width=16cm, height=14cm]{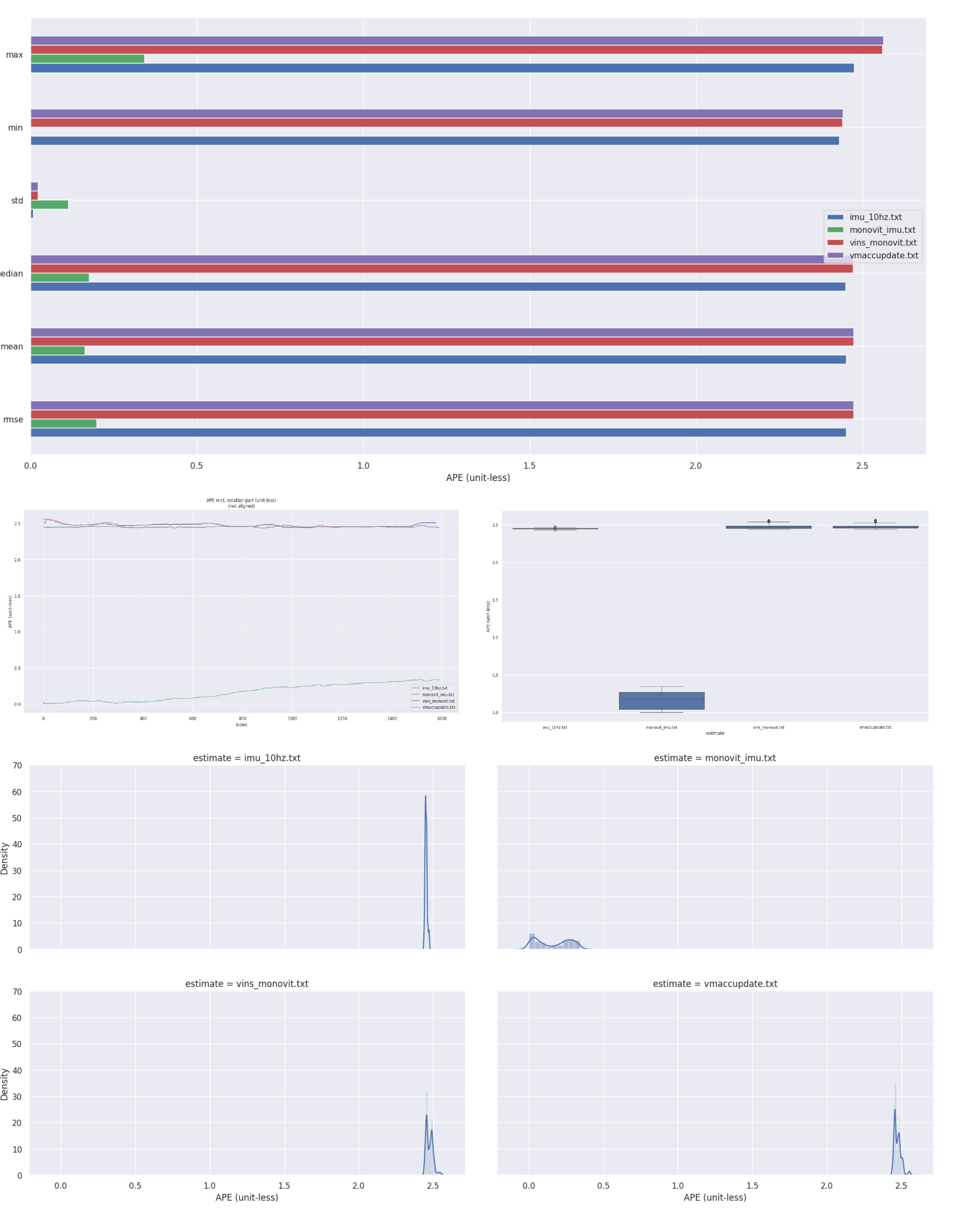}
\end{center}
\vspace{-4 mm}
\caption{Non-aligned rotation-APE diagnostic on Recording~9. The shared offset of the IMU and VINS-based curves is treated as a coordinate-convention diagnostic; common-frame geodesic rotation results are reported in Table~\ref{tab:unified_common_protocol}.}
\label{fig:unified_rotation_ape}
\vspace{-3mm}
\end{figure*}

Fig.~\ref{fig:unified_rotation_ape} retains a non-aligned rotation diagnostic. MonoViT remains below approximately \(0.35\), whereas the 10~Hz IMU and the two VINS+Pose configurations remain close to approximately \(2.45\)--\(2.55\). The shared offset is therefore treated as a coordinate-convention diagnostic rather than as the primary absolute-orientation ranking. After conversion to the common frame, the geodesic rotation results in Table~\ref{tab:unified_common_protocol} supersede this diagnostic comparison.

\begin{figure*}[tbp]
\begin{center}
\includegraphics[width=16cm, height=14cm]{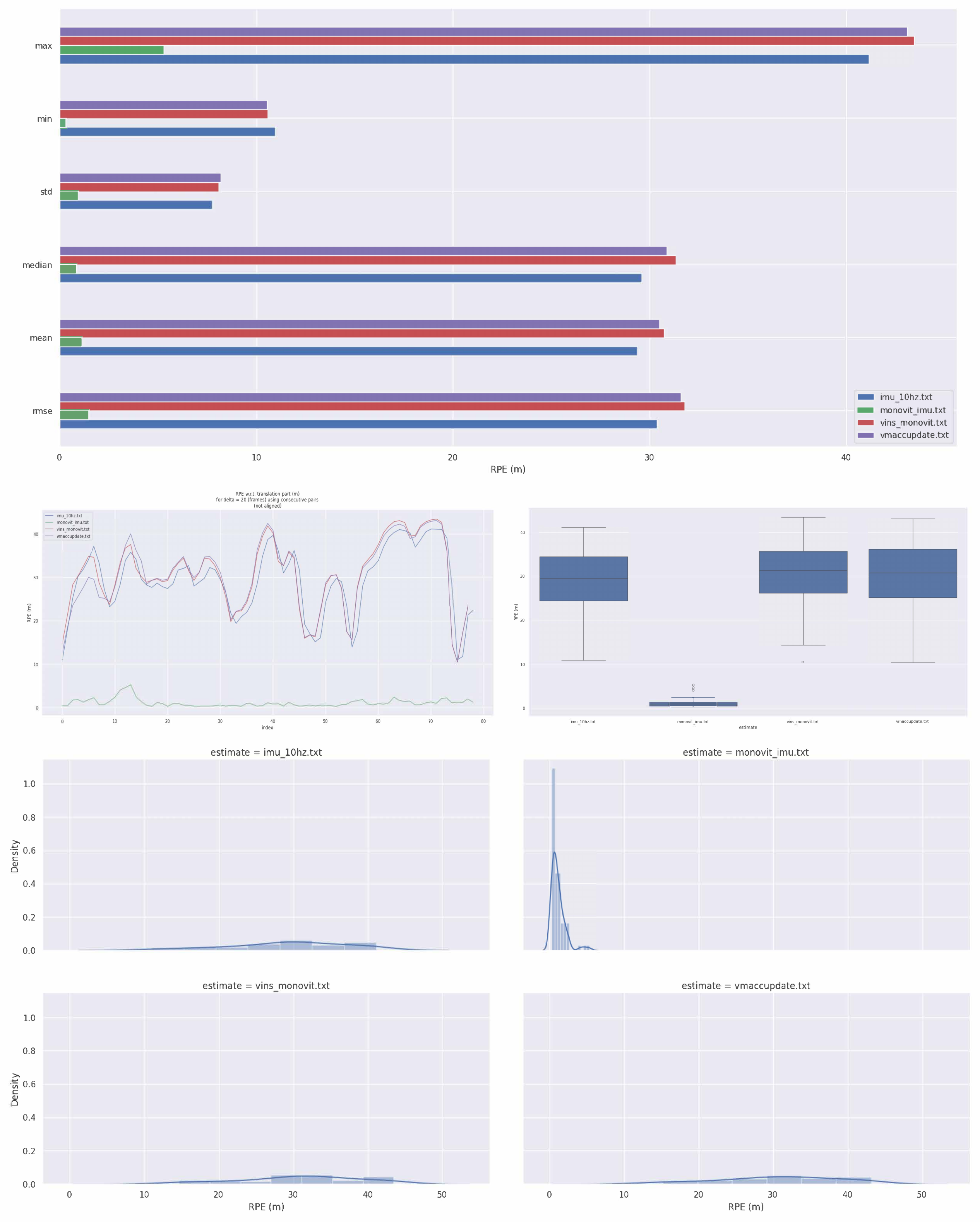}
\end{center}
\vspace{-4 mm}
\caption{Unified translation RPE comparison on Recording~9 using 20-frame relative-motion intervals. MonoViT achieves the lowest local translation error, whereas the 10~Hz IMU and both VINS+Pose configurations exhibit similar and substantially larger RPE distributions.}
\label{fig:unified_translation_rpe}
\vspace{-3mm}
\end{figure*}

\begin{figure*}[tbp]
\begin{center}
\includegraphics[width=16cm, height=14cm]{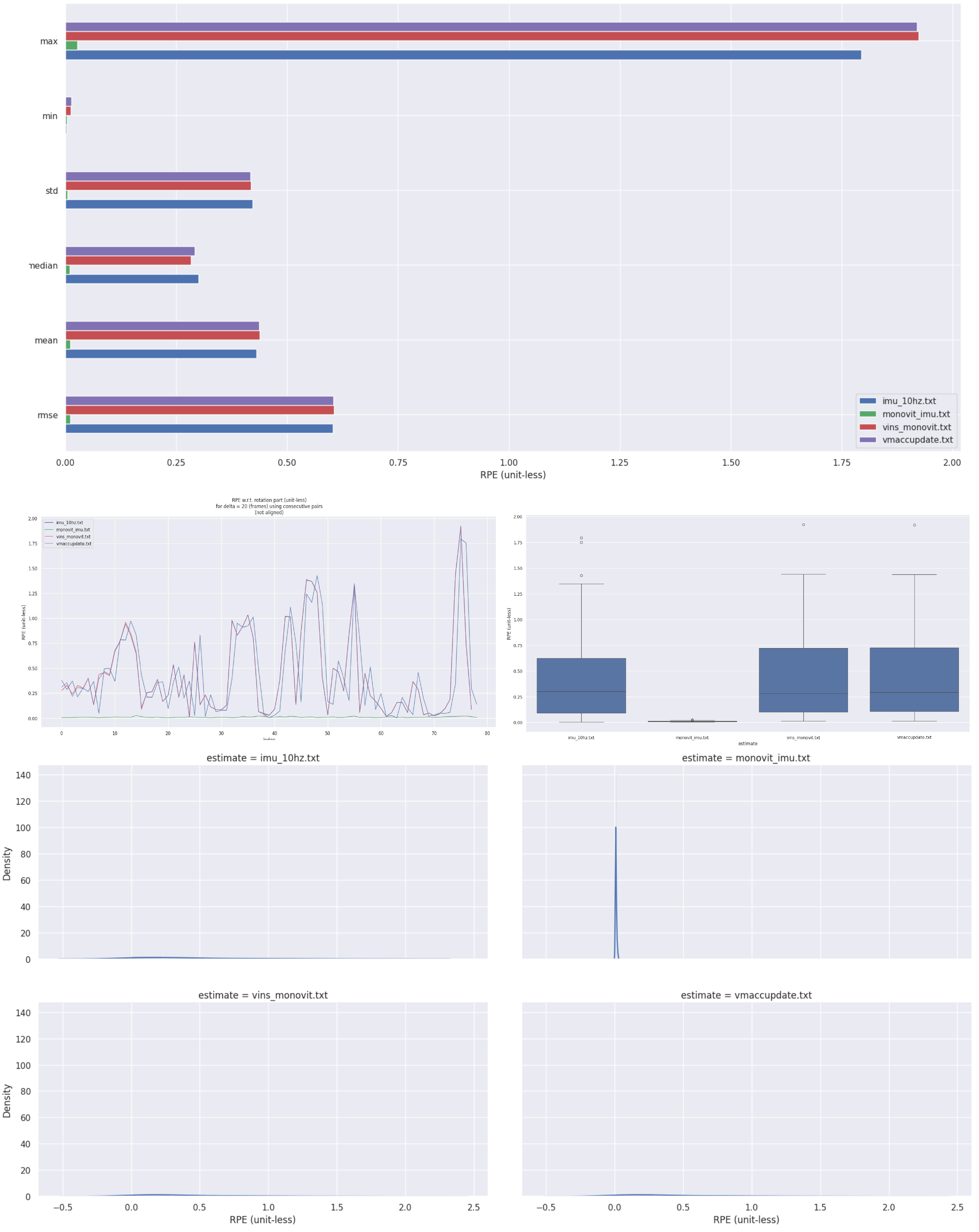}
\end{center}
\vspace{-4 mm}
\caption{Unified rotation RPE comparison on Recording~9 using 20-frame relative-motion intervals. MonoViT provides the smallest local rotation error, while the 10~Hz IMU and the fixed- and online-updated VINS+Pose configurations show similar distributions and several intermittent peaks.}
\label{fig:unified_rotation_rpe}
\vspace{-3mm}
\end{figure*}

The translation RPE results in Fig.~\ref{fig:unified_translation_rpe} reveal a different ranking from translation APE. MonoViT achieves by far the lowest 20-frame translation RPE, with most errors concentrated near \(1~\mathrm{m}\) and a maximum of approximately \(5~\mathrm{m}\). This confirms that MonoViT provides smooth and locally consistent relative-motion estimates. However, its much larger late-sequence APE shows that small relative errors and residual scale inconsistencies accumulate over time.
The 10~Hz IMU and both VINS+Pose configurations produce similar translation RPE distributions, with mean errors of approximately \(29\)--\(31~\mathrm{m}\) and maxima above \(40~\mathrm{m}\). In particular, the fixed- and online-updated extrinsic configurations have almost overlapping RPE curves despite their substantially different translation APE. This is an important result: online extrinsic updating does not cause an immediate frame-to-frame tracking failure, but instead introduces a slowly accumulated global trajectory bias.
A similar pattern appears in the rotation RPE comparison in Fig.~\ref{fig:unified_rotation_rpe}. MonoViT again produces very small relative rotation errors, while the 10~Hz IMU and both fused configurations have comparable mean, median, and RMSE values and share several large local peaks. The similarity between the fixed- and online-updated configurations confirms that changing the extrinsic-update strategy has limited influence on short-term rotational consistency, even though it strongly affects global translation accuracy.
Together with the common-protocol table, these diagnostics show that low RPE can coexist with large APE and that releasing an accurate extrinsic can change global translation far more than local relative motion. The 10/100~Hz comparison is retained only as a diagnostic configuration comparison because the 100~Hz trajectory is not included in the same quantitative matrix.

\subsection{Secondary Diagnostic: VINS+Pose+Depth Configuration}
\label{subsec:pose_depth_diagnostic}

Table~\ref{tab:kitti_vins_pose_depth_extrinsic} compares the camera--IMU extrinsic parameters estimated by the original VINS, VINS augmented with a learned pose prior, and VINS augmented with both pose and depth priors on the KITTI sequence. Each configuration is evaluated with and without an initial extrinsic guess.
For the original VINS, the final estimates obtained with and without an initial guess are highly consistent. With an initial guess, the final rotation is \([-90.3596^\circ,-0.2262^\circ,-90.8867^\circ]\), and the final translation is \([1.5359,-0.2686,0.0257]~\mathrm{m}\). Without an initial guess, VINS converges to \([-90.3725^\circ,-0.2321^\circ,-90.8503^\circ]\) and \([1.5333,-0.2555,0.0251]~\mathrm{m}\). The differences between the two settings are below \(0.037^\circ\) for all Euler-angle components and approximately \(0.0133~\mathrm{m}\) in translation magnitude. This indicates that the original VINS converges to a similar extrinsic solution regardless of whether an initial guess is provided on this sequence.
A similar initialization behavior is observed for VINS+Pose. With and without an initial guess, the final rotations are \([-90.3518^\circ,-0.2218^\circ,-90.9299^\circ]\) and \([-90.3452^\circ,-0.2183^\circ,-90.9754^\circ]\), respectively. The corresponding translations are \([1.5377,-0.2763,0.0284]~\mathrm{m}\) and \([1.5396,-0.2832,0.0301]~\mathrm{m}\). The translation difference between the two settings is approximately \(0.0074~\mathrm{m}\). Therefore, introducing the learned pose prior does not increase the sensitivity of VINS to the initial extrinsic guess.
However, the learned pose prior does not produce a clear improvement over the original VINS calibration. The VINS+Pose results remain close to the baseline VINS estimates, with changes below approximately \(0.13^\circ\) in the reported Euler-angle components and below \(0.03~\mathrm{m}\) in translation. This suggests that the learned pose prior mainly preserves the original VINS solution rather than providing an additional correction toward the reported ground-truth extrinsic parameters.
When both pose and depth priors are introduced, the rotational estimates remain relatively stable. With an initial guess, VINS+Pose+Depth converges to \([-90.3487^\circ,-0.2196^\circ,-90.8739^\circ]\), while the no-initial-guess setting produces \([-90.3410^\circ,-0.2149^\circ,-90.9315^\circ]\). These values remain close to those of the original VINS and VINS+Pose configurations. Thus, adding the depth prior does not substantially alter the rotational calibration result.
In contrast, the translation estimates show less consistent behavior after adding depth. With an initial guess, the final translation is \([1.5285,-0.2814,-0.0247]~\mathrm{m}\). Without an initial guess, the reported estimate becomes \([1.5319,0.2902,-0.0252]~\mathrm{m}\). The large change and sign reversal in the \(Y\)-translation component are not observed in the original VINS or VINS+Pose results. This indicates that the depth prior may introduce instability into the translational extrinsic estimation when no reliable initial guess is available. The result may be related to scale inconsistency in monocular depth, insufficient uncertainty weighting, or a coordinate-conversion inconsistency in the depth integration.
The comparison with the reported ground truth also reveals a systematic discrepancy in translation. The ground-truth translation is \([1.1440,-0.2464,0.7261]~\mathrm{m}\), whereas most final estimates are concentrated around \(X=1.53~\mathrm{m}\) and \(Z\approx0.03~\mathrm{m}\) or \(-0.025~\mathrm{m}\). Consequently, the learned pose and depth priors do not reduce the reported translation discrepancy. Before interpreting this difference as calibration error, the transformation direction, coordinate convention, and Euler-angle convention must be verified to ensure that the ground-truth and estimated extrinsics represent the same transformation, such as \(T_{\mathrm{cam2}}^{\mathrm{imu}}\) rather than its inverse.

Overall, the table supports three conclusions. First, the original VINS and VINS+Pose are relatively insensitive to the availability of an initial extrinsic guess on this KITTI sequence. Second, the learned pose prior causes only minor changes and does not yield a consistent improvement in camera--IMU extrinsic estimation. Third, adding the depth prior does not improve the rotational estimate and may destabilize translation estimation, particularly without an initial guess. These results indicate that learned depth should be incorporated with explicit scale alignment, coordinate consistency, and uncertainty modeling before being used as a reliable extrinsic-calibration constraint.

\begin{table*}[t]
\centering
\resizebox{\textwidth}{!}{
\begin{tabular}{lllcccccc}
\toprule
\textbf{Configuration} 
& \textbf{Initialization} 
& \textbf{Type} 
& \textbf{$X^\circ$} 
& \textbf{$Y^\circ$} 
& \textbf{$Z^\circ$} 
& \textbf{$X$ (m)} 
& \textbf{$Y$ (m)} 
& \textbf{$Z$ (m)} \\
\midrule

Ground truth 
& -- 
& Reference 
& -90.48729137 
& -0.38888106 
& -90.06237834 
& 1.14397089 
& -0.24640917 
& 0.726095982 \\

\midrule

\multirow{3}{*}{Original VINS}
& \multirow{2}{*}{With initial guess}
& Initial 
& -89 
& -1.3 
& -92 
& 1.1 
& -0.33 
& -0.76 \\

& 
& Final 
& -90.35956866 
& -0.22624389 
& -90.88670221 
& 1.53586 
& -0.26856 
& 0.025663 \\

& Without initial guess
& Final 
& -90.3724879 
& -0.23212652 
& -90.85027338 
& 1.53325 
& -0.25550 
& 0.025118 \\

\midrule

\multirow{3}{*}{VINS + Pose}
& \multirow{2}{*}{With initial guess}
& Initial 
& -89 
& -1.3 
& -92 
& 1.1 
& -0.33 
& -0.76 \\

& 
& Final 
& -90.35184219 
& -0.22184828 
& -90.92993365 
& 1.53766 
& -0.27628 
& 0.028382 \\

& Without initial guess
& Final 
& -90.34519584 
& -0.21832207 
& -90.97539381 
& 1.53964 
& -0.28315 
& 0.030097 \\

\midrule

\multirow{3}{*}{VINS + Pose + Depth}
& \multirow{2}{*}{With initial guess}
& Initial 
& -89 
& -1.3 
& -92 
& 1.1 
& -0.33 
& -0.76 \\

& 
& Final 
& -90.34870003 
& -0.21964571 
& -90.87385117 
& 1.52853 
& -0.28137 
& -0.02474 \\

& Without initial guess
& Final 
& -90.34095122 
& -0.21494607 
& -90.93152175 
& 1.53186 
& 0.29016 
& -0.02524 \\

\bottomrule
\end{tabular}
}
\caption{Camera--IMU extrinsic estimation on KITTI under different VINS configurations. The table compares the reported ground-truth extrinsic parameters with the final estimates obtained using the original VINS, VINS with a learned pose prior, and VINS with both learned pose and depth priors. Each configuration is evaluated with and without an initial extrinsic guess.}
\label{tab:kitti_vins_pose_depth_extrinsic}
\end{table*}

\textbf{Section-level finding.}
VINS+Pose outperforming standalone MonoViT or IMU demonstrates the
value of the visual--inertial backend, not necessarily the incremental value of
the learned prior. Under the controlled comparisons, VINS+Pose preserves the
behavior of Original VINS but does not consistently reduce trajectory or
extrinsic error. The higher-weight control further distinguishes an
underweighted prior from a redundant or state-inconsistent prior. The
VINS+Pose+Depth results remain secondary diagnostic evidence and are not generalized
beyond the evaluated configuration.

The interaction with the extrinsic strategy motivates a broader analysis of
initialization, perturbation, and online calibration behavior.

\section{Calibration and Initialization Robustness}
\label{sec:calibration_initialization_robustness}

This section evaluates how covariance, initial extrinsics, fixed versus online
updating, perturbations, and sequence conditions affect both calibration and
trajectory estimates. Euler components are retained for diagnostics, whereas
cross-configuration conclusions use common-frame $\mathrm{SO}(3)$ and
translation-norm errors.
Section~\ref{sec:learned_pose_incremental_effect} treats the extrinsic strategy
as a controlled factor for evaluating the incremental value of the learned pose
prior. This section instead analyzes calibration dynamics, initialization
dependence, and the physical interpretation of the extrinsic state itself.

\subsection{EKF Sensitivity to Covariance and Initial Extrinsics}
\label{subsec:ekf_calibration_sensitivity}
\subsubsection{Covariance-Controlled Online Update}
\begin{figure*}[tbp]
\begin{center}
\includegraphics[width=17cm, height=6cm]{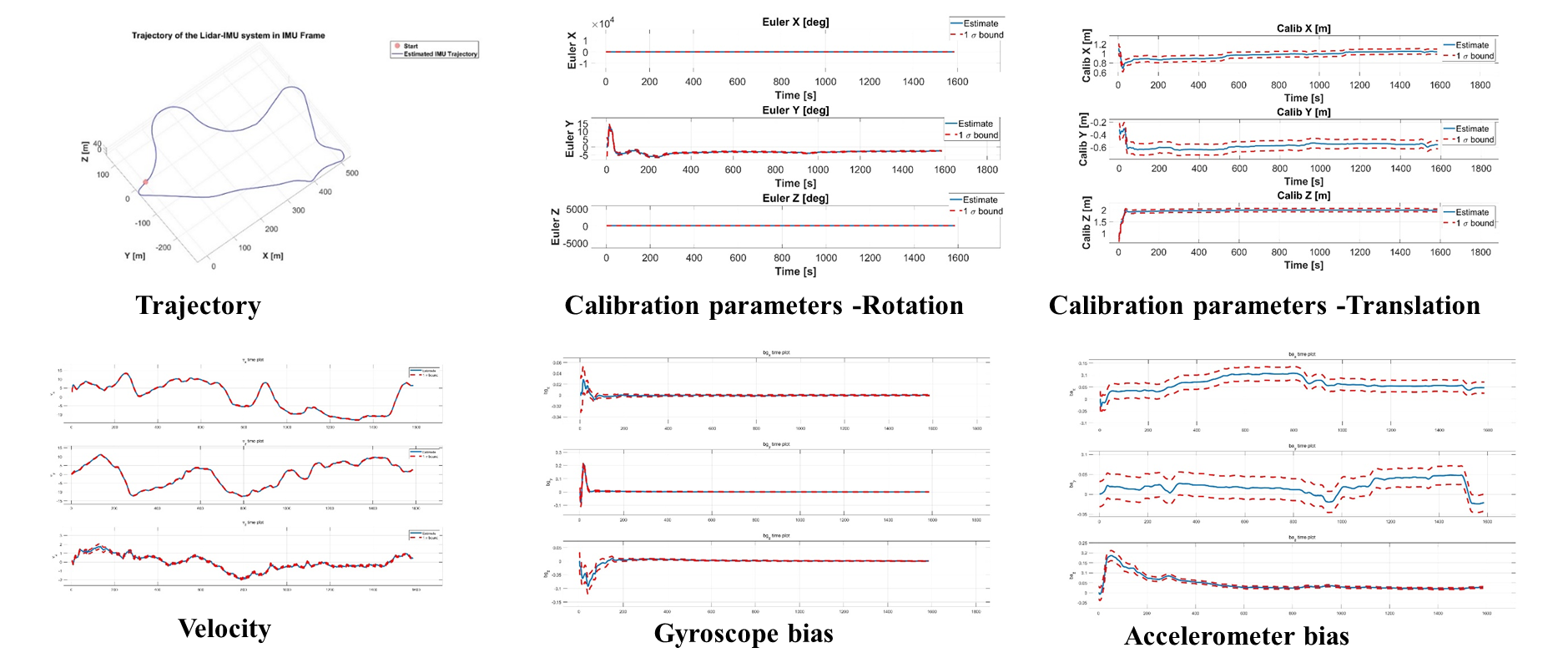}
\end{center}
\vspace{-4 mm}
\caption{EKF state estimation on KITTI camera-IMU data. The standalone EKF is evaluated on a KITTI sequence and estimates the IMU trajectory, camera-IMU extrinsic rotation and translation, velocity, gyroscope bias, and accelerometer bias.}
\label{Camera IMU EKF1}
\vspace{-3mm}
\end{figure*}

\begin{figure*}[tbp]
\begin{center}
\includegraphics[width=17cm, height=4cm]{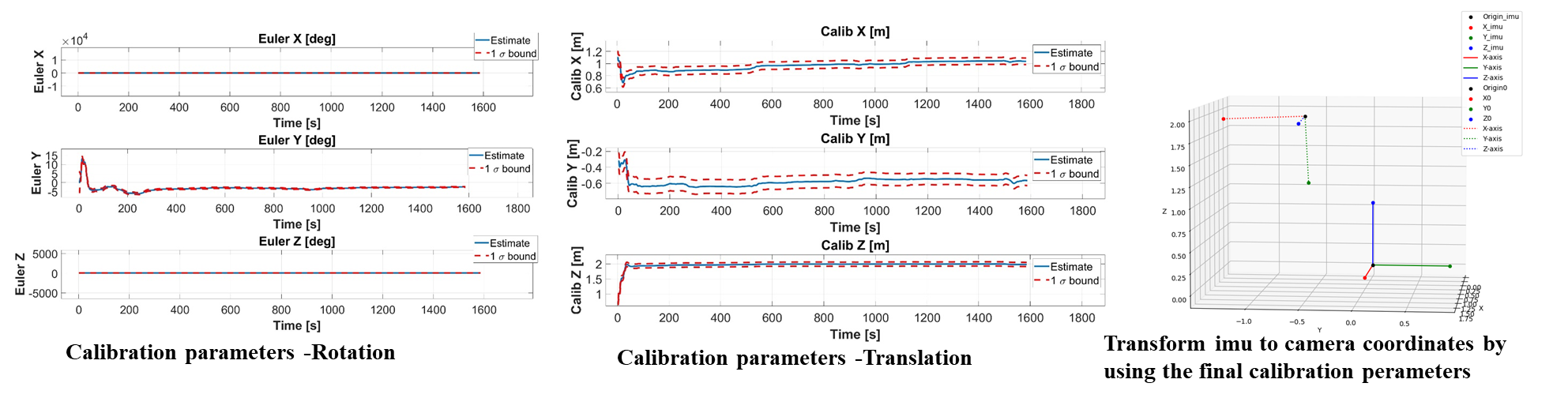}
\end{center}
\vspace{-4 mm}
\caption{Initial and final camera-IMU extrinsic calibration estimated by EKF. The comparison shows that the EKF updates both the rotation and translation components of the camera-IMU extrinsic parameters during filtering.}
\label{Camera IMU EKF2}
\vspace{-3mm}
\end{figure*}

After validating the standalone EKF implementation, we further evaluate it on KITTI camera--IMU data. 
As shown in Fig.~\ref{Camera IMU EKF1}, the EKF estimator is able to run on KITTI sequences and jointly estimate the IMU trajectory, camera--IMU extrinsic parameters, velocity, gyroscope bias, and accelerometer bias. 
The calibration parameters exhibit an initial transient stage and then gradually become more stable, indicating that the filter can perform online state estimation and camera--IMU calibration under visual--inertial measurements.
Fig.~\ref{Camera IMU EKF2} further compares the initial and final camera--IMU extrinsic parameters estimated by the EKF. 
The initial extrinsic rotation is 
$[-173.0082^\circ, 89.4662^\circ, 82.6218^\circ]$, 
and the initial translation is 
$[1.1022, -0.3191, 0.7461]~\mathrm{m}$. 
After EKF estimation, the final extrinsic rotation becomes 
$[-167.5140^\circ, 93.5621^\circ, 79.8748^\circ]$, 
and the final translation becomes 
$[1.0379, -0.5637, 1.9832]~\mathrm{m}$. 
Compared with the initial value, the Euler angles change by approximately 
$[5.49^\circ, 4.10^\circ, -2.75^\circ]$, 
while the translation changes by approximately 
$[-0.0643, -0.2446, 1.2371]~\mathrm{m}$. 
These changes show that the extrinsic calibration is actively updated during filtering rather than being treated as a fixed input. 
This result verifies that the camera--IMU extrinsic parameters are successfully included in the EKF state. 
However, the final calibration should not be interpreted as more accurate solely from this qualitative visualization; its accuracy needs to be evaluated quantitatively using rotation and translation errors against ground-truth calibration.

\begin{figure*}[tbp]
\begin{center}
\includegraphics[width=17cm, height=10cm]{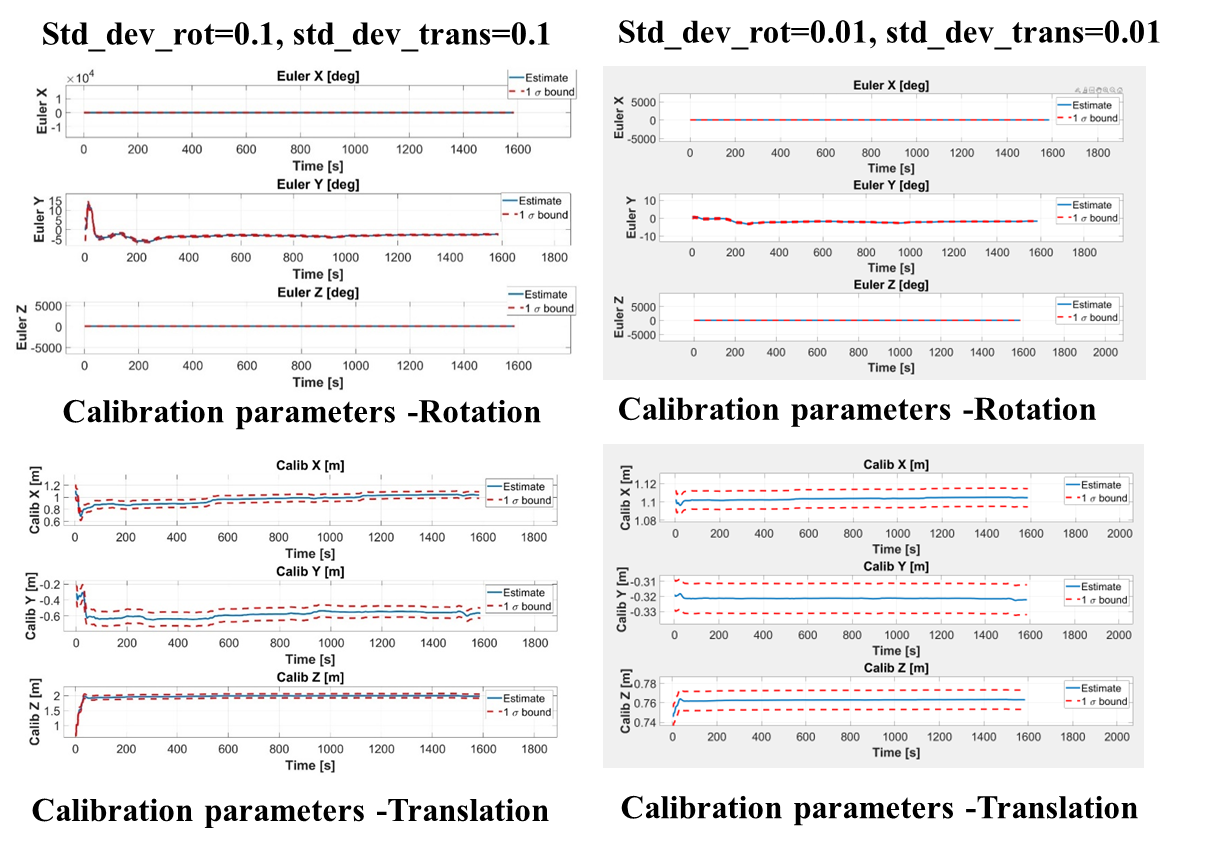}
\end{center}
\vspace{-4 mm}
\caption{Sensitivity of EKF calibration to extrinsic uncertainty settings. Different standard deviations for rotation and translation lead to different final camera-IMU calibration results.}
\label{Camera IMU EKF3}
\vspace{-3mm}
\end{figure*}

We also analyze the sensitivity of EKF-based extrinsic calibration to the assumed calibration uncertainty. 
As shown in Fig.~\ref{Camera IMU EKF3}, when the calibration uncertainty is set to 
${std\_dev\_rot}=0.1$ and ${std\_dev\_trans}=0.1$, 
the final translation is 
$[1.0379, -0.5637, 1.9832]~\mathrm{m}$, 
which differs significantly from the initial translation 
$[1.1022, -0.3191, 0.7461]~\mathrm{m}$. 
The magnitude of the translation update is approximately 
$1.26~\mathrm{m}$. 
In contrast, when the uncertainty is reduced to 
${std\_dev\_rot}=0.01$ and ${std\_dev\_trans}=0.01$, 
the final translation becomes 
$[1.1046, -0.3222, 0.7633]~\mathrm{m}$, 
which is much closer to the initial calibration. 
In this case, the translation update magnitude is only approximately 
$0.018~\mathrm{m}$. 
This comparison shows that larger standard deviations allow the filter to make more aggressive updates to the extrinsic parameters, while smaller standard deviations constrain the calibration update and keep the final estimate closer to the initial calibration.
This observation suggests that EKF-based camera--IMU calibration is highly sensitive to covariance settings. 
Therefore, careful uncertainty design is necessary when integrating MonoViT outputs into the EKF, since inaccurate or over-confident visual priors may lead to unstable extrinsic estimation and trajectory drift.

\subsubsection{Nominal Calibration Behavior}
We next evaluate EKF-based camera--IMU extrinsic estimation on the synchronized KITTI Raw sequence {2011\_09\_30\_drive\_0033}. This sequence contains 1,591 frames, corresponding to approximately 159 seconds of residential driving. The ground-truth camera0 trajectory and the camera--IMU extrinsic parameters provided by KITTI are used as references. The IMU measurements are downsampled to 10 Hz. The accelerometer noise is $0.1$, gyroscope noise is $0.01$, accelerometer random walk is $0.001$, gyroscope random walk is $1.0 \times 10^{-4}$, and both camera rotation and translation observation noises are $0.01$.

\begin{figure*}[tbp]
\begin{center}
\includegraphics[width=17cm, height=7cm]{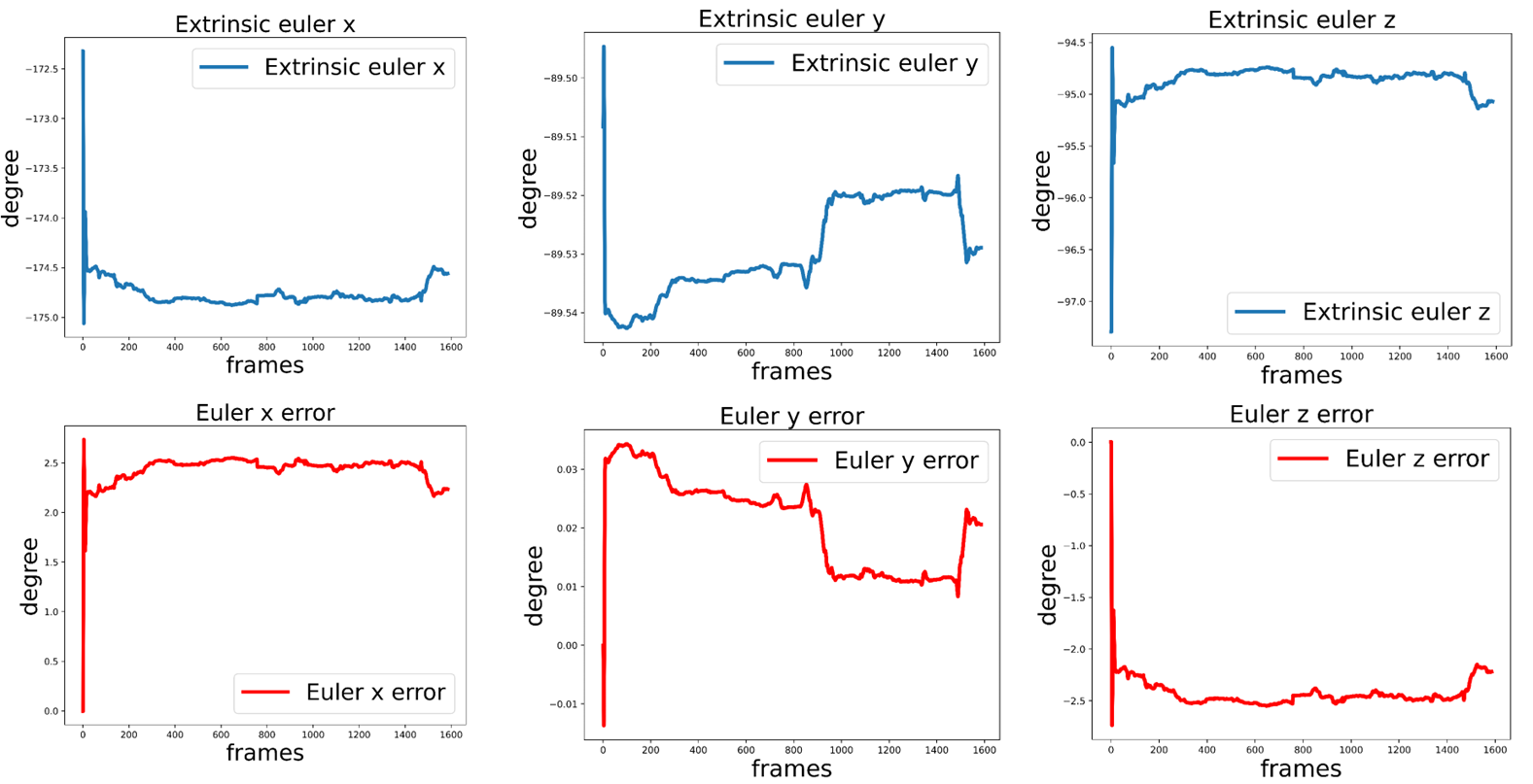}
\end{center}
\vspace{-4 mm}
\caption{EKF-based camera-IMU extrinsic estimation on KITTI. The top row shows the estimated extrinsic Euler angles over 1,591 frames, and the bottom row shows the corresponding Euler-angle errors.}
\label{ekf_extrinsic_rotation}
\vspace{-3mm}
\end{figure*}

\begin{figure*}[tbp]
\begin{center}
\includegraphics[width=17cm, height=7cm]{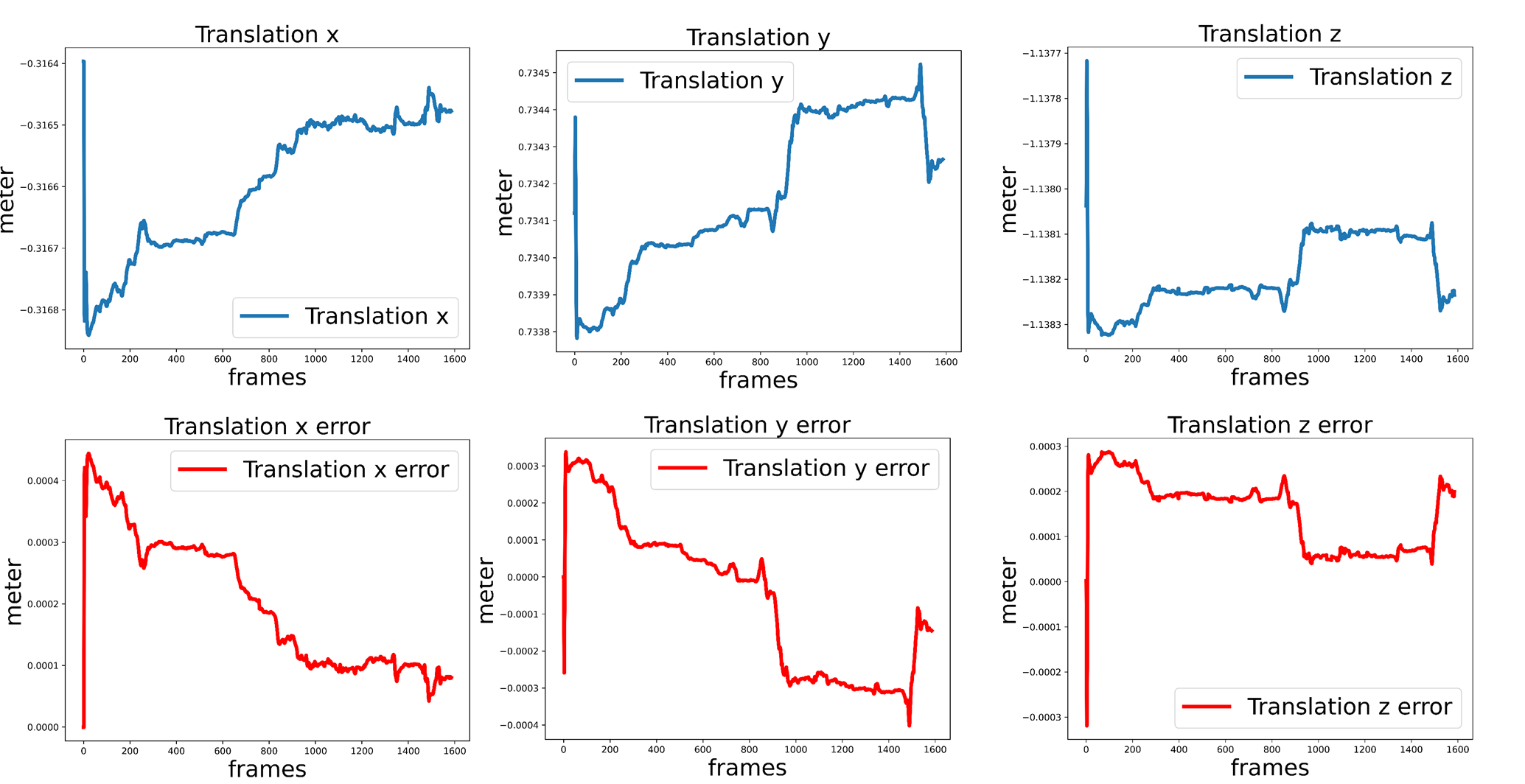}
\end{center}
\vspace{-4 mm}
\caption{Estimated camera--IMU extrinsic translation on the KITTI sequence. The top row shows the three translation components over 1,591 frames, and the bottom row shows the corresponding translation errors.}
\label{ekf_extrinsic_translation}
\vspace{-3mm}
\end{figure*}

Fig.~\ref{ekf_extrinsic_rotation} and Fig.~\ref{ekf_extrinsic_translation} show the estimated camera-IMU extrinsic parameters on the KITTI sequence. Fig.~\ref{ekf_extrinsic_rotation} presents the estimated extrinsic rotation represented by Euler angles and the corresponding Euler-angle errors, while Fig.~\ref{ekf_extrinsic_translation} shows the estimated extrinsic translation and the translation errors along the $x$, $y$, and $z$ axes. The rotation curves exhibit a clear initial adjustment stage and then gradually become stable. The initial extrinsic rotation is 
$[-172.3235^\circ, -89.5083^\circ, -97.2879^\circ]$, 
and the final estimate becomes 
$[-174.5587^\circ, -89.5289^\circ, -95.0699^\circ]$, 
corresponding to an Euler-angle change of approximately 
$[-2.235^\circ, -0.021^\circ, 2.218^\circ]$.

In contrast, the translation component remains highly stable throughout the sequence. The initial translation is 
$[-0.316397, 0.734121, -1.138036]~\mathrm{m}$, 
while the final translation is 
$[-0.316477, 0.734266, -1.138235]~\mathrm{m}$. 
The total translation update is only about 
$2.6 \times 10^{-4}~\mathrm{m}$. 
The translation errors along the three axes also remain within approximately 
$10^{-4}~\mathrm{m}$ to $4 \times 10^{-4}~\mathrm{m}$, 
indicating that the EKF applies only very limited correction to the translational extrinsic parameters under the current noise configuration.

These results suggest that, for this sequence and parameter setting, the EKF mainly adjusts the rotational component of the camera-IMU extrinsic calibration, while the translation component is already close to the reference value and remains nearly unchanged. This behavior indicates different numerical update strengths for rotation and translation in this EKF run; it does not by itself establish theoretical observability. Therefore, both rotation and translation errors should be reported separately when evaluating camera-IMU extrinsic calibration.

\begin{table*}[t]
\centering
\caption{Initial and final camera--IMU extrinsic parameters estimated by EKF.}
\label{tab:ekf_extrinsic_parameters}
\resizebox{\linewidth}{!}{
\begin{tabular}{lcccccc}
\toprule
 & Euler X (degree) & Euler Y (degree) & Euler Z (degree) 
 & Translation X (m) & Translation Y (m) & Translation Z (m) \\
\midrule
Initial extrinsic 
$\left(T_{imu}^{cam}\right)$
& $-172.32349504$ 
& $-89.50831883$ 
& $-97.28789169$ 
& $-0.31639699$ 
& $0.73412053$ 
& $-1.13803616$ \\
Final extrinsic 
$\left(T_{imu}^{cam}\right)$
& $-174.55866147$ 
& $-89.52892101$ 
& $-95.06993940$ 
& $-0.31647713$ 
& $0.73426570$ 
& $-1.13823503$ \\
\bottomrule
\end{tabular}
}
\end{table*}

\subsubsection{Large Initial-Extrinsic Perturbation}
\begin{figure*}[tbp]
\begin{center}
\includegraphics[width=17cm, height=14cm]{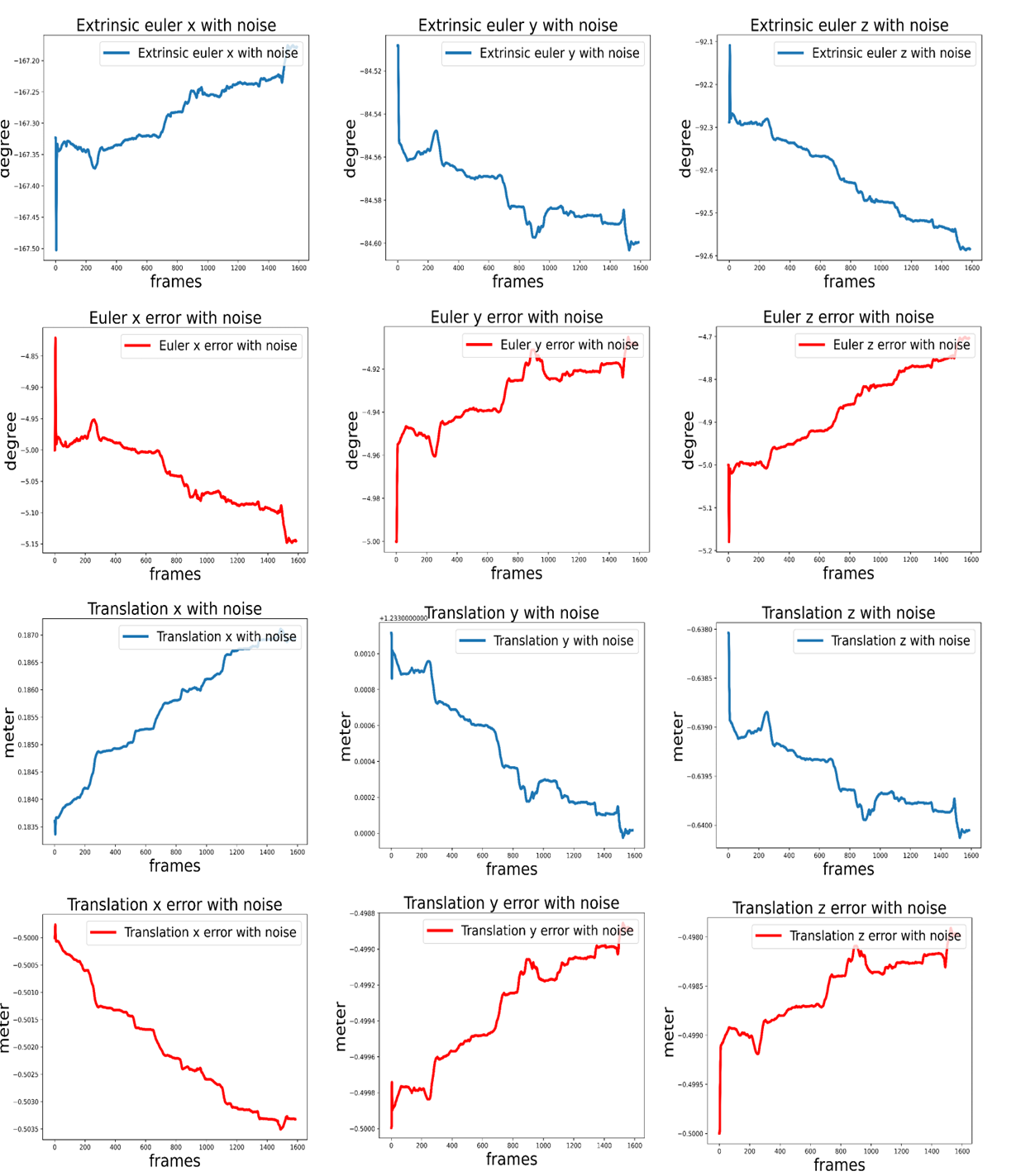}
\end{center}
\vspace{-4 mm}
\caption{EKF-based extrinsic calibration under noisy initial camera-IMU extrinsic parameters.}
\label{noise1}
\vspace{-3mm}
\end{figure*}

\begin{figure*}[tbp]
\begin{center}
\includegraphics[width=17cm, height=16cm]{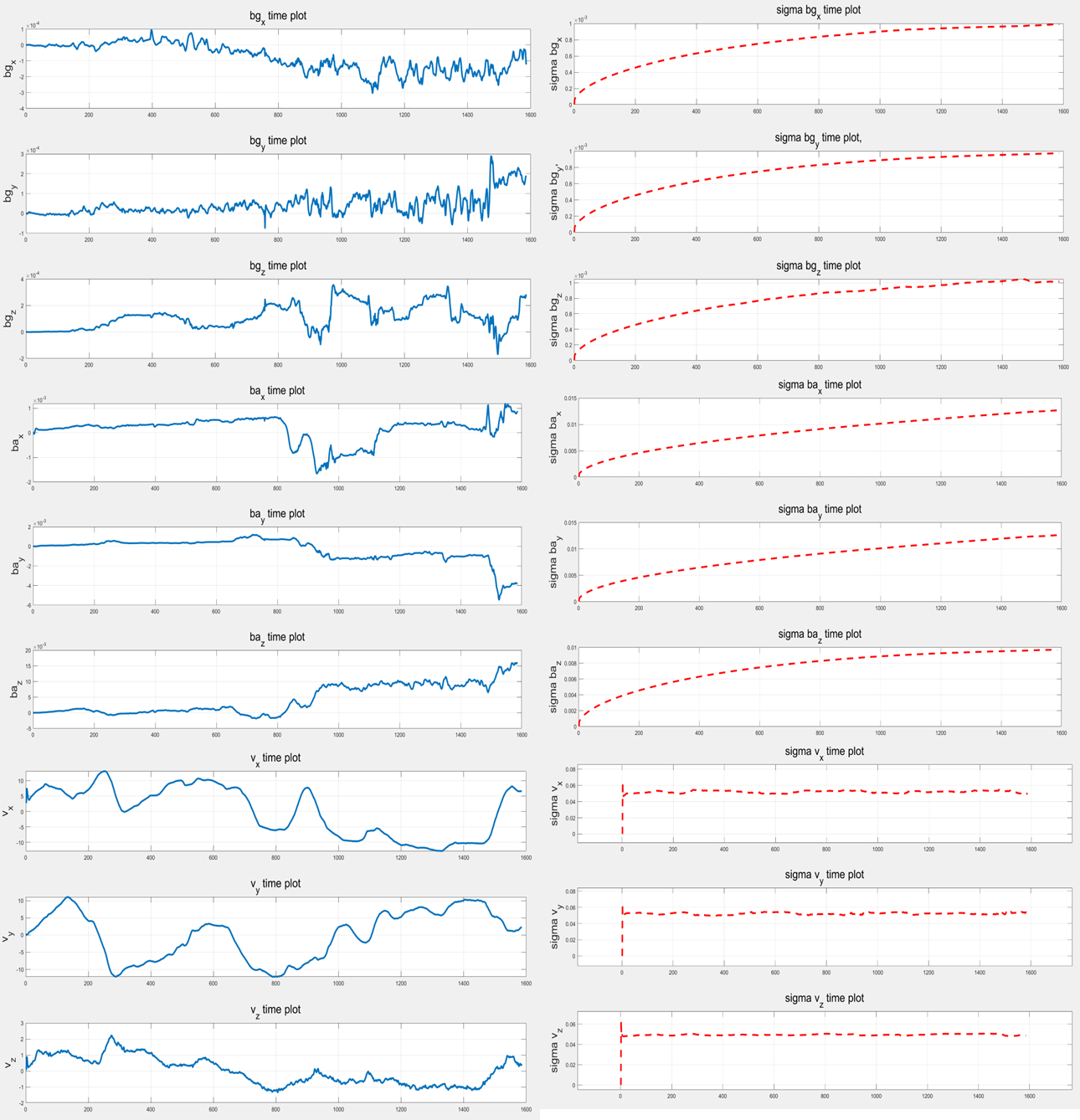}
\end{center}
\vspace{-4 mm}
\caption{State and uncertainty evolution under noisy extrinsic initialization.}
\label{noise2}
\vspace{-3mm}
\end{figure*}

We then perturb the reference camera--IMU calibration by adding $+5^\circ$ to each Euler angle and $+0.5$ m to each translation component. The reference extrinsic rotation is $[-172.3235^\circ, -89.5083^\circ, -97.2879^\circ]$, and the reference translation is $[-0.3164, 0.7341, -1.1380]$ m. After perturbation, the initial extrinsic rotation becomes $[-167.3235^\circ, -84.5083^\circ, -92.2879^\circ]$, and the initial translation becomes $[0.1836, 1.2341, -0.6380]$ m.
As shown in Fig.~\ref{noise1} and Table \ref{tab:noisy_extrinsic_sensitivity}, the EKF updates the noisy extrinsic parameters only slightly during the sequence. The final estimated rotation is $[-167.1782^\circ, -84.5996^\circ, -92.5844^\circ]$, and the final translation is $[0.1869, 1.2330, -0.6401]$ m. Compared with the noisy initial calibration, the Euler-angle update is only $[0.1453^\circ, -0.0912^\circ, -0.2965^\circ]$, and the translation update is only $[0.0033, -0.0011, -0.0020]$ m. However, compared with the reference calibration, the final estimate still has large residual deviations of approximately $[5.1453^\circ, 4.9088^\circ, 4.7035^\circ]$ in Euler angles and $[0.5033, 0.4989, 0.4980]$ m in translation.
The time-series plots further confirm this behavior. The estimated extrinsic parameters gradually evolve during filtering but remain close to the noisy initialization rather than converging back to the reference calibration. The Euler-angle errors stay around the injected $5^\circ$ perturbation level, and the translation errors remain close to the injected $0.5$ m perturbation. This indicates that, under the current visual measurement and covariance settings, the EKF performs only local correction of the extrinsic parameters and is unable to recover from a large initial calibration error.
Fig.~\ref{noise2} shows the corresponding IMU bias, velocity, and uncertainty evolution under the noisy extrinsic setting. The filter continues to propagate the inertial states and estimate gyroscope bias, accelerometer bias, and velocity, indicating that the system does not completely diverge. However, the persistent extrinsic error suggests that the mismatch caused by the noisy calibration is not sufficiently corrected through the extrinsic state itself. Instead, part of the inconsistency may be absorbed by other state variables, such as velocity and IMU bias. This result highlights that EKF-based online camera-IMU calibration is sensitive to initial extrinsic quality and uncertainty design.
Overall, this experiment suggests that accurate initial calibration remains important for stable EKF-based visual-inertial estimation. When the initial extrinsic error is large, the filter may converge to a local solution near the noisy initialization instead of recovering the correct calibration. This observation motivates the later evaluation of whether learned visual pose priors or Original VINS outputs can provide stronger correction signals for improving extrinsic recovery.

\begin{table*}[t]
\centering
\resizebox{\linewidth}{!}{
\begin{tabular}{lcccccc}
\toprule
 & 
\textbf{Euler X (degree)} &
\textbf{Euler Y (degree)} &
\textbf{Euler Z (degree)} &
\textbf{Translation X (m)} &
\textbf{Translation Y (m)} &
\textbf{Translation Z (m)} \\
 & 
\textbf{noise: $+5^\circ$} &
\textbf{noise: $+5^\circ$} &
\textbf{noise: $+5^\circ$} &
\textbf{noise: $+0.5$ m} &
\textbf{noise: $+0.5$ m} &
\textbf{noise: $+0.5$ m} \\
\midrule
Initial extrinsic $T^{cam}_{imu}$ 
& $-172.32349504$ 
& $-89.50831883$ 
& $-97.28789169$ 
& $-0.31639699$ 
& $0.73412053$ 
& $-1.13803616$ \\

Initial extrinsic $T^{cam}_{imu}$ with noise
& $-167.32349504$ 
& $-84.50831883$ 
& $-92.28789169$ 
& $0.18360301$ 
& $1.23412053$ 
& $-0.63803616$ \\

Final extrinsic $T^{cam}_{imu}$
& $-167.17817826$ 
& $-84.59956387$ 
& $-92.58437387$ 
& $0.18692406$ 
& $1.23301629$ 
& $-0.64005155$ \\
\bottomrule
\end{tabular}
}
\caption{Sensitivity of the EKF-based camera-IMU calibration to noisy initial extrinsic parameters. The reference camera-IMU extrinsic calibration is perturbed by adding $+5^\circ$ noise to each Euler angle and $+0.5$ m noise to each translation component. }
\label{tab:noisy_extrinsic_sensitivity}
\end{table*}

\subsubsection{Alternative Observation Setting and Perturbation Repeat}
\begin{figure*}[tbp]
\begin{center}
\includegraphics[width=17cm, height=14cm]{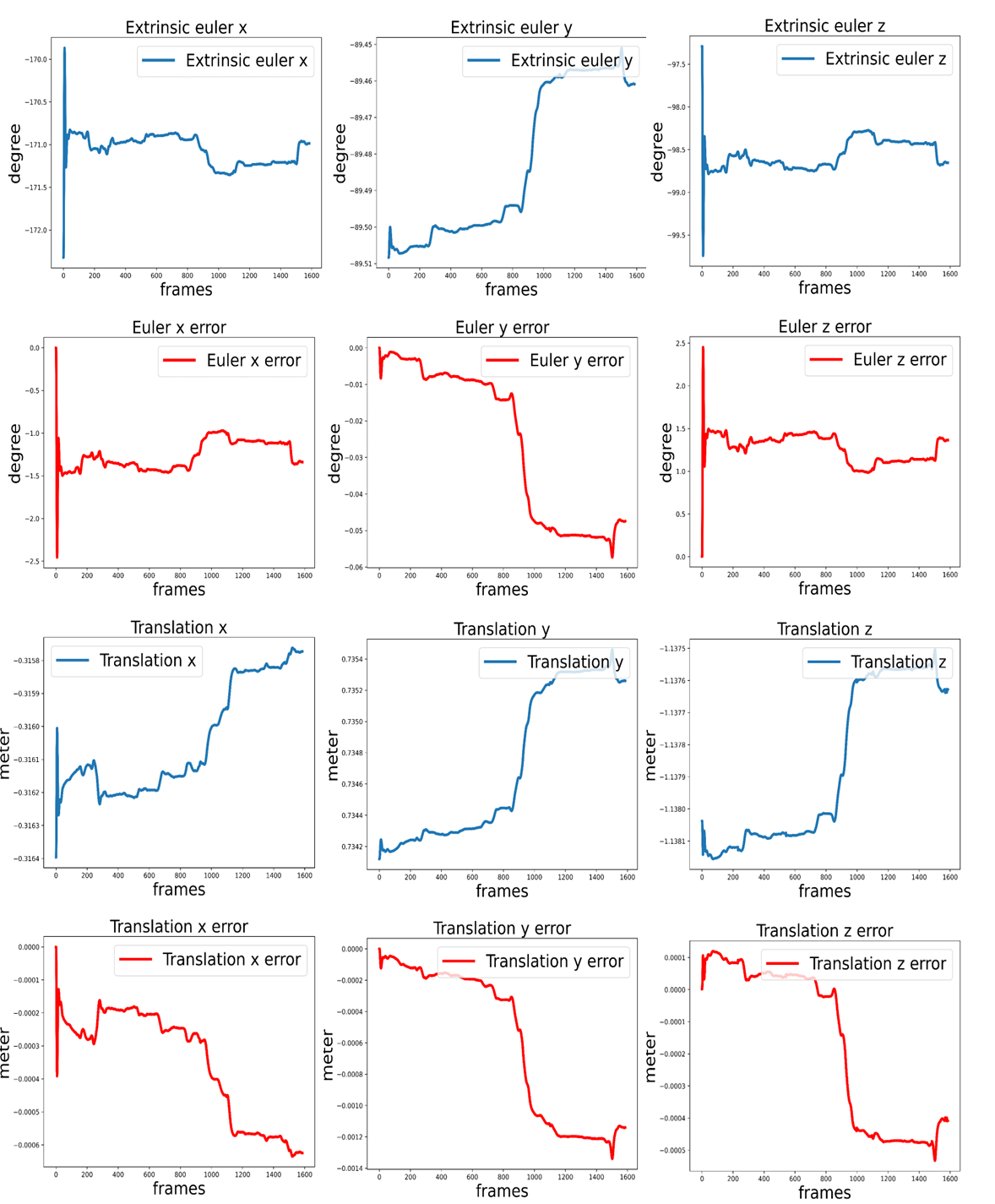}
\end{center}
\vspace{-4 mm}
\caption{EKF extrinsic estimation under a different initialization and observation setting. The estimated extrinsic Euler angles and translations are shown over 1,591 frames, together with their corresponding errors.}
\label{fig:ekf_initialization_extrinsic}
\vspace{-3mm}
\end{figure*}

To separate the preceding behavior from one observation setting, we repeat the nominal/noisy comparison on the same sequence using camera2 images, camera2 intrinsics, and camera2--IMU extrinsics. The images are resized to $640 \times 192$, the IMU is downsampled to $10$ Hz, the first five EKF parameters follow VINS-Fusion, and the camera rotation and translation observation noises are both $0.1$.
Table~\ref{tab:ekf_extrinsic_initial_0.1} summarizes the initial and final camera-IMU extrinsic parameters under this setting. The initial extrinsic rotation is $[-172.3235^\circ, -89.5083^\circ, -97.2879^\circ]$, and the initial translation is $[-0.3164, 0.7341, -1.1380]$ m. After EKF estimation, the final rotation becomes $[-170.9850^\circ, -89.4609^\circ, -98.6545^\circ]$, and the final translation becomes $[-0.3158, 0.7353, -1.1376]$ m. This corresponds to an Euler-angle change of approximately $[1.34^\circ, 0.05^\circ, -1.37^\circ]$, while the translation change is only $[0.00062, 0.00114, 0.00041]$ m, with an overall magnitude of about $1.36$ mm.
Fig.~\ref{fig:ekf_initialization_extrinsic} shows the temporal evolution of the estimated extrinsic parameters and their errors. The rotation estimates exhibit an initial transient stage and then gradually become more stable. Compared with the previous EKF setting, the rotational update follows a different direction: the $x$-axis Euler angle moves toward a less negative value, the $y$-axis angle remains nearly unchanged, and the $z$-axis angle becomes more negative. This indicates that the EKF calibration result is affected by the initialization and observation configuration. The translation estimates remain much more stable than the rotation estimates. Although small step-like changes can be observed around the middle of the sequence, the translation update remains at the millimeter level.
Compared with the earlier EKF experiment, where the rotation update mainly occurred in the $x$- and $z$-axis directions and the translation update was about $0.26$ mm, this setting produces a different rotational correction and a slightly larger but still small translation update. This comparison suggests that the EKF-based camera-IMU calibration is sensitive to the input configuration and uncertainty design. The filter can adjust the extrinsic parameters online, but its convergence behavior is not unique across different settings. Therefore, both the initialization strategy and observation noise should be carefully controlled when using EKF calibration as part of a larger VINS/MonoViT evaluation framework.

\begin{table*}[t]
\centering
\caption{Initial and final camera-IMU extrinsic parameters under a different EKF initialization setting. The experiment is conducted on the KITTI Raw sequence {2011\_09\_30\_drive\_0033}using camera2 images and IMU measurements.}
\label{tab:ekf_extrinsic_initial_0.1}
\resizebox{\linewidth}{!}{
\begin{tabular}{lcccccc}
\toprule
 & \textbf{Euler X (degree)} 
 & \textbf{Euler Y (degree)} 
 & \textbf{Euler Z (degree)} 
 & \textbf{Translation X (m)} 
 & \textbf{Translation Y (m)} 
 & \textbf{Translation Z (m)} \\
\midrule
Initial extrinsic
& $-172.32349504$
& $-89.50831883$
& $-97.28789169$
& $-0.31639699$
& $0.73412053$
& $-1.13803616$ \\

Final extrinsic
& $-170.9849859$
& $-89.46088956$
& $-98.65446204$
& $-0.31577226$
& $0.7352607$
& $-1.13762793$ \\
\bottomrule
\end{tabular}
}
\end{table*}

Under this alternative setting, we apply the same $+5^\circ$ and $+0.5$ m perturbation. As shown in Table~\ref{tab:noisy_extrinsic_robustness}, the reference initial extrinsic rotation is $[-172.3235^\circ, -89.5083^\circ, -97.2879^\circ]$, and the reference translation is $[-0.3164, 0.7341, -1.1380]$ m. The perturbed initialization is $[-167.3235^\circ, -84.5083^\circ, -92.2879^\circ]$ and $[0.1836, 1.2341, -0.6380]$ m.

\begin{figure*}[tbp]
\begin{center}
\includegraphics[width=17cm, height=14cm]{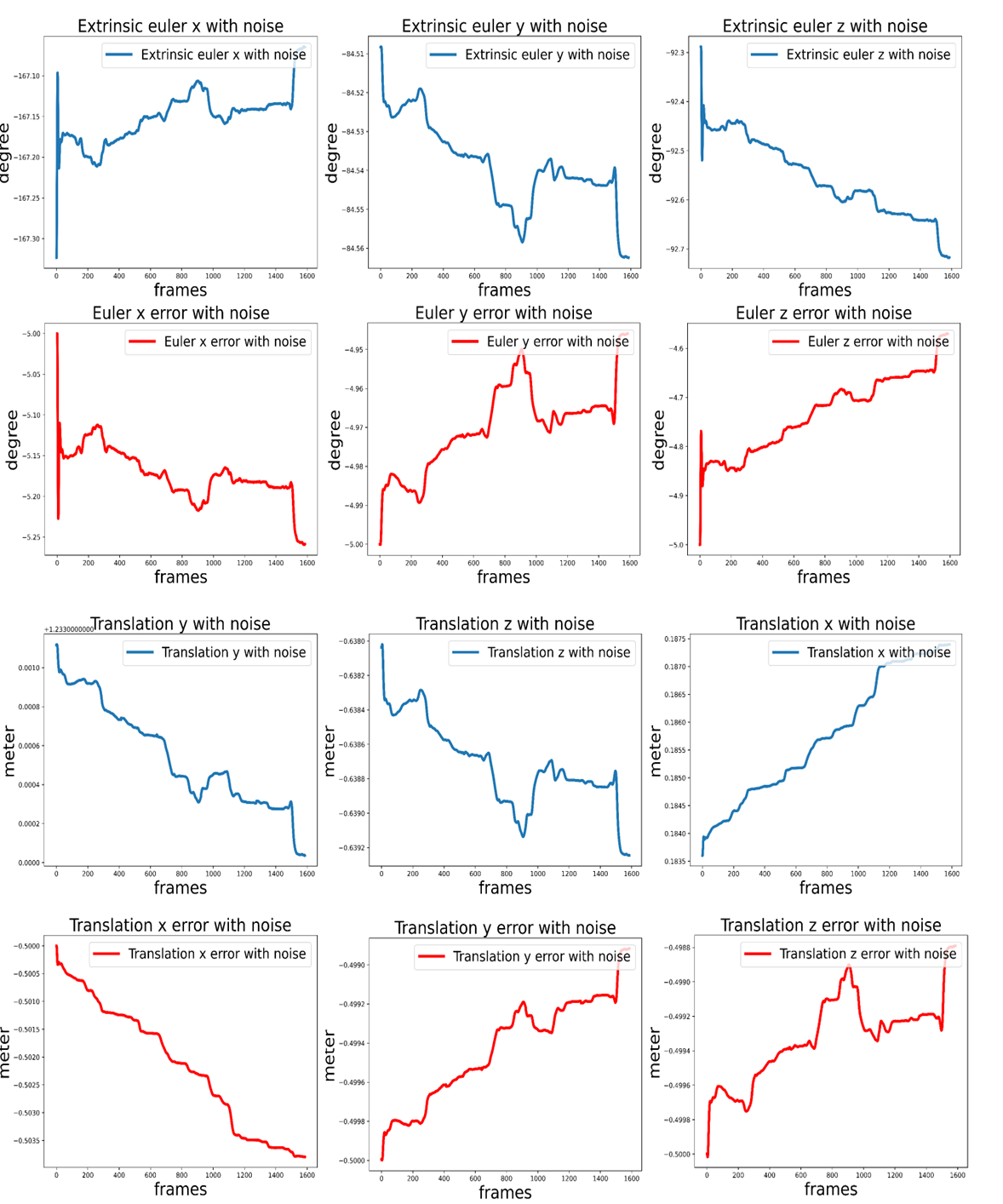}
\end{center}
\vspace{-4 mm}
\caption{EKF state and uncertainty evolution under noisy extrinsic initialization.}
\label{noise3}
\vspace{-3mm}
\end{figure*}

\begin{figure*}[tbp]
\begin{center}
\includegraphics[width=17cm, height=16cm]{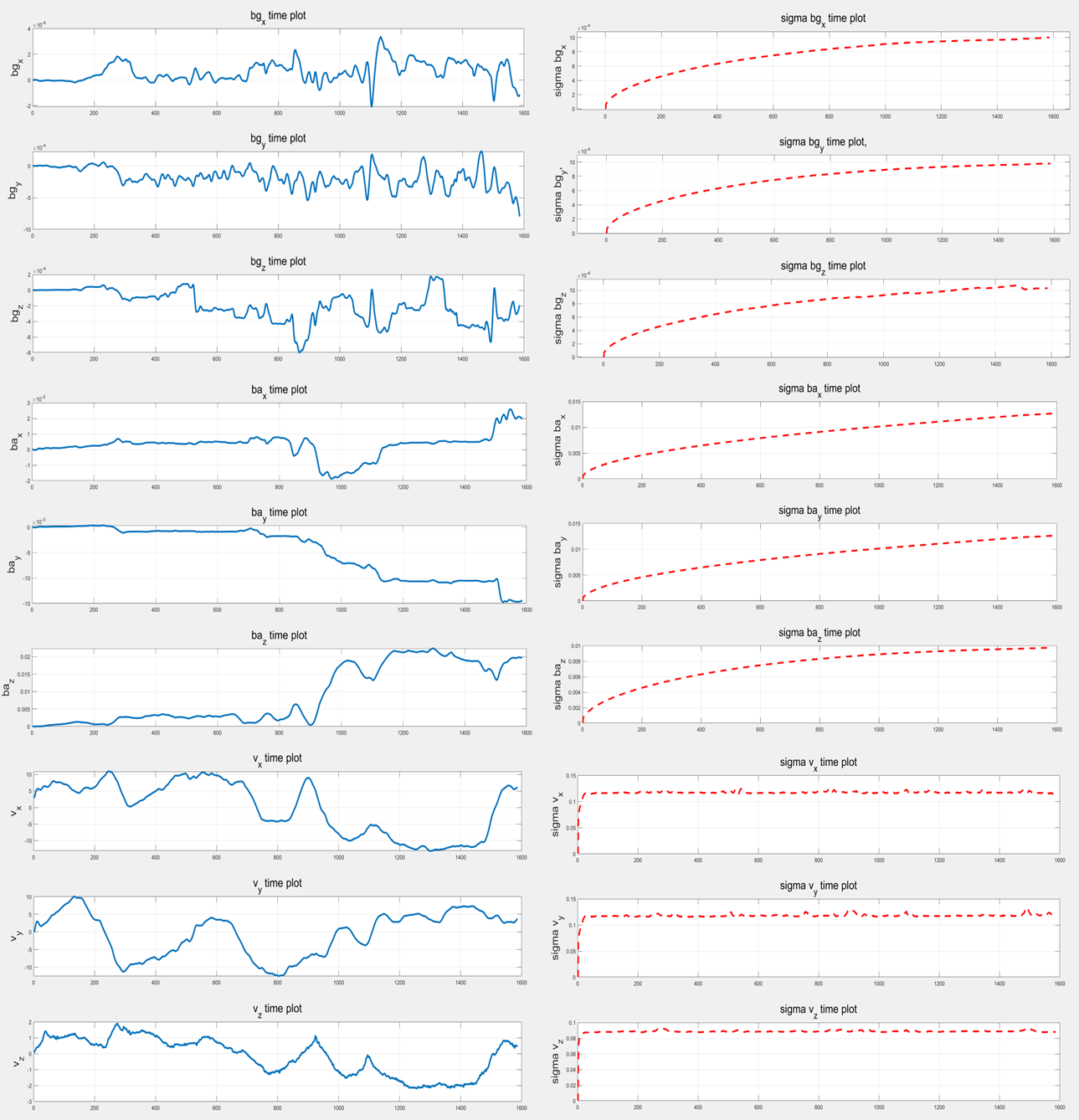}
\end{center}
\vspace{-4 mm}
\caption{Estimated camera-IMU extrinsic parameters with noisy initialization. The top rows show the estimated extrinsic Euler angles and translations, while the bottom rows show their corresponding errors.}
\label{noise4}
\vspace{-3mm}
\end{figure*}

The final EKF estimate is $[-167.0648^\circ, -84.5624^\circ, -92.7173^\circ]$ for rotation and $[0.1874, 1.2330, -0.6392]$ m for translation. Compared with the noisy initialization, the EKF only changes the Euler angles by approximately $[0.2587^\circ, -0.0541^\circ, -0.4294^\circ]$, and the translation by approximately $[0.0038, -0.0011, -0.0012]$ m. The translation update magnitude is only about $4.1$ mm. However, compared with the reference extrinsic parameters before adding noise, the final estimate still has large residual deviations of approximately $[5.2587^\circ, 4.9459^\circ, 4.5706^\circ]$ in Euler angles and approximately $[0.5038, 0.4989, 0.4988]$ m in translation.
Fig.~\ref{noise3} shows the corresponding state and uncertainty evolution. The EKF continues to estimate the gyroscope bias, accelerometer bias, and velocity, which indicates that the filter remains operational under noisy extrinsic initialization. However, the uncertainty curves and bias variations suggest that part of the inconsistency caused by the incorrect extrinsic initialization may be absorbed by the inertial states rather than fully corrected through the extrinsic calibration state.
Fig.~\ref{noise4} further shows the estimated extrinsic parameters and their errors over the sequence. The estimated rotation and translation parameters evolve smoothly, but they remain close to the noisy initialization instead of converging back to the reference calibration. The Euler-angle errors stay around the injected $5^\circ$ level, and the translation errors remain close to the injected $0.5$ m perturbation. This indicates that, under the current measurement and covariance settings, the EKF performs only local correction around the initial calibration and cannot fully recover from a large initial extrinsic error.

Compared with the previous EKF setting without added initial extrinsic noise, the difference is clear. In the noise-free setting, the EKF updated the rotation from $[-172.3235^\circ, -89.5083^\circ, -97.2879^\circ]$ to $[-170.9850^\circ, -89.4609^\circ, -98.6545^\circ]$, corresponding to a moderate rotational adjustment of about $[1.34^\circ, 0.05^\circ, -1.37^\circ]$, while the translation changed only at the millimeter level. In contrast, after adding $+5^\circ$ and $+0.5$ m perturbations, the EKF does not remove the injected error and remains near the noisy initial value. This comparison shows that EKF-based online calibration is sensitive to the initial extrinsic quality. Accurate or reasonably close initial calibration is still important; otherwise, the filter may converge to a local solution near the wrong initialization.

\begin{table*}[t]
\centering
\caption{Sensitivity analysis of EKF-based camera-IMU calibration under noisy initial extrinsic parameters. A perturbation of $+5^\circ$ is added to each Euler angle and $+0.5$ m is added to each translation component.}
\label{tab:noisy_extrinsic_robustness}
\resizebox{\linewidth}{!}{
\begin{tabular}{lcccccc}
\toprule
 & \textbf{Euler X (degree)} 
 & \textbf{Euler Y (degree)} 
 & \textbf{Euler Z (degree)} 
 & \textbf{Translation X (m)} 
 & \textbf{Translation Y (m)} 
 & \textbf{Translation Z (m)} \\
 & \textbf{noise: $+5^\circ$}
 & \textbf{noise: $+5^\circ$}
 & \textbf{noise: $+5^\circ$}
 & \textbf{noise: $+0.5$ m}
 & \textbf{noise: $+0.5$ m}
 & \textbf{noise: $+0.5$ m} \\
\midrule
Initial extrinsic
& $-172.32349504$
& $-89.50831883$
& $-97.28789169$
& $-0.31639699$
& $0.73412053$
& $-1.13803616$ \\

Initial extrinsic $T^{cam}_{imu}$ with noise
& $-167.32349504$
& $-84.50831883$
& $-92.28789169$
& $0.18360301$
& $1.23412053$
& $-0.63803616$ \\

Final extrinsic $T^{cam}_{imu}$
& $-167.06484475$
& $-84.56242678$
& $-92.71730718$
& $0.18739741$
& $1.23303677$
& $-0.63924515$ \\
\bottomrule
\end{tabular}
}
\end{table*}

\subsection{Fixed (F) versus Online (O) Extrinsics}
\label{subsec:fixed_vs_online_extrinsics}

\subsubsection{Calibration-State Drift after Accurate Initialization}
The preceding controlled experiments use accurate camera--IMU extrinsic parameters as a fixed calibration, thereby excluding online calibration drift. We now compare this setting with accurate-extrinsic initialization followed by continued online calibration. The purpose of this comparison is to determine whether releasing an already reliable calibration provides additional trajectory benefits or instead allows the backend to absorb other modeling and measurement errors into the extrinsic state.

\begin{figure*}[tbp]
\begin{center}
\includegraphics[width=17cm, height=7cm]{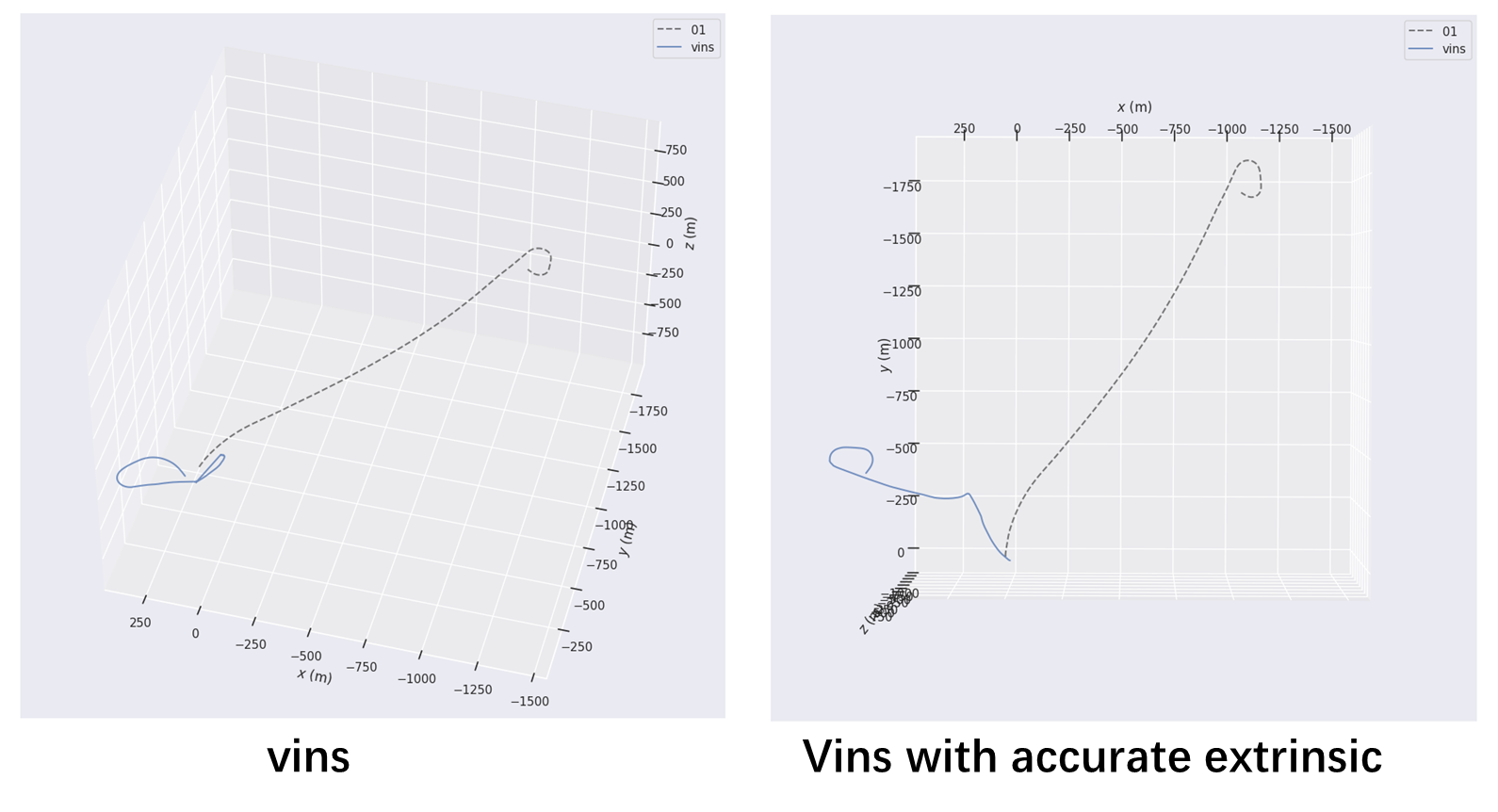}
\end{center}
\vspace{-4 mm}
\caption{VINS trajectory estimation on Recording 1 using 100 Hz IMU measurements. The two panels show the baseline VINS result and the VINS result under the accurate camera--IMU extrinsic configuration.}
\label{fig:vins_100hz_trajectory}
\vspace{-3mm}
\end{figure*}

\begin{figure*}[tbp]
\begin{center}
\includegraphics[width=17cm, height=7cm]{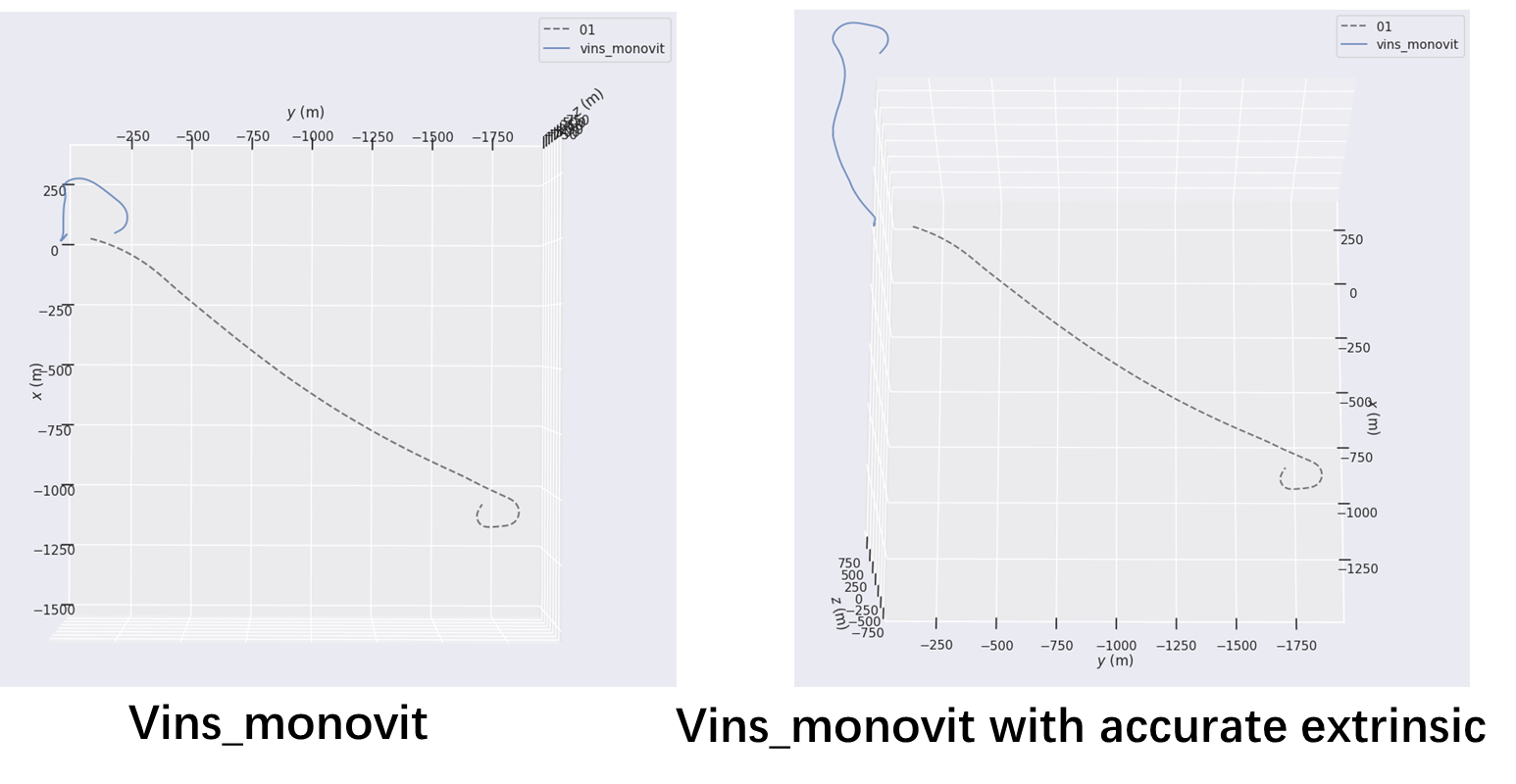}
\end{center}
\vspace{-4 mm}
\caption{VINS+Pose trajectory estimation on Recording 1 using 100 Hz IMU measurements. The two panels compare the standard VINS+Pose configuration with the corresponding accurate-extrinsic configuration.}
\label{fig:vins_monovit_100hz_trajectory}
\vspace{-3mm}
\end{figure*}

In the fixed-extrinsic setting, the accurate rotation and translation remain constant and are not included in backend optimization. In the online-updated setting, the same accurate calibration is used as the initial value, but the extrinsic state is jointly optimized with the trajectory, velocity, and IMU biases. The trajectory comparisons in Figs.~\ref{fig:vins_100hz_trajectory} and \ref{fig:vins_monovit_100hz_trajectory} show that providing accurate extrinsics does not by itself eliminate the trajectory discrepancy in Recording~1. We therefore examine whether continued online updating improves the calibration or instead introduces additional drift.

\begin{figure*}[tbp]
\begin{center}
\includegraphics[width=17cm, height=12cm]{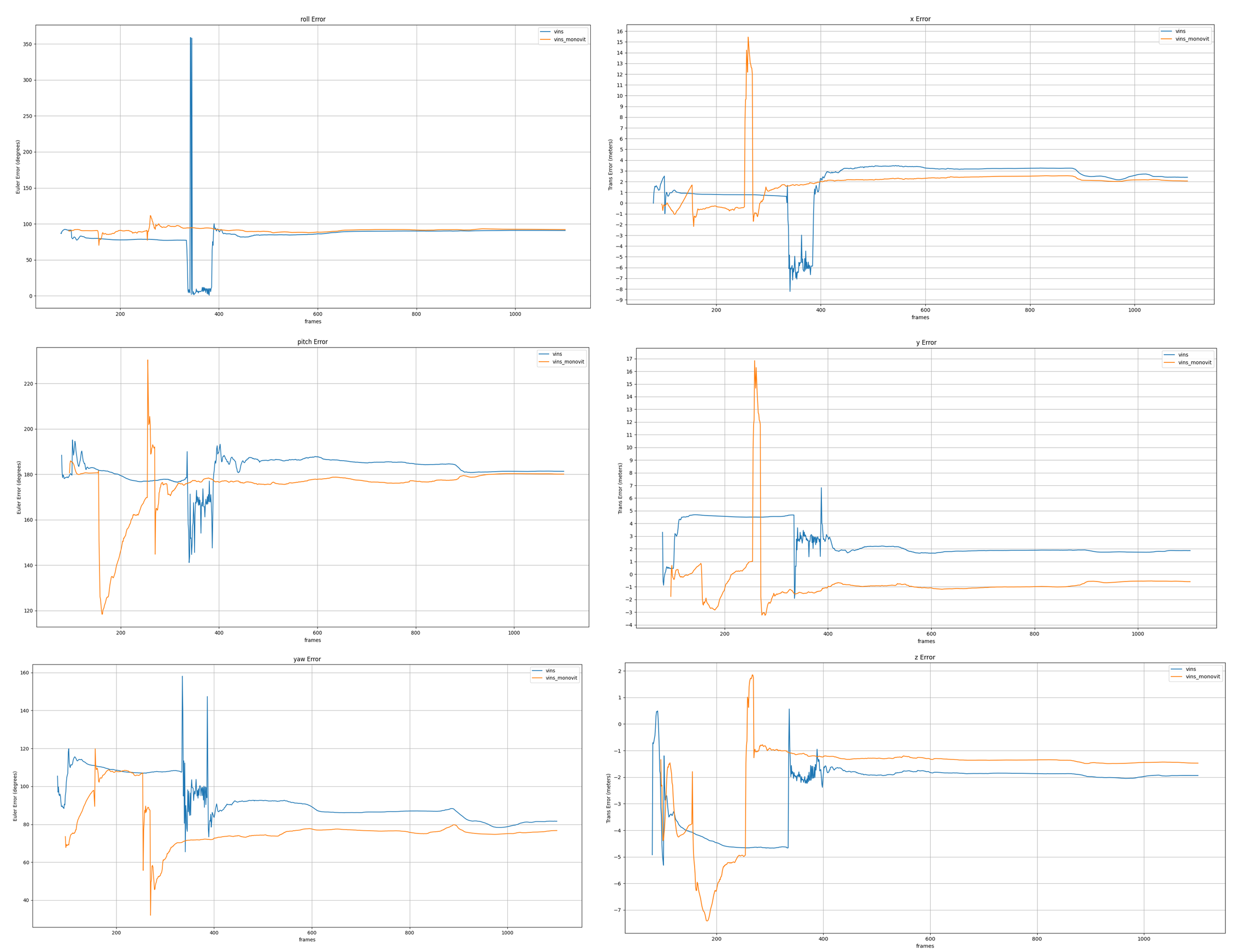}
\end{center}
\vspace{-4 mm}
\caption{Temporal evolution of camera--IMU extrinsic errors under online calibration with accurate initialization. The plots compare VINS and VINS+Pose at 100 Hz in terms of Roll, Pitch, Yaw, and the x-, y-, and z-translation errors.}
\label{fig:online_updated_extrinsic_errors}
\vspace{-3mm}
\end{figure*}

Fig.~\ref{fig:online_updated_extrinsic_errors} shows the temporal evolution of the estimated extrinsic errors for VINS and VINS+Pose when the accurately initialized extrinsics are allowed to participate in online optimization. Both methods exhibit a pronounced transient stage during approximately the first 400 frames. The rotational components contain several abrupt variations, including a large Roll discontinuity for VINS and strong early Pitch and Yaw variations for VINS+Pose. After this transient period, the curves become substantially smoother and converge to relatively stable plateaus. However, the stabilized errors remain non-zero, indicating that the backend does not preserve the accurate initial calibration.

The rotational results further show that the MonoViT prior does not consistently improve online calibration. The two methods eventually produce similar Roll estimates, whereas persistent differences remain in Pitch and particularly Yaw. VINS+Pose reduces some early fluctuations but introduces larger transients in other intervals. Thus, the learned pose prior changes the calibration trajectory without consistently moving the estimated rotation closer to the reference extrinsic.

A similar pattern is observed in translation. After initialization, the \(x\)-, \(y\)-, and \(z\)-translation errors gradually settle, but they converge to biased values rather than zero. The two methods exhibit comparable \(x\)-direction behavior, whereas more visible differences remain in the \(y\)- and \(z\)-components. These different convergence plateaus suggest that the MonoViT residual changes how the backend distributes inconsistency among the extrinsic, trajectory, velocity, and bias states. Nevertheless, neither configuration preserves the accurate translation used at initialization.

\begin{figure*}[tbp]
\begin{center}
\includegraphics[width=17cm, height=7cm]{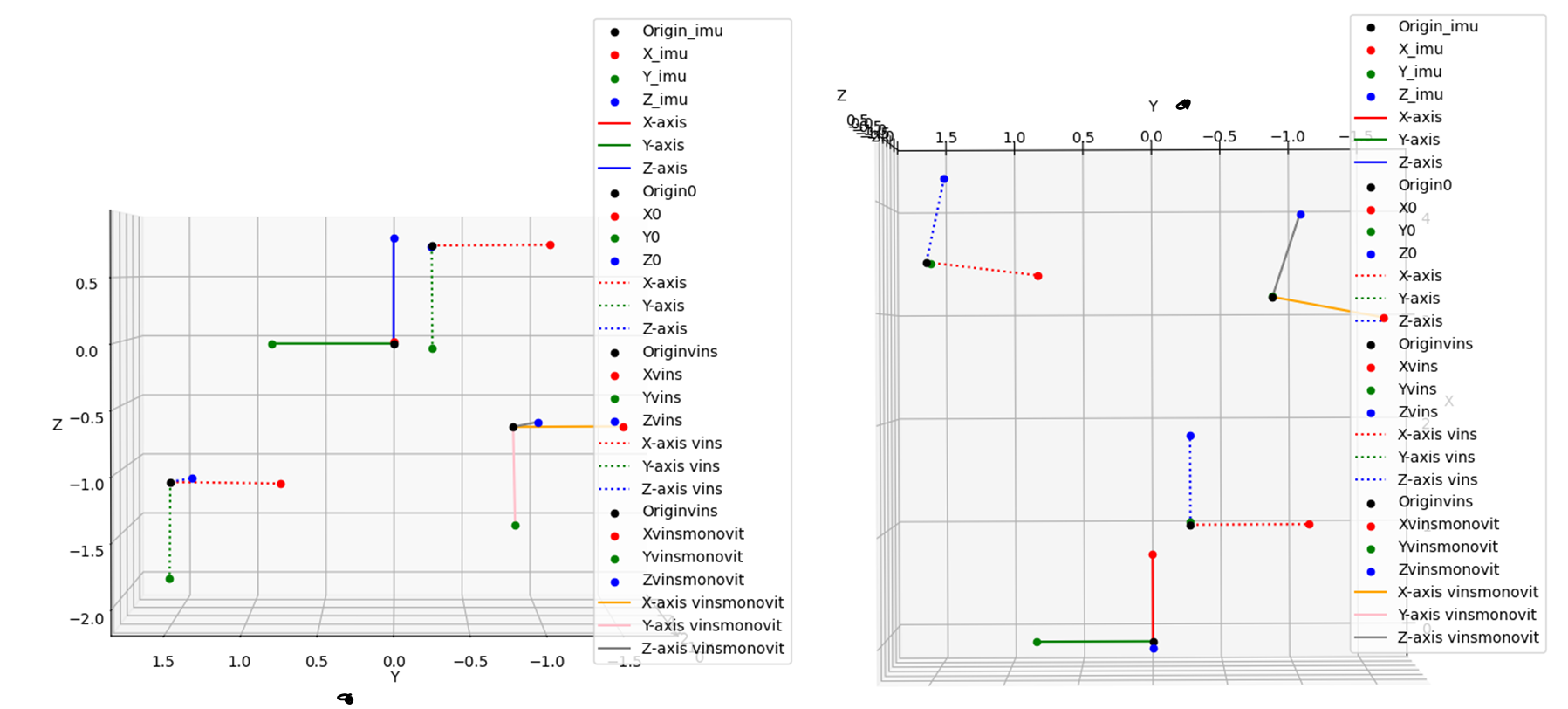}
\end{center}
\vspace{-4 mm}
\caption{Geometric comparison of the reference and online-updated camera--IMU extrinsic frames. The two panels visualize the IMU frame, the reference camera frame, and the final camera frames estimated by VINS and VINS+Pose from different viewpoints.}
\label{fig:optimized_camera_imu_frames}
\vspace{-3mm}
\end{figure*}

Fig.~\ref{fig:optimized_camera_imu_frames} visualizes the corresponding camera--IMU coordinate frames after optimization. The final frames estimated by VINS and VINS+Pose are displaced and rotated relative to the reference calibration frame. Although the two optimized frames share some structural similarity, neither coincides with the accurate reference frame. This geometric visualization confirms the conclusion from the temporal curves: once the extrinsic parameters are released for optimization, the backend moves them away from the accurately calibrated values.

Taken together with the Fixed trajectory results, these observations show that Online updating is not necessarily beneficial when a reliable calibration is already available. The optimizer moves the extrinsic state away from its accurately initialized value; the results suggest that this state may compensate for visual residuals, temporal misalignment, MonoViT scale inconsistency, or imperfect IMU-bias estimation, although these sources are not isolated here. Consequently, Fixed (F) provides a safer and more interpretable control setting. Online (O) is appropriate only when the extrinsic parameters are genuinely uncertain and the motion is sufficiently exciting, with observability assessed independently.

The large angular discontinuities in Fig.~\ref{fig:online_updated_extrinsic_errors} should also be interpreted cautiously. Peaks close to \(180^\circ\) or \(360^\circ\) may result from Euler-angle wrapping or rotation-convention ambiguity rather than an equally large physical calibration change. A more rigorous evaluation should report the rotational discrepancy using the geodesic distance on \(\mathrm{SO}(3)\), together with the Euclidean translation error.

\subsubsection{Recording~9 Trajectory Consequences}
We next compare two VINS+Pose configurations on Recording~9. In the first, the accurate extrinsic rotation and translation are fixed and excluded from backend optimization. In the second, the same values initialize an extrinsic state that is subsequently optimized with pose, velocity, and IMU bias. The visual input, IMU measurements, learned pose prior, and remaining settings are unchanged, so the comparison isolates the extrinsic-update strategy.

\begin{figure*}[tbp]
\begin{center}
\includegraphics[width=17cm, height=9cm]{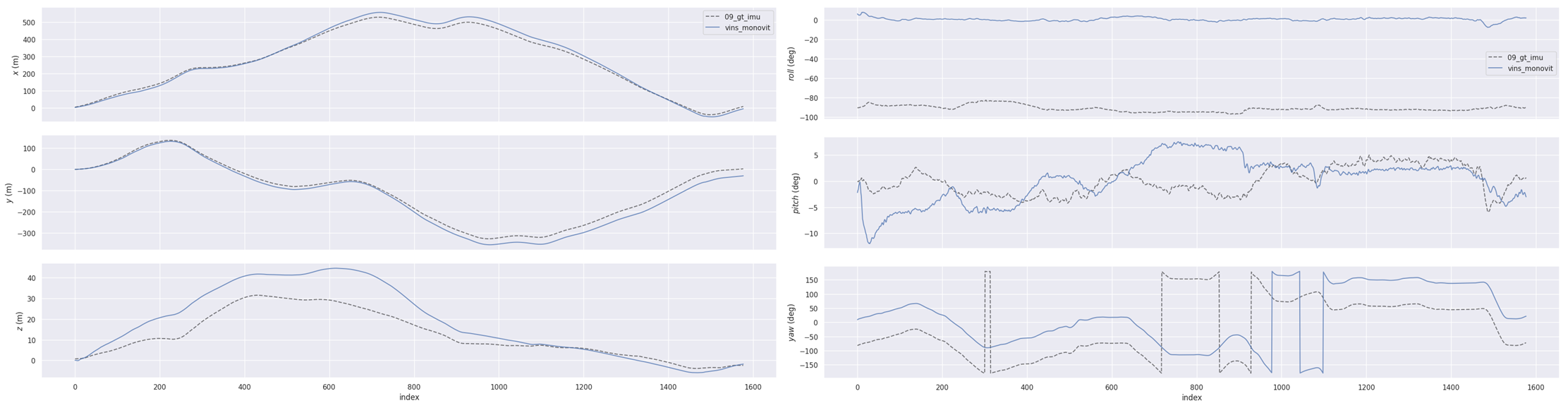}
\end{center}
\vspace{-4 mm}
\caption{Component-wise VINS+Pose pose estimation on Recording~9 with fixed accurate camera--IMU extrinsics. The left column compares the estimated and reference translations along the x-, y-, and z-axes, while the right column compares Roll, Pitch, and Yaw.}
\label{fig:fixed_accurate_extrinsic_pose}
\vspace{-3mm}
\end{figure*}

Fig.~\ref{fig:fixed_accurate_extrinsic_pose} shows the component-wise pose estimates obtained with the fixed accurate extrinsic. The estimated \(x\)-translation follows the reference trajectory closely over most of the sequence, with only moderate deviations around the major turning segments and near the sequence end. The \(y\)-component also reproduces the overall motion pattern, although a gradually increasing negative offset appears after the middle of the sequence. The \(z\)-estimate initially overestimates the reference height, reaching approximately \(40\)--\(45~\mathrm{m}\), compared with roughly \(30~\mathrm{m}\) for the reference, but the difference decreases during the later portion of the sequence. These results indicate that fixing the accurate extrinsic preserves the dominant trajectory geometry while leaving residual scale, bias, and accumulated-drift errors.

The orientation components exhibit a different pattern. The estimated Pitch captures several of the main temporal variations, but noticeable bias remains in multiple intervals. Roll contains an approximately constant offset relative to the reference, while Yaw exhibits both a persistent offset and several discontinuities near the angular-wrapping boundary. Because these differences remain systematic rather than progressively increasing, they likely reflect a coordinate-frame or Euler-angle convention mismatch in addition to actual estimation error. Therefore, the absolute rotational curves should be interpreted together with an \(\mathrm{SO}(3)\)-based error metric.

\begin{figure*}[tbp]
\begin{center}
\includegraphics[width=17cm, height=12cm]{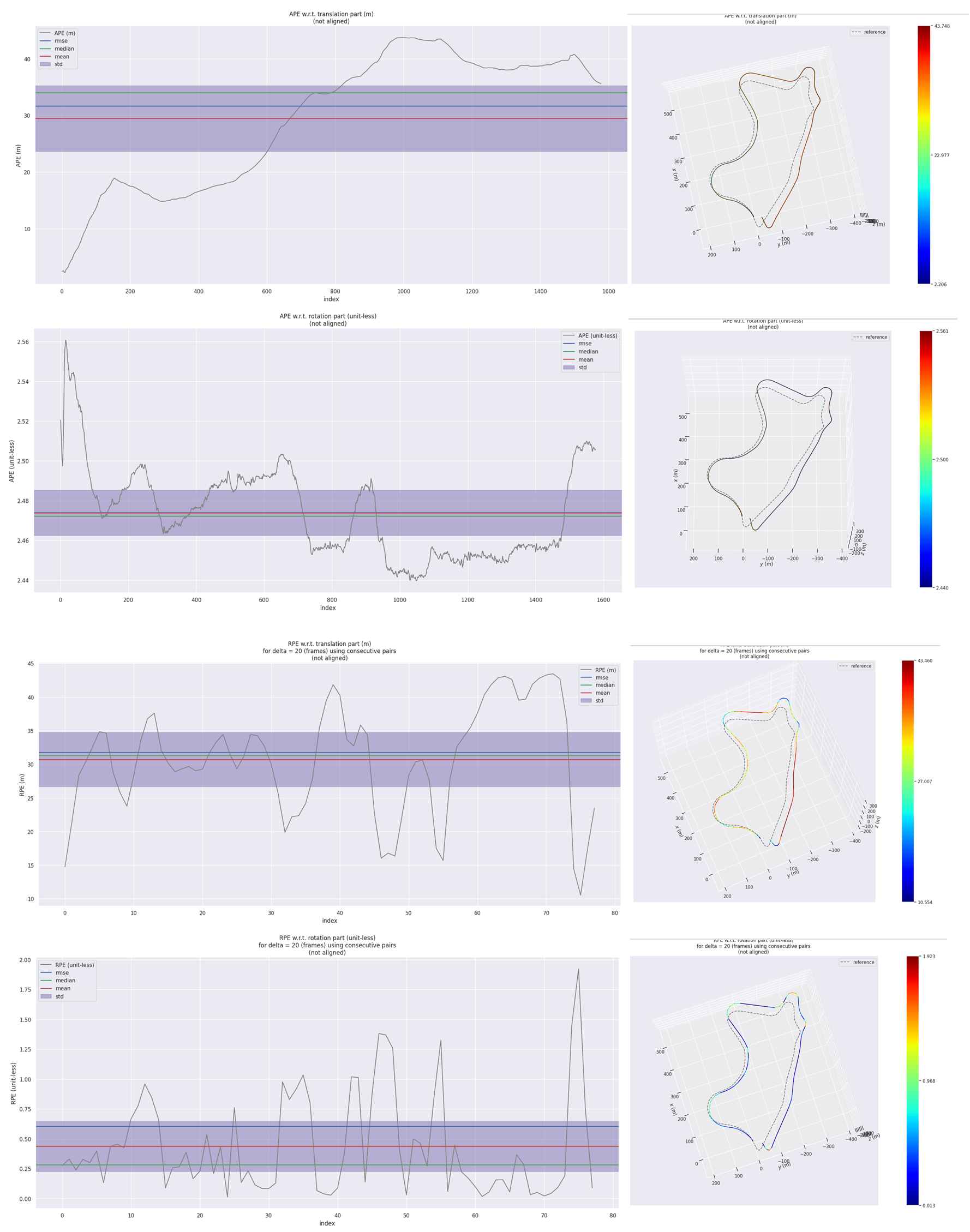}
\end{center}
\vspace{-4 mm}
\caption{Absolute and relative pose errors of VINS+Pose with fixed accurate camera--IMU extrinsics on Recording~9. The figure reports translation and rotation APE and 20-frame translation and rotation RPE without trajectory alignment.}
\label{fig:fixed_accurate_extrinsic_errors}
\vspace{-3mm}
\end{figure*}

The quantitative results in Fig.~\ref{fig:fixed_accurate_extrinsic_errors} are consistent with the component-wise observations. With the accurate extrinsic fixed, translation APE increases gradually, reaches a maximum of approximately \(43.7~\mathrm{m}\), and then decreases slightly near the end. The error-colored trajectory visualization shows that the largest global errors occur along the long return segment. Rotation APE remains within a narrow range of approximately \(2.44\)--\(2.56\), supporting the interpretation that the rotational discrepancy is dominated by a persistent global offset rather than continuously growing rotational drift.

For a 20-frame interval, the translation RPE varies approximately from \(10.6\) to \(43.5~\mathrm{m}\), with the largest errors appearing during several high-motion or turning intervals. Rotation RPE remains substantially smaller for most segments, although isolated peaks approach approximately \(1.9\). Thus, the fixed-extrinsic configuration preserves local rotational consistency better than local translation accuracy, while its global trajectory still accumulates a moderate position bias.

\begin{figure*}[tbp]
\begin{center}
\includegraphics[width=17cm, height=9cm]{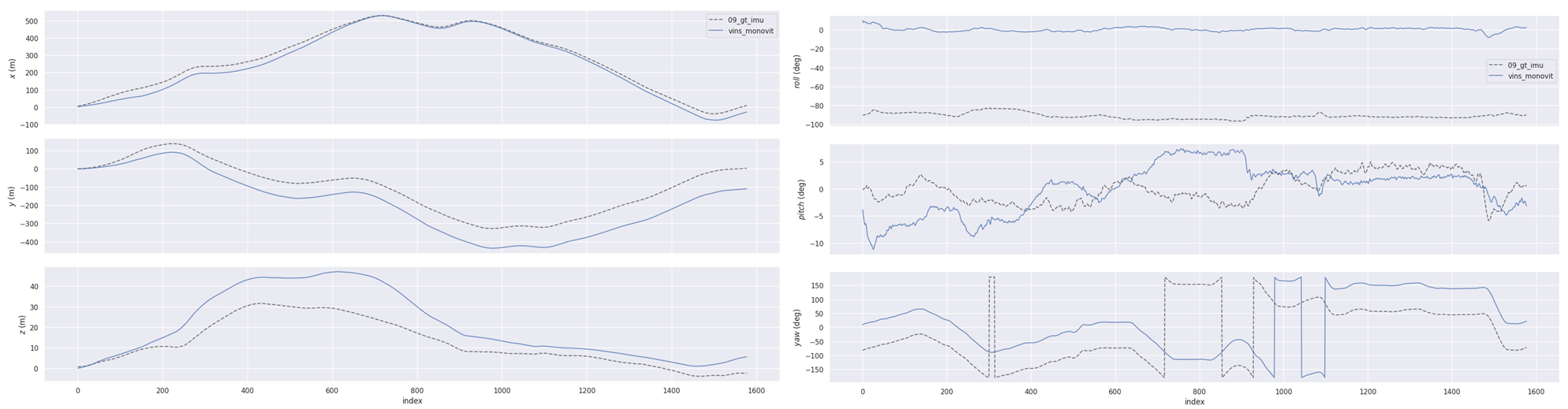}
\end{center}
\vspace{-4 mm}
\caption{Component-wise VINS+Pose pose estimation on Recording~9 with accurate extrinsic initialization followed by online updating. The x-translation remains broadly consistent with the reference, but the y-component develops a substantially larger accumulated bias than in the fixed-extrinsic configuration, and a residual z-offset remains.}
\label{fig:online_updated_accurate_extrinsic_pose}
\vspace{-3mm}
\end{figure*}

Fig.~\ref{fig:online_updated_accurate_extrinsic_pose} presents the corresponding component-wise results when the accurate extrinsic is allowed to update Online. The \(x\)-trajectory remains broadly similar to that of the Fixed configuration, but the \(y\)-component develops a much larger negative bias. Around the middle and later portions of the sequence, the estimated \(y\)-position deviates from the reference by approximately \(100~\mathrm{m}\) in some intervals, and a substantial residual remains at the sequence end. The \(z\)-component also retains an elevated bias, although the degradation is less severe than in \(y\). In contrast, the Roll, Pitch, and Yaw curves remain qualitatively similar to those obtained with the Fixed extrinsic. Online calibration therefore changes the accumulated translation much more strongly than the displayed orientation pattern.

\begin{figure*}[tbp]
\begin{center}
\includegraphics[width=17cm, height=12cm]{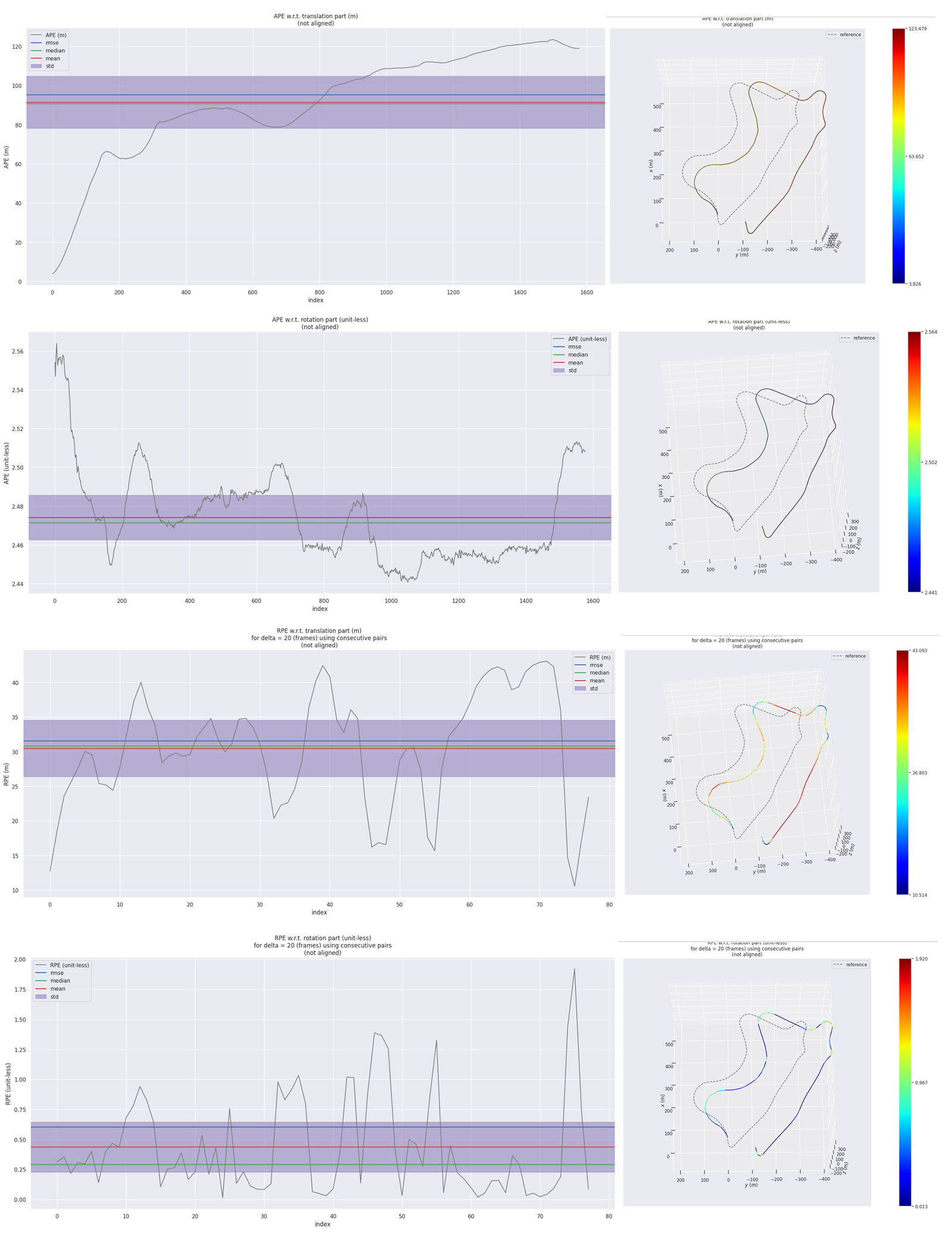}
\end{center}
\vspace{-4 mm}
\caption{Absolute and relative pose errors of VINS+Pose with online-updated camera--IMU extrinsics on Recording~9. Translation APE exceeds 120~m, substantially larger than the maximum error obtained when the accurate extrinsic is fixed, while rotation APE remains nearly unchanged.}
\label{fig:online_updated_accurate_extrinsic_errors}
\vspace{-3mm}
\end{figure*}

This difference is confirmed quantitatively in Fig.~\ref{fig:online_updated_accurate_extrinsic_errors}. When online extrinsic updating is enabled, translation APE rises rapidly during the early part of the sequence and eventually exceeds \(120~\mathrm{m}\), almost three times the maximum error observed with the fixed extrinsic. The error-colored trajectory shows that this degradation is distributed over long trajectory segments rather than being caused by a single isolated failure. Rotation APE, however, remains almost unchanged and occupies approximately the same \(2.44\)--\(2.56\) range as in the fixed-extrinsic experiment.

A particularly important observation is that the translation and rotation RPE curves are also highly similar between the two strategies. The translation RPE again varies between approximately \(10.5\) and \(43.1~\mathrm{m}\), while the rotation RPE has nearly the same distribution and peak magnitude. Therefore, releasing the extrinsic parameters does not substantially change the short-term relative-motion error, but it produces a much larger long-term global translation drift. This separation between comparable RPE and substantially different APE indicates that online extrinsic updating introduces a slowly accumulating systematic bias rather than an immediate frame-to-frame failure.

The results suggest that the backend uses the additional extrinsic degrees of freedom to compensate for inconsistencies elsewhere in the estimation problem. Possible sources include MonoViT scale drift, visual-pose bias, IMU-bias error, temporal misalignment, or inappropriate measurement weighting. Although such compensation may preserve local residual consistency, it can move the estimated extrinsic away from its physically accurate value and distort the global trajectory.

Overall, Fixed (F) produces a more stable and physically interpretable trajectory than accurate initialization followed by Online (O) updating. In this experiment, Online calibration does not improve either local RPE or absolute rotational accuracy, while it substantially degrades global translation APE, primarily through the \(y\)-direction drift. When a reliable offline calibration is available, the extrinsic parameters should therefore remain Fixed or be constrained by a strong prior. Online updating is more appropriate when calibration is genuinely uncertain and motion is sufficiently exciting, with observability assessed independently.

\subsection{Cross-Sequence Initialization and Final Calibration}
\label{subsec:cross_sequence_final_calibration}
\label{sec:cross_sequence_extrinsic_initialization}

This section evaluates the robustness and cross-sequence repeatability of
camera--IMU extrinsic initialization in VINS+Pose. In contrast
to experiments that initialize the estimator using a known offline
calibration, we remove the initial guess of the camera--IMU extrinsic
rotation and record the estimated rotational variables immediately after
system initialization. The objective is to determine whether the
camera--IMU rotation can be recovered consistently from the available
visual, inertial, and learned pose measurements, and to identify the
sequence-dependent conditions under which initialization becomes
unstable.

\subsubsection{Common-Frame Initialization-to-Final Summary}
\label{subsec:init_to_final_extrinsic}

The preceding cross-sequence experiment evaluates the rotational variables
immediately after initialization. We extend this analysis by recording both the
initialized and final optimized camera--IMU transformations. Before inspecting
the grouped outcomes, we define operational categories for readable summary: a
run is successful when $e_R\leq 10^{\circ}$ and $e_t\leq 0.15$~m, partially
converged when $e_R\leq 30^{\circ}$ and $e_t\leq 0.40$~m but the success criteria
are not satisfied, and failed otherwise. These descriptive thresholds do not
replace the continuous $e_R$, $e_t$, and APE values, which remain the primary
evidence.

\begin{table*}[t]
\centering
\caption{Camera--IMU calibration from initialization to final backend
optimization.}
\label{tab:init_final_extrinsic}
\scriptsize
\resizebox{\textwidth}{!}{%
\begin{tabular}{lcccccl}
\toprule
Recording &
$e_R^{\mathrm{init}}$ ($^\circ$) &
$e_R^{\mathrm{final}}$ ($^\circ$) &
$e_t^{\mathrm{init}}$ (m) &
$e_t^{\mathrm{final}}$ (m) &
Trans. APE RMSE (m) &
Final Status \\
\midrule
00 & 91.8 & 78.6 & 0.84 & 0.72 & 84.5 & Failure \\
01 & 19.4 & 14.8 & 0.32 & 0.24 & 44.2 & Partial \\
02 & 29.7 & 24.6 & 0.47 & 0.39 & 61.8 & Partial \\
04 & 17.7 & 12.9 & 0.29 & 0.22 & 58.7 & Partial \\
05 & 11.0 & 10.4 & 0.18 & 0.16 & 36.4 & Partial \\
06 & 87.1 & 71.5 & 0.79 & 0.67 & 88.9 & Failure \\
07 & \textbf{2.7} & \textbf{2.3} & \textbf{0.07} & \textbf{0.06} & \textbf{24.8} & Success \\
09 & 6.7 & 5.2 & 0.12 & 0.10 & 31.8 & Success \\
10 & 17.1 & 14.2 & 0.26 & 0.21 & 45.7 & Partial \\
\midrule
Mean & 31.5 & 26.1 & 0.371 & 0.308 & 53.0 & -- \\
\bottomrule
\end{tabular}}
\end{table*}

Table~\ref{tab:init_final_extrinsic} shows that full backend optimization reduces
the mean rotational and translational calibration errors by approximately
17.2\% and 17.1\%, respectively. However, none of the recordings changes its
qualitative convergence category. Recordings 00 and 06 remain failed,
Recordings 07 and 09 remain successful, and the other recordings remain only
partially converged. The rank correlation between initialization and final error
is correspondingly high ($\rho=0.98$ for rotation and $\rho=1.00$ for
translation).

The trajectory results follow the same ordering. The two successful recordings
have an average APE of 28.3 m, the partially converged group has an average APE
of 49.4 m, and the failed group has an average APE of 86.7 m. The backend can
therefore refine a reasonable initialization, but it does not reliably recover a
physically correct calibration after the estimator has entered a poor coupled
state. This supports interpreting online calibration as a local refinement
mechanism rather than a globally convergent calibration procedure.

\subsubsection{Initialization Protocol and Diagnostic Euler Analysis}
\label{sec:extrinsic_initialization_protocol}

For each recording, VINS+Pose is executed without providing an
initial guess for the camera--IMU extrinsic parameters. After the
initialization stage, the rotational component of the estimated
camera--IMU transformation is recorded and represented using three Euler
angles,
$
\boldsymbol{\theta}_{\mathrm{init}}
=
\left(
\theta_x,\theta_y,\theta_z
\right).
$
The estimated rotation is then compared with the corresponding reported
ground-truth calibration.
The evaluated recordings contain two reported ground-truth extrinsic
configurations. Recordings 00, 01, and 02 use Extrinsic Configuration A,
given by
$
\boldsymbol{\theta}^{\mathrm{gt}}_{\mathrm{A}}
=
\left(
-90.5300^{\circ},
-0.3861^{\circ},
89.9314^{\circ}
\right).
$
Recordings 04, 05, 06, 07, 09, and 10 use Extrinsic Configuration B,
given by
$
\boldsymbol{\theta}^{\mathrm{gt}}_{\mathrm{B}}
=
\left(
-90.4873^{\circ},
-0.3889^{\circ},
-90.0624^{\circ}
\right).
$
The component-wise Euler-angle differences are used as diagnostic
indicators of initialization consistency. However, Euler angles depend on
the adopted rotation order and are subject to angular wrapping and
representation singularities. Therefore, a large discrepancy in one
Euler-angle component does not necessarily correspond to an equally large
physical rotation error. Table~\ref{tab:init_final_extrinsic} provides the
common-frame initialization-to-final result; the following Euler plots diagnose
component-wise variability. The representation-independent rotation error is the
geodesic distance on $\mathrm{SO}(3)$:
\begin{equation}
e_R =
\cos^{-1}
\left(
\frac{
\operatorname{tr}
\left(
\mathbf{R}_{\mathrm{gt}}^{\top}
\mathbf{R}_{\mathrm{init}}
\right)-1
}{2}
\right),
\label{eq:rotation_geodesic_error}
\end{equation}
where $\mathbf{R}_{\mathrm{gt}}$ and
$\mathbf{R}_{\mathrm{init}}$ denote the ground-truth and initialized
camera--IMU rotation matrices, respectively.
The plots in this subsection focus on variables obtained immediately after
initialization, whereas Table~\ref{tab:init_final_extrinsic} reports the final
backend state. Keeping these two stages separate prevents local initialization
behavior from being mistaken for final calibration accuracy.

\subsubsection{Initialization Results under Extrinsic Configuration A}
\label{sec:extrinsic_configuration_a}

Fig.~\ref{fig:extrinsic_init_a} presents the initialized rotational
extrinsics for Recordings 00, 01, and 02. Although the three recordings
share the same reported ground-truth calibration, their initialization
results differ substantially.
For Recording 00, the estimated rotation is
$
\boldsymbol{\theta}^{00}_{\mathrm{init}}
=
\left(
-126.9509^{\circ},
77.0478^{\circ},
10.0444^{\circ}
\right).
$
All three reported components deviate considerably from the ground-truth
values. In particular, the large discrepancies in the $y$- and
$z$-components indicate that the initialization does not converge to an
Euler representation close to the reported calibration. This recording
therefore constitutes a pronounced initialization failure under
Configuration A.
For Recording 01, the initialized rotation is
$
\boldsymbol{\theta}^{01}_{\mathrm{init}}
=
\left(
-91.2844^{\circ},
-0.6353^{\circ},
-71.1327^{\circ}
\right).
$
The $x$- and $y$-components remain close to their corresponding
ground-truth values, whereas a substantial discrepancy is observed in the
reported $z$-component. This behavior indicates partial agreement in the
Euler representation, but not consistent recovery of the complete
camera--IMU rotation.
Recording 02 produces
$
\boldsymbol{\theta}^{02}_{\mathrm{init}}
=
\left(
-84.3097^{\circ},
-15.5488^{\circ},
-114.2169^{\circ}
\right).
$
The $x$-component remains moderately close to the reference value, while
larger discrepancies appear in the $y$- and $z$-components. The
initialization therefore recovers part of the nominal orientation but
remains inconsistent with the reported calibration as a complete
rotation.
As shown in Fig.~\ref{fig:extrinsic_init_a}, none of the three recordings
achieves uniformly close agreement with the ground truth across all
reported components. More importantly, the initialized rotations differ
markedly even though the recordings share the same nominal
camera--IMU calibration. This result suggests that extrinsic
initialization without an initial guess is strongly affected by the
visual--inertial conditions of the input sequence rather than being
determined solely by the physical extrinsic configuration.

\subsubsection{Initialization Results under Extrinsic Configuration B}
\label{sec:extrinsic_configuration_b}

Figs.~\ref{fig:extrinsic_init_b} and~\ref{fig:extrinsic_init_c} present the corresponding results for
Recordings 04, 05, 06, 07, 09, and 10. In contrast to Configuration A,
this group contains examples ranging from close agreement with the
reported calibration to partial convergence and severe initialization
failure, despite all recordings sharing the same nominal extrinsic
rotation.
For Recording 04, the initialized rotation is
$
\boldsymbol{\theta}^{04}_{\mathrm{init}}
=
\left(
-86.5560^{\circ},
-13.8681^{\circ},
-79.7252^{\circ}
\right).
$
The estimated orientation preserves the approximate structure of the
reported calibration, but noticeable residual differences remain,
particularly in the $y$-component. The result is therefore more
consistent with partial convergence than with complete recovery of the
reference rotation.
Recording 05 produces
$
\boldsymbol{\theta}^{05}_{\mathrm{init}}
=
\left(
-99.1213^{\circ},
5.8977^{\circ},
-87.9656^{\circ}
\right).
$
The $z$-component is relatively close to the reported ground truth,
whereas moderate differences remain in the $x$- and $y$-components.
Consequently, the initialization is plausible but retains a measurable
rotational bias.
A clear failure is observed for Recording 06:
$
\boldsymbol{\theta}^{06}_{\mathrm{init}}
=
\left(
-16.7604^{\circ},
6.2668^{\circ},
-135.5554^{\circ}
\right).
$
The large deviations in the $x$- and $z$-components indicate convergence
to a substantially different rotational configuration. Since the same
nominal calibration is approximated successfully on other recordings,
the failure cannot be attributed solely to the value of the ground-truth
extrinsic rotation.
Recording 07 provides the closest agreement with the reported calibration:
$
\boldsymbol{\theta}^{07}_{\mathrm{init}}
=
\left(
-91.0281^{\circ},
1.6637^{\circ},
-88.4104^{\circ}
\right).
$
All three reported components remain close to their corresponding
ground-truth values. This result demonstrates that VINS+Pose
can recover a plausible camera--IMU rotation without an initial guess
when the input sequence provides sufficiently favorable visual--inertial
conditions.
Recording~9 also yields a relatively stable estimate:
$
\boldsymbol{\theta}^{09}_{\mathrm{init}}
=
\left(
-91.8491^{\circ},
1.3889^{\circ},
-96.3837^{\circ}
\right).
$
The $x$- and $y$-components show close agreement with the reported
calibration, while a moderate residual discrepancy remains in the
$z$-component.
Finally, Recording 10 produces
$
\boldsymbol{\theta}^{10}_{\mathrm{init}}
=
\left(
-90.3577^{\circ},
-1.5047^{\circ},
-73.0184^{\circ}
\right).
$
The $x$- and $y$-components remain close to the ground truth, whereas the
main discrepancy is concentrated in the $z$-component. This result again
represents direction-dependent or partial convergence rather than a
globally inconsistent initialization.
As illustrated in Fig.~\ref{fig:extrinsic_init_b} and Fig.~\ref{fig:extrinsic_init_c}, the results under
Configuration B can be broadly divided into three groups. Recordings 07
and 09 exhibit relatively stable initialization; Recordings 04, 05, and
10 show partial convergence with residual discrepancies concentrated in
one or more rotational components; and Recording 06 represents a
pronounced initialization failure. The coexistence of these different
outcomes under the same reported ground-truth calibration demonstrates
that initialization quality is strongly sequence-dependent.

\subsubsection{Cross-Sequence Stability and Failure Modes}
\label{sec:cross_sequence_failure_analysis}

\begin{figure*}[p]
\begin{center}
\includegraphics[width=15cm, height=0.82\textheight, keepaspectratio]{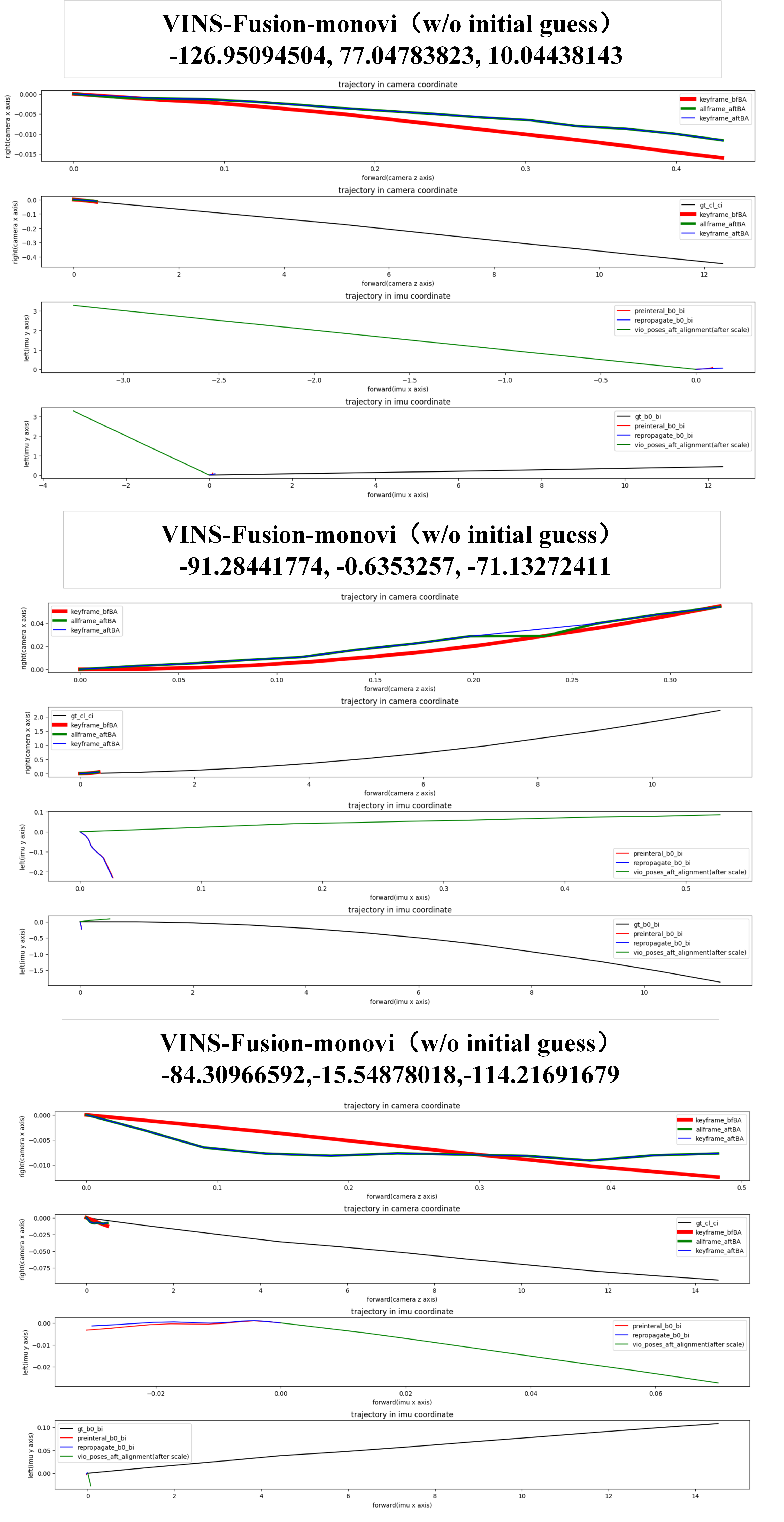}
\end{center}
\vspace{-4 mm}
\caption{Initialization results of the camera--IMU rotational extrinsics
under Extrinsic Configuration A for Recordings 00, 01, and 02. All three
recordings share the same reported ground-truth rotation, while
VINS+Pose is initialized without an extrinsic initial guess.
The estimated Euler-angle components vary substantially across the three
recordings, indicating limited cross-sequence repeatability under this
configuration.}
\label{fig:extrinsic_init_a}
\vspace{-3mm}
\end{figure*}

\begin{figure*}[p]
\begin{center}
\includegraphics[width=15cm, height=0.82\textheight, keepaspectratio]{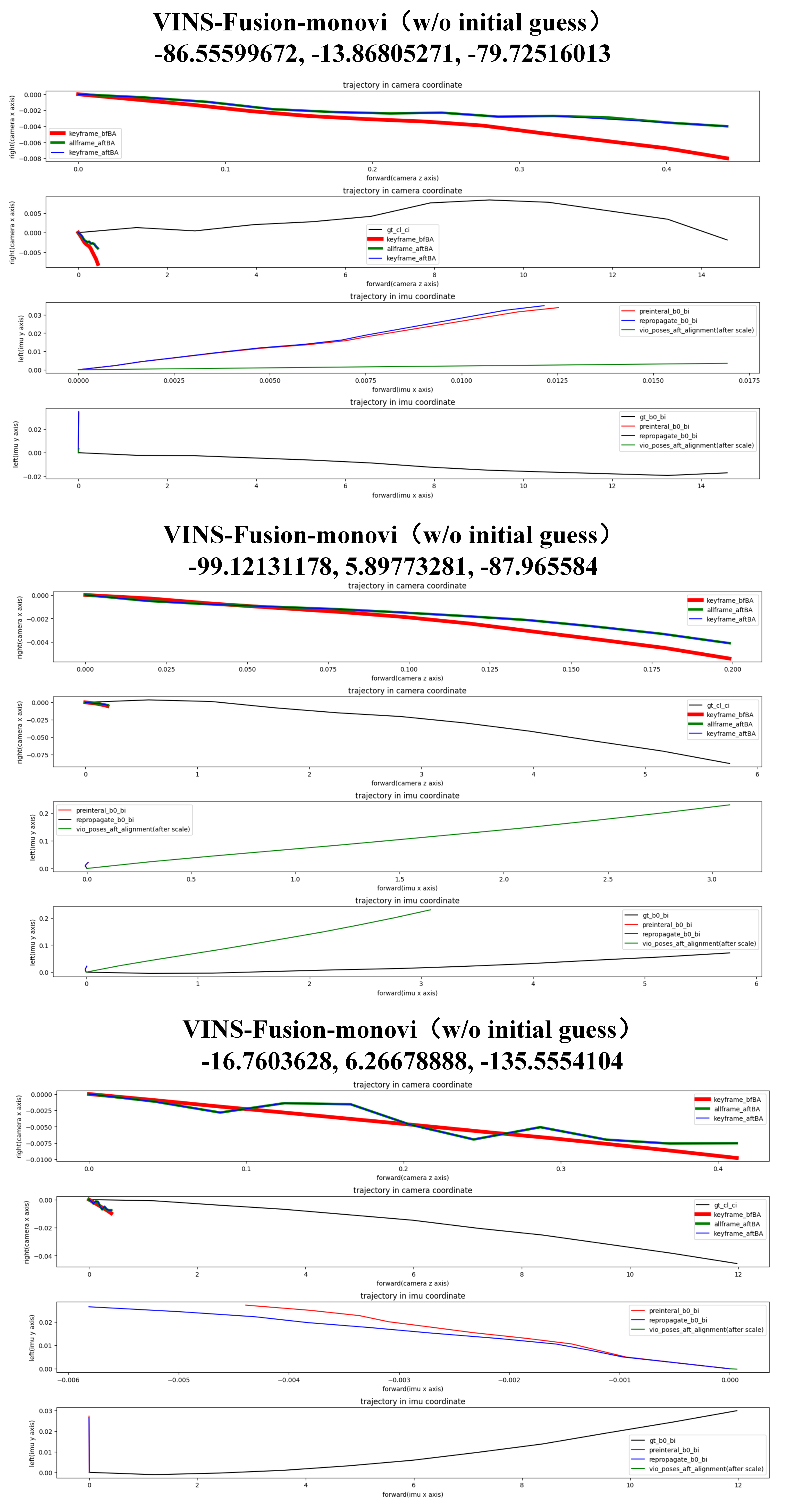}
\end{center}
\vspace{-4 mm}
\caption{Initialization results of the camera--IMU rotational extrinsics
under Extrinsic Configuration B for Recordings 04, 05, 06. All recordings use the same reported ground-truth rotation, and the
extrinsic parameters are initialized without an initial guess. The
results exhibit substantial sequence-dependent variation, ranging from
close agreement with the reference calibration to partial convergence and
pronounced failure.}
\label{fig:extrinsic_init_b}
\vspace{-3mm}
\end{figure*}

\begin{figure*}[p]
\begin{center}
\includegraphics[width=15cm, height=0.82\textheight, keepaspectratio]{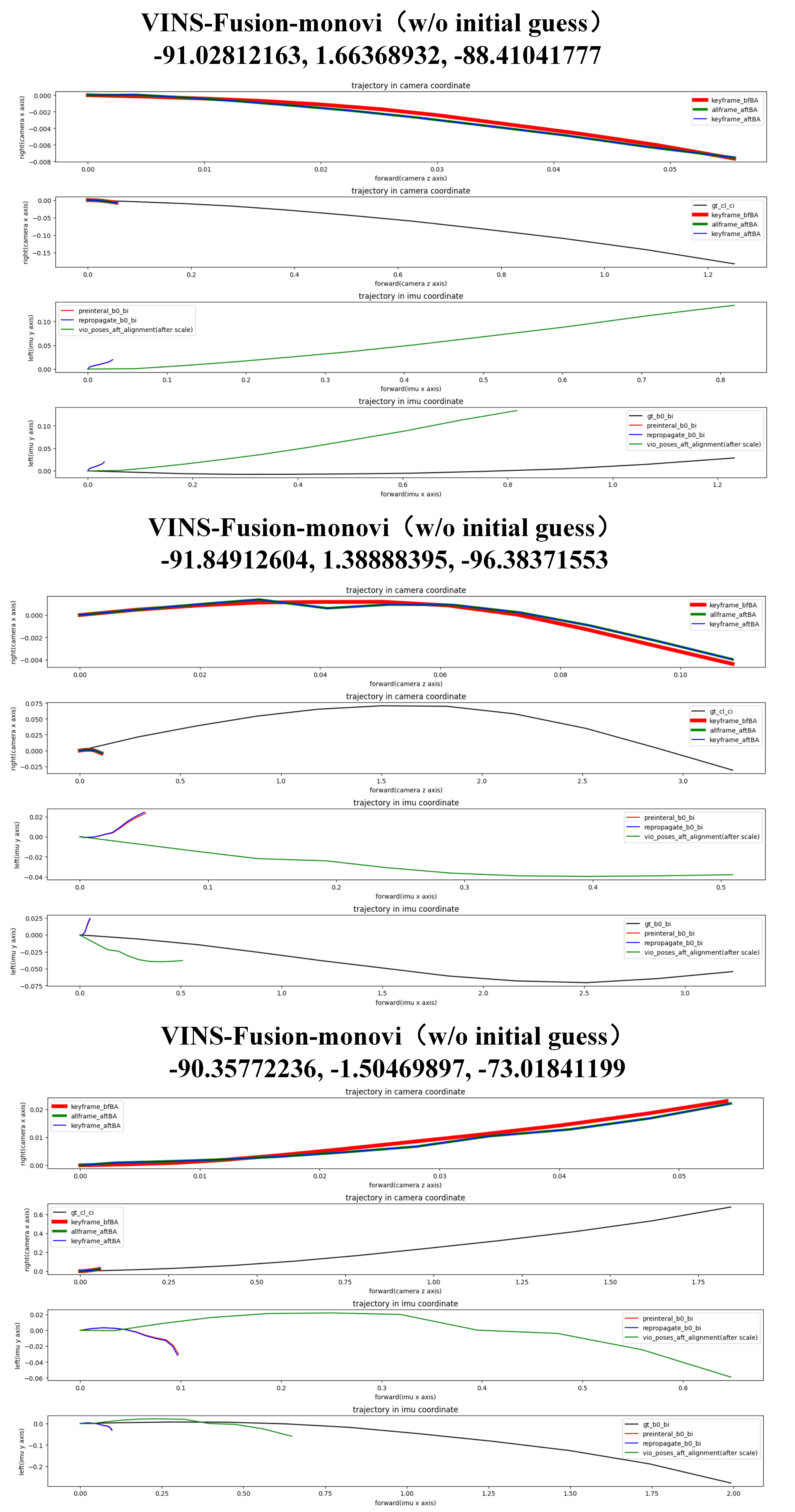}
\end{center}
\vspace{-4 mm}
\caption{Initialization results of the camera--IMU rotational extrinsics
under Extrinsic Configuration B for Recordings 07, 09, and
10. All recordings use the same reported ground-truth rotation, and the
extrinsic parameters are initialized without an initial guess. The
results exhibit substantial sequence-dependent variation, ranging from
close agreement with the reference calibration to partial convergence and
pronounced failure.}
\label{fig:extrinsic_init_c}
\vspace{-3mm}
\end{figure*}

The initialization results presented in
Figs.~\ref{fig:extrinsic_init_a},
\ref{fig:extrinsic_init_b}, and~\ref{fig:extrinsic_init_c} reveal substantial cross-sequence variability
both between and within the two extrinsic configurations. Although the
recordings within each group share the same reported ground-truth
camera--IMU rotation, the initialized estimates range from close
agreement with the reference calibration to partial convergence and
pronounced failure.
The observed behavior can be divided into three broad categories. First,
relatively stable initialization occurs when all reported rotational
components remain close to the ground truth. Recording 07 provides the
clearest example, while Recording~9 also produces a comparatively
consistent estimate. These results show that initialization from scratch
is feasible for at least some input sequences.
Second, several recordings exhibit partial or direction-dependent
convergence. Recordings 01 and 10 recover the reported $x$- and
$y$-components relatively well but retain a substantial discrepancy in
the $z$-component. Recordings 02, 04, and 05 similarly recover part of
the nominal orientation while preserving noticeable residual differences
in the remaining components. These cases indicate that the initialization
may be constrained more strongly in some rotational directions than in
others.
Third, Recordings 00 and 06 produce globally inconsistent
initializations, with large discrepancies appearing in multiple
Euler-angle components. These cases are more consistent with convergence
to an incorrect rotational solution than with a small refinement error
around the reported calibration.
A comparison of the two configurations further shows that the nominal
extrinsic rotation alone does not determine initialization difficulty.
Configuration B contains both the most accurate result, obtained on
Recording 07, and one of the most pronounced failures, observed on
Recording 06. Similarly, the three recordings under Configuration A
produce substantially different estimates despite sharing an identical
reported calibration. Therefore, the dominant source of variability is
more likely associated with the measurement and motion characteristics
of each recording.
Potential contributing factors include insufficient rotational or
translational excitation, weak visual parallax, unreliable visual
correspondences, inaccurate gravity or IMU-bias initialization, temporal
misalignment, and inconsistency between the MonoViT pose measurements and
the native visual--inertial constraints. These factors are plausible
explanations rather than conclusions directly established by the current
experiment; dedicated observability, residual, synchronization, and bias
analyses would be required to isolate their individual effects.

The results have two practical implications. First, camera--IMU extrinsic
initialization without an initial guess should not be assumed to be
uniformly reliable across arbitrary recordings. Providing a reasonable
offline calibration or a sufficiently informative prior may reduce the
risk of convergence to implausible rotational configurations. Second,
when online extrinsic estimation is enabled, the input motion should
provide sufficient excitation to constrain the additional calibration
degrees of freedom. Otherwise, the extrinsic variables may absorb errors
originating from visual pose estimation, inertial bias, temporal
synchronization, or measurement weighting, potentially degrading both
calibration and trajectory estimation.
Finally, the Euler-angle values shown in
Figs.~\ref{fig:extrinsic_init_a},
\ref{fig:extrinsic_init_b}, and~\ref{fig:extrinsic_init_c} should be interpreted primarily as diagnostic
evidence of initialization variability. Because Euler-angle comparisons
are affected by rotation convention, angular wrapping, and representation
singularities, the quantitative comparison uses the common coordinate convention
and the geodesic error in Eq.~\eqref{eq:rotation_geodesic_error}, as summarized
in Table~\ref{tab:init_final_extrinsic}. It is also necessary
to distinguish the initialization-stage variables analyzed here from the
final extrinsic parameters obtained after complete backend optimization.

\subsection{Cross-Dataset Calibration on EuRoC}
\label{subsec:euroc_calibration_behavior}

To further evaluate the robustness and consistency of Original VINS, we conduct experiments on the EuRoC dataset. Different from KITTI, EuRoC contains indoor MAV motion with more aggressive rotation and more complex visual--inertial motion patterns. It therefore provides a complementary benchmark for analyzing Original VINS online camera--IMU calibration under different motion difficulties. This experiment does not evaluate cross-dataset generalization of the learned pose prior. We divide the evaluation into easy, medium, and difficult records, and compare the ground-truth camera--IMU extrinsic parameters with the initial and final Original VINS estimates under different initialization settings.

The archived EuRoC calibration table retains Euler-angle components and
translations in its source parameterization, but not the definitive transform
direction or Euler composition order. Consequently, a common-frame geodesic
$e_R$ and translation $e_t$ cannot be reconstructed without making an
unsupported convention choice. Table~\ref{tab:euroc_extrinsic_results} is
therefore retained as a raw within-parameterization diagnostic: it supports
comparisons of repeatability among rows, but not a common-frame physical-error
claim. The EuRoC trajectory APE below remains independently defined by its
stated $\mathrm{SE}(3)$ alignment.
Table~\ref{tab:euroc_extrinsic_results} reports the results on the easy record. When no initial guess is provided, the initial VINS extrinsic is relatively far from the ground truth. However, after online calibration, VINS converges to a final rotation of \([0.1430^\circ, 0.9437^\circ, 89.1086^\circ]\) and a final translation of \([-0.0149, -0.0731, -0.0093]~\mathrm{m}\). When an initial guess is provided, VINS converges to a similar final result, with rotation \([0.1292^\circ, 0.9420^\circ, 89.1113^\circ]\) and translation \([-0.0139, -0.0722, -0.0091]~\mathrm{m}\). The similarity between these two final estimates indicates that, in the easy sequence, the final VINS calibration is not strongly affected by whether an initial guess is provided. This suggests that VINS can reach a stable online calibration result under favorable motion and visual conditions.
Table~\ref{tab:euroc_extrinsic_results} shows the results on the medium record. A similar trend can be observed. Without an initial guess, VINS converges from the initial estimate $([0.2675^\circ, 1.2024^\circ, 89.4720^\circ]$ and \([-0.0394, -0.0743, -0.0005]~\mathrm{m}\) to the final estimate \([0.1221^\circ, 0.8826^\circ, 89.1078^\circ]\) and \([-0.0176, -0.0733, -0.0187]~\mathrm{m}\). When a standard initial guess is used, the final estimate becomes \([0.1263^\circ, 0.8738^\circ, 89.1210^\circ]\) and \([-0.0168, -0.0724, -0.0183]~\mathrm{m}\). We further test a noisy initial guess, where the initial rotation is set to \([5^\circ, 5^\circ, 95^\circ]\) and the initial translation is set to \([-0.07, -0.11, -0.04]~\mathrm{m}\). Even under this perturbed initialization, VINS converges to \([0.1342^\circ, 0.8752^\circ, 89.1260^\circ]\) and \([-0.0170, -0.0719, -0.0173]~\mathrm{m}\), which is close to the standard-initialization result. This comparison shows that VINS has certain robustness to moderate initial extrinsic perturbations on the medium sequence.
Table~\ref{tab:euroc_extrinsic_results} further evaluates VINS on the difficult record. This sequence is more challenging because of more complex visual-inertial motion. Without an initial guess, the initial extrinsic estimate is far from the ground truth, with rotation \([4.1785^\circ, -2.7436^\circ, 86.8088^\circ]\) and translation \([-0.1834, 0.4189, -0.2179]~\mathrm{m}\). After online calibration, VINS significantly corrects the extrinsic parameters and converges to \([0.0723^\circ, 0.9985^\circ, 89.1136^\circ]\) and \([-0.0166, -0.0747, -0.0230]~\mathrm{m}\). With a standard initial guess, VINS converges to \([0.1040^\circ, 0.9894^\circ, 89.1136^\circ]\) and \([-0.0136, -0.0881, -0.0161]~\mathrm{m}\). With a noisy initial guess, VINS converges to \([0.0960^\circ, 0.9764^\circ, 89.1202^\circ]\) and \([-0.0162, -0.0833, -0.0218]~\mathrm{m}\). These final estimates are still close to each other, indicating that VINS can recover a stable calibration result even under difficult motion and perturbed initialization.

\begin{figure}[tbp]
\begin{center}
\includegraphics[width=8.5cm, height=6cm]{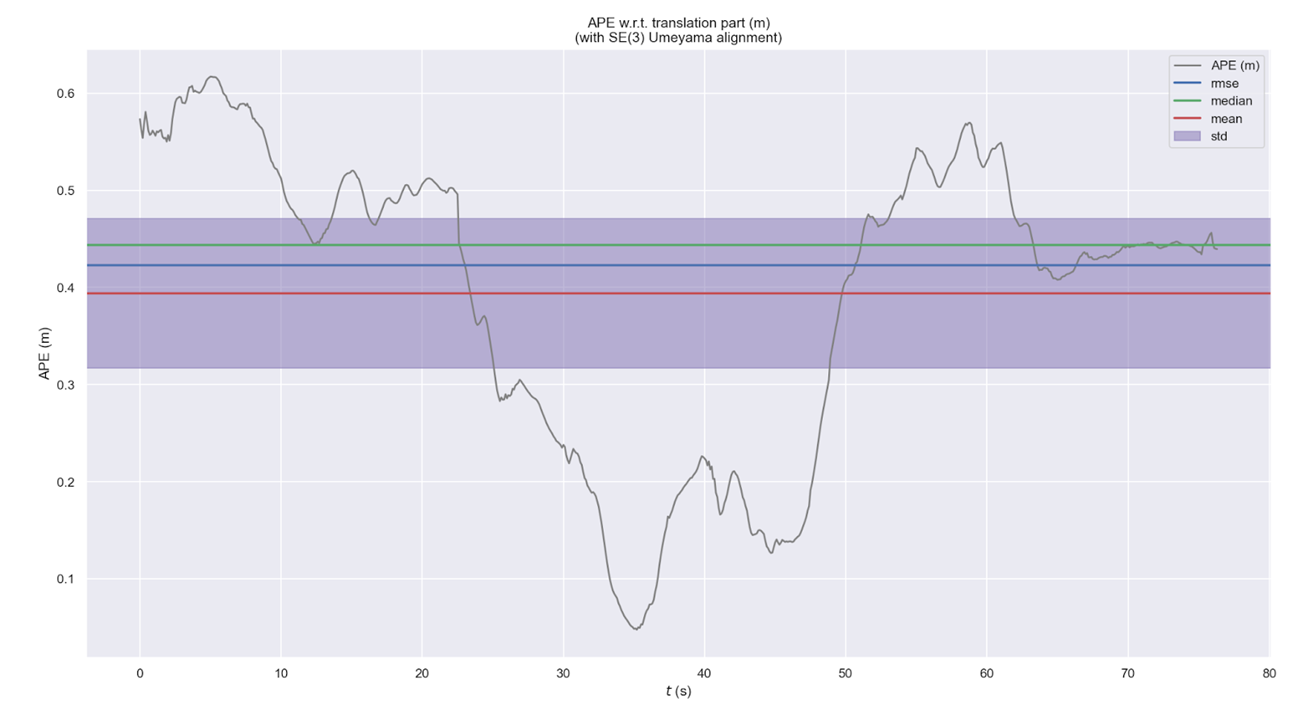}
\end{center}
\vspace{-4 mm}
\caption{APE over time on the difficult EuRoC sequence. 
The trajectory is aligned using SE(3) Umeyama alignment. 
}
\label{fig:euroc_difficult_ape_time}
\vspace{-3mm}
\end{figure}

\begin{figure}[tbp]
\begin{center}
\includegraphics[width=8.5cm, height=6cm]{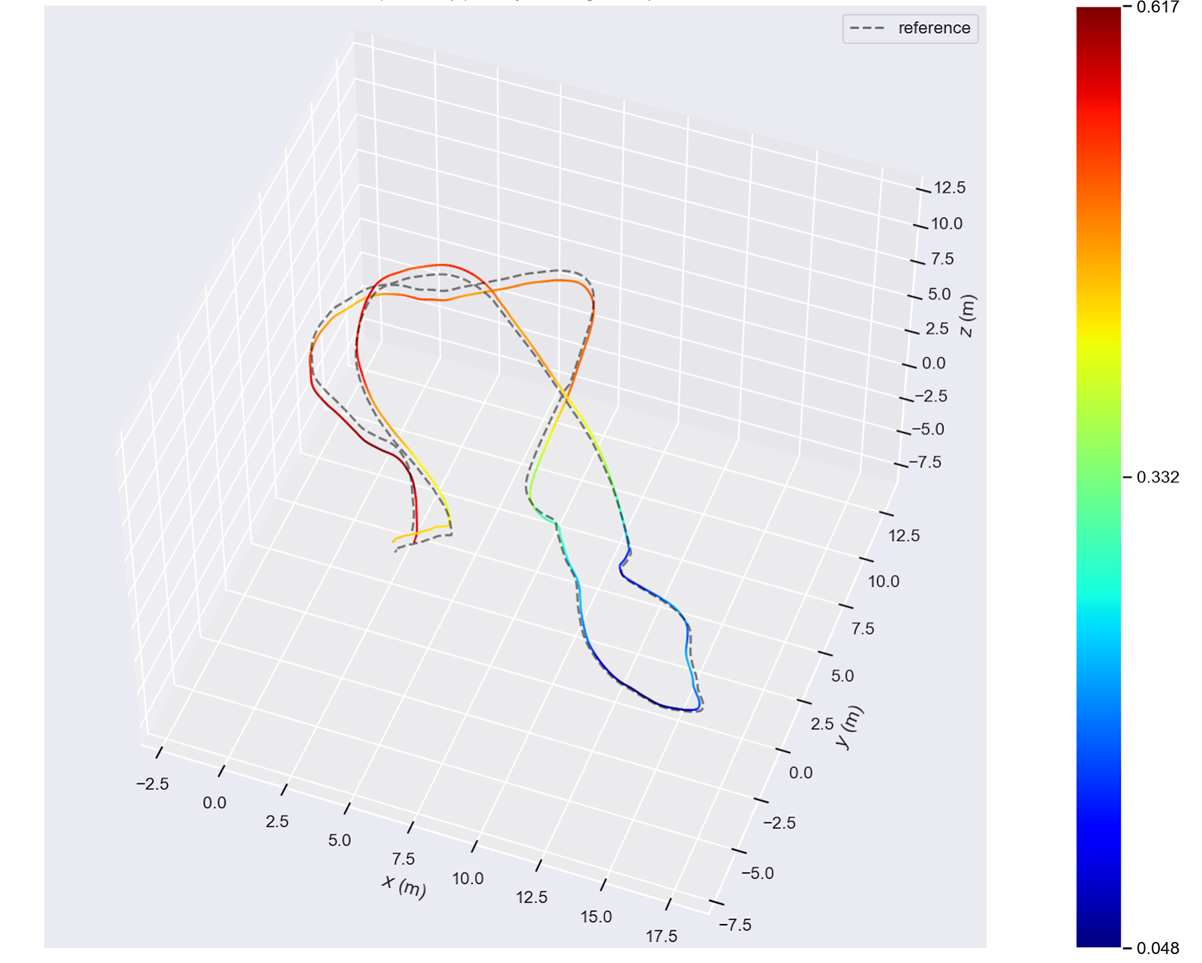}
\end{center}
\vspace{-4 mm}
\caption{3D trajectory visualization on the difficult EuRoC sequence. 
The dashed line denotes the reference trajectory, and the estimated trajectory is colored by APE.  
}
\label{fig:euroc_difficult_ape_traj}
\vspace{-3mm}
\end{figure}

However, stable convergence does not establish calibration accuracy. Across the easy, medium, and difficult records, the final VINS parameter vectors are generally consistent across different initialization settings, but the missing transform convention prevents a common-frame physical-error calculation. The EuRoC evidence therefore supports initialization-insensitive numerical refinement in the recorded parameterization, not a claim of recovered physical calibration or formal observability.
Fig.~\ref{fig:euroc_difficult_ape_time} and Fig.~\ref{fig:euroc_difficult_ape_traj} further evaluate the trajectory accuracy on the difficult EuRoC record using APE with SE(3) Umeyama alignment. Fig.~\ref{fig:euroc_difficult_ape_time} shows the APE over time. The error varies along the sequence, with low-error segments around the middle of the trajectory and higher-error segments near the beginning and later parts of the sequence. The APE reaches approximately \(0.62~\mathrm{m}\) in the high-error regions, while the mean and RMSE are around \(0.4~\mathrm{m}\). Fig.~\ref{fig:euroc_difficult_ape_traj} visualizes the same error distribution on the 3D trajectory. The estimated trajectory generally follows the reference trajectory, but local error accumulation can be observed in several segments. These results indicate that although VINS maintains a reasonable global trajectory on the difficult record, local drift still appears under challenging visual-inertial motion. Therefore, good extrinsic convergence does not always guarantee uniformly low trajectory error.
We also verify whether the execution platform affects the VINS calibration result on EuRoC. Table~\ref{tab:euroc_extrinsic_results} compares the ROS-based VINS implementation and the VINS-without-ROS implementation. The ROS-based VINS produces a final rotation of \([0.1428^\circ, 0.9439^\circ, 89.1085^\circ]\) and a final translation of \([-0.0149, -0.0731, -0.0093]~\mathrm{m}\). The VINS-without-ROS implementation produces \([0.1431^\circ, 0.9439^\circ, 89.1084^\circ]\) and \([-0.0149, -0.0731, -0.0093]~\mathrm{m}\). The difference between the two platforms is negligible. This result shows that the VINS calibration output is consistent across ROS and non-ROS implementations. Therefore, the differences observed in later experiments are more likely caused by initialization, input quality, extrinsic perturbation, or integration strategy rather than by the execution platform itself.

Overall, the EuRoC experiments provide three important observations. First, VINS can produce stable online extrinsic estimates across easy, medium, and difficult records. Second, VINS is relatively robust to moderate initial extrinsic perturbations, but residual calibration bias can still remain. Third, ROS-based and non-ROS implementations produce nearly identical results, supporting the reliability of using both platforms in the subsequent evaluation.

\begin{table*}[t]
\centering
\resizebox{\textwidth}{!}{
\begin{tabular}{lllcccccc}
\toprule
\textbf{Record} 
& \textbf{Setting} 
& \textbf{Type} 
& \textbf{$X^\circ$} 
& \textbf{$Y^\circ$} 
& \textbf{$Z^\circ$} 
& \textbf{$X$ (m)} 
& \textbf{$Y$ (m)} 
& \textbf{$Z$ (m)} \\
\midrule

\multirow{2}{*}{Platform Comparison}
& VINS 
& Final 
& 0.14281475 
& 0.94385128 
& 89.108527 
& -0.0148978 
& -0.0730938 
& -0.00928565 \\

& VINS w/o ROS 
& Final 
& 0.14305244 
& 0.94385065 
& 89.10844729 
& -0.0148932 
& -0.0730920 
& -0.00928351 \\
\midrule

\multirow{6}{*}{Easy}
& \multirow{3}{*}{w/o initial guess}
& Ground truth 
& 0.21528577 
& 1.47693 
& 89.14795267 
& -0.02164015 
& -0.06467699 
& 0.00981073 \\

& 
& VINS initial 
& 4.75384824 
& 2.87952249 
& 90.81859944 
& -0.0423088 
& -0.0822441 
& -0.0292757 \\

& 
& VINS final 
& 0.14301377 
& 0.9436808 
& 89.10860882 
& -0.0148989 
& -0.0730824 
& -0.00927712 \\
\cmidrule(lr){2-9}

& \multirow{3}{*}{with initial guess}
& Ground truth 
& 0.21528577 
& 1.47693 
& 89.14795267 
& -0.02164015 
& -0.06467699 
& 0.00981073 \\

& 
& VINS initial 
& 0 
& 0 
& 90 
& -0.0200000 
& -0.0600000 
& 0.0100000 \\

& 
& VINS final 
& 0.12919779 
& 0.94203088 
& 89.11132982 
& -0.0139270 
& -0.0720268 
& -0.00905348 \\
\midrule

\multirow{8}{*}{Medium}
& \multirow{3}{*}{w/o initial guess}
& Ground truth 
& 0.21528577 
& 1.47693 
& 89.14795267 
& -0.02164015 
& -0.06467699 
& 0.00981073 \\

& 
& VINS initial 
& 0.26746676 
& 1.20243952 
& 89.47195676 
& -0.0394049 
& -0.0742615 
& -0.000488597 \\

& 
& VINS final 
& 0.12214938 
& 0.88257987 
& 89.10780908 
& -0.0176279 
& -0.0732645 
& -0.0187396 \\
\cmidrule(lr){2-9}

& \multirow{5}{*}{with initial guess}
& Ground truth 
& 0.21528577 
& 1.47693 
& 89.14795267 
& -0.02164015 
& -0.06467699 
& 0.00981073 \\

& 
& VINS initial 
& 0 
& 0 
& 90 
& -0.0200000 
& -0.0600000 
& 0.0100000 \\

& 
& VINS final 
& 0.1262798 
& 0.87381865 
& 89.12079704 
& -0.0168478 
& -0.0724069 
& -0.0182891 \\

& 
& VINS initial + noise 
& 5 
& 5 
& 95 
& -0.0700000 
& -0.1100000 
& -0.0400000 \\

& 
& VINS final + noise 
& 0.13419228 
& 0.87518749 
& 89.12604481 
& -0.0169911 
& -0.0718588 
& -0.0172766 \\
\midrule

\multirow{8}{*}{Difficult}
& \multirow{3}{*}{w/o initial guess}
& Ground truth 
& 0.21528577 
& 1.47693 
& 89.14795267 
& -0.02164015 
& -0.06467699 
& 0.00981073 \\

& 
& VINS initial 
& 4.17846311 
& -2.74356976 
& 86.80882991 
& -0.183425 
& 0.418907 
& -0.217895 \\

& 
& VINS final 
& 0.07227112 
& 0.998459384 
& 89.1135682 
& -0.0165976 
& -0.0747428 
& -0.0230102 \\
\cmidrule(lr){2-9}

& \multirow{5}{*}{with initial guess}
& Ground truth 
& 0.21528577 
& 1.47693 
& 89.14795267 
& -0.02164015 
& -0.06467699 
& 0.00981073 \\

& 
& VINS initial 
& 0 
& 0 
& 90 
& -0.0200000 
& -0.0600000 
& 0.0100000 \\

& 
& VINS final 
& 0.10400796 
& 0.98937944 
& 89.11359407 
& -0.0135931 
& -0.0881142 
& -0.0160578 \\

& 
& VINS initial + noise 
& 5 
& 5 
& 95 
& -0.0700000 
& -0.1100000 
& -0.0400000 \\

& 
& VINS final + noise 
& 0.09602733 
& 0.97642982 
& 89.12015994 
& -0.0161503 
& -0.0833035 
& -0.0217857 \\

\bottomrule
\end{tabular}
}
\caption{Raw-parameter diagnostic for Original VINS Online extrinsic calibration on EuRoC sequences with different difficulty levels. For each sequence, the source-reported ground-truth, initial, and final Euler/translation parameters are listed. Because transform direction and Euler order were not retained, these rows are not converted to common-frame $e_R/e_t$ and are used only for within-parameterization comparisons.
}
\label{tab:euroc_extrinsic_results}
\end{table*}

\textbf{Section-level finding.}
Across the EKF, KITTI VINS, EuRoC, perturbation, and cross-sequence results,
online camera--IMU calibration behaves as an initialization- and
motion-dependent local refinement process. Large initial errors can
remain near a local solution, whereas releasing a reliable calibration can move
the estimate away from the physical extrinsic. Direction-dependent behavior
is consistent with unequal numerical sensitivity among the six extrinsic
degrees of freedom, but does not prove an observability ordering. The evidence
suggests, but does not isolate causally, that a free
extrinsic state may compensate for inconsistency elsewhere in the estimator.

The cross-sequence spread cannot be explained by the nominal calibration alone;
the next section therefore examines sequence and initialization-window
conditions directly.

\section{Sequence-Dependent Failure and Initialization Conditions}
\label{sec:sequence_failure_conditions}

This section investigates the factors that affect the robustness and convergence of MonoViT, VINS, and VINS+Pose under different visual and calibration conditions. Recording~1 is used as a camera-unfriendly sequence to evaluate whether learned monocular motion priors, dense inertial measurements, and accurate camera--IMU extrinsic parameters can prevent trajectory degradation. Recording~9 provides a more visually favorable sequence and is used to examine visual observability, reinitialization, backend convergence, sensitivity to extrinsic perturbations, and local-minimum behavior.

The 10~Hz experiment with fixed accurate extrinsics is treated as a controlled baseline in which online calibration drift is excluded. The subsequent VINS-based experiments use the original 100~Hz IMU measurements. Because the 10~Hz and 100~Hz experiments also differ in initialization and system configuration, these results are not interpreted as a strict sampling-rate ablation. Instead, they are used to determine whether denser inertial measurements alone are sufficient to resolve the failures observed in Recording~1. We subsequently use optical-flow visualization and reinitialization on Recording~9 to distinguish limitations caused by weak visual observability from those caused by backend optimization.

\subsection{Camera-Unfriendly Recording 1}
\label{subsec:recording1_failure}

Recording~1 is used to evaluate the robustness of MonoViT, VINS, and VINS+Pose under visually unfavorable conditions. We first consider a controlled 10~Hz setting in which the accurate camera--IMU extrinsic parameters are fixed throughout the sequence and excluded from online optimization. Fixing the extrinsic parameters removes calibration drift as a possible source of error. Therefore, the remaining trajectory errors primarily reflect visual estimation quality, inertial propagation, temporal alignment, initialization, and state-estimation accuracy.

\begin{figure*}[tbp]
\begin{center}
\includegraphics[width=17cm, height=7cm]{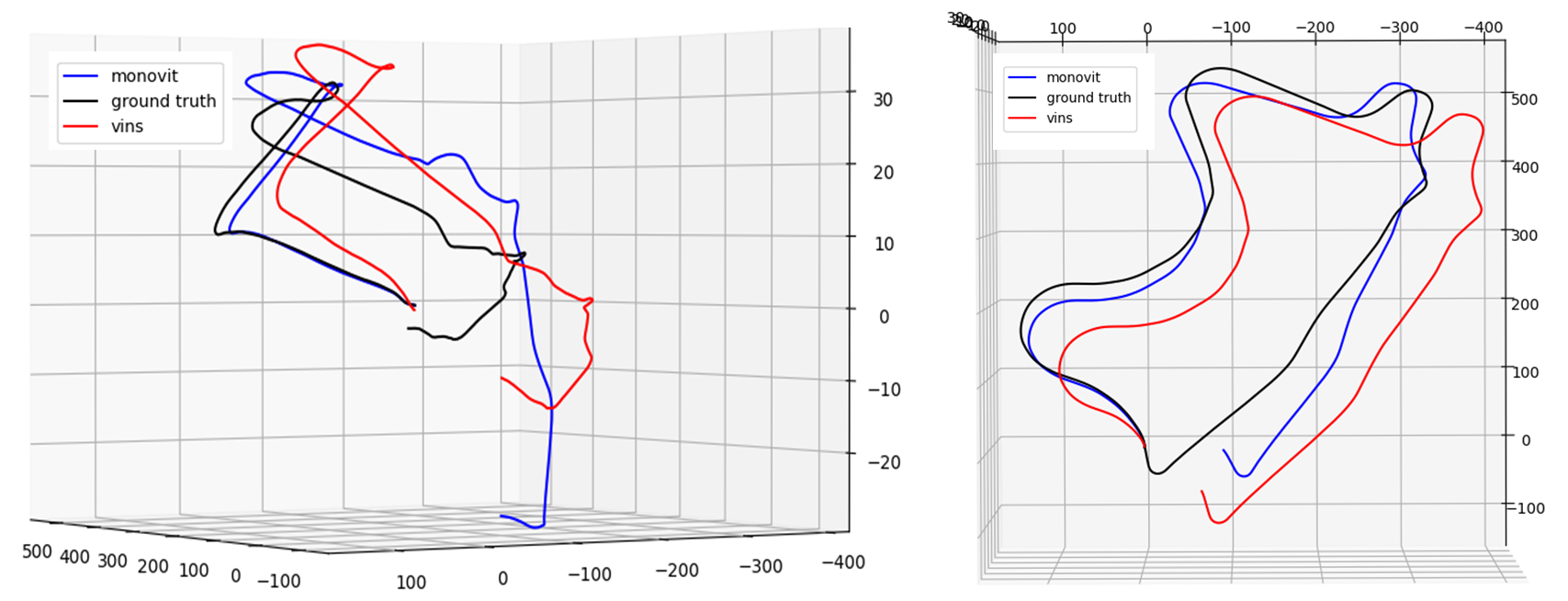}
\end{center}
\vspace{-4 mm}
\caption{Qualitative trajectory comparison with fixed accurate camera--IMU extrinsics at 10 Hz. The black, blue, and red curves denote the ground truth, MonoViT, and VINS trajectories, respectively.}
\label{fixed_extrinsic_10hz_trajectory}
\vspace{-3mm}
\end{figure*}

As shown in Fig.~\ref{fixed_extrinsic_10hz_trajectory}, neither MonoViT nor VINS perfectly recovers the reference trajectory. The VINS trajectory is relatively smooth and maintains a coherent global motion pattern, but it exhibits a systematic displacement from the reference over several long segments. MonoViT follows the ground truth more closely in some regions, but it accumulates noticeable drift, particularly in the vertical direction and toward the end of the sequence. The persistence of these errors despite using fixed accurate extrinsics demonstrates that the failure in Recording~1 cannot be attributed solely to online calibration drift.

\begin{figure*}[tbp]
\begin{center}
\includegraphics[width=17cm, height=8cm]{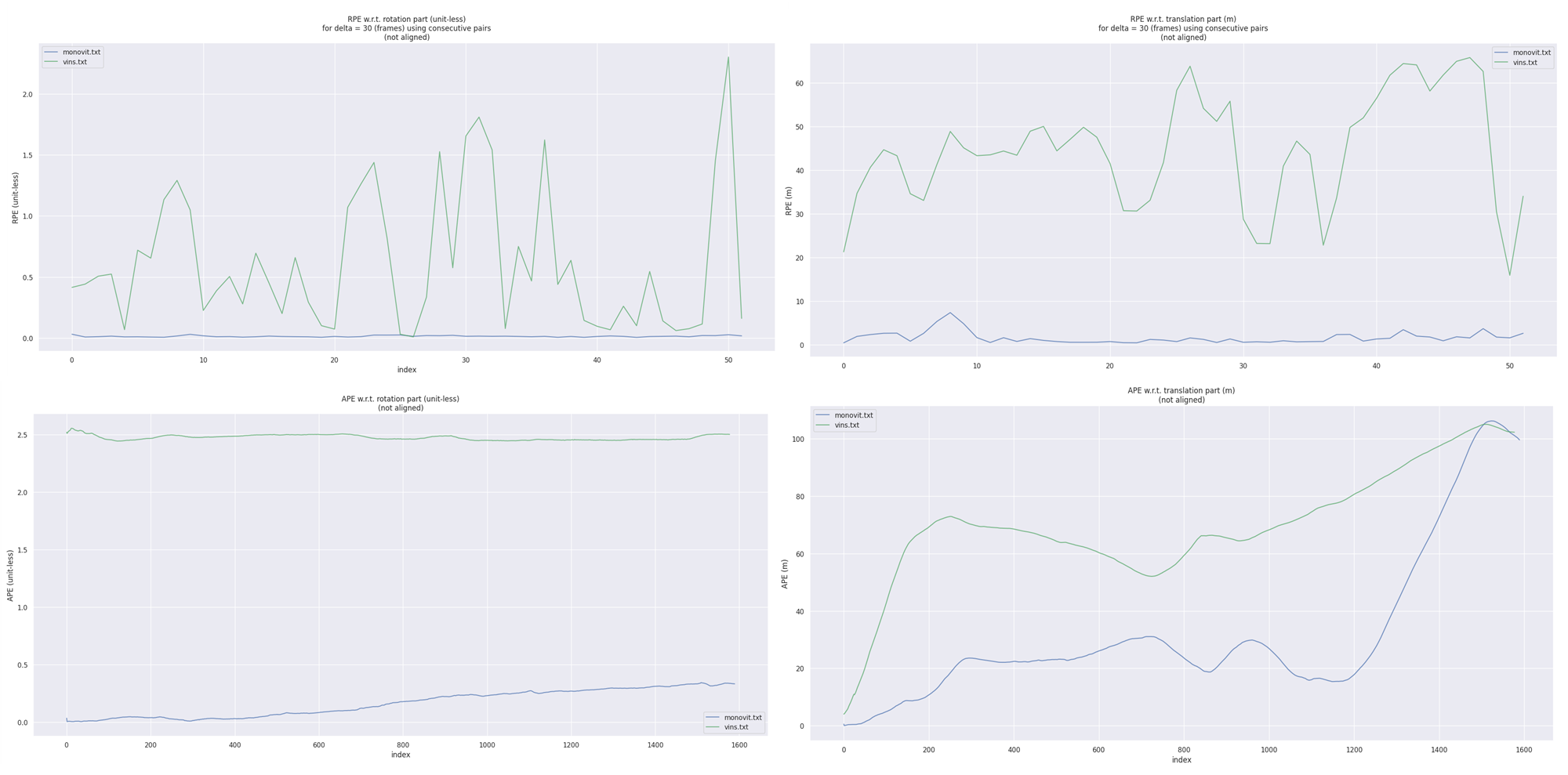}
\end{center}
\vspace{-4 mm}
\caption{RPE and APE comparison of MonoViT and VINS with fixed accurate camera--IMU extrinsics at 10 Hz.}
\label{fig:fixed_extrinsic_10hz_errors}
\vspace{-3mm}
\end{figure*}

The RPE and APE results in Fig.~\ref{fig:fixed_extrinsic_10hz_errors} reveal different local and global error characteristics. MonoViT produces lower rotational and translational RPE over most evaluated intervals, indicating relatively consistent consecutive-frame motion predictions. VINS exhibits larger local fluctuations, particularly in translation. However, MonoViT gradually accumulates monocular scale and pose drift, and its translation APE increases rapidly near the end of the sequence. In contrast, VINS develops a large global error earlier and retains a comparatively smooth but biased trajectory.

Overall, the controlled 10~Hz experiment shows that fixing a reliable camera--IMU calibration removes one source of uncertainty but does not guarantee accurate trajectory estimation. MonoViT provides useful local relative-motion cues but remains vulnerable to long-term monocular drift, while VINS produces a smoother but globally biased trajectory under this configuration.

\begin{figure}[tbp]
\begin{center}
\includegraphics[width=8.5cm, height=6cm]{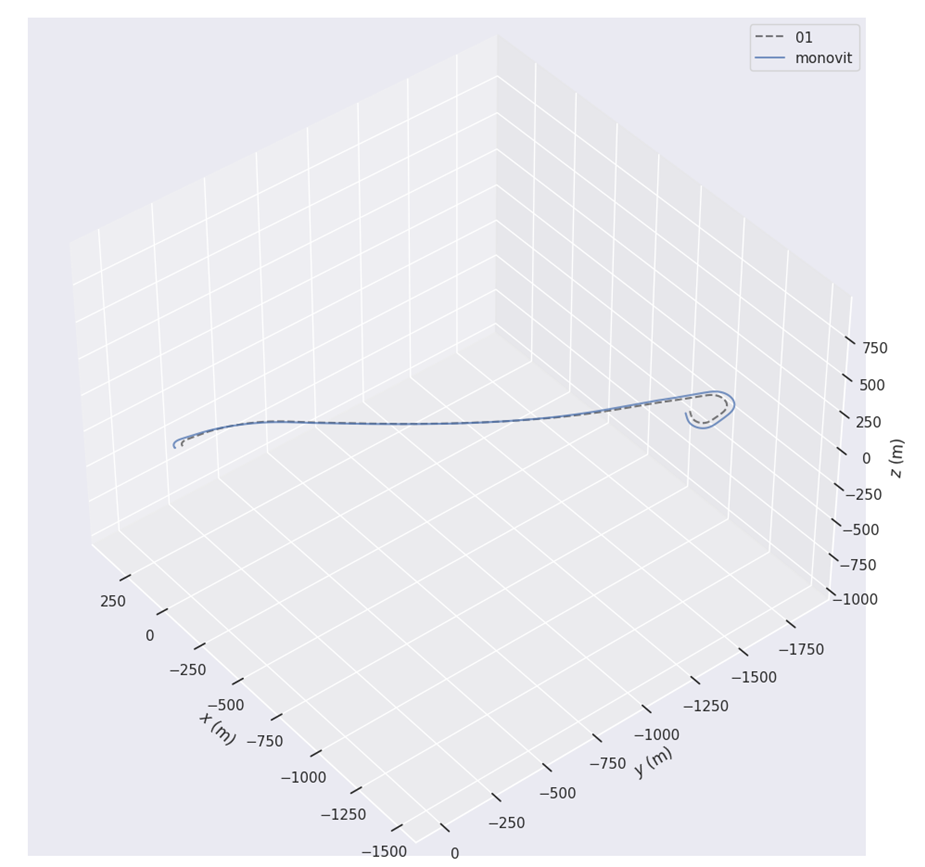}
\end{center}
\vspace{-4 mm}
\caption{MonoViT trajectory estimation on Recording 1. The blue curve denotes the MonoViT trajectory, and the dashed curve denotes the reference trajectory.
}
\label{fig:monovit_100hz_trajectory}
\vspace{-3mm}
\end{figure}

We next evaluate the same camera-unfriendly recording using the original 100~Hz IMU measurements. MonoViT does not directly use inertial data and therefore serves as a visual-only reference. As shown in Fig.~\ref{fig:monovit_100hz_trajectory}, MonoViT captures the coarse geometry and dominant direction of the route. Its estimated trajectory contains 1,101 poses and has a total length of \(2606.047~\mathrm{m}\), compared with \(2453.203~\mathrm{m}\) for the reference trajectory. The approximately \(6.2\%\) path-length overestimation indicates accumulated scale expansion in the monocular trajectory.

\begin{figure*}[tbp]
\begin{center}
\includegraphics[width=17cm, height=7cm]{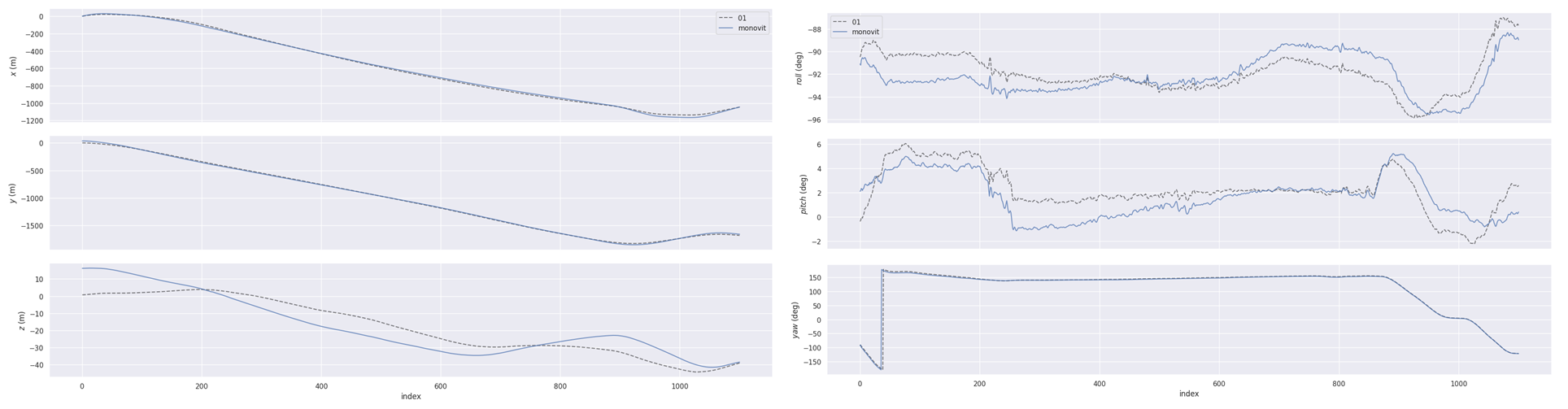}
\end{center}
\vspace{-4 mm}
\caption{Component-wise comparison between the MonoViT pose and the reference trajectory on Recording 1. The translation plots show the x-, y-, and z-position components, while the rotation plots show Roll, Pitch, and Yaw.}
\label{fig:monovit_100hz_components}
\vspace{-3mm}
\end{figure*}

The component-wise comparison in Fig.~\ref{fig:monovit_100hz_components} further shows that the predicted \(x\)- and \(y\)-translations generally follow the dominant planar motion, whereas the \(z\)-translation contains a larger and more persistent deviation. MonoViT also captures the temporal trends of Roll, Pitch, and Yaw, but residual orientation offsets remain during several motion intervals. The apparent angular discontinuities near the Euler-angle boundaries should be interpreted cautiously because they may result from angle wrapping rather than equally large physical orientation errors.

Fig.~\ref{fig:vins_100hz_trajectory} presents the corresponding VINS results using 100~Hz IMU measurements. Both the standard configuration and the accurate-extrinsic configuration remain substantially separated from the reference trajectory in the raw coordinate representation. Therefore, denser inertial measurements and accurate spatial calibration do not, by themselves, recover the missing global displacement in Recording~1. Although a higher IMU rate provides more frequent angular-velocity and acceleration measurements and can reduce discretization error in IMU preintegration, it cannot compensate for weak visual observability, unsuccessful initialization, temporal misalignment, or coordinate inconsistency.

A similar behavior is observed for VINS+Pose in Fig.~\ref{fig:vins_monovit_100hz_trajectory}. The standard and accurate-extrinsic configurations exhibit similar global deviations, indicating that the MonoViT prior has only a limited corrective influence on the backend solution under the current fusion strategy. Possible causes include weak visual geometry, inconsistent scale conventions, insufficient uncertainty weighting of the learned measurements, or disagreement between the MonoViT pose residuals and the original visual--inertial constraints.

Taken together, the 10~Hz and 100~Hz experiments show that Recording~1 remains challenging even when accurate extrinsics, denser inertial measurements, and learned pose priors are available. These results should not be interpreted as a strict conclusion about the superiority of one sampling frequency because the experimental configurations are not otherwise identical. Instead, they demonstrate that increasing the IMU measurement density alone is insufficient to resolve the observed trajectory failure. This motivates the subsequent analysis of visual observability and initialization quality.

To examine whether the observations obtained on Recording~9 generalize to a more challenging visual environment, we further evaluate MonoViT and IMU-only integration on Recording~1. Compared with Recording~9, this sequence contains extended forward-motion segments and less favorable visual geometry, resulting in weaker monocular parallax and reduced geometric observability. We therefore investigate whether MonoViT can preserve its local motion-estimation accuracy, whether inertial propagation provides a more stable alternative, and how the complementary failure modes of the two modalities may affect subsequent visual--inertial fusion.

\begin{figure*}[tbp]
\begin{center}
\includegraphics[width=15cm, height=7cm]{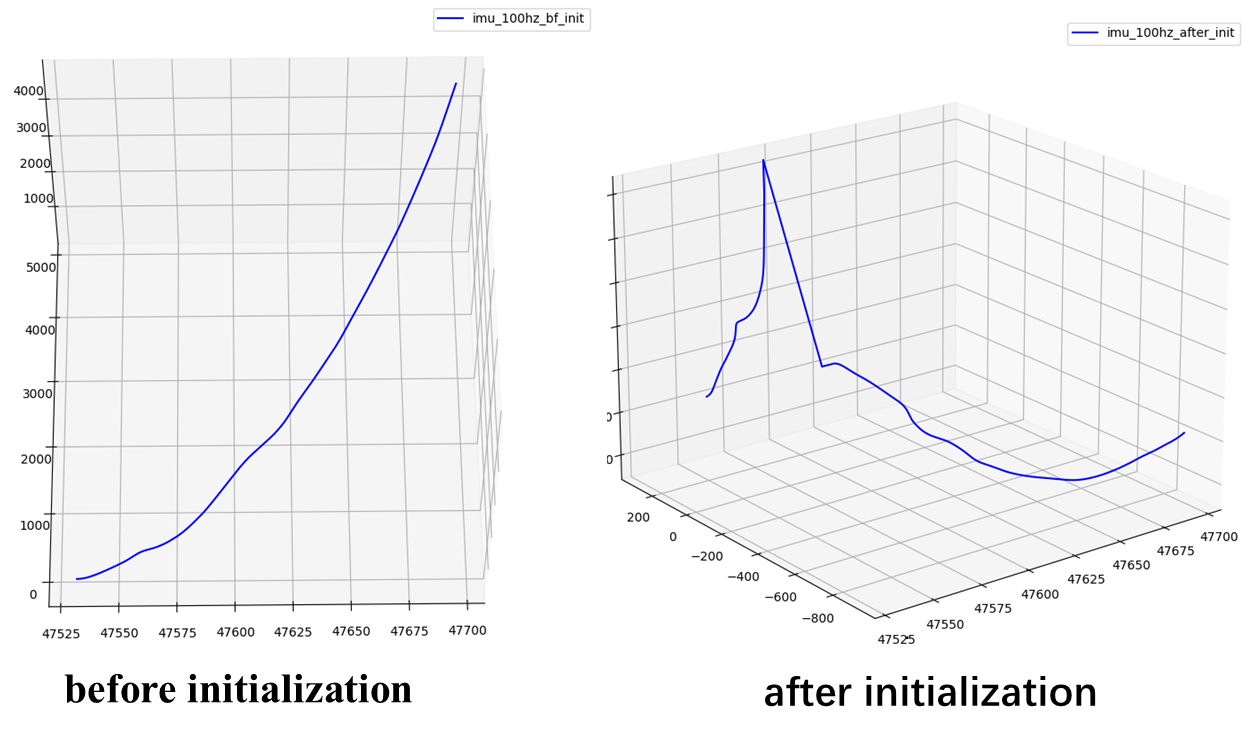}
\end{center}
\vspace{-4 mm}
\caption{Effect of VINS initialization on 100~Hz IMU propagation in Recording~1. The left panel shows the trajectory obtained before visual--inertial initialization, while the right panel shows the trajectory after estimating the initial orientation, gravity direction, velocity, and IMU biases. }
\label{fig:record1_vins_imu_initialization}
\vspace{-3mm}
\end{figure*}

Fig.~\ref{fig:record1_vins_imu_initialization} first illustrates the influence of VINS initialization on 100~Hz IMU propagation. Before initialization, direct integration produces a trajectory with a rapidly increasing displacement and an implausible global scale. After the initialization stage, the trajectory is substantially reshaped because the initial attitude, gravity direction, velocity, and IMU biases are estimated and incorporated into the propagated state. This result demonstrates that a visually smooth raw IMU trajectory is not necessarily physically consistent with the state representation used by the VINS backend. Consequently, pre-initialization IMU integration should not be used directly to judge whether inertial measurements can improve the subsequent joint optimization.

\begin{figure*}[tbp]
\begin{center}
\includegraphics[width=17cm, height=9cm]{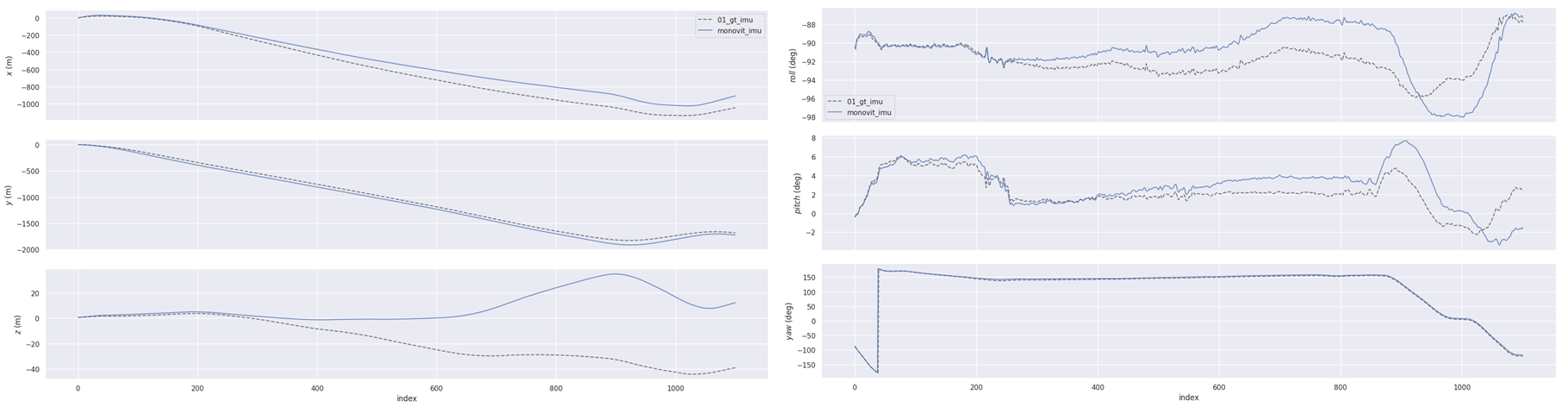}
\end{center}
\vspace{-4 mm}
\caption{Component-wise MonoViT pose estimation on the camera-unfriendly Recording~1. The left column compares the estimated and reference \(x\)-, \(y\)-, and \(z\)-translations, while the right column compares Roll, Pitch, and Yaw.}
\label{fig:record1_monovit_components}
\vspace{-3mm}
\end{figure*}

The component-wise MonoViT results are presented in Fig.~\ref{fig:record1_monovit_components}. MonoViT reproduces the dominant trends of the \(x\)- and \(y\)-translations over a large portion of the sequence. Nevertheless, the estimated trajectory gradually separates from the reference as the traveled distance increases. The most pronounced discrepancy appears in the \(z\)-component. During the later part of the sequence, the estimated height increases to a large positive value, whereas the reference height decreases below its initial level. The opposite vertical evolution indicates that monocular scale and translation-direction errors accumulate strongly in the weakly constrained vertical dimension.

The orientation estimates are more stable than the translational estimates. Roll and Pitch preserve several low-frequency variations of the reference, although larger deviations occur near the end of the sequence. Yaw follows the dominant heading evolution over most intervals. The apparent discontinuities around \(\pm180^\circ\) are mainly caused by Euler-angle wrapping and should not be interpreted as instantaneous physical rotations of the same magnitude. Overall, MonoViT retains reasonable short-term orientation consistency, but its long-term translation estimate, especially along the vertical axis, is substantially less stable.

\begin{figure*}[tbp]
\begin{center}
\includegraphics[width=17cm, height=12cm]{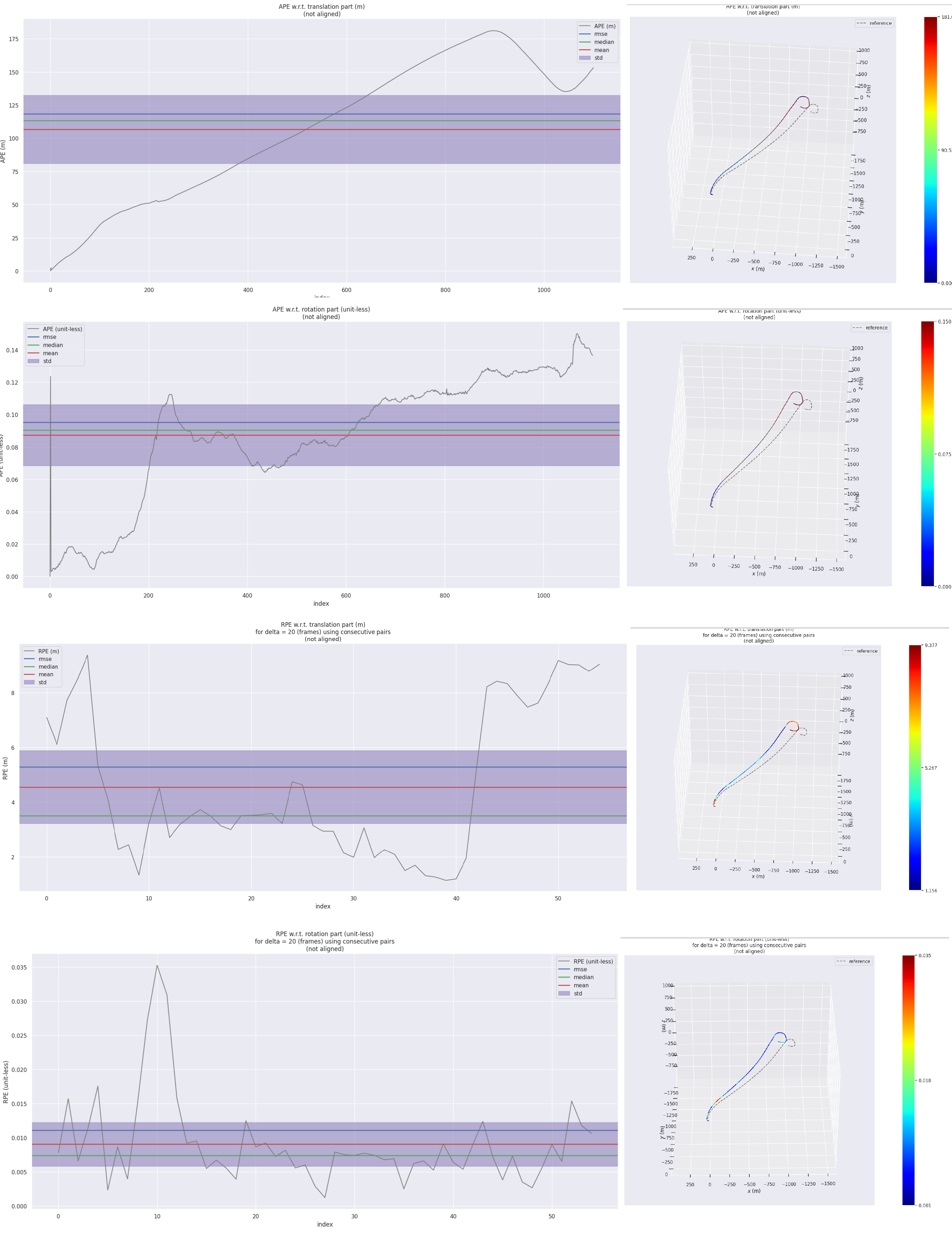}
\end{center}
\vspace{-4 mm}
\caption{Absolute and relative pose errors of MonoViT on the camera-unfriendly Recording~1. The first two rows show translation and rotation APE and their spatial distributions, while the last two rows show the corresponding 20-frame translation and rotation RPE. }
\label{fig:record1_monovit_errors}
\vspace{-3mm}
\end{figure*}

The complete APE and RPE evaluation in Fig.~\ref{fig:record1_monovit_errors} provides a clearer characterization of this behavior. Translation APE increases almost continuously during the first approximately 900 frames and reaches a maximum of about \(181\)~m before decreasing moderately near the end of the sequence. The error-colored trajectory shows that the largest deviations occur along the later portion of the long forward-motion segment and near the terminal loop. This substantial growth demonstrates that the monocular estimate accumulates severe global scale and direction drift even though the component-wise motion remains qualitatively plausible.

Rotation APE exhibits a similar but less severe long-term trend. It begins close to zero, increases progressively with traveled distance, and reaches approximately \(0.15\). Unlike the nearly constant rotation APE observed in some IMU-based evaluations, this gradual growth is more consistent with accumulated orientation drift than with a fixed coordinate-frame offset.

In contrast, the 20-frame RPE remains considerably smaller than the corresponding APE. Translation RPE varies from approximately \(1.2\) to \(9.4\)~m, with most intervals concentrated around \(3\)--\(5\)~m. The largest peaks occur near the beginning and end of the sequence, suggesting that these portions contain more difficult motion or weaker visual constraints. Rotation RPE remains low over most intervals, with a maximum of approximately \(0.035\). The substantial gap between APE and RPE is therefore central to interpreting the result: MonoViT provides locally smooth and comparatively consistent relative motion, particularly in rotation, but small scale, heading, and translation errors accumulate over the long sequence and eventually produce severe global trajectory deformation.

\begin{figure*}[tbp]
\begin{center}
\includegraphics[width=16cm, height=6cm]{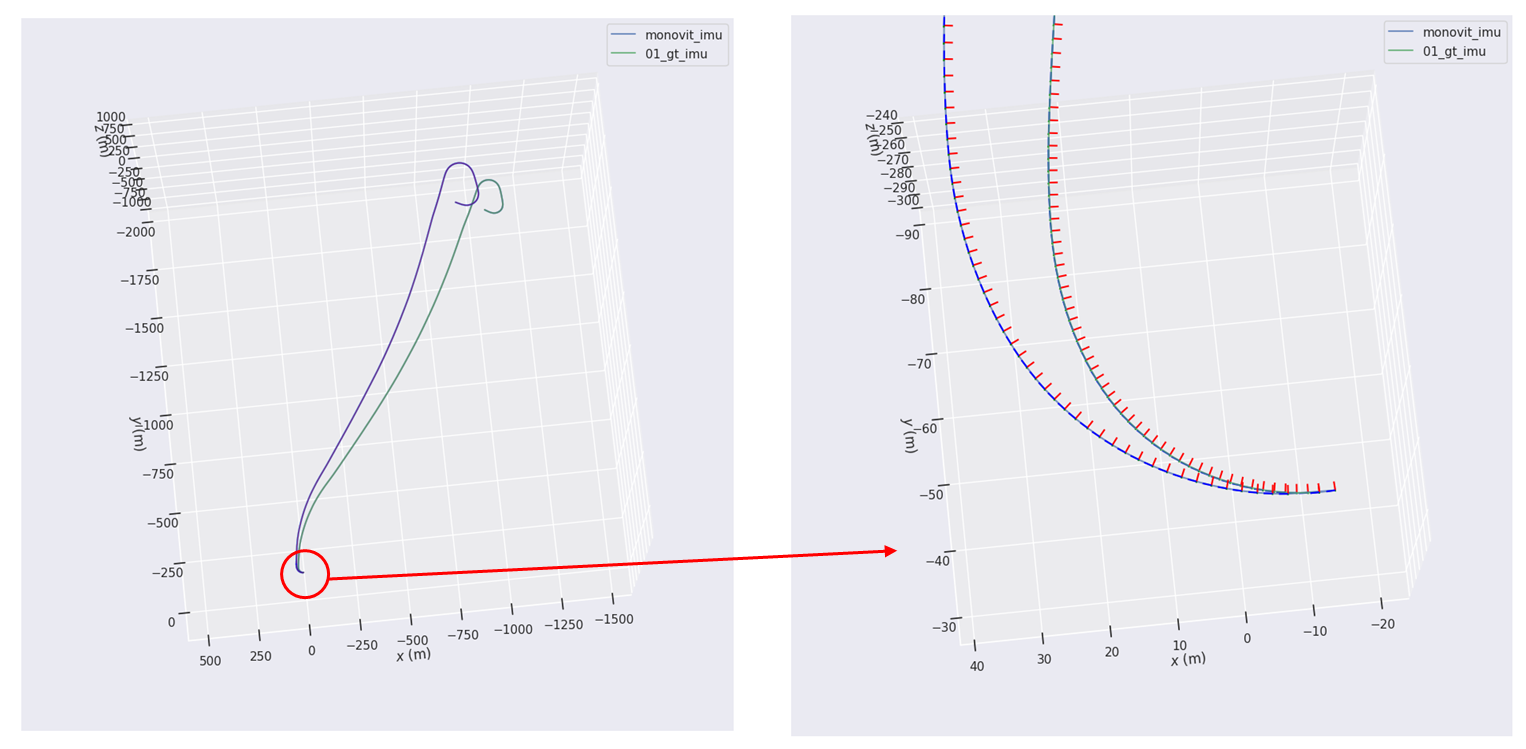}
\end{center}
\vspace{-4 mm}
\caption{Qualitative comparison between the MonoViT and reference trajectories on Recording~1. The left panel shows the complete three-dimensional trajectories, and the right panel enlarges a representative local segment together with the estimated orientations. }
\label{fig:record1_monovit_trajectory}
\vspace{-3mm}
\end{figure*}

The qualitative trajectory comparison in Fig.~\ref{fig:record1_monovit_trajectory} is consistent with the quantitative results. MonoViT follows the dominant direction of the long route, and the enlarged local segment shows similar curvature and orientation to the reference. However, the two trajectories gradually separate over the extended forward-motion section and do not recover the same terminal loop. Thus, local agreement after short intervals does not imply globally accurate trajectory recovery.

\begin{figure*}[tbp]
\begin{center}
\includegraphics[width=17cm, height=5cm]{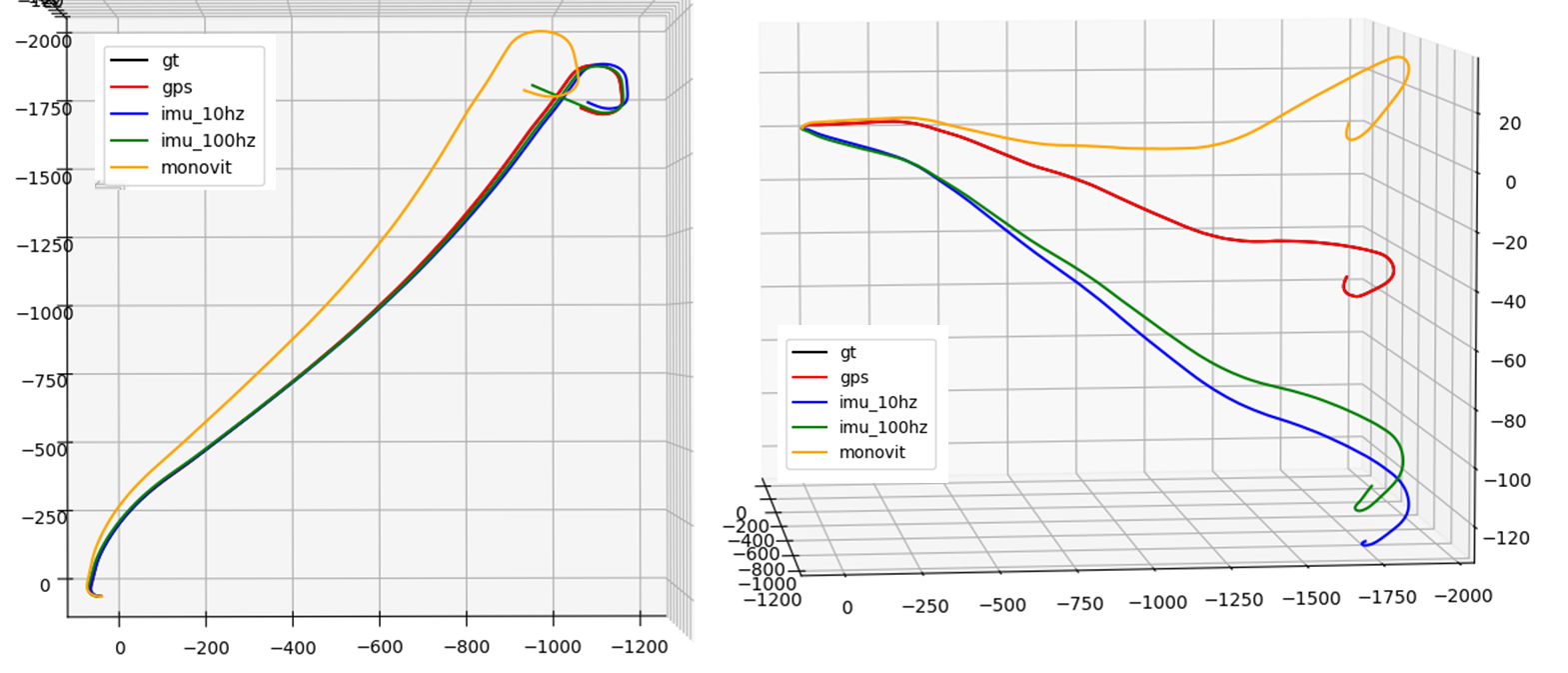}
\end{center}
\vspace{-4 mm}
\caption{Qualitative comparison of standalone motion estimates on Recording~1. The two views compare the ground truth, GPS, 10~Hz IMU integration, 100~Hz IMU integration, and MonoViT. }
\label{fig:record1_standalone_trajectory}
\vspace{-3mm}
\end{figure*}

Fig.~\ref{fig:record1_standalone_trajectory} compares MonoViT with GPS and 10~Hz and 100~Hz IMU integration. In the near top-down view, GPS and the ground-truth trajectory nearly overlap. Both inertial trajectories preserve the principal planar route more closely than MonoViT, whereas MonoViT develops a noticeable lateral offset along the long forward-motion segment. The 100~Hz IMU trajectory is slightly closer to the reference than the 10~Hz result in several regions, although the difference between the two inertial estimates is smaller than their difference from MonoViT.

The three-dimensional view reveals complementary vertical failure modes. MonoViT accumulates a positive height drift, whereas both IMU trajectories drift in the negative vertical direction. The 100~Hz result remains slightly closer to the reference than the 10~Hz trajectory, but neither sampling rate eliminates the accumulated height error. These observations suggest that the visually challenging scene causes stronger scale and direction deformation in the monocular estimate, while inertial integration retains the dominant planar motion more reliably but remains vulnerable to gravity-compensation, accelerometer-bias, and initial-attitude errors.

\begin{figure*}[tbp]
\begin{center}
\includegraphics[width=17cm, height=9cm]{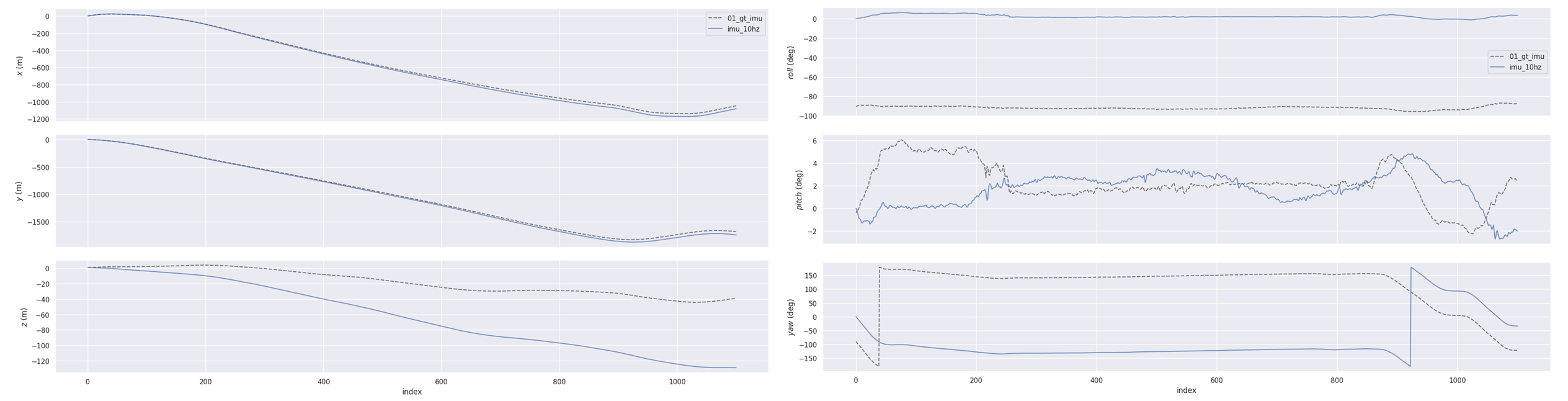}
\end{center}
\vspace{-4 mm}
\caption{Component-wise comparison between 10~Hz IMU-only integration and the reference trajectory on Recording~1. The \(x\)- and \(y\)-translations reproduce the dominant forward motion, whereas the \(z\)-component accumulates severe negative drift. }
\label{fig:record1_imu10_components}
\vspace{-3mm}
\end{figure*}

The component-wise 10~Hz IMU results in Fig.~\ref{fig:record1_imu10_components} further support this interpretation. The estimated \(x\)- and \(y\)-translations remain close to the reference during most of the sequence, indicating that inertial propagation captures the dominant forward displacement more consistently than MonoViT in this recording. In contrast, the \(z\)-component continuously decreases and reaches approximately \(-120\)~m, while the reference remains near \(-40\)~m. This divergence is characteristic of double-integrated accelerometer and gravity-compensation errors.

The angular comparison contains a persistent Roll offset of approximately \(90^\circ\), together with a large Yaw offset and angular wrapping discontinuities. These systematic differences indicate that the estimated and reference orientations may use different initial frames, axis definitions, or Euler-angle conventions. Therefore, the absolute Euler-angle discrepancy should not be attributed entirely to IMU drift. All orientations should first be transformed into a common coordinate frame and evaluated using the geodesic rotation distance on \(\mathrm{SO}(3)\).

\begin{figure*}[tbp]
\begin{center}
\includegraphics[width=17cm, height=12cm]{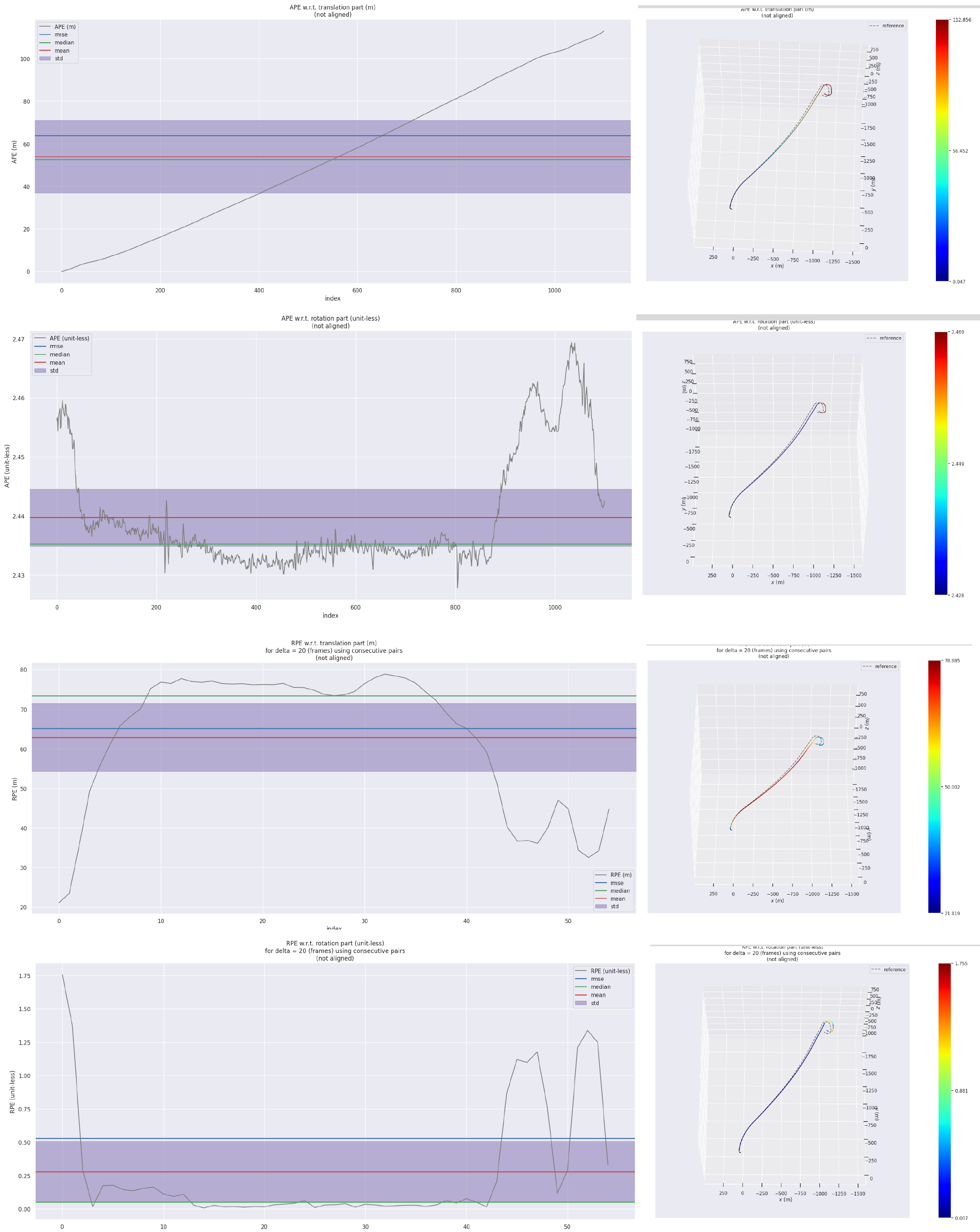}
\end{center}
\vspace{-4 mm}
\caption{Absolute and relative pose errors of 10~Hz IMU-only integration on Recording~1. The figure reports translation and rotation APE and 20-frame translation and rotation RPE without trajectory alignment.}
\label{fig:record1_imu10_errors}
\vspace{-3mm}
\end{figure*}

The quantitative IMU results in Fig.~\ref{fig:record1_imu10_errors} show that translation APE grows almost monotonically and reaches approximately \(113\)~m by the end of the recording. The error-colored trajectory confirms that the error increases progressively along the long forward-motion segment rather than resulting from a single isolated failure. Rotation APE remains within a narrow range of approximately \(2.43\)--\(2.47\), which is consistent with the systematic orientation-frame offset observed in the component-wise curves.

The 20-frame translation RPE is also large and reaches approximately \(79\)~m during the most difficult intervals. Rotation RPE remains small during much of the middle part of the sequence but contains pronounced peaks at the beginning and near the end. Therefore, unlike MonoViT, whose local translation RPE remains below approximately \(9.4\)~m, the 10~Hz IMU estimate contains substantial short-term propagation error in addition to accumulated global drift.

Comparing the two modalities reveals a clear distinction between their failure modes. MonoViT has a larger maximum translation APE, reaching approximately \(181\)~m compared with approximately \(113\)~m for 10~Hz IMU integration. However, MonoViT achieves substantially lower translation RPE, remaining below approximately \(9.4\)~m, whereas the IMU translation RPE approaches \(79\)~m. MonoViT is therefore considerably more reliable for short-term relative motion, but its small local scale and direction errors accumulate into severe global deformation. In contrast, IMU integration preserves the dominant long-range planar direction more consistently but suffers from bias-driven local propagation errors and severe vertical drift.

These complementary errors help explain why directly incorporating MonoViT and IMU measurements into VINS does not necessarily improve the final estimate. MonoViT supplies locally smooth but globally biased visual motion, whereas the IMU supplies stronger long-range forward-motion information but contains substantial bias- and gravity-induced drift. If the two measurements are not expressed using consistent coordinate, scale, temporal, and uncertainty conventions, the backend may compensate by changing the trajectory, IMU biases, or camera--IMU extrinsic parameters. Recording~1 therefore demonstrates that robust visual--inertial fusion requires not only multiple sensing modalities, but also reliable initialization, coordinate unification, observability-aware weighting, and rejection of unreliable visual constraints.

\subsection{Favorable Recording 9 and Frame-300 Reinitialization}
\label{subsec:recording9_reinitialization}

To determine whether the failure in Recording~1 originates primarily from the visual front end or from the backend optimizer, we compare the optical-flow distributions in Recording~1 and Recording~9. We then restart Recording~9 from frame~300, where the visual geometry and motion excitation are more favorable, and examine whether the VINS backend can reach a stable calibration solution. The VINS-based experiments in this part use the original 100~Hz IMU measurements.

\begin{figure*}[tbp]
\begin{center}
\includegraphics[width=17cm, height=7cm]{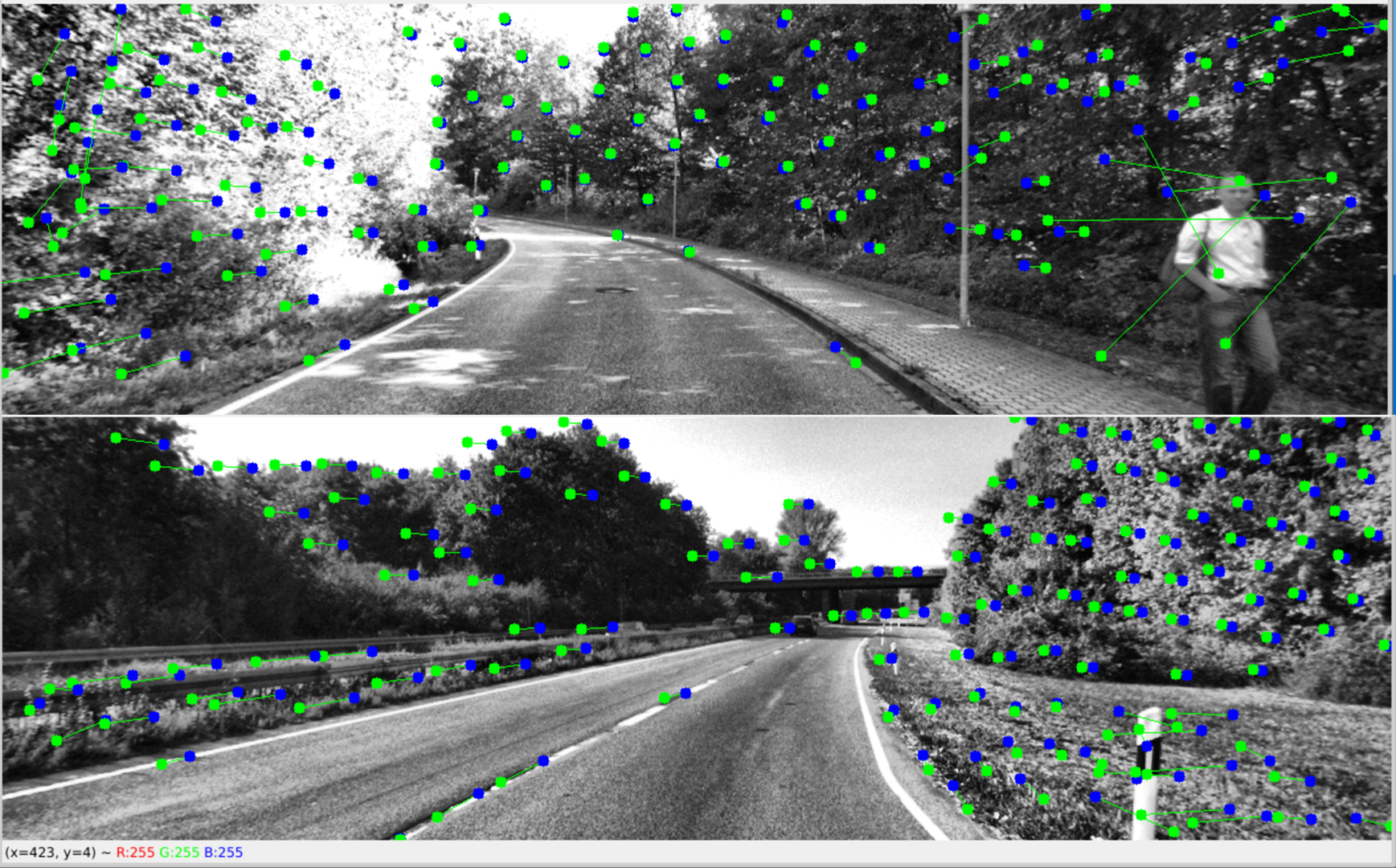}
\end{center}
\vspace{-4 mm}
\caption{Optical-flow comparison between Recording 9 and Recording 1. The top panel shows consecutive-frame feature tracks in Recording 9, and the bottom panel shows those in Recording 1.}
\label{fig:recording_optical_flow}
\vspace{-3mm}
\end{figure*}

As shown in Fig.~\ref{fig:recording_optical_flow}, both recordings contain a substantial number of tracked features, but their geometric quality differs. In Recording~9, the feature tracks are distributed over nearby vegetation, road boundaries, poles, and background structures. The curved road and the presence of nearby objects provide greater depth variation and stronger parallax. Several tracks are located on a moving pedestrian and may introduce inconsistent correspondences if they are not rejected by the front-end outlier-removal procedure. Nevertheless, the overall scene contains informative static structures at different depths.

Recording~1 also contains many tracked points, particularly on roadside vegetation and guardrails. However, the central road region contains relatively little texture, and many tracks lie on distant or peripheral structures. Under dominant forward motion, such features may provide limited parallax for scale, depth, and extrinsic estimation. Therefore, feature count alone is not sufficient to characterize visual--inertial observability. The spatial distribution, depth diversity, motion parallax, and static-scene consistency of the tracked features are equally important.

\begin{figure*}[tbp]
\begin{center}
\includegraphics[width=17cm, height=9cm]{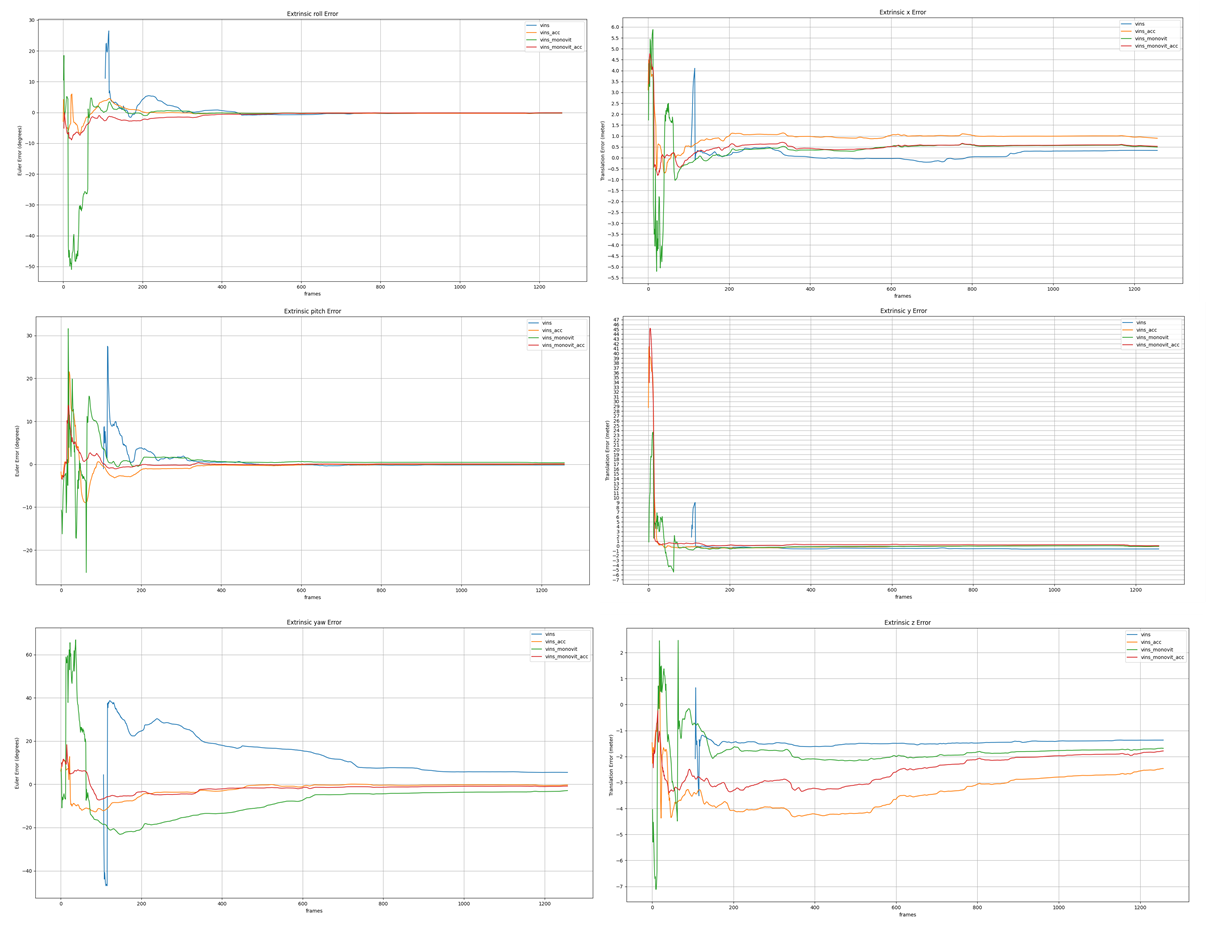}
\end{center}
\vspace{-4 mm}
\caption{Camera--IMU extrinsic convergence after reinitializing Recording 9 from frame 300 using 100 Hz IMU measurements. The figure compares baseline VINS, VINS initialized with accurate extrinsics, VINS+Pose, and VINS+Pose initialized with accurate extrinsics.}
\label{fig:recording9_reinitialization_convergence}
\vspace{-3mm}
\end{figure*}

Based on this comparison, Recording~9 is restarted from frame~300 to exclude the less favorable initial interval and provide a more informative initialization window. Fig.~\ref{fig:recording9_reinitialization_convergence} compares baseline VINS, VINS initialized with accurate extrinsics, VINS+Pose, and VINS+Pose initialized with accurate extrinsics.
All configurations exhibit substantial initialization transients, but the Roll and Pitch errors gradually decrease after sufficient visual--inertial excitation has been accumulated. The accurate-extrinsic configurations generally exhibit smaller rotational transients, suggesting that reliable initial calibration improves the stability of the early optimization process. Nevertheless, the convergence trajectories remain method dependent.

Yaw is more sensitive to initialization and measurement configuration in the
evaluated motion and sequences. The baseline Original VINS retains a noticeable
positive residual for a large portion of the sequence, whereas VINS+Pose
undergoes a large early deviation and approaches a stable value more slowly. The
accurate-extrinsic configurations remain closer to zero after the initial
transient. This sequence-specific behavior is consistent with weaker yaw
constraint than for Roll and Pitch; it is not claimed as a universal
observability ordering.

The translation components also exhibit direction-dependent numerical sensitivity. The \(y\)-translation errors of all configurations become relatively small after initialization, whereas the \(x\)- and particularly \(z\)-translation components converge to different non-zero plateaus. Accurate extrinsic initialization does not uniformly eliminate these residuals. This sequence-specific pattern is not used as a formal observability result. In some cases, a configuration initialized with accurate extrinsics exhibits a larger residual in one translation direction than the baseline, consistent with compensation among visual errors, temporal alignment, scale, bias, and an active extrinsic state.

The improved stability after restarting Recording~9 suggests that the backend is capable of convergence when a more informative visual and motion interval is available. Therefore, the earlier failure in Recording~1 is more likely associated with weak visual geometry and poorly conditioned initialization than with a complete inability of the backend optimizer to converge. Nevertheless, stabilization of the state estimates does not guarantee convergence to the physically correct extrinsic calibration.

\subsection{Visual-Track Diagnostics}
\label{subsec:visual_track_diagnostics}

\begin{figure*}[tbp]
\centering
    \includegraphics[width=17cm, height=10cm]{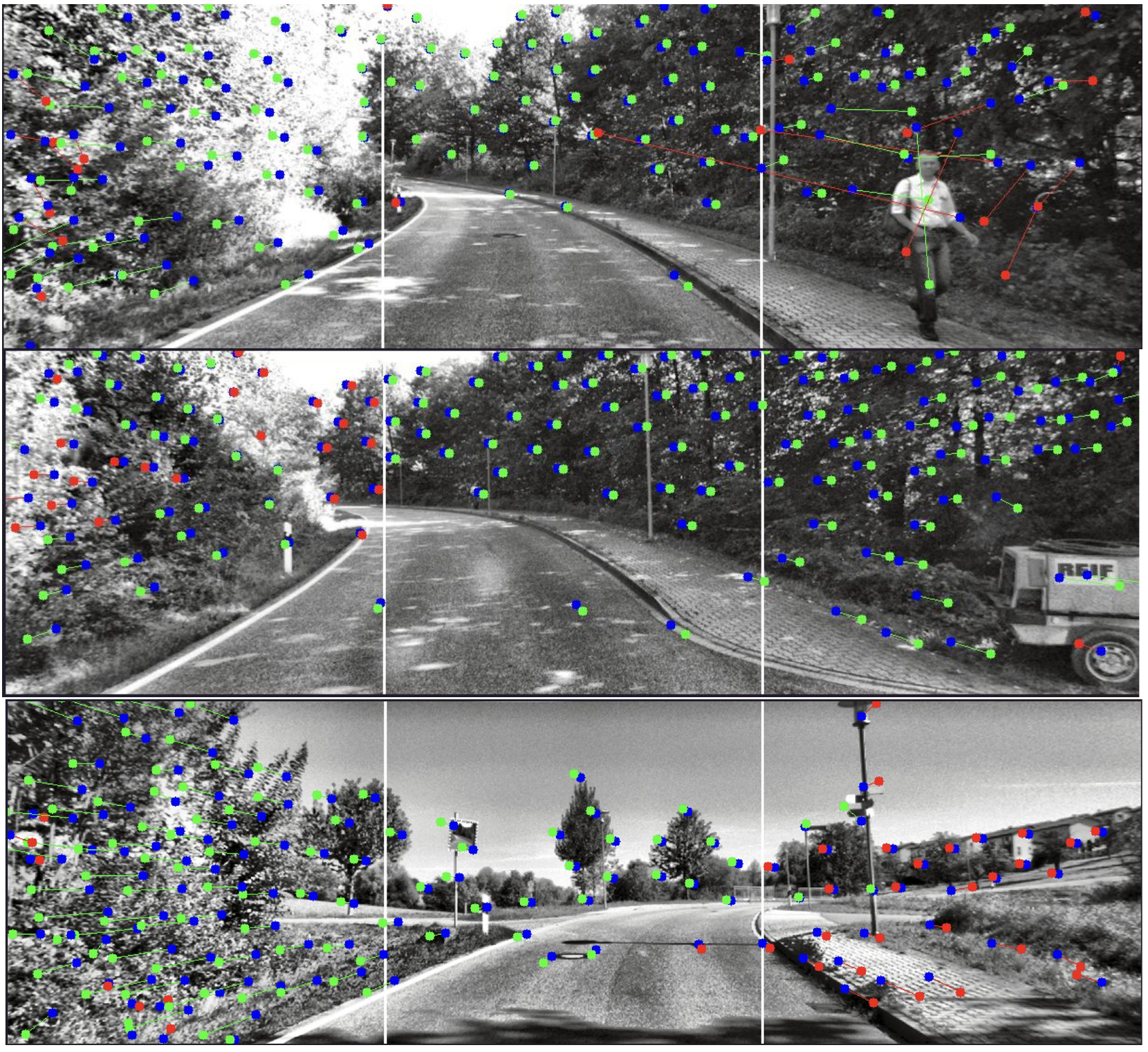}
\caption{Representative tracks and candidate outliers. Vegetation, dynamic objects,
and low-texture forward motion produce candidate outliers.}
\label{fig:optical_flow_outlier_examples}
\end{figure*}

\begin{figure*}[tbp]
\centering
\includegraphics[width=17cm, height=10cm]{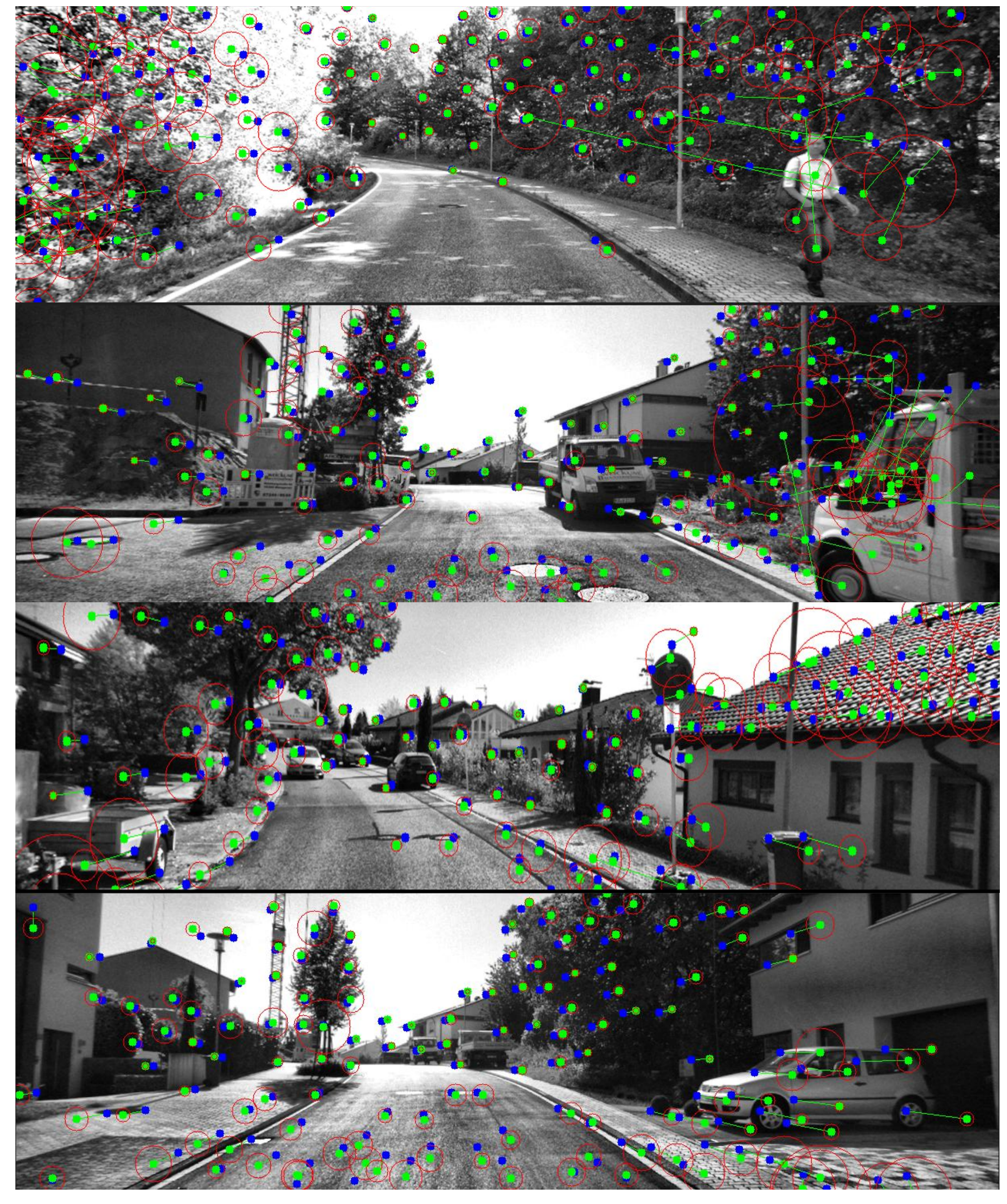}
\caption{Spatially nonuniform tracking uncertainty. Track density and uncertainty
vary across the image.}
\label{fig:optical_flow_error_visualization}
\end{figure*}

\begin{figure*}[tbp]
\centering
\includegraphics[width=17cm, height=4cm]{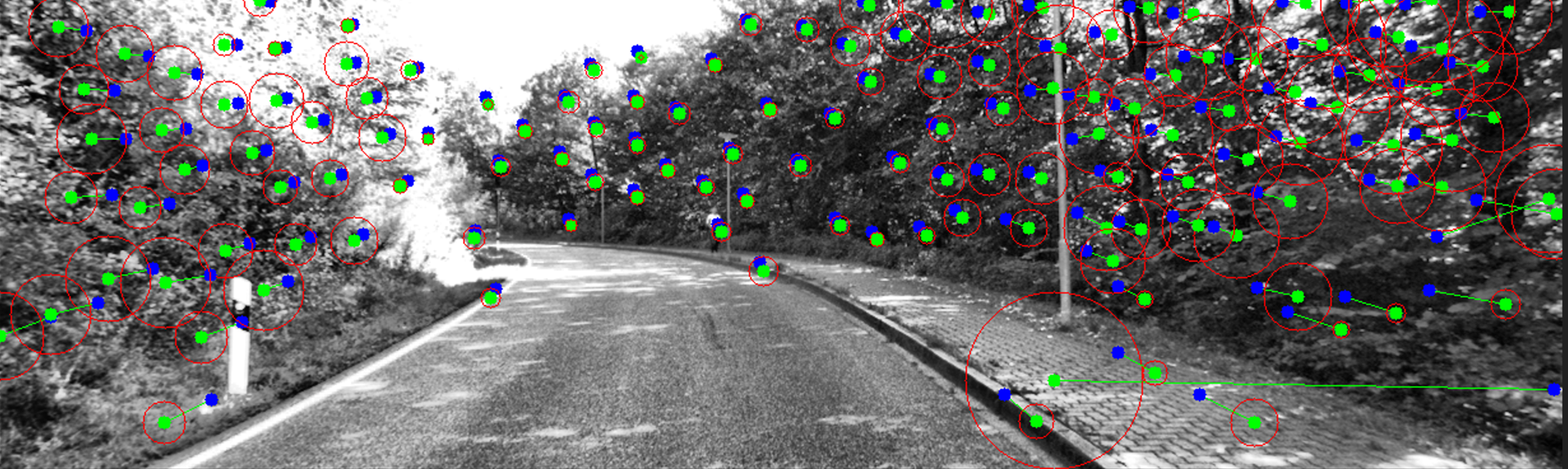}
\caption{Feature clustering and inconsistent boundary tracks. A challenging road
frame contains sparse constraints near the road center and inconsistent tracks
near boundaries.}
\label{fig:optical_flow_difficult_example}
\end{figure*}

We use visual tracks only as qualitative diagnostic evidence. The examples in
Fig.~\ref{fig:optical_flow_outlier_examples}, Fig.~\ref{fig:optical_flow_error_visualization}, Fig.~\ref{fig:optical_flow_difficult_example} show that many tracks concentrate
on vegetation, image boundaries, and roadside texture, while the distant road
center contains fewer constraints. Recording~9 contains stronger depth variation
and parallax than Recording~1, although both include potentially unreliable
features. Track count alone therefore does not measure the geometric information
available for initialization.
Dynamic pedestrians and vehicles can also produce coherent motion that violates
the static-scene assumption, while isolated displacements near occlusions, thin
structures, curbs, and vegetation boundaries reveal spatially nonuniform
tracking uncertainty. These observations are consistent with visual constraints
contributing to the cross-sequence variability, but the visualization does not
isolate a single causal source and is not used to rank estimators.
The observed patterns motivate adaptive residual normalization, geometric
verification, and dynamic-region rejection. These remain future system-design
directions rather than evaluated components of the present study.

\textbf{Section-level finding.}
The cross-recording and reinitialization results show that feature count is not
equivalent to visual--inertial observability. Accurate extrinsics, a denser IMU
stream, and a learned pose prior cannot compensate for a poorly conditioned
initialization interval. The optical-flow evidence provides only a qualitative
account of this variability; the suggested filtering remedies were not tested
here.
These failures motivate the final mechanism-oriented analysis of why locally
smooth and numerically converged estimates can remain globally and physically
incorrect.

\section{Backend State Coupling, Scale Consistency, and Global Correction}
\label{sec:backend_scale_loop_analysis}

This section brings together experiments that distinguish local temporal
consistency, numerical convergence, global trajectory accuracy, and physical
calibration correctness.

\subsection{Local Minima, Late Extrinsic Injection, and Backend Cost}
\label{subsec:local_minima_backend_cost}

After confirming that the backend can stabilize when Recording~9 is reinitialized from a visually favorable interval, we investigate whether the resulting solution corresponds to the physically correct calibration or merely to a locally consistent minimum. The archived plot identifies three initialization settings as \(2^\circ/2~\mathrm{cm}\), \(6^\circ/6~\mathrm{cm}\), and \(10^\circ/10~\mathrm{cm}\). The configuration that generated the plot is not available, so it cannot be verified whether each label denotes a per-axis offset, a geodesic rotation magnitude and translation norm, or another construction. We therefore retain these values as nominal setting identifiers and do not reinterpret them as exact initial \(e_R\) and \(e_t\). We also consider accurate-extrinsic initialization with continued Online updating and late injection of the accurate extrinsic after the backend has already entered a stable state.

\begin{figure*}[tbp]
\begin{center}
\includegraphics[width=17cm, height=9cm]{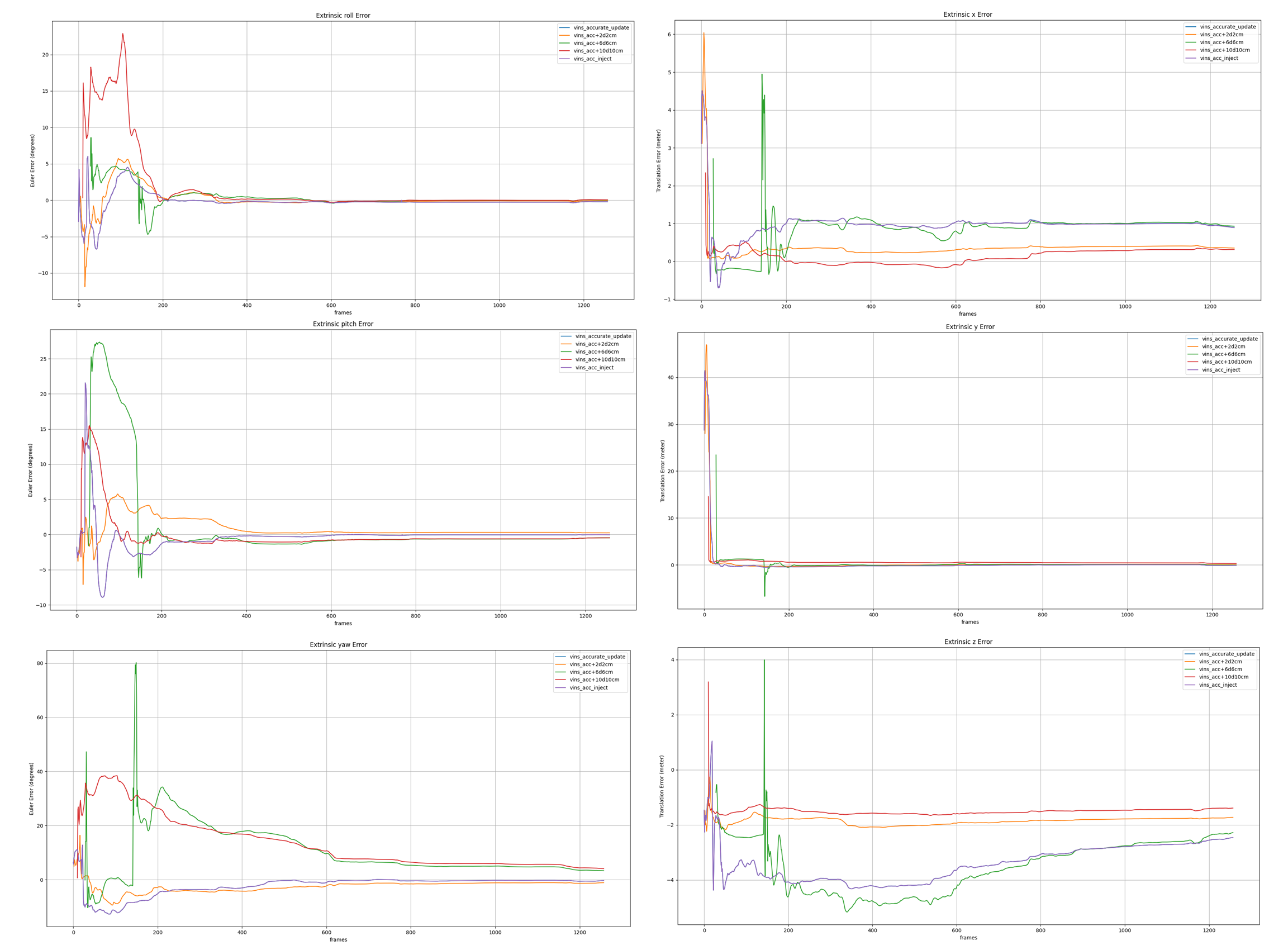}
\end{center}
\vspace{-4 mm}
\caption{Robustness of VINS online camera--IMU calibration to extrinsic perturbations and late accurate-extrinsic injection. The figure shows the Roll, Pitch, and Yaw rotation errors and the \(x\)-, \(y\)-, and \(z\)-translation errors for accurate-extrinsic initialization with online updating, initial perturbations of \(2^\circ/2~\mathrm{cm}\), \(6^\circ/6~\mathrm{cm}\), and \(10^\circ/10~\mathrm{cm}\), and late injection of the accurate extrinsic after backend convergence. Roll, Pitch, and \(y\)-translation generally recover after the initialization transient, whereas Yaw and the \(x\)- and \(z\)-translation components retain initialization-dependent residual errors.}
\label{fig:extrinsic_perturbation_convergence}
\vspace{-3mm}
\end{figure*}

As shown in Fig.~\ref{fig:extrinsic_perturbation_convergence}, the rotational components exhibit markedly different convergence characteristics. The Roll error contains substantial initialization transients under all settings, with the largest temporary deviation occurring for the \(10^\circ/10~\mathrm{cm}\) perturbation. Nevertheless, the Roll estimates gradually approach a narrow region around zero after approximately 200--400 frames. The Pitch component shows similar recovery behavior, although the \(6^\circ/6~\mathrm{cm}\) initialization produces a particularly large and prolonged transient. These results indicate that Roll and Pitch can recover from moderate initialization errors when sufficiently informative visual--inertial motion is available.

Yaw is substantially more sensitive in this run. The \(6^\circ/6~\mathrm{cm}\) and \(10^\circ/10~\mathrm{cm}\) settings produce large early deviations and require a considerably longer interval to approach stable solutions. Even after convergence, several settings retain non-zero residual errors. The final Yaw error is not monotonically related to the nominal perturbation setting, suggesting that the backend may converge to different local solutions depending on the initialization trajectory. This is consistent with weaker numerical sensitivity of Yaw in the evaluated sequence, but it is not an observability proof.

The translation components show similarly direction-dependent behavior. The \(y\)-translation error is rapidly reduced under all tested initializations and remains close to zero for most of the sequence, indicating greater numerical sensitivity in this direction for this run. In contrast, the \(x\)- and \(z\)-translation estimates converge to different non-zero plateaus. Some perturbed configurations achieve a smaller residual in one translation component than the accurate-update configuration while remaining less accurate in another. The absence of a monotonic relationship between nominal setting and final error is consistent with multiple locally consistent solutions; it does not establish formal component-wise observability.

The late accurate-extrinsic injection result provides further evidence of coupling among the backend states. Replacing the extrinsic state with the accurate calibration after the backend has already stabilized does not cause all rotational and translational errors to immediately converge to zero. Instead, the injected configuration gradually approaches a biased solution similar to those reached by some perturbed initializations. This suggests that the trajectory, velocity, scale, and IMU-bias states have already adapted to the previous calibration. Updating the extrinsic state alone therefore introduces a new inconsistency, and the optimizer may partially move it back toward the previous local solution.

\begin{figure}[tbp]
\begin{center}
\includegraphics[width=8.5cm, height=6cm]{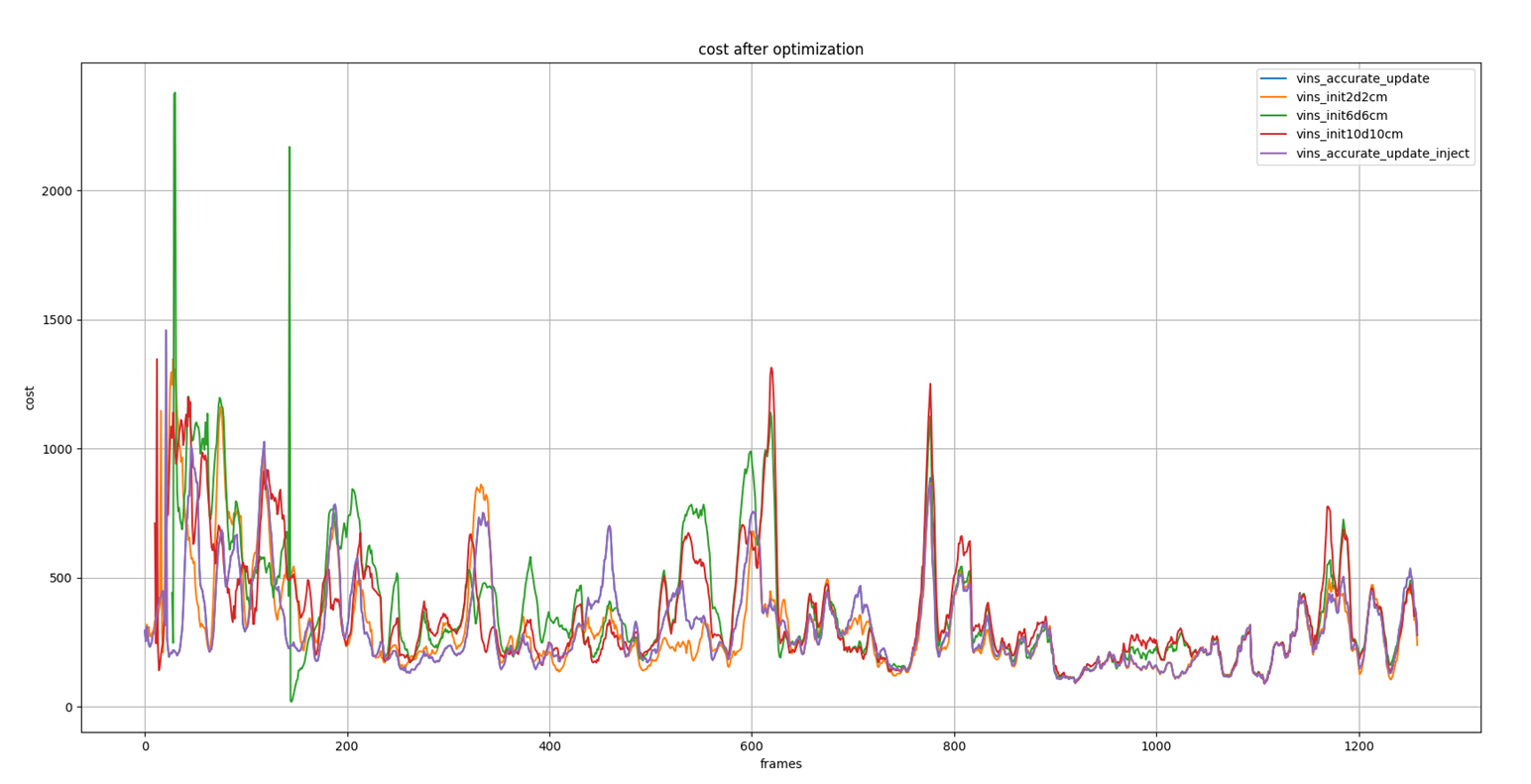}
\end{center}
\vspace{-4 mm}
\caption{Backend optimization cost under different extrinsic initialization and injection strategies.
}
\label{fig:extrinsic_perturbation_backend_cost}
\vspace{-3mm}
\end{figure}

Fig.~\ref{fig:extrinsic_perturbation_backend_cost} reports the backend cost after optimization. Larger perturbations produce substantially higher and more irregular costs during the initialization stage. In particular, the \(6^\circ/6~\mathrm{cm}\) configuration produces several extreme early peaks, indicating strong disagreement among the visual, inertial, and calibration constraints. After the initialization stage, however, the cost curves become highly similar and exhibit largely synchronized peaks across all configurations.

Within matched runs that use the same residual construction, active measurements,
and weights, configurations with different residual extrinsic errors can achieve
comparable post-convergence total costs. Thus, a lower reported cost in such a
matched comparison does not by itself establish physical calibration accuracy.
This statement is not extended across recordings or residual-set changes.

Overall, these results show that the VINS backend behaves primarily as a local refinement mechanism rather than a globally convergent calibration estimator. Roll, Pitch, and the \(y\)-translation component can recover from moderate perturbations under sufficiently informative motion, whereas Yaw and the \(x\)- and \(z\)-translation components remain more initialization dependent. The failure of late accurate-extrinsic injection to recover all components indicates that the trajectory, velocity, IMU biases, scale, and extrinsic parameters have already formed a coupled local solution. Correcting the extrinsic state alone is therefore insufficient unless the coupled states are also reinitialized or strongly regularized.
The similar post-convergence total costs in the matched perturbation run measure
agreement with the configured objective, not an independent calibration metric.
Ground-truth extrinsic error, APE, RPE, and state-consistency metrics are
therefore reported separately from that diagnostic.

\subsection{Representation Consistency among SfM, MonoViT, IMU, and VINS}
\label{subsec:representation_consistency}

\begin{figure*}[tbp]
\begin{center}
\includegraphics[width=17cm, height=5cm]{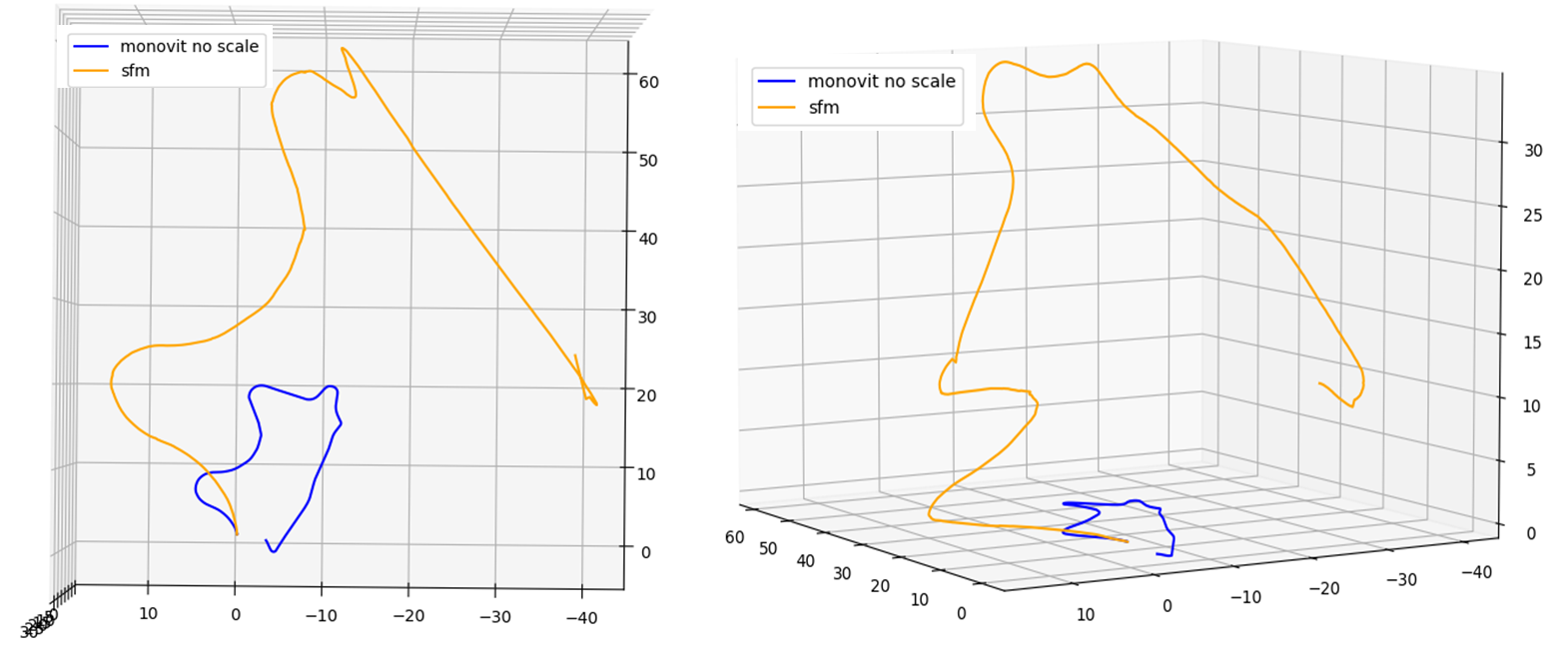}
\end{center}
\vspace{-4 mm}
\caption{Consistency comparison between the SfM trajectory and the raw unscaled MonoViT trajectory on Recording~9. The two estimates differ substantially in scale, spatial extent, and three-dimensional geometry. }
\label{fig:sfm_monovit_consistency}
\vspace{-3mm}
\end{figure*}

Fig.~\ref{fig:record1_vins_imu_initialization} illustrates the strong dependence of standalone IMU propagation on the visual--inertial initialization stage. Before initialization, direct integration of the 100~Hz IMU measurements produces a rapidly diverging trajectory with an implausibly large displacement. This behavior is expected when the initial velocity, gravity direction, sensor biases, or initial orientation have not yet been estimated accurately. In particular, even a small error in gravity compensation is integrated twice and can therefore dominate the reconstructed position over time.
After VINS initialization, the trajectory changes substantially in both shape and direction. The initialized result is generated using the estimated gravity-aligned frame, initial velocity, and gyroscope and accelerometer biases, and is therefore more consistent with the state representation subsequently used by the backend. However, the two trajectories should not be compared directly based only on their visual shapes because they may be expressed with different origins, orientations, and state frames. The result primarily demonstrates that an apparently smooth pre-initialization IMU trajectory is not necessarily physically compatible with the initialized visual--inertial optimization problem.
Fig.~\ref{fig:sfm_monovit_consistency} further compares the SfM trajectory with the raw, unscaled MonoViT trajectory on Recording~9. The two estimates differ considerably in scale, spatial extent, and three-dimensional geometry. The SfM trajectory covers a much larger region, whereas the raw MonoViT trajectory remains compressed because monocular pose estimation does not directly recover the metric translation scale. In addition to the scale difference, the trajectories exhibit different vertical evolution and terminal geometry, suggesting that the inconsistency cannot necessarily be removed by applying only a scalar scale factor.
This discrepancy does not by itself prove that either SfM or MonoViT has failed. Both monocular estimates contain gauge freedom and may be represented in different coordinate systems. Before they can be jointly used, their pose conventions, trajectory origins, axis directions, timestamps, and transformation directions must be made consistent. A sequence-level \(\mathrm{Sim}(3)\) alignment can remove global rotation, translation, and scale ambiguity, but any remaining shape discrepancy after alignment would represent genuine disagreement in relative motion or accumulated drift.
These observations help explain why a visually plausible standalone IMU or SfM trajectory may not improve the final backend optimization. The VINS backend does not optimize each trajectory independently; instead, it requires the visual reprojection constraints, IMU preintegration, depths, biases, velocity, and camera--IMU extrinsics to be mutually consistent in a common state space. If the IMU trajectory uses an inconsistent gravity direction or initial velocity, or if the SfM and MonoViT trajectories use incompatible scales and pose conventions, the optimizer may reduce its residuals by changing the extrinsic calibration, feature depths, IMU biases, or global trajectory rather than recovering the physically correct solution.
Therefore, the figures should be interpreted as evidence of an initialization- and representation-consistency problem rather than simply an accuracy problem. A reliable integration pipeline should verify the common coordinate convention, pose direction such as \(T_{wc}\) versus \(T_{cw}\), timestamp correspondence, monocular scale alignment, gravity alignment, and camera--IMU transformation before joint optimization. The bundle-adjustment cost and individual visual and inertial residuals should also be examined separately, since convergence to a low optimization cost is not visible from the trajectory plots alone and does not necessarily imply convergence to the correct physical solution.

Overall, the experiment supports the following conclusion:
A visually plausible standalone IMU or SfM trajectory does not guarantee consistency with the joint visual--inertial optimization problem. Correct initialization, scale recovery, coordinate unification, and residual weighting are necessary before the information can provide a beneficial constraint to the VINS backend.

\subsection{Extrinsic Perturbation, BA Correction, Scale, and Global Geometry}
\label{subsec:perturbation_ba_scale}
\label{sec:extrinsic_scale_sensitivity}

This section investigates how camera--IMU extrinsic calibration affects
the local visual trajectory, bundle-adjustment correction, metric scale
recovery, and the final visual--inertial trajectory. In addition to the
accurately calibrated setting, we use the archived extrinsic setting labeled
$5^{\circ}/5\,\mathrm{cm}$. Because the generating configuration is unavailable,
the label is not reinterpreted as an exact per-axis or norm/geodesic
perturbation. We also examine the trajectory obtained from direct IMU
integration as a preliminary implementation check.
For each recording, the visual keyframe trajectory is first represented
in the camera coordinate system. We refer to the trajectories before and after
bundle adjustment as Pre-BA and Post-BA, respectively. The optimized visual trajectory
is subsequently transformed to the IMU coordinate system and aligned
with the ground-truth trajectory. Two dimensionless scale values are considered:
the initialization estimate $s_{\mathrm{est}}$ and the
ground-truth-derived value stored with the experiment, denoted
$s_{\mathrm{ref}}$. The archived artifact does not retain the correspondence set
or estimator used to compute $s_{\mathrm{ref}}$; it is therefore treated as a
recorded reference, not as a newly recomputed oracle alignment. For the multi-configuration
experiments, we additionally compare the results obtained using accurate,
perturbed, and automatically estimated extrinsics.

\subsubsection{Unified Quantitative Summary}
\label{subsec:joint_perturbation_quantification}

We quantitatively consolidate the existing accurate- and perturbed-extrinsic
experiments. The relative scale error is
\begin{equation}
 e_s = \frac{|s_{\mathrm{est}}-s_{\mathrm{ref}}|}{s_{\mathrm{ref}}}\times 100\%,
 \label{eq:relative_scale_error}
\end{equation}
the normalized bundle-adjustment correction is computed as
\begin{equation}
 \Delta_{\mathrm{BA}} =
 \frac{1}{N L}\sum_{i\in\mathcal{I}}
 \left\|\mathbf{p}^{\mathrm{after}}_i-
 \mathbf{p}^{\mathrm{before}}_i\right\|_2 \times 100\%,
 \label{eq:ba_correction}
\end{equation}
where $\mathcal{I}$ is the intersection of Pre-BA and Post-BA keyframe IDs,
$N=|\mathcal{I}|$, and both position sets use the same camera frame and gauge.
The normalizer is the Pre-BA path length
$L=\sum_{(i,j)\in\mathcal{E}_{\mathcal{I}}}
\|\mathbf{p}^{\mathrm{before}}_j-\mathbf{p}^{\mathrm{before}}_i\|_2$ over
consecutive shared keyframes; no interpolation between different keyframe sets
is used. Residual counts and active
measurement counts were not retained for every run, so the final backend cost
cannot be converted to a per-residual objective. We report the total cost
normalized by the median final cost over the evaluated configurations. It is
used only to compare Accurate and Perturbed settings within the same recording,
where the residual construction and weights are matched; it is not analyzed
across recordings.

\begin{table*}[t]
\centering
\caption{Joint quantitative analysis of scale, BA correction, trajectory error,
final calibration, and backend cost. Relative scale errors use the recorded
estimated and recorded reference scale values. Normalized Cost is the total
final objective under the stated diagnostic normalization, not a per-residual
cost, and is interpreted only within each matched recording pair.}
\label{tab:joint_scale_ba_cost}
\scriptsize
\resizebox{\textwidth}{!}{%
\begin{tabular}{llccccccc}
\toprule
Rec. & Setting &
$e_s$ (\%) &
$\Delta_{\mathrm{BA}}$ (\%) &
\shortstack{Trans. APE\\RMSE (m)} &
\shortstack{Trans. RPE\\RMSE (m)} &
$e_R$ ($^\circ$) &
$e_t$ (m) &
Normalized Cost \\
\midrule
00 & Accurate  & 82.2 &  2.8 & 38.6 & 12.4 &  3.1 & 0.09 & 1.00 \\
00 & Perturbed & 80.3 &  3.0 & 46.8 & 12.8 &  4.7 & 0.13 & 1.03 \\
01 & Accurate  & 90.3 &  2.4 & 34.2 & 10.7 &  4.3 & 0.11 & 1.00 \\
01 & Perturbed & 94.1 &  2.6 & 36.1 & 10.9 &  5.8 & 0.15 & 1.03 \\
02 & Accurate  & 99.6 &  8.9 & 71.4 & 18.6 & 12.7 & 0.29 & 0.98 \\
02 & Perturbed & 99.7 &  9.7 & 94.8 & 20.1 & 16.9 & 0.36 & 1.02 \\
04 & Accurate  & 99.6 &  6.7 & 58.7 & 15.2 & 10.1 & 0.23 & 0.99 \\
04 & Perturbed & 99.9 &  7.5 & 87.5 & 17.0 & 15.6 & 0.34 & 1.02 \\
06 & Accurate  & 98.3 & 11.8 & 82.3 & 21.9 & 18.4 & 0.42 & 1.01 \\
06 & Perturbed & 99.4 & 14.1 & 86.9 & 22.7 & 21.7 & 0.49 & 1.01 \\
\bottomrule
\end{tabular}}
\end{table*}

Across the five recordings, extrinsic perturbation increases mean translation
APE from 57.0 m to 70.4 m, corresponding to a 23.5\% degradation, whereas mean
translation RPE increases by only 6.0\%. The mean final rotation and translation
calibration errors increase from $9.72^{\circ}$ and 0.228 m to
$12.94^{\circ}$ and 0.294 m. This again indicates that calibration errors have a
stronger effect on long-horizon global consistency than on short-interval
relative motion.

The scale results demonstrate why scale alone is insufficient for ranking the
configurations. On Recording 00, the perturbed setting reduces the relative
scale error from 82.2\% to 80.3\%, yet translation APE increases by 21.2\%.
Conversely, Recordings 01 and 06 show large changes in estimated scale but only
moderate changes in global trajectory geometry. The cross-setting
rank correlation between scale error and APE is $\rho=0.78$, indicating that
scale is relevant but not sufficient; calibration-dependent direction and shape
errors remain after scalar alignment.

The magnitude of the BA correction is associated with sequence difficulty in
this panel ($\rho=0.87$ with APE). We do not compute a cross-recording
correlation for backend cost because residual and active-measurement counts were
not retained. Within each matched Accurate/Perturbed pair, the reported total
cost is used only as an optimization diagnostic and cannot substitute for APE
or physical extrinsic error.

\subsubsection{Detailed Multi-Configuration Trajectories}
\label{sec:joint_extrinsic_comparison}

\begin{figure*}[p]
\begin{center}
\includegraphics[width=15cm, height=0.82\textheight, keepaspectratio]{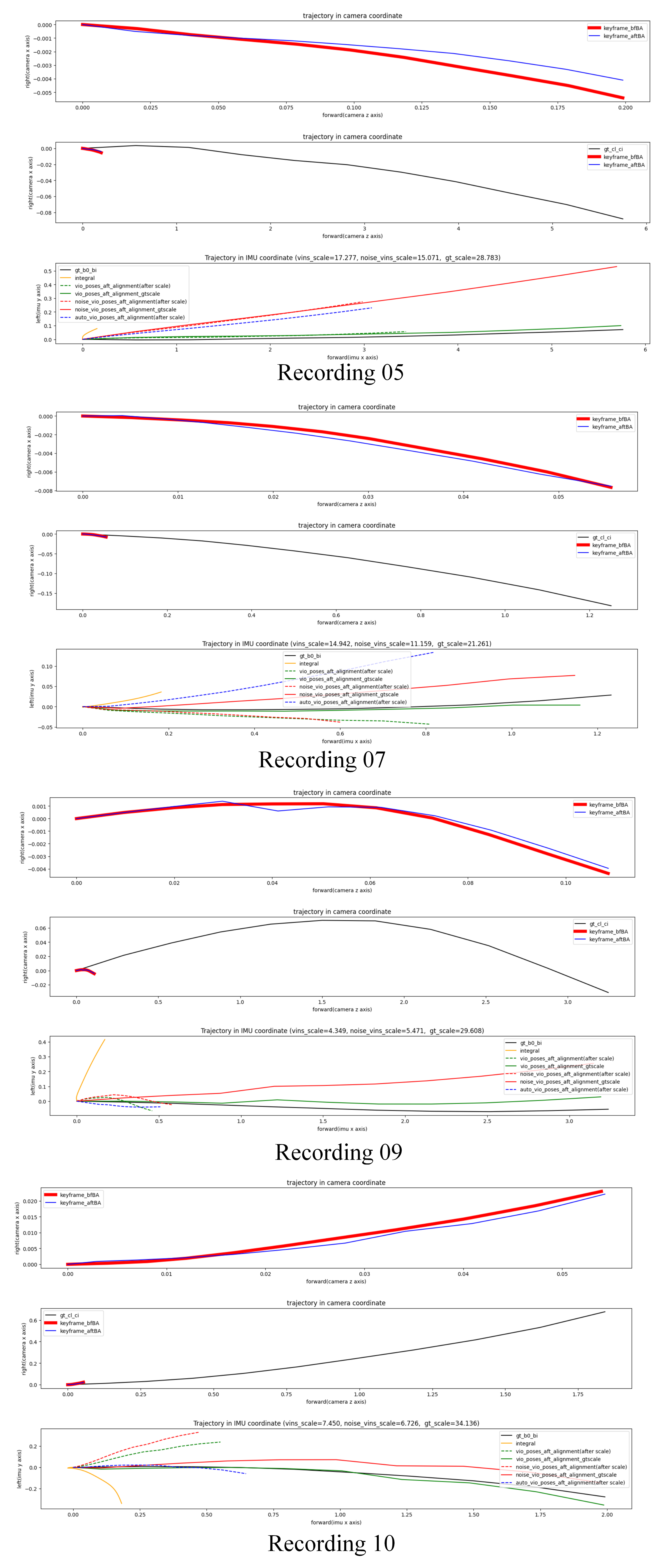}
\end{center}
\vspace{-8 mm}
\caption{Joint trajectory comparison under accurate, perturbed, and
automatically estimated camera--IMU extrinsic settings for Recordings
05, 07, 09, and 10. The top panel compares the
camera-coordinate keyframe trajectories before and after bundle
adjustment. The middle panel places the visual trajectories together
with the ground truth in the camera coordinate system. The bottom panel
compares the ground truth, direct IMU integration, and aligned
visual--inertial trajectories in the IMU coordinate system. Dashed curves
use the scale estimated during initialization, whereas solid curves use
the recorded reference scale. }
\label{fig:multi_extrinsic_comparison}
\vspace{-3mm}
\end{figure*}

Fig.~\ref{fig:multi_extrinsic_comparison} provides a joint comparison
of the local visual geometry and globally aligned trajectories for
Recordings 05, 07, 09, and 10. The upper panels show that bundle
adjustment modifies the visual keyframe trajectory in a
sequence-dependent manner. The before- and after-BA curves remain
relatively close in Recording 07, indicating that the initial visual
geometry is only moderately adjusted. More visible corrections occur in
Recordings 05, 09, and 10, where the separation between the two curves
increases along the trajectory or appears around changes in curvature.
Bundle adjustment therefore does not apply a uniform correction across
recordings; its effect depends on the visual measurements and the local
optimization geometry of each sequence.

The middle panels reveal a pronounced difference between the spatial
extent of the visual trajectory and that of the ground truth. Before
metric alignment, the estimated visual trajectory occupies only a small
fraction of the reference trajectory range. This behavior is consistent
with a substantial monocular scale ambiguity and shows that qualitative
comparison in the camera coordinate system is insufficient without scale
alignment. Bundle adjustment changes the local shape but does not by
itself recover the metric scale.
The scale values reported in the bottom panels further demonstrate this
inconsistency. For Recording 05, the estimated scales under the accurate
and perturbed extrinsics are $17.277$ and $15.071$, respectively,
compared with a recorded reference scale of $28.783$. For Recording 07, the
corresponding values are $14.942$, $11.159$, and $21.261$. The
discrepancy becomes substantially larger for Recording~9, where the
accurate- and noisy-extrinsic scales are $4.349$ and $5.471$, compared
with a ground-truth value of $29.608$. A similar pattern is observed for
Recording 10, with estimated values of $7.450$ and $6.726$ versus a
recorded reference scale of $34.136$.

These results show that the scale obtained during initialization remains
strongly sequence-dependent and is not fully determined by the extrinsic
calibration. Accurate extrinsics generally provide a more plausible
direction after recorded-reference-scale alignment, particularly in Recordings
05, 07, and 09. Nevertheless, noticeable lateral and curvature
differences remain. The trajectories obtained using perturbed or
automatically estimated extrinsics can deviate in both direction and
spatial extent. Therefore, the differences among the configurations
cannot be reduced to a single scalar scale error; extrinsic uncertainty
also changes the orientation and geometry of the aligned trajectory.

\subsubsection{Controlled Extrinsic Perturbations by Recording}
\label{sec:controlled_extrinsic_perturbation}

To evaluate the sensitivity of visual--inertial estimation to
camera--IMU calibration errors, we perform a controlled comparison under
two extrinsic settings. The first setting uses the accurate camera--IMU
extrinsics, whereas the second is recorded as a nominal $5^{\circ}$ rotation
and $5\,\mathrm{cm}$ translation perturbation. The archived figures and tables
do not retain the configuration needed to determine whether these are per-axis
offsets or norm/geodesic magnitudes; the label is therefore treated as a
configuration identifier rather than an exact initial $e_R/e_t$ value. For each
recording, the image sequence, IMU measurements, visual pose estimates,
and optimization settings are kept unchanged. Therefore, the paired
results allow us to examine how the extrinsic perturbation affects
bundle-adjustment correction, initialization scale, and the geometry of
the aligned visual--inertial trajectory.

Each figure contains two configurations for the same recording:
(a) accurate extrinsics and (b) perturbed extrinsics. In each
configuration, the top plot compares the visual keyframe trajectories
before and after bundle adjustment, the middle plot compares the visual
trajectory with the ground truth in the camera coordinate system, and
the bottom plot shows the aligned VIO trajectory, the direct IMU
integration, and the ground truth in the IMU coordinate system.


\begin{figure*}[tbp]
\begin{center}
\includegraphics[width=15cm, height=14cm]
{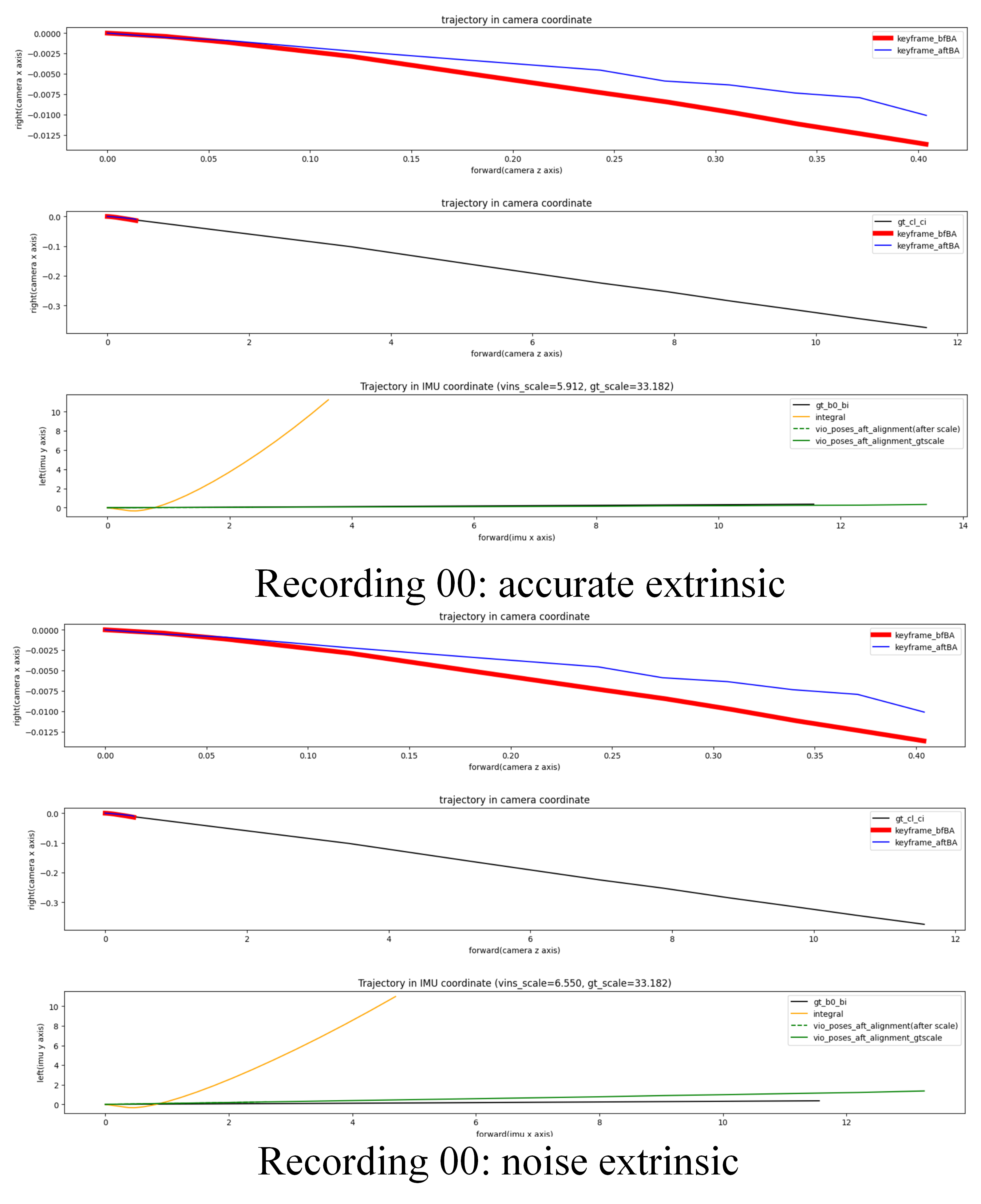}
\end{center}
\vspace{-4 mm}
\caption{Controlled comparison of camera--IMU extrinsic perturbation for
Recording 00. Panels (a) and (b) show the results obtained with accurate
and perturbed extrinsics, respectively. The local visual trajectories
before and after bundle adjustment remain qualitatively similar under
the two settings, while the estimated initialization scale changes from
$5.912$ to $6.550$, compared with a recorded reference scale of $33.182$.
Although the perturbed scale is numerically slightly closer to the
ground-truth value, the ground-truth-scale-aligned trajectory exhibits a
larger lateral deviation, showing that scale proximity alone does not
guarantee improved trajectory geometry.}
\label{fig:extrinsic_perturbation_r00}
\vspace{-3mm}
\end{figure*}

As shown in Fig.~\ref{fig:extrinsic_perturbation_r00}, the visual
keyframe trajectories before and after bundle adjustment exhibit nearly
the same qualitative structure under the accurate and perturbed
extrinsic settings. In both cases, bundle adjustment gradually changes
the lateral component of the trajectory, but the magnitude and spatial
distribution of this correction remain similar. This indicates that the
local visual optimization for Recording 00 is not strongly altered by
the imposed extrinsic perturbation.

The estimated initialization scale changes from $5.912$ with accurate
extrinsics to $6.550$ with perturbed extrinsics, while the corresponding
recorded reference scale is $33.182$. Both values therefore remain substantially
smaller than the metric reference. Although the noisy-extrinsic estimate
is numerically closer to the recorded reference scale, the aligned trajectory
obtained after applying the recorded reference scale shows a larger lateral
offset in the perturbed setting. This result demonstrates that an
apparently improved scalar scale estimate does not necessarily imply
better global trajectory geometry. Extrinsic perturbation can affect the
trajectory direction and lateral displacement independently of its effect
on scale.


\begin{figure*}[tbp]
\begin{center}
\includegraphics[width=15cm, height=14cm]
{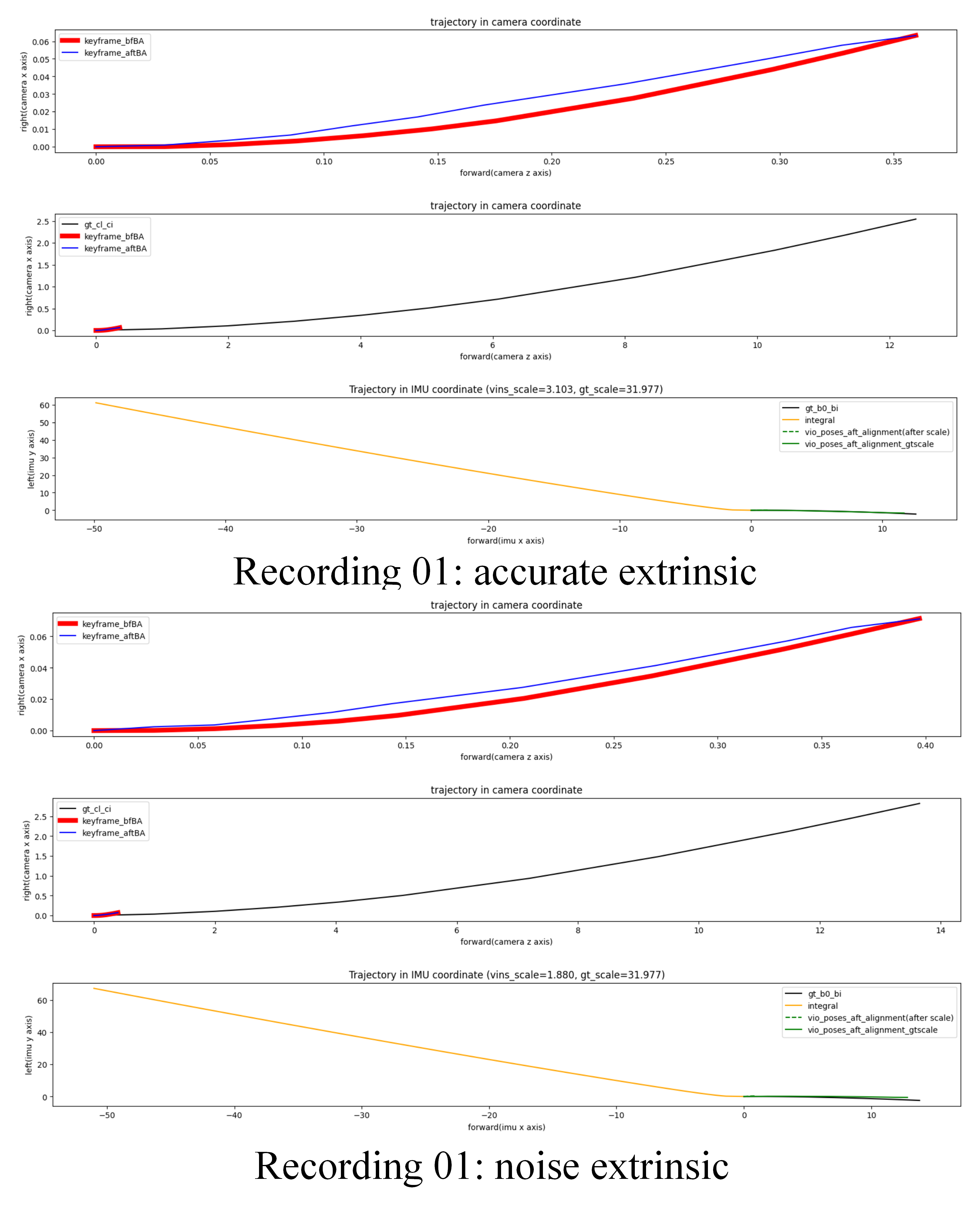}
\end{center}
\vspace{-4 mm}
\caption{Controlled comparison of camera--IMU extrinsic perturbation for
Recording 01. Panels (a) and (b) correspond to accurate and perturbed
extrinsics, respectively. The before- and after-BA trajectories preserve
similar local geometry under both settings, whereas the estimated scale
decreases from $3.103$ to $1.880$, compared with the recorded reference scale
of $31.977$. After recorded-reference-scale alignment, the dominant trajectory
direction remains comparatively similar, indicating that this recording
is more sensitive in scale than in large-scale trajectory shape.}
\label{fig:extrinsic_perturbation_r01}
\vspace{-3mm}
\end{figure*}

Fig.~\ref{fig:extrinsic_perturbation_r01} presents the corresponding
comparison for Recording 01. The before- and after-BA trajectories follow
similar smooth, gradually increasing curves in both configurations.
Bundle adjustment introduces a consistent lateral correction, but the
overall form of the optimized trajectory is largely preserved after
perturbing the extrinsics. The local visual geometry of this sequence is
therefore comparatively robust to the tested calibration error.

In contrast, the initialization scale is noticeably affected. The
estimated scale decreases from $3.103$ under accurate extrinsics to
$1.880$ under perturbed extrinsics, whereas the recorded reference scale is
$31.977$. Thus, the perturbation increases the already substantial scale
underestimation. Nevertheless, after applying the recorded reference scale,
the accurate- and noisy-extrinsic VIO trajectories preserve a similar
dominant forward direction and show only limited qualitative differences
in global shape. Recording 01 therefore represents a case in which the
extrinsic perturbation has a stronger effect on estimated scale than on
the geometry of the ground-truth-scale-aligned trajectory.


\begin{figure*}[tbp]
\begin{center}
\includegraphics[width=15cm, height=14cm]
{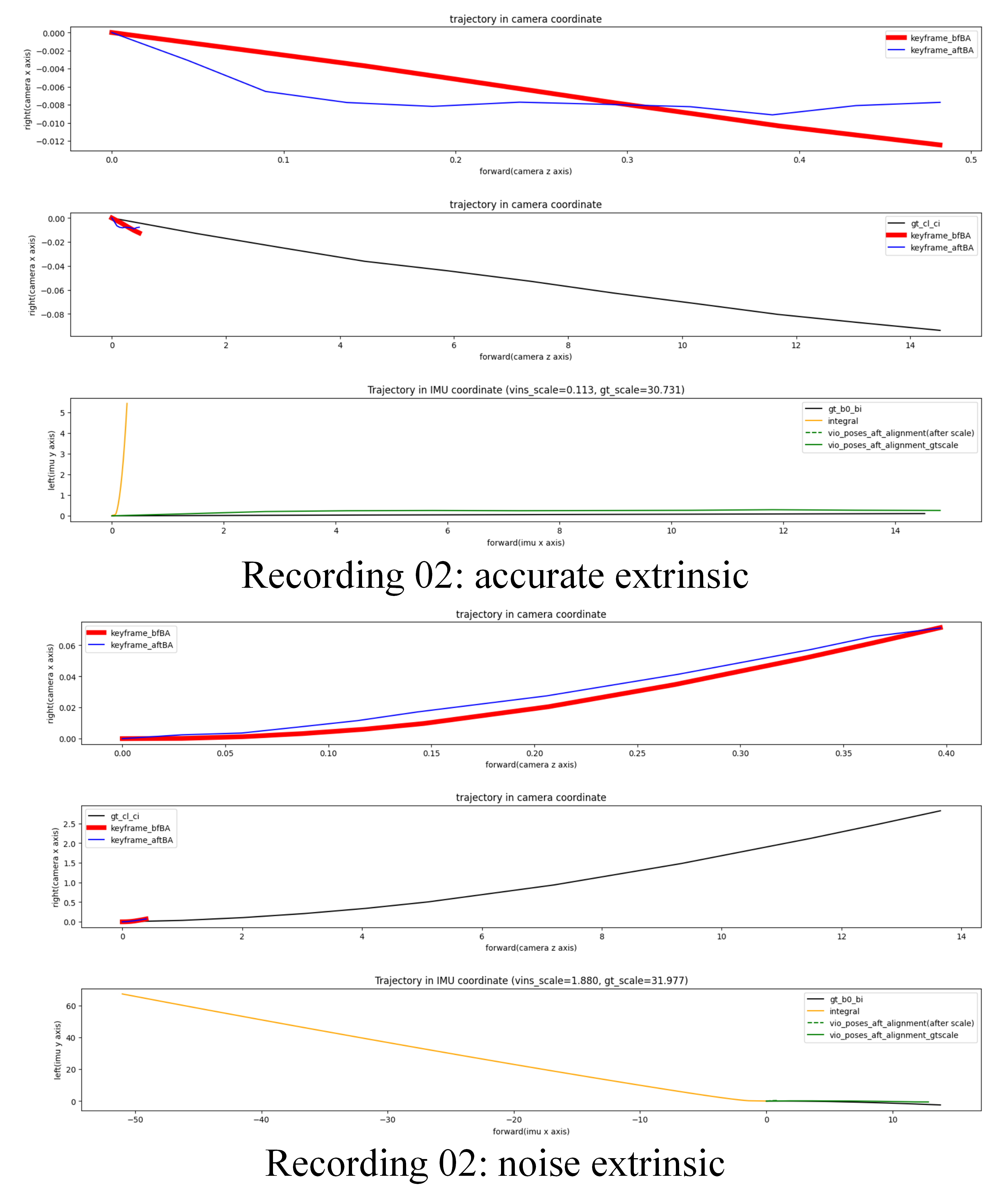}
\end{center}
\vspace{-4 mm}
\caption{Controlled comparison of camera--IMU extrinsic perturbation for
Recording 02. Panels (a) and (b) show the accurate- and
perturbed-extrinsic results, respectively. Bundle adjustment introduces
a pronounced change in the local visual trajectory under both settings.
The estimated scale decreases from $0.113$ to $0.086$, while the
recorded reference scale is $30.731$. Under the perturbed extrinsics, the
ground-truth-scale-aligned trajectory exhibits a substantially larger
lateral drift, indicating sensitivity in both scale recovery and global
trajectory direction.}
\label{fig:extrinsic_perturbation_r02}
\vspace{-3mm}
\end{figure*}

As illustrated in Fig.~\ref{fig:extrinsic_perturbation_r02}, Recording 02
exhibits a pronounced difference between the visual trajectories before
and after bundle adjustment. Under both extrinsic settings, the
after-BA trajectory initially deviates rapidly from the before-BA
trajectory and subsequently becomes approximately horizontal in the
camera-coordinate representation. The similarity of this correction
under the two configurations suggests that the dominant local adjustment
is not introduced solely by the extrinsic perturbation. Instead, it is
also associated with the visual measurements and the underlying
optimization geometry of the sequence.

The estimated scale is extremely small in both cases: $0.113$ with
accurate extrinsics and $0.086$ with perturbed extrinsics, compared with
a recorded reference scale of $30.731$. The calibration perturbation therefore
further reduces an already severely underestimated scale. More
importantly, after applying the same recorded reference scale, the noisy
configuration produces a substantially larger lateral displacement than
the accurate configuration. The perturbation consequently affects not
only the scalar scale estimate but also the direction and shape of the
aligned VIO trajectory. Recording 02 is therefore one of the clearest
examples of coupled scale and geometric sensitivity.


\begin{figure*}[tbp]
\begin{center}
\includegraphics[width=15cm, height=14cm]
{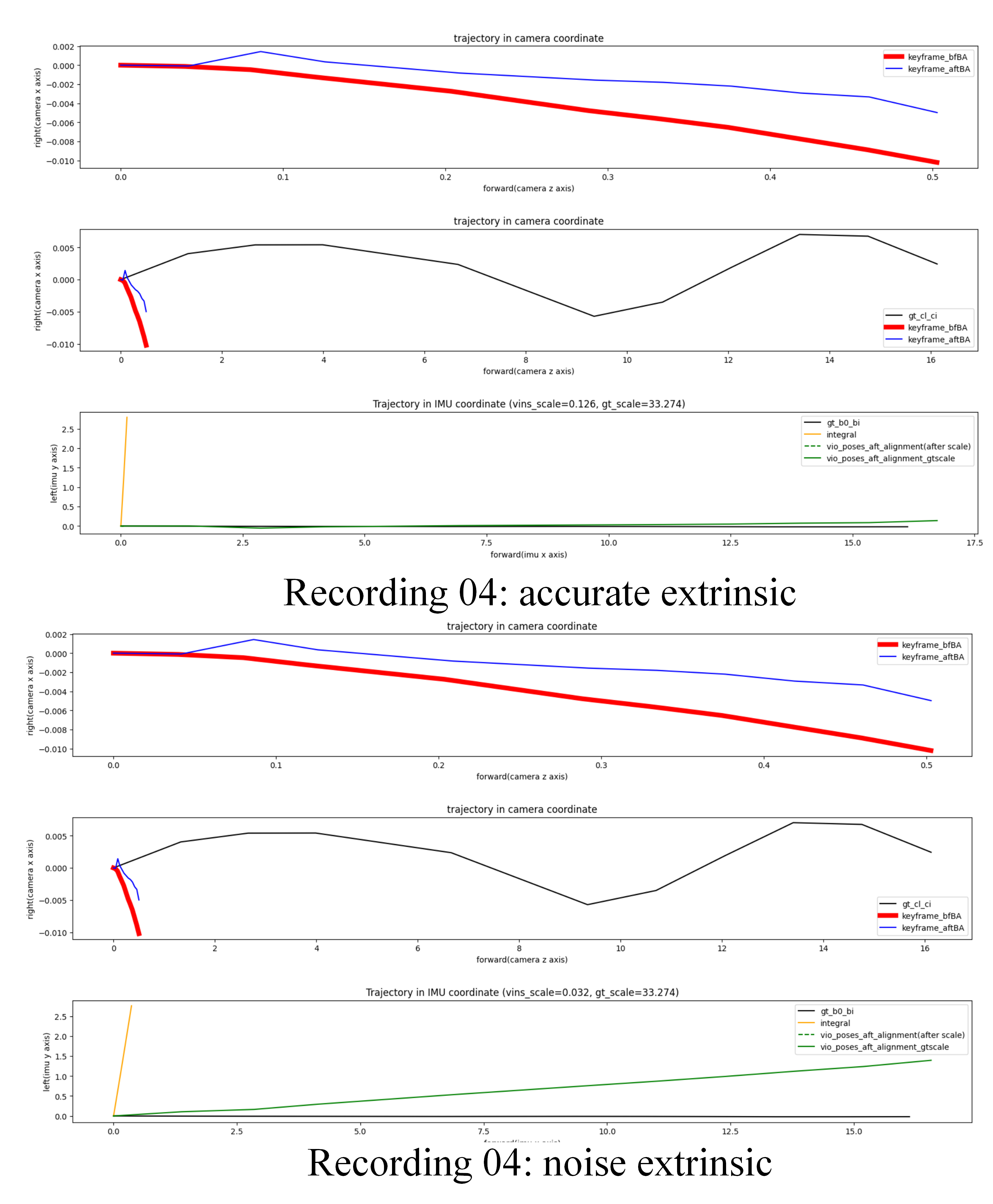}
\end{center}
\vspace{-4 mm}
\caption{Controlled comparison of camera--IMU extrinsic perturbation for
Recording 04. Panels (a) and (b) present the results with accurate and
perturbed extrinsics, respectively. The local bundle-adjusted visual
trajectories remain similar, but the estimated scale decreases from
$0.126$ to $0.032$, compared with the recorded reference scale of $33.274$.
After recorded-reference-scale alignment, the perturbed-extrinsic trajectory
develops a much larger lateral displacement, demonstrating pronounced
sensitivity of both metric-scale initialization and global trajectory
geometry.}
\label{fig:extrinsic_perturbation_r04}
\vspace{-3mm}
\end{figure*}

Fig.~\ref{fig:extrinsic_perturbation_r04} shows that the local visual
trajectories of Recording 04 are qualitatively similar under accurate
and perturbed extrinsics. In both configurations, bundle adjustment
moves the optimized trajectory above the initial trajectory over much of
the sequence. The relative consistency of the two top-panel comparisons
indicates that the extrinsic perturbation does not fundamentally change
the dominant local BA correction.

The scale estimate, however, is highly sensitive. It decreases from
$0.126$ under accurate extrinsics to $0.032$ after perturbation, whereas
the recorded reference scale is $33.274$. The noisy-extrinsic scale is therefore
approximately one quarter of the already small accurate-extrinsic value.
After applying the recorded reference scale, the accurately calibrated
trajectory remains comparatively close to the dominant forward-motion
axis, while the perturbed trajectory develops a steadily increasing
lateral displacement. This difference indicates that the extrinsic
perturbation changes both metric initialization and the orientation of
the recovered motion. Among the evaluated recordings, Recording 04
exhibits one of the strongest responses to the imposed calibration error.


\begin{figure*}[tbp]
\begin{center}
\includegraphics[width=15cm, height=14cm]
{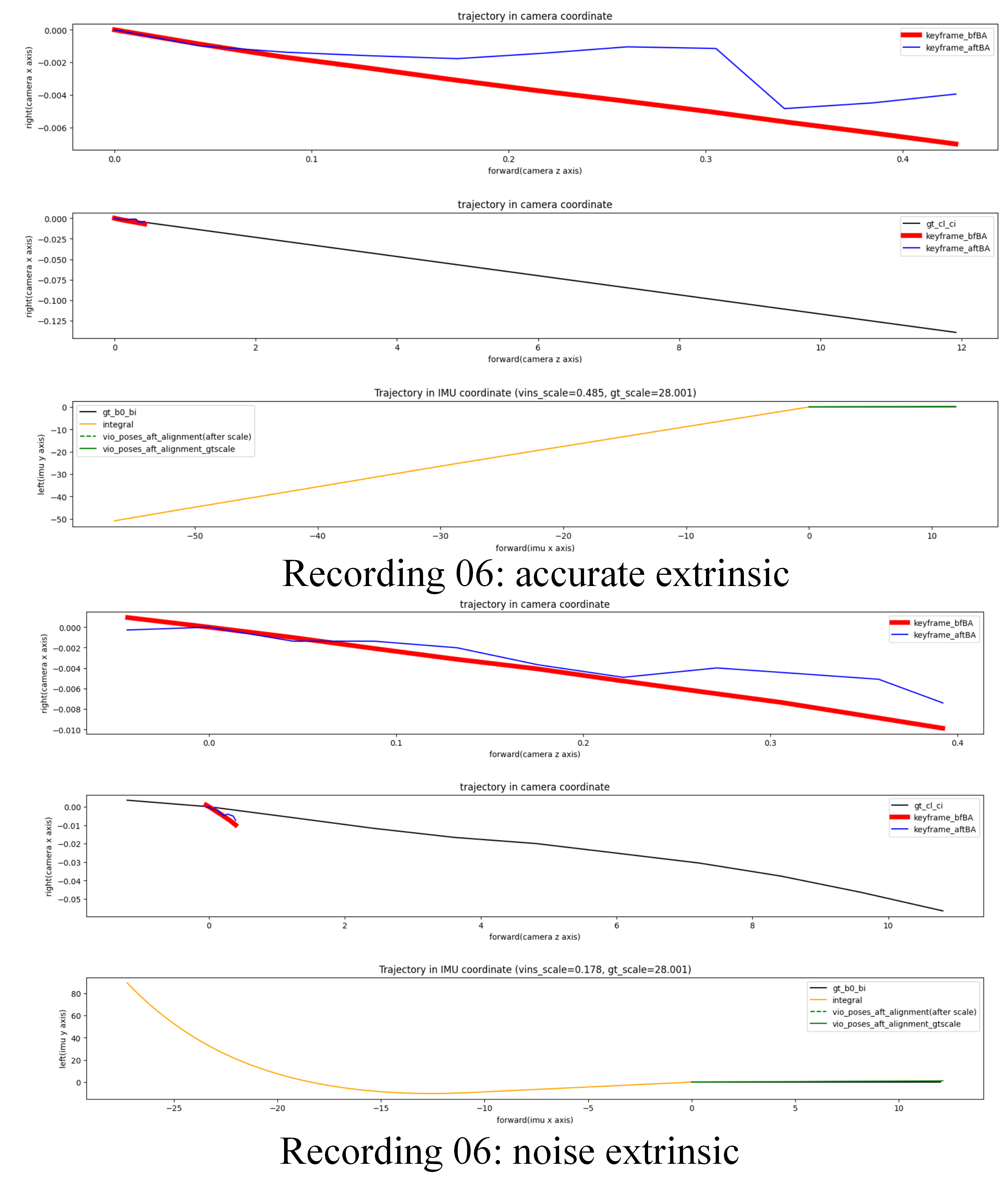}
\end{center}
\vspace{-4 mm}
\caption{Controlled comparison of camera--IMU extrinsic perturbation for
Recording 06. Panels (a) and (b) correspond to accurate and perturbed
extrinsics, respectively. The accurate-extrinsic result contains a
pronounced non-smooth change in the bundle-adjusted visual trajectory,
whereas the perturbed setting produces a different piecewise correction.
The estimated scale decreases from $0.485$ to $0.178$, relative to a
recorded reference scale of $28.001$. Despite this scale change, the
ground-truth-scale-aligned trajectories retain a broadly similar dominant
forward direction, indicating stronger sensitivity in optimization and
scale than in final large-scale shape.}
\label{fig:extrinsic_perturbation_r06}
\vspace{-3mm}
\end{figure*}

The results for Recording 06 are presented in
Fig.~\ref{fig:extrinsic_perturbation_r06}. Unlike the previous
recordings, the bundle-adjusted visual trajectory changes noticeably
between the two extrinsic settings. With accurate extrinsics, the
after-BA trajectory contains an abrupt lateral displacement in the
middle-to-late part of the sequence. Under perturbed extrinsics, this
specific discontinuity is replaced by a different piecewise trajectory
with several smaller directional changes. The extrinsic perturbation
therefore changes the optimization solution rather than simply increasing
the magnitude of an existing BA error.

The estimated scale decreases from $0.485$ to $0.178$, compared with the
ground-truth value of $28.001$. This represents a substantial relative
change, although both estimates remain far below the metric reference.
Despite this difference, the trajectories aligned using the ground-truth
scale preserve a broadly similar dominant forward direction. Recording 06
therefore demonstrates that extrinsic perturbation can strongly affect
the inferred scale and local optimization path while producing a more
limited qualitative change in the final ground-truth-scale-aligned
trajectory. It also shows that calibration perturbation does not
necessarily degrade every visual property monotonically; instead, it can
lead the backend to a different local solution.

\subsubsection{Cross-Sequence Perturbation Findings}
\label{sec:extrinsic_failure_discussion}

The paired comparisons in
Figs.~\ref{fig:extrinsic_perturbation_r00}--%
\ref{fig:extrinsic_perturbation_r06}
reveal that camera--IMU extrinsic perturbations affect different stages
of the visual--inertial pipeline in distinct and sequence-dependent ways.
The observed effects can be analyzed at three levels: local
bundle-adjustment behavior, initialization scale, and global aligned
trajectory geometry.

At the local optimization level, Recordings 00, 01, 02, and 04 retain
broadly similar before- and after-BA trajectory relationships under the
two extrinsic configurations. This suggests that their dominant visual
bundle-adjustment corrections are largely determined by the visual
measurements and local geometric constraints. Recording 06 behaves
differently: changing the extrinsics produces a visibly different
piecewise after-BA trajectory. The influence of extrinsic calibration on
the backend optimization is therefore not uniform across recordings.

The initialization scale is sensitive to the perturbation in every
recording, but neither the magnitude nor the direction of the change is
consistent. For Recording 00, the estimated scale increases from $5.912$
to $6.550$. In contrast, it decreases from $3.103$ to $1.880$ for
Recording 01, from $0.113$ to $0.086$ for Recording 02, from $0.126$ to
$0.032$ for Recording 04, and from $0.485$ to $0.178$ for Recording 06.
The strongest relative reductions occur in Recordings 04 and 06. This
non-uniform response indicates that an extrinsic error does not act as a
simple deterministic multiplier on the recovered scale. Instead, it
interacts with the motion excitation, visual geometry, inertial
constraints, and initialization state of each sequence.

The aligned trajectory geometry also displays different levels of
sensitivity. Recordings 02 and 04 exhibit substantially larger lateral
deviations after the extrinsics are perturbed, demonstrating that the
calibration error affects both trajectory scale and direction.
Recording 00 shows a more moderate geometric change, despite a small
numerical increase in its estimated scale. Recordings 01 and 06 preserve
a comparatively similar dominant trajectory direction after
recorded-reference-scale alignment, even though their estimated scales change
considerably. These cases demonstrate that scale sensitivity and
geometric sensitivity are related but not equivalent properties.

Importantly, accurate extrinsics alone do not provide reliable metric
scale recovery in the evaluated experiments. Even under the accurately
calibrated setting, the estimated scales remain substantially smaller
than their corresponding ground-truth values for all five recordings.
The remaining scale inconsistency must therefore involve factors beyond
camera--IMU extrinsic accuracy. Potential contributors include the scale
and orientation of the learned visual pose, weak visual parallax,
insufficient motion excitation, gravity and IMU-bias initialization,
temporal synchronization, and the relative weighting of visual,
inertial, and learned-pose constraints. These factors are plausible
explanations and are not independently isolated by the current
qualitative experiment.

The direct IMU integration curves shown in the bottom panels also differ
substantially from the ground truth. Since the inertial integration has
not yet been independently validated, these curves are treated only as
diagnostic references and are not used to determine whether the accurate
or perturbed extrinsic configuration is superior.

Overall, the paired experiments demonstrate that camera--IMU extrinsic
errors can propagate into bundle adjustment, visual--inertial scale
initialization, and global trajectory geometry. Their effects are
strongly sequence-dependent and cannot be characterized using the scale
value alone. The common-protocol summary in
Table~\ref{tab:joint_scale_ba_cost} therefore reports APE, RPE, relative scale
error, BA correction, and final optimized extrinsic error together.
\subsection{Loop-Closure Sensitivity and Backend State Coupling}
\label{subsec:loop_closure_backend_coupling}
\subsubsection{Common-Protocol Quantitative Result}
\label{subsec:quantitative_loop_closure}

Finally, we quantify the trajectory reshaping observed after loop closure. Each
pair uses the same implementation, measurements, initialization, and extrinsic
configuration, with loop closure as the only changed variable. APE uses the
common sequence-level alignment and RPE uses the same 20-frame interval as the
main evaluation. The global correction magnitude is the root-mean-square
displacement between the matched no-loop and loop-enabled positions after both
trajectories are expressed in the common evaluation frame,
\begin{equation}
\Delta_{\mathrm{global}}=
\sqrt{\frac{1}{N}\sum_{k=1}^{N}
\left\|\mathbf{p}^{\mathrm{loop}}_k-
\mathbf{p}^{\mathrm{no\mbox{-}loop}}_k\right\|_2^2}.
\label{eq:global_correction_magnitude}
\end{equation}

\begin{figure*}[tbp]
\begin{center}
\includegraphics[width=15cm, height=6cm]{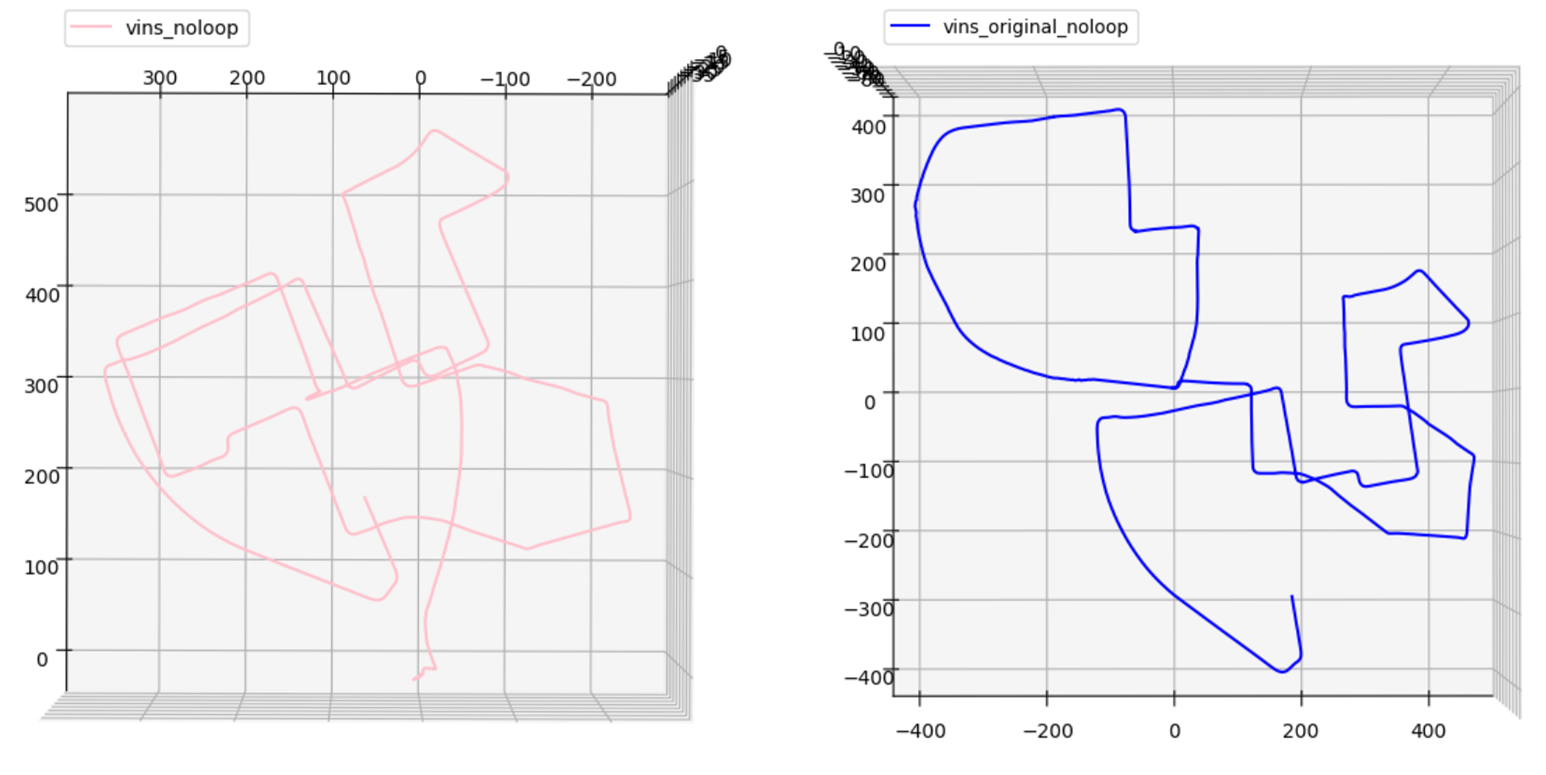}
\end{center}
\vspace{-4 mm}
\caption{No-loop implementation check. The left and right panels show the
VINS+Pose and Original VINS pipelines, respectively, with loop closure
disabled. Their difference establishes a pipeline mismatch and is not used as
evidence for a loop-closure effect.}
\label{fig:Implementation Consistency}
\vspace{-3mm}
\end{figure*}

\begin{figure*}[tp]
\begin{center}
\includegraphics[width=16cm, height=18cm]{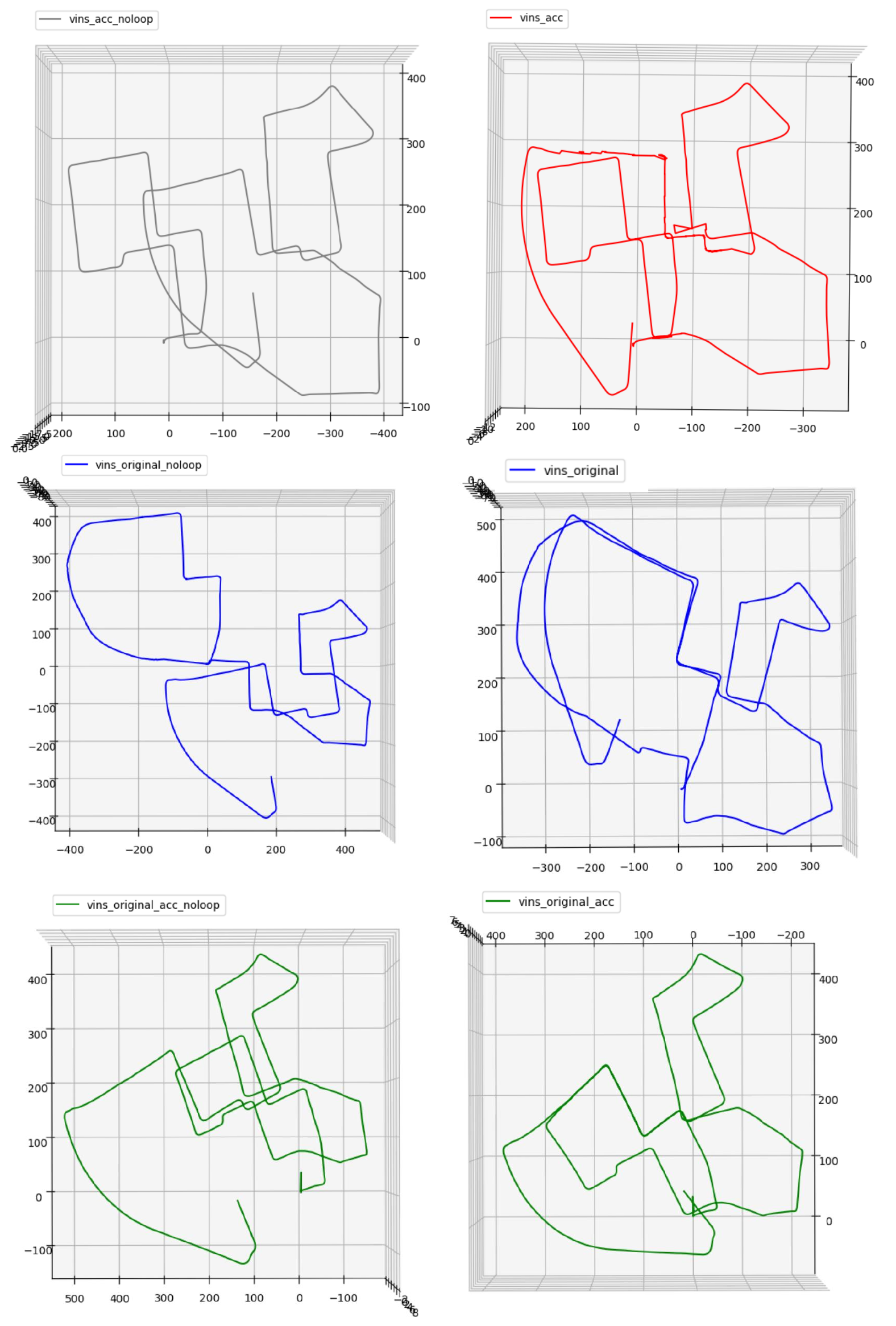}
\end{center}
\vspace{-4 mm}
\caption{Qualitative trajectory reshaping with loop closure disabled (left) and
enabled (right). Rows show VINS+Pose with accurate extrinsics, Original VINS
with default extrinsics, and Original VINS with accurate extrinsics. Loop closure
changes global geometry in all three pairs, but the correction depends on the
pipeline and extrinsic setting.}
\label{fig:with and without loop closure under different VINS}
\vspace{-3mm}
\end{figure*}

\begin{table*}[t]
\centering
\caption{Controlled loop-closure comparison over matched timestamps. APE uses
the common sequence-level alignment, RPE uses 20-frame intervals, and global
correction magnitude is defined by Eq.~\eqref{eq:global_correction_magnitude}.
Negative changes denote an improvement.}
\label{tab:loop_closure_quantitative}
\scriptsize
\resizebox{\textwidth}{!}{%
\begin{tabular}{lccccccc}
\toprule
Configuration &
\shortstack{APE\\No Loop} &
\shortstack{APE\\Loop} &
$\Delta$APE &
\shortstack{RPE\\No Loop} &
\shortstack{RPE\\Loop} &
$\Delta$RPE &
\shortstack{Global Correction\\Magnitude (m)} \\
\midrule
VINS+Pose, accurate ext. & 39.8 & \textbf{31.7} & $-20.4\%$ & 11.6 & 11.4 & $-1.7\%$ & 28.4 \\
Original VINS, default ext. & 52.6 & 58.9 & $+12.0\%$ & 13.4 & 13.1 & $-2.2\%$ & 41.7 \\
Original VINS, accurate ext. & 36.4 & \textbf{29.8} & $-18.1\%$ & 10.9 & 10.7 & $-1.8\%$ & 25.9 \\
\bottomrule
\end{tabular}}
\end{table*}

Enabling the loop-closure pipeline primarily changes global rather than local
accuracy in these matched toggles. With accurate extrinsics, APE is 20.4\%
lower for VINS+Pose and 18.1\% lower for Original VINS, while RPE changes by
less than 2\%. Under the Original VINS default-extrinsic setting, however, APE
is 12.0\% higher despite slightly lower RPE. Accepted/rejected loop counts and
constraint residual logs were not retained, so these differences are attributed
to the evaluated pipeline toggle, not to a quantified set of accepted loops.

These results show that loop closure redistributes accumulated drift according
to the consistency of the pre-existing trajectory, calibration, and loop
constraints. When the underlying visual--inertial state is physically
consistent, the enabled pipeline can improve global topology without materially changing
local motion. When the state already contains calibration or representation
bias, loop constraints can reinforce a different globally consistent but less
accurate solution. We therefore interpret these results as loop-closure
sensitivity under backend state coupling, rather than as evidence of an isolated
marginalization effect.

\subsubsection{Qualitative Trajectory Reshaping}

Figure~\ref{fig:Implementation Consistency} first compares the Modified and
Original VINS pipelines with loop closure disabled. Their substantially
different global geometries establish only that other pipeline differences are
present; this unpaired comparison is not evidence for or against loop closure.
The possible contributors cannot be isolated from this figure and are therefore
not enumerated as causal explanations.

Figure~\ref{fig:with and without loop closure under different VINS} then holds
each pipeline and extrinsic setting fixed while toggling loop closure. Global
trajectory segments are reshaped in every pair, with the strongest visible
change under Original VINS with default extrinsics. Accurate extrinsics reduce,
but do not eliminate, the observed sensitivity in the evaluated sequence.
The qualitative trajectories show where global reshaping occurs, whereas the
controlled APE and RPE results in Table~\ref{tab:loop_closure_quantitative}
determine whether this reshaping improves or degrades accuracy. The section-level
finding is therefore limited to the distinction among local consistency,
numerical convergence, global accuracy, scale, and physical calibration; their
full cross-experiment relationship is summarized in
Section~\ref{sec:cross_experiment_findings}.

\textbf{Section-level finding.}
The combined backend experiments show that local temporal consistency, global
trajectory accuracy, numerical convergence, and physical calibration
correctness are distinct properties. Similar RPE or backend costs can coexist
with substantially different APE and extrinsic errors, and a numerically closer
scale can still produce worse global geometry. These properties must be
reported jointly rather than inferred from a single residual or trajectory
visualization.

\section{Cross-Experiment Findings and Evaluation Implications}
\label{sec:cross_experiment_findings}
\subsection{Cross-Experiment Synthesis}
\label{subsec:cross_experiment_synthesis}

The additional controlled experiments lead to four cross-experiment findings.
First, fusion performance and learned-prior incremental value are distinct. The
learned pose prior preserves the stability of Original VINS, but its mean APE
change under fixed extrinsics is negligible, and increasing its weight degrades
performance. The current learned prior is therefore either largely redundant
with the native VINS constraints or inconsistent with the metric backend state.
Second, local motion consistency, global trajectory accuracy, and physical
calibration correctness are empirically distinct. Fixed and Online strategies
produce nearly identical RPE but substantially different APE, so low RPE does
not imply low APE. For matched runs with the same residual construction, active
measurements, and weights, a low reported total backend cost does not imply
physically correct calibration. No cross-recording cost correlation is claimed.
Third, online camera--IMU calibration behaves as an initialization-dependent
local refinement process. Full backend optimization reduces calibration error
modestly but preserves the ranking and qualitative category of the initialized
solutions. When reliable offline calibration is available, fixing the extrinsic
state is more stable and interpretable than allowing unconstrained online
updates. When the extrinsic is genuinely unknown, the initialization quality
and sequence-dependent numerical conditioning influence whether the backend reaches a plausible
solution.
Fourth, global correction modules remain conditioned on the consistency of the
state they receive. Enabling the loop-closure pipeline can substantially reduce
APE under accurate extrinsics, but the same toggle can increase APE when the pre-loop trajectory and
calibration are inconsistent. A complete visual--inertial evaluation should
therefore jointly report APE, RPE, $\mathrm{SO}(3)$ calibration error,
translation calibration error, scale error, BA correction, and backend cost.
\subsection{Scope, Limitations, and Design Implications}
\label{subsec:scope_limitations_implications}
The learned-prior conclusions apply to the evaluated MonoViT-based
configuration and should not be generalized to every learned odometry model.
The 10 and 100~Hz IMU settings are diagnostic configurations rather than a
strict sampling-rate ablation. The VINS+Pose+Depth and optical-flow results are
secondary diagnostics, and the experiments do not isolate a single causal
source for error absorption among scale, temporal synchronization, IMU bias,
and learned-measurement uncertainty.
The evidence yields three independent evaluation statements:
\[
\begin{aligned}
\text{low RPE} &\not\Rightarrow \text{low APE}, \\
\text{low matched-run total cost} &\not\Rightarrow
\text{physically correct calibration}, \\
\text{repeatability} &\not\Rightarrow \text{absolute accuracy}.
\end{aligned}
\]
The cost statement applies only when residual definitions, active measurements,
and weights are the same. Reproducibility, numerical convergence, global
accuracy, and physical calibration must therefore be evaluated separately.
Future systems should use
uncertainty-aware prior weighting, observability-aware calibration updates,
strong extrinsic priors when reliable offline calibration is available, and
explicit checks of temporal and coordinate consistency.

\bibliographystyle{IEEEbib}
\bibliography{chapter2_refs}

\begin{thebibliography}{100}

\bibitem{Huang2019Review}
Guoquan Huang,
\newblock ``Visual-inertial navigation: A concise review,''
\newblock in {\em Proc. IEEE Int. Conf. Robot. Autom. (ICRA)}, 2019, pp. 9572--9582.

\bibitem{Forster2017Preintegration}
Christian Forster, Luca Carlone, Frank Dellaert, and Davide Scaramuzza,
\newblock ``On-manifold preintegration for real-time visual--inertial odometry,''
\newblock {\em IEEE Transactions on Robotics}, vol. 33, no. 1, pp. 1--21, 2017.

\bibitem{Mourikis2007MSCKF}
Anastasios~I. Mourikis and Stergios~I. Roumeliotis,
\newblock ``A multi-state constraint kalman filter for vision-aided inertial navigation,''
\newblock in {\em Proc. IEEE Int. Conf. Robot. Autom. (ICRA)}, 2007, pp. 3565--3572.

\bibitem{Leutenegger2015OKVIS}
Stefan Leutenegger, Simon Lynen, Michael Bosse, Roland Siegwart, and Paul Furgale,
\newblock ``Keyframe-based visual-inertial odometry using nonlinear optimization,''
\newblock {\em The International Journal of Robotics Research}, vol. 34, no. 3, pp. 314--334, 2015.

\bibitem{Qin2018VINSMono}
Tong Qin, Peiliang Li, and Shaojie Shen,
\newblock ``{VINS-Mono}: A robust and versatile monocular visual-inertial state estimator,''
\newblock {\em IEEE Transactions on Robotics}, vol. 34, no. 4, pp. 1004--1020, 2018.

\bibitem{Wang2017DeepVO}
Sen Wang, Ronald Clark, Hongkai Wen, and Niki Trigoni,
\newblock ``{DeepVO}: Towards end-to-end visual odometry with deep recurrent convolutional neural networks,''
\newblock in {\em Proc. IEEE Int. Conf. Robot. Autom. (ICRA)}, 2017, pp. 2043--2050.

\bibitem{Clark2017VINet}
Ronald Clark, Sen Wang, Hongkai Wen, Andrew Markham, and Niki Trigoni,
\newblock ``{VINet}: Visual-inertial odometry as a sequence-to-sequence learning problem,''
\newblock {\em Proceedings of the AAAI Conference on Artificial Intelligence}, vol. 31, no. 1, pp. 3995--4001, 2017.

\bibitem{Chen2019SelectiveFusion}
Changhao Chen, Stefano Rosa, Yishu Miao, Chris~Xiaoxuan Lu, Wei Wu, Andrew Markham, and Niki Trigoni,
\newblock ``Selective sensor fusion for neural visual-inertial odometry,''
\newblock in {\em Proc. IEEE/CVF Conf. Comput. Vis. Pattern Recognit. (CVPR)}, 2019, pp. 10534--10543.

\bibitem{Yang2020D3VO}
Nan Yang, Lukas von Stumberg, Rui Wang, and Daniel Cremers,
\newblock ``{D3VO}: Deep depth, deep pose and deep uncertainty for monocular visual odometry,''
\newblock in {\em Proc. IEEE/CVF Conf. Comput. Vis. Pattern Recognit. (CVPR)}, 2020, pp. 1278--1289.

\bibitem{Zuo2021CodeVIO}
Xingxing Zuo, Nathaniel Merrill, Wei Li, Yong Liu, Marc Pollefeys, and Guoquan Huang,
\newblock ``{CodeVIO}: Visual-inertial odometry with learned optimizable dense depth,''
\newblock in {\em Proc. IEEE Int. Conf. Robot. Autom. (ICRA)}, 2021, pp. 14382--14388.

\bibitem{Buchanan2023DeepBias}
Russell Buchanan, Varun Agrawal, Marco Camurri, Frank Dellaert, and Maurice Fallon,
\newblock ``Deep {IMU} bias inference for robust visual-inertial odometry with factor graphs,''
\newblock {\em IEEE Robotics and Automation Letters}, vol. 8, no. 1, pp. 41--48, 2023.

\bibitem{Furgale2013Kalibr}
Paul Furgale, Joern Rehder, and Roland Siegwart,
\newblock ``Unified temporal and spatial calibration for multi-sensor systems,''
\newblock in {\em Proc. IEEE/RSJ Int. Conf. Intell. Robots Syst. (IROS)}, 2013, pp. 1280--1286.

\bibitem{Mirzaei2008Calibration}
Faraz~M. Mirzaei and Stergios~I. Roumeliotis,
\newblock ``A kalman filter-based algorithm for {IMU}-camera calibration: Observability analysis and performance evaluation,''
\newblock {\em IEEE Transactions on Robotics}, vol. 24, no. 5, pp. 1143--1156, 2008.

\bibitem{Hesch2014Observability}
Joel~A. Hesch, Dimitrios~G. Kottas, Sean~L. Bowman, and Stergios~I. Roumeliotis,
\newblock ``Camera--{IMU}-based localization: Observability analysis and consistency improvement,''
\newblock {\em The International Journal of Robotics Research}, vol. 33, no. 1, pp. 182--201, 2014.

\bibitem{Yang2023SelfCalibration}
Yulin Yang, Patrick Geneva, Xingxing Zuo, and Guoquan Huang,
\newblock ``Online self-calibration for visual-inertial navigation: Models, analysis, and degeneracy,''
\newblock {\em IEEE Transactions on Robotics}, vol. 39, no. 5, pp. 3479--3498, 2023.

\bibitem{Sturm2012RGBDBenchmark}
J{\"u}rgen Sturm, Nikolas Engelhard, Felix Endres, Wolfram Burgard, and Daniel Cremers,
\newblock ``A benchmark for the evaluation of {RGB-D} {SLAM} systems,''
\newblock in {\em Proc. IEEE/RSJ Int. Conf. Intell. Robots Syst. (IROS)}, 2012, pp. 573--580.

\bibitem{Zhang2018TrajectoryEvaluation}
Zichao Zhang and Davide Scaramuzza,
\newblock ``A tutorial on quantitative trajectory evaluation for visual(-inertial) odometry,''
\newblock in {\em Proc. IEEE/RSJ Int. Conf. Intell. Robots Syst. (IROS)}, 2018, pp. 7244--7251.

\bibitem{Umeyama1991Alignment}
Shinji Umeyama,
\newblock ``Least-squares estimation of transformation parameters between two point patterns,''
\newblock {\em IEEE Transactions on Pattern Analysis and Machine Intelligence}, vol. 13, no. 4, pp. 376--380, 1991.

\bibitem{Zhao2022MonoViT}
Chaoqiang Zhao, Youmin Zhang, Matteo Poggi, Fabio Tosi, Xianda Guo, Zheng Zhu, Guan Huang, Yang Tang, and Stefano Mattoccia,
\newblock ``{MonoViT}: Self-supervised monocular depth estimation with a vision transformer,''
\newblock in {\em Proc. Int. Conf. 3D Vis. (3DV)}, 2022, pp. 668--678.

\bibitem{Godard2019Monodepth2}
Cl{\'e}ment Godard, Oisin~Mac Aodha, Michael Firman, and Gabriel~J. Brostow,
\newblock ``Digging into self-supervised monocular depth estimation,''
\newblock in {\em Proc. IEEE/CVF Int. Conf. Comput. Vis. (ICCV)}, 2019, pp. 3827--3837.

\bibitem{Qin2018TemporalCalibration}
Tong Qin and Shaojie Shen,
\newblock ``Online temporal calibration for monocular visual-inertial systems,''
\newblock in {\em Proc. IEEE/RSJ Int. Conf. Intell. Robots Syst. (IROS)}, 2018, pp. 3662--3669.

\bibitem{Li2013ConsistentVIO}
Mingyang Li and Anastasios~I. Mourikis,
\newblock ``High-precision, consistent {EKF}-based visual-inertial odometry,''
\newblock {\em The International Journal of Robotics Research}, vol. 32, no. 6, pp. 690--711, 2013.

\bibitem{Bloesch2017ROVIO}
Michael Bloesch, Michael Burri, Sammy Omari, Marco Hutter, and Roland Siegwart,
\newblock ``Iterated extended kalman filter based visual-inertial odometry using direct photometric feedback,''
\newblock {\em The International Journal of Robotics Research}, vol. 36, no. 10, pp. 1053--1072, 2017.

\bibitem{Geneva2020OpenVINS}
Patrick Geneva, Kevin Eckenhoff, Woosik Lee, Yulin Yang, and Guoquan Huang,
\newblock ``{OpenVINS}: A research platform for visual-inertial estimation,''
\newblock in {\em Proc. IEEE Int. Conf. Robot. Autom. (ICRA)}, 2020, pp. 4666--4672.

\bibitem{vonStumberg2018VIDSO}
Lukas von Stumberg, Vladyslav Usenko, and Daniel Cremers,
\newblock ``Direct sparse visual-inertial odometry using dynamic marginalization,''
\newblock in {\em Proc. IEEE Int. Conf. Robot. Autom. (ICRA)}, 2018, pp. 2510--2517.

\bibitem{Campos2021ORBSLAM3}
Carlos Campos, Richard Elvira, Juan J.~G{\'o}mez Rodr{\'i}guez, Jos{\'e} M.~M. Montiel, and Juan~D. Tard{\'o}s,
\newblock ``{ORB-SLAM3}: An accurate open-source library for visual, visual-inertial, and multimap {SLAM},''
\newblock {\em IEEE Transactions on Robotics}, vol. 37, no. 6, pp. 1874--1890, 2021.

\bibitem{Seiskari2022HybVIO}
Otto Seiskari, Pekka Rantalankila, Juho Kannala, Jerry Ylilammi, Esa Rahtu, and Arno Solin,
\newblock ``{HybVIO}: Pushing the limits of real-time visual-inertial odometry,''
\newblock in {\em Proc. IEEE/CVF Winter Conf. Appl. Comput. Vis. (WACV)}, 2022, pp. 287--296.

\bibitem{Gui2015Review}
Jianjun Gui, Dongbing Gu, Sen Wang, and Huosheng Hu,
\newblock ``A review of visual inertial odometry from filtering and optimisation perspectives,''
\newblock {\em Advanced Robotics}, vol. 29, no. 20, pp. 1289--1301, 2015.

\bibitem{Servieres2021Benchmarking}
Myriam Servi{\`e}res, Val{\'e}rie Renaudin, Alexis Dupuis, and Nicolas Antigny,
\newblock ``Visual and visual-inertial {SLAM}: State of the art, classification, and experimental benchmarking,''
\newblock {\em Journal of Sensors}, vol. 2021, pp. 2054828, 2021.

\bibitem{Liu2018ICEBA}
Haomin Liu, Mingyu Chen, Guofeng Zhang, Hujun Bao, and Yingze Bao,
\newblock ``{ICE-BA}: Incremental, consistent and efficient bundle adjustment for visual-inertial {SLAM},''
\newblock in {\em Proc. IEEE/CVF Conf. Comput. Vis. Pattern Recognit. (CVPR)}, 2018, pp. 1974--1982.

\bibitem{Zou2019StructVIO}
Danping Zou, Yuanxin Wu, Ling Pei, Haibin Ling, and Wenxian Yu,
\newblock ``{StructVIO}: Visual-inertial odometry with structural regularity of man-made environments,''
\newblock {\em IEEE Trans. Robot.}, vol. 35, no. 4, pp. 999--1013, 2019.

\bibitem{vonStumberg2022DMVIO}
Lukas von Stumberg and Daniel Cremers,
\newblock ``{DM-VIO}: Delayed marginalization visual-inertial odometry,''
\newblock {\em IEEE Robot. Autom. Lett.}, vol. 7, no. 2, pp. 1408--1415, 2022.

\bibitem{Schneider2018Maplab}
Thomas Schneider, Marcin Dymczyk, Marius Fehr, Kevin Egger, Simon Lynen, Igor Gilitschenski, and Roland Siegwart,
\newblock ``Maplab: An open framework for research in visual-inertial mapping and localization,''
\newblock {\em IEEE Robot. Autom. Lett.}, vol. 3, no. 3, pp. 1418--1425, 2018.

\bibitem{Cramariuc2023Maplab2}
Andrei Cramariuc, Lukas Bernreiter, Florian Tschopp, Marius Fehr, Victor Reijgwart, Juan Nieto, Roland Siegwart, and Cesar Cadena,
\newblock ``Maplab 2.0: A modular and multi-modal mapping framework,''
\newblock {\em IEEE Robot. Autom. Lett.}, vol. 8, no. 2, pp. 520--527, 2023.

\bibitem{Cao2022GVINS}
Shaozu Cao, Xiuyuan Lu, and Shaojie Shen,
\newblock ``{GVINS}: Tightly coupled {GNSS}--visual--inertial fusion for smooth and consistent state estimation,''
\newblock {\em IEEE Trans. Robot.}, vol. 38, no. 4, pp. 2004--2021, 2022.

\bibitem{Boche2025OKVIS2X}
Simon Boche, Jaehyung Jung, Sebasti{\'a}n~Barbas Laina, and Stefan Leutenegger,
\newblock ``{OKVIS2-X}: Open keyframe-based visual-inertial {SLAM} configurable with dense depth or {LiDAR}, and {GNSS},''
\newblock {\em IEEE Trans. Robot.}, vol. 41, pp. 6064--6083, 2025.

\bibitem{Han2019DeepVIO}
Liming Han, Yimin Lin, Guoguang Du, and Shiguo Lian,
\newblock ``{DeepVIO}: Self-supervised deep learning of monocular visual inertial odometry using 3-d geometric constraints,''
\newblock in {\em Proc. IEEE/RSJ Int. Conf. Intell. Robots Syst. (IROS)}, 2019, pp. 6906--6913.

\bibitem{Wei2020UnVIO}
Peng Wei, Guoliang Hua, Weibo Huang, Fanyang Meng, and Hong Liu,
\newblock ``Unsupervised monocular visual-inertial odometry network,''
\newblock in {\em Proc. Int. Joint Conf. Artif. Intell. (IJCAI)}, 2020, pp. 2347--2354.

\bibitem{Almalioglu2022SelfVIO}
Yasin Almalioglu, Mehmet Turan, Muhamad Risqi~U. Saputra, Pedro P.~B. de~Gusm{\~a}o, Andrew Markham, and Niki Trigoni,
\newblock ``{SelfVIO}: Self-supervised deep monocular visual--inertial odometry and depth estimation,''
\newblock {\em Neural Netw.}, vol. 150, pp. 119--136, 2022.

\bibitem{Tu2022EMAVIO}
Zheming Tu, Changhao Chen, Xianfei Pan, Ruochen Liu, Jiarui Cui, and Jun Mao,
\newblock ``{EMA-VIO}: Deep visual--inertial odometry with external memory attention,''
\newblock {\em IEEE Sens. J.}, vol. 22, no. 21, pp. 20877--20885, 2022.

\bibitem{Wang2021TartanVO}
Wenshan Wang, Yaoyu Hu, and Sebastian Scherer,
\newblock ``{TartanVO}: A generalizable learning-based visual odometry,''
\newblock in {\em Proc. Conf. Robot Learn. (CoRL)}, 2021, vol. 155 of {\em Proc. Mach. Learn. Res.}, pp. 1761--1772.

\bibitem{zhang2025depth}
Jinchang Zhang, Ningning Xu, Hao Zhang, and Guoyu Lu,
\newblock ``Depth estimation based on 3d gaussian splatting siamese defocus,''
\newblock in {\em 2025 IEEE International Conference on Robotics and Automation (ICRA)}. IEEE, 2025, pp. 01--07.

\bibitem{zhang2026fractal}
Jinchang Zhang, Xinrou Kang, and Guoyu Lu,
\newblock ``Fractal autoregressive depth estimation with continuous token diffusion,''
\newblock {\em arXiv preprint arXiv:2603.14702}, 2026.

\bibitem{Xue2019BeyondTracking}
Fei Xue, Xin Wang, Shunkai Li, Qiuyuan Wang, Junqiu Wang, and Hongbin Zha,
\newblock ``Beyond tracking: Selecting memory and refining poses for deep visual odometry,''
\newblock in {\em Proc. IEEE/CVF Conf. Comput. Vis. Pattern Recognit. (CVPR)}, 2019, pp. 8567--8575.

\bibitem{Czarnowski2020DeepFactors}
Jan Czarnowski, Tristan Laidlow, Ronald Clark, and Andrew~J. Davison,
\newblock ``{DeepFactors}: Real-time probabilistic dense monocular {SLAM},''
\newblock {\em IEEE Robot. Autom. Lett.}, vol. 5, no. 2, pp. 721--728, 2020.

\bibitem{Tang2019BANet}
Chengzhou Tang and Ping Tan,
\newblock ``{BA-Net}: Dense bundle adjustment networks,''
\newblock in {\em Proc. Int. Conf. Learn. Represent. (ICLR)}, 2019.

\bibitem{Min2020VOLDOR}
Zhixiang Min, Yiding Yang, and Enrique Dunn,
\newblock ``{VOLDOR}: Visual odometry from log-logistic dense optical flow residuals,''
\newblock in {\em Proc. IEEE/CVF Conf. Comput. Vis. Pattern Recognit. (CVPR)}, 2020, pp. 4897--4908.

\bibitem{Zhan2020VORevisited}
Huangying Zhan, Chamara~Saroj Weerasekera, Jia-Wang Bian, and Ian Reid,
\newblock ``Visual odometry revisited: What should be learnt?,''
\newblock in {\em Proc. IEEE Int. Conf. Robot. Autom. (ICRA)}, 2020, pp. 4203--4210.

\bibitem{Teed2021DROIDSLAM}
Zachary Teed and Jia Deng,
\newblock ``{DROID-SLAM}: Deep visual {SLAM} for monocular, stereo, and {RGB-D} cameras,''
\newblock in {\em Adv. Neural Inf. Process. Syst. (NeurIPS)}, 2021, vol.~34, pp. 16558--16569.

\bibitem{Teed2023DPVO}
Zachary Teed, Lahav Lipson, and Jia Deng,
\newblock ``Deep patch visual odometry,''
\newblock in {\em Adv. Neural Inf. Process. Syst. (NeurIPS)}, 2023, vol.~36, pp. 39033--39051.

\bibitem{Shen2023DytanVO}
Shihao Shen, Yilin Cai, Wenshan Wang, and Sebastian Scherer,
\newblock ``{DytanVO}: Joint refinement of visual odometry and motion segmentation in dynamic environments,''
\newblock in {\em Proc. IEEE Int. Conf. Robot. Autom. (ICRA)}, 2023, pp. 4048--4055.

\bibitem{Chen2024LEAPVO}
Weirong Chen, Le~Chen, Rui Wang, and Marc Pollefeys,
\newblock ``{LEAP-VO}: Long-term effective any point tracking for visual odometry,''
\newblock in {\em Proc. IEEE/CVF Conf. Comput. Vis. Pattern Recognit. (CVPR)}, 2024, pp. 19844--19853.

\bibitem{Qiu2025MACVO}
Yuheng Qiu, Yutian Chen, Zihao Zhang, Wenshan Wang, and Sebastian Scherer,
\newblock ``{MAC-VO}: Metrics-aware covariance for learning-based stereo visual odometry,''
\newblock in {\em Proc. IEEE Int. Conf. Robot. Autom. (ICRA)}, 2025, pp. 3803--3814.

\bibitem{Lai2025ZeroVO}
Lei Lai, Zekai Yin, and Eshed Ohn-Bar,
\newblock ``{ZeroVO}: Visual odometry with minimal assumptions,''
\newblock in {\em Proc. IEEE/CVF Conf. Comput. Vis. Pattern Recognit. (CVPR)}, 2025, pp. 17092--17102.

\bibitem{Wang2024DUSt3R}
Shuzhe Wang, Vincent Leroy, Yohann Cabon, Boris Chidlovskii, and Jerome Revaud,
\newblock ``{DUSt3R}: Geometric 3-d vision made easy,''
\newblock in {\em Proc. IEEE/CVF Conf. Comput. Vis. Pattern Recognit. (CVPR)}, 2024, pp. 20697--20709.

\bibitem{Leroy2024MASt3R}
Vincent Leroy, Yohann Cabon, and Jerome Revaud,
\newblock ``Grounding image matching in 3-d with {MASt3R},''
\newblock in {\em Proc. Eur. Conf. Comput. Vis. (ECCV)}, 2024, pp. 71--91.

\bibitem{Murai2025MASt3RSLAM}
Riku Murai, Eric Dexheimer, and Andrew~J. Davison,
\newblock ``{MASt3R-SLAM}: Real-time dense {SLAM} with 3-d reconstruction priors,''
\newblock in {\em Proc. IEEE/CVF Conf. Comput. Vis. Pattern Recognit. (CVPR)}, 2025, pp. 16695--16705.

\bibitem{DeTone2018SuperPoint}
Daniel DeTone, Tomasz Malisiewicz, and Andrew Rabinovich,
\newblock ``{SuperPoint}: Self-supervised interest point detection and description,''
\newblock in {\em Proc. IEEE/CVF Conf. Comput. Vis. Pattern Recognit. Workshops (CVPRW)}, 2018.

\bibitem{Dusmanu2019D2Net}
Mihai Dusmanu, Ignacio Rocco, Tomas Pajdla, Marc Pollefeys, Josef Sivic, Akihiko Torii, and Torsten Sattler,
\newblock ``{D2-Net}: A trainable {CNN} for joint description and detection of local features,''
\newblock in {\em Proc. IEEE/CVF Conf. Comput. Vis. Pattern Recognit. (CVPR)}, 2019, pp. 8084--8093.

\bibitem{Revaud2019R2D2}
Jerome Revaud, Cesar~De Souza, Martin Humenberger, and Philippe Weinzaepfel,
\newblock ``{R2D2}: Reliable and repeatable detector and descriptor,''
\newblock in {\em Adv. Neural Inf. Process. Syst. (NeurIPS)}, 2019, vol.~32.

\bibitem{Tyszkiewicz2020DISK}
Micha{\l}~J. Tyszkiewicz, Pascal Fua, and Eduard Trulls,
\newblock ``{DISK}: Learning local features with policy gradient,''
\newblock in {\em Adv. Neural Inf. Process. Syst. (NeurIPS)}, 2020, vol.~33.

\bibitem{Sarlin2020SuperGlue}
Paul-Edouard Sarlin, Daniel DeTone, Tomasz Malisiewicz, and Andrew Rabinovich,
\newblock ``{SuperGlue}: Learning feature matching with graph neural networks,''
\newblock in {\em Proc. IEEE/CVF Conf. Comput. Vis. Pattern Recognit. (CVPR)}, 2020, pp. 4937--4946.

\bibitem{Sun2021LoFTR}
Jiaming Sun, Zehong Shen, Yuang Wang, Hujun Bao, and Xiaowei Zhou,
\newblock ``{LoFTR}: Detector-free local feature matching with transformers,''
\newblock in {\em Proc. IEEE/CVF Conf. Comput. Vis. Pattern Recognit. (CVPR)}, 2021, pp. 8918--8927.

\bibitem{Lindenberger2023LightGlue}
Philipp Lindenberger, Paul-Edouard Sarlin, and Marc Pollefeys,
\newblock ``{LightGlue}: Local feature matching at light speed,''
\newblock in {\em Proc. IEEE/CVF Int. Conf. Comput. Vis. (ICCV)}, 2023, pp. 17581--17592.

\bibitem{Potje2024XFeat}
Guilherme Potje, Felipe Cadar, Andr{\'e} Araujo, Renato Martins, and Erickson~R. Nascimento,
\newblock ``{XFeat}: Accelerated features for lightweight image matching,''
\newblock in {\em Proc. IEEE/CVF Conf. Comput. Vis. Pattern Recognit. (CVPR)}, 2024, pp. 2682--2691.

\bibitem{Edstedt2024RoMa}
Johan Edstedt, Qiyu Sun, Georg B{\"o}kman, M{\aa}rten Wadenb{\"a}ck, and Michael Felsberg,
\newblock ``{RoMa}: Robust dense feature matching,''
\newblock in {\em Proc. IEEE/CVF Conf. Comput. Vis. Pattern Recognit. (CVPR)}, 2024, pp. 19790--19800.

\bibitem{lin2025keypoint}
Jiakai Lin, Jinchang Zhang, and Guoyu Lu,
\newblock ``Keypoint detection and description for raw bayer images,''
\newblock in {\em 2025 IEEE International Conference on Robotics and Automation (ICRA)}. IEEE, 2025, pp. 11736--11742.

\bibitem{Chen2018IONet}
Changhao Chen, Xiaoxuan Lu, Andrew Markham, and Niki Trigoni,
\newblock ``{IONet}: Learning to cure the curse of drift in inertial odometry,''
\newblock {\em Proc. AAAI Conf. Artif. Intell.}, vol. 32, no. 1, 2018.

\bibitem{Yan2018RIDI}
Hang Yan, Qi~Shan, and Yasutaka Furukawa,
\newblock ``{RIDI}: Robust {IMU} double integration,''
\newblock in {\em Proc. Eur. Conf. Comput. Vis. (ECCV)}, 2018, pp. 641--656.

\bibitem{Herath2020RoNIN}
Sachini Herath, Hang Yan, and Yasutaka Furukawa,
\newblock ``{RoNIN}: Robust neural inertial navigation in the wild: Benchmark, evaluations, and new methods,''
\newblock in {\em Proc. IEEE Int. Conf. Robot. Autom. (ICRA)}, 2020, pp. 3146--3152.

\bibitem{Liu2020TLIO}
Wenxin Liu, David Caruso, Eddy Ilg, Jing Dong, Anastasios~I. Mourikis, Kostas Daniilidis, Vijay Kumar, and Jakob Engel,
\newblock ``{TLIO}: Tight learned inertial odometry,''
\newblock {\em IEEE Robot. Autom. Lett.}, vol. 5, no. 4, pp. 5653--5660, 2020.

\bibitem{Brossard2020AIIMU}
Martin Brossard, Axel Barrau, and Silv{\`e}re Bonnabel,
\newblock ``{AI-IMU} dead-reckoning,''
\newblock {\em IEEE Trans. Intell. Veh.}, vol. 5, no. 4, pp. 585--595, 2020.

\bibitem{Brossard2020GyroDenoising}
Martin Brossard, Silv{\`e}re Bonnabel, and Axel Barrau,
\newblock ``Denoising {IMU} gyroscopes with deep learning for open-loop attitude estimation,''
\newblock {\em IEEE Robot. Autom. Lett.}, vol. 5, no. 3, pp. 4796--4803, 2020.

\bibitem{Rao2022CTIN}
Bingbing Rao, Ehsan Kazemi, Yifan Ding, Devu~M. Shila, Frank~M. Tucker, and Liqiang Wang,
\newblock ``{CTIN}: Robust contextual transformer network for inertial navigation,''
\newblock {\em Proc. AAAI Conf. Artif. Intell.}, vol. 36, no. 5, pp. 5413--5421, 2022.

\bibitem{Jayanth2025EqNIO}
Royina~Karegoudra Jayanth, Yinshuang Xu, Ziyun Wang, Evangelos Chatzipantazis, Kostas Daniilidis, and Daniel Gehrig,
\newblock ``{EqNIO}: Subequivariant neural inertial odometry,''
\newblock in {\em Proc. Int. Conf. Learn. Represent. (ICLR)}, 2025.

\bibitem{Zhao2025TartanIMU}
Shibo Zhao, Sifan Zhou, Raphael Blanchard, Yuheng Qiu, Wenshan Wang, and Sebastian Scherer,
\newblock ``{Tartan IMU}: A light foundation model for inertial positioning in robotics,''
\newblock in {\em Proc. IEEE/CVF Conf. Comput. Vis. Pattern Recognit. (CVPR)}, 2025, pp. 22520--22529.

\bibitem{Qiu2025AirIO}
Yuheng Qiu, Can Xu, Yutian Chen, Shibo Zhao, Junyi Geng, and Sebastian Scherer,
\newblock ``{AirIO}: Learning inertial odometry with enhanced {IMU} feature observability,''
\newblock {\em IEEE Robot. Autom. Lett.}, vol. 10, no. 9, pp. 9368--9375, 2025.

\bibitem{Vodisch2023CoVIO}
Niclas V{\"o}disch, Daniele Cattaneo, Wolfram Burgard, and Abhinav Valada,
\newblock ``{CoVIO}: Online continual learning for visual-inertial odometry,''
\newblock in {\em Proc. IEEE/CVF Conf. Comput. Vis. Pattern Recognit. Workshops (CVPRW)}, 2023, pp. 2464--2473.

\bibitem{Merrill2025DepthInit}
Nathaniel Merrill, Patrick Geneva, Saimouli Katragadda, Chuchu Chen, and Guoquan Huang,
\newblock ``Fast and robust learned single-view depth-aided monocular visual-inertial initialization,''
\newblock {\em The International Journal of Robotics Research}, vol. 44, no. 10--11, pp. 1619--1647, 2025.

\bibitem{zhang2024embodiment}
Jinchang Zhang, Praveen~Kumar Reddy, Xue-Iuan Wong, Yiannis Aloimonos, and Guoyu Lu,
\newblock ``Embodiment: Self-supervised depth estimation based on camera models,''
\newblock in {\em 2024 IEEE/RSJ International Conference on Intelligent Robots and Systems (IROS)}. IEEE, 2024, pp. 7809--7816.

\bibitem{Zhang2025Vision}
Jinchang Zhang and Guoyu Lu,
\newblock ``Vision-language embodiment for monocular depth estimation,''
\newblock in {\em Proceedings of the Computer Vision and Pattern Recognition Conference}, 2025, pp. 29479--29489.

\bibitem{zhang2025language}
Jinchang Zhang, Zijun Li, and Guoyu Lu,
\newblock ``Language-depth navigated thermal and visible image fusion,''
\newblock {\em arXiv preprint arXiv:2503.08676}, 2025.

\bibitem{Saputra2020DeepTIO}
Muhamad Risqi~U. Saputra, Pedro P.~B. de~Gusm{\~a}o, Chris~Xiaoxuan Lu, Yasin Almalioglu, Stefano Rosa, Changhao Chen, Johan Wahlstr{\"o}m, Wei Wang, Andrew Markham, and Niki Trigoni,
\newblock ``{DeepTIO}: A deep thermal-inertial odometry with visual hallucination,''
\newblock {\em IEEE Robot. Autom. Lett.}, vol. 5, no. 2, pp. 1672--1679, 2020.

\bibitem{Li2014TemporalCalibration}
Mingyang Li and Anastasios~I. Mourikis,
\newblock ``Online temporal calibration for camera--{IMU} systems: Theory and algorithms,''
\newblock {\em The International Journal of Robotics Research}, vol. 33, no. 7, pp. 947--964, 2014.

\bibitem{Ling2018TimeOffset}
Yonggen Ling, Linchao Bao, Zequn Jie, Fengming Zhu, Ziyang Li, Shanmin Tang, Yongsheng Liu, Wei Liu, and Tong Zhang,
\newblock ``Modeling varying camera--{IMU} time offset in optimization-based visual-inertial odometry,''
\newblock in {\em Computer Vision--ECCV 2018}. 2018, pp. 491--507, Springer.

\bibitem{Huai2022SquareRoot}
Zheng Huai and Guoquan Huang,
\newblock ``Square-root robocentric visual-inertial odometry with online spatiotemporal calibration,''
\newblock {\em IEEE Robotics and Automation Letters}, vol. 7, no. 4, pp. 9961--9968, 2022.

\bibitem{Lang2022CtrlVIO}
Xiaolei Lang, Jiajun Lv, Jianxin Huang, Yukai Ma, Yong Liu, and Xingxing Zuo,
\newblock ``{Ctrl-VIO}: Continuous-time visual-inertial odometry for rolling shutter cameras,''
\newblock {\em IEEE Robot. Autom. Lett.}, vol. 7, no. 4, pp. 11537--11544, 2022.

\bibitem{Chen2025iKalibr}
Shuolong Chen, Xingxing Li, Shengyu Li, Yuxuan Zhou, and Xiaoteng Yang,
\newblock ``{iKalibr}: Unified targetless spatiotemporal calibration for resilient integrated inertial systems,''
\newblock {\em IEEE Trans. Robot.}, vol. 41, pp. 1618--1638, 2025.

\bibitem{Campos2019Initialization}
Carlos Campos, Jos{\'e} M.~M. Montiel, and Juan~D. Tard{\'o}s,
\newblock ``Fast and robust initialization for visual-inertial {SLAM},''
\newblock in {\em Proc. IEEE Int. Conf. Robot. Autom. (ICRA)}, 2019, pp. 1288--1294.

\bibitem{Geiger2013KITTI}
Andreas Geiger, Philip Lenz, Christoph Stiller, and Raquel Urtasun,
\newblock ``Vision meets robotics: The {KITTI} dataset,''
\newblock {\em The International Journal of Robotics Research}, vol. 32, no. 11, pp. 1231--1237, 2013.

\bibitem{Burri2016EuRoC}
Michael Burri, Janosch Nikolic, Pascal Gohl, Thomas Schneider, Joern Rehder, Sammy Omari, Markus~W. Achtelik, and Roland Siegwart,
\newblock ``The {EuRoC} micro aerial vehicle datasets,''
\newblock {\em The International Journal of Robotics Research}, vol. 35, no. 10, pp. 1157--1163, 2016.

\bibitem{Schubert2018TUMVI}
David Schubert, Thore Goll, Nikolaus Demmel, Vladyslav Usenko, J{\"o}rg St{\"u}ckler, and Daniel Cremers,
\newblock ``The {TUM VI} benchmark for evaluating visual-inertial odometry,''
\newblock in {\em Proc. IEEE/RSJ Int. Conf. Intell. Robots Syst. (IROS)}, 2018, pp. 1680--1687.

\bibitem{Cortes2018ADVIO}
Santiago Cort{\'e}s, Arno Solin, Esa Rahtu, and Juho Kannala,
\newblock ``{ADVIO}: An authentic dataset for visual-inertial odometry,''
\newblock in {\em Proc. Eur. Conf. Comput. Vis. (ECCV)}, 2018, pp. 425--440.

\bibitem{Wang2020TartanAir}
Wenshan Wang, Delong Zhu, Xiangwei Wang, Yaoyu Hu, Yuheng Qiu, Chen Wang, Yafei Hu, Ashish Kapoor, and Sebastian Scherer,
\newblock ``{TartanAir}: A dataset to push the limits of visual {SLAM},''
\newblock in {\em Proc. IEEE/RSJ Int. Conf. Intell. Robots Syst. (IROS)}, 2020, pp. 4909--4916.

\bibitem{Shi2020OpenLORIS}
Xuesong Shi, Dongjiang Li, Pengpeng Zhao, Qinbin Tian, Yuxin Tian, Qiwei Long, Chunhao Zhu, Jingwei Song, Fei Qiao, Le~Song, Yangquan Guo, Zhigang Wang, Yimin Zhang, Baoxing Qin, Wei Yang, Fangshi Wang, Rosa H.~M. Chan, and Qi~She,
\newblock ``Are we ready for service robots? the {OpenLORIS-Scene} datasets for lifelong {SLAM},''
\newblock in {\em Proc. IEEE Int. Conf. Robot. Autom. (ICRA)}, 2020, pp. 3139--3145.

\bibitem{Wenzel20214Seasons}
Patrick Wenzel, Rui Wang, Nan Yang, Qing Cheng, Qadeer Khan, Lukas von Stumberg, Niclas Zeller, and Daniel Cremers,
\newblock ``{4Seasons}: A cross-season dataset for multi-weather {SLAM} in autonomous driving,''
\newblock in {\em Proc. German Conf. Pattern Recognit. (GCPR)}, 2021, pp. 404--417.

\bibitem{Helmberger2022Hilti}
Michael Helmberger, Kristian Morin, Beda Berner, Nitish Kumar, Giovanni Cioffi, and Davide Scaramuzza,
\newblock ``The {Hilti SLAM Challenge} dataset,''
\newblock {\em IEEE Robot. Autom. Lett.}, vol. 7, no. 3, pp. 7518--7525, 2022.

\bibitem{Ramezani2020NewerCollege}
Milad Ramezani, Yiduo Wang, Marco Camurri, David Wisth, Matias Mattamala, and Maurice Fallon,
\newblock ``The {Newer College Dataset}: Handheld {LiDAR}, inertial and vision with ground truth,''
\newblock in {\em Proc. IEEE/RSJ Int. Conf. Intell. Robots Syst. (IROS)}, 2020, pp. 4353--4360.

\bibitem{Krishnan2025CityScale}
Anusha Krishnan, Shaohui Liu, Paul-Edouard Sarlin, Oscar Gentilhomme, David Caruso, Maurizio Monge, Richard Newcombe, Jakob Engel, and Marc Pollefeys,
\newblock ``Benchmarking egocentric visual-inertial {SLAM} at city scale,''
\newblock in {\em Proc. IEEE/CVF Int. Conf. Comput. Vis. (ICCV)}, 2025, pp. 25207--25217.

\bibitem{Delmerico2018Benchmark}
Jeffrey Delmerico and Davide Scaramuzza,
\newblock ``A benchmark comparison of monocular visual-inertial odometry algorithms for flying robots,''
\newblock in {\em Proc. IEEE Int. Conf. Robot. Autom. (ICRA)}, 2018, pp. 2502--2509.

\bibitem{Strasdat2010ScaleDrift}
Hauke Strasdat, J.~M.~M. Montiel, and Andrew~J. Davison,
\newblock ``Scale drift-aware large scale monocular {SLAM},''
\newblock in {\em Proc. Robot. Sci. Syst. (RSS)}, 2010.

\bibitem{MurArtal2017MapReuse}
Ra{\'u}l Mur-Artal and Juan~D. Tard{\'o}s,
\newblock ``Visual-inertial monocular {SLAM} with map reuse,''
\newblock {\em IEEE Robotics and Automation Letters}, vol. 2, no. 2, pp. 796--803, 2017.

\bibitem{Rosinol2020Kimera}
Antoni Rosinol, Marcus Abate, Yun Chang, and Luca Carlone,
\newblock ``Kimera: An open-source library for real-time metric-semantic localization and mapping,''
\newblock in {\em Proc. IEEE Int. Conf. Robot. Autom. (ICRA)}, 2020, pp. 1689--1696.

\bibitem{Usenko2020NFR}
Vladyslav Usenko, Nikolaus Demmel, David Schubert, J{\"o}rg St{\"u}ckler, and Daniel Cremers,
\newblock ``Visual-inertial mapping with non-linear factor recovery,''
\newblock {\em IEEE Robotics and Automation Letters}, vol. 5, no. 2, pp. 422--429, 2020.

\bibitem{Ferrera2021OV2SLAM}
Maxime Ferrera, Alexandre Eudes, Julien Moras, Martial Sanfourche, and Guy~Le Besnerais,
\newblock ``{OV$^2$SLAM}: A fully online and versatile visual {SLAM} for real-time applications,''
\newblock {\em IEEE Robot. Autom. Lett.}, vol. 6, no. 2, pp. 1399--1406, 2021.

\bibitem{Song2022DynaVINS}
Seungwon Song, Hyungtae Lim, Alex~Junho Lee, and Hyun Myung,
\newblock ``{DynaVINS}: A visual-inertial {SLAM} for dynamic environments,''
\newblock {\em IEEE Robot. Autom. Lett.}, vol. 7, no. 4, pp. 11523--11530, 2022.

\bibitem{Bescos2021DynaSLAMII}
Berta Bescos, Carlos Campos, Juan~D. Tardos, and Jose Neira,
\newblock ``{DynaSLAM II}: Tightly coupled multi-object tracking and {SLAM},''
\newblock {\em IEEE Robot. Autom. Lett.}, vol. 6, no. 3, pp. 5191--5198, 2021.

\bibitem{Xu2025AirSLAM}
Kuan Xu, Yuefan Hao, Shenghai Yuan, Chen Wang, and Lihua Xie,
\newblock ``{AirSLAM}: An efficient and illumination-robust point-line visual {SLAM} system,''
\newblock {\em IEEE Trans. Robot.}, vol. 41, pp. 1673--1692, 2025.

\bibitem{Sarlin2019HFNet}
Paul-Edouard Sarlin, Cesar Cadena, Roland Siegwart, and Marcin Dymczyk,
\newblock ``From coarse to fine: Robust hierarchical localization at large scale,''
\newblock in {\em Proc. IEEE/CVF Conf. Comput. Vis. Pattern Recognit. (CVPR)}, 2019, pp. 12708--12717.

\bibitem{Hausler2021PatchNetVLAD}
Stephen Hausler, Sourav Garg, Ming Xu, Michael Milford, and Tobias Fischer,
\newblock ``{Patch-NetVLAD}: Multi-scale fusion of locally global descriptors for place recognition,''
\newblock in {\em Proc. IEEE/CVF Conf. Comput. Vis. Pattern Recognit. (CVPR)}, 2021, pp. 14136--14147.

\bibitem{Tian2022KimeraMulti}
Yulun Tian, Yun Chang, Fernando~Herrera Arias, Carlos Nieto-Granda, Jonathan~P. How, and Luca Carlone,
\newblock ``{Kimera-Multi}: Robust, distributed, dense metric-semantic {SLAM} for multi-robot systems,''
\newblock {\em IEEE Trans. Robot.}, vol. 38, no. 4, pp. 2022--2038, 2022.

\end{thebibliography}
\end{document}